\documentclass{article}
\PassOptionsToPackage{square,comma,numbers,compress}{natbib}

\usepackage[preprint]{neurips_2023}

\usepackage[utf8]{inputenc} %
\usepackage[T1]{fontenc}    %
\usepackage[hidelinks,colorlinks=true]{hyperref}       %
\usepackage[dvipsnames]{xcolor}
\usepackage{url}            %
\usepackage{booktabs}       %
\usepackage{amsfonts}       %
\usepackage{nicefrac}       %
\usepackage{microtype}      %
\usepackage{xcolor}         %
\usepackage{graphicx}
\graphicspath{ {figures/} }
\usepackage{array}
\usepackage{times,graphicx,tabulary,multirow,xspace}
\usepackage{subfig}
\usepackage{changepage}
\usepackage{color, colortbl}
\definecolor{LightGray}{rgb}{0.92,0.92,0.92}
\definecolor{Gray1}{rgb}{0.95,0.95,0.95}
\definecolor{Gray2}{rgb}{0.9,0.9,0.9}

\definecolor{redhl}{HTML}{ea9999}
\definecolor{greenhl}{HTML}{d9ead3}
\definecolor{bluehl}{HTML}{c9daf8}
\definecolor{yellowhl}{HTML}{fff2cc}
\makeatletter
\DeclareRobustCommand\onedot{\futurelet\@let@token\@onedot}
\def\@onedot{\ifx\@let@token.\else.\null\fi\xspace}

\def\eg{\emph{e.g}\onedot} 
\def\ie{\emph{i.e}\onedot} 
 
\def\etc{\emph{etc}\onedot}

\makeatother

\newcommand{\modelname}{GPT-4V\xspace}
\newcommand{\modelnamefull}{GPT-4V(ision)\xspace}

\title{The Dawn of LMMs: \\Preliminary Explorations with \modelnamefull}

\author{
{\bf Zhengyuan Yang$^{*}$, Linjie Li$^{*}$, Kevin Lin$^{*}$, Jianfeng Wang$^{*}$, Chung-Ching Lin$^{*}$,} \\ 
\textbf{Zicheng Liu, Lijuan Wang$^{*\spadesuit}$} \\
Microsoft Corporation\\
\and
\footnotesize{
$^*$~Core Contributor \;
$^{\spadesuit}$~Project Lead \;
}
}

\begin{document}

\maketitle

\vspace{20pt}
\begin{abstract}
Large multimodal models (LMMs) extend large language models (LLMs) with multi-sensory skills, such as visual understanding, to achieve stronger generic intelligence. In this paper, we analyze the latest model, \modelnamefull~\cite{gpt4,gpt4v,gpt4vcontribution,gpt4vblog}\footnote{This report explores \modelnamefull~with the vision capability and refers to the model as ``\modelname,'' following the OpenAI reports~\cite{gpt4v,gpt4}. We refer to the text-only version of the model as ``GPT-4 (no vision)''~\cite{gpt4}.}, to deepen the understanding of LMMs. The analysis focuses on the intriguing tasks that \modelname~can perform, containing test samples to probe the quality and genericity of \modelname's capabilities, its supported inputs and working modes, and the effective ways to prompt the model. In our approach to exploring \modelname, we curate and organize a collection of carefully designed qualitative samples spanning a variety of domains and tasks. Observations from these samples demonstrate that \modelname's unprecedented ability in processing arbitrarily interleaved multimodal inputs and the genericity of its capabilities together make \modelname~a powerful multimodal generalist system. Furthermore, \modelname's unique capability of understanding visual markers drawn on input images can give rise to new 
human-computer interaction methods such as visual referring prompting.
We conclude the report with in-depth discussions on the emerging application scenarios and the future research directions for \modelname-based systems. We hope that this preliminary exploration will inspire future research on the next-generation multimodal task formulation, new ways to exploit and enhance LMMs to solve real-world problems, and gaining better understanding of multimodal foundation models.
Finally, we acknowledge that the model under our study is solely the product of OpenAI's innovative work, and they should be fully credited for its development. Please see the GPT-4V contributions paper~\cite{gpt4vcontribution} for the authorship and credit attribution: \url{https://cdn.openai.com/contributions/gpt-4v.pdf}.
\end{abstract}
% \vspace{20pt}

{
  \hypersetup{linkcolor=black}
  \tableofcontents
  \label{sec:toc}
}

\clearpage
{
\hypersetup{linkcolor=black}
\addcontentsline{toc}{section}{List of Figures}
\listoffigures
\label{sec:lof}
}
\clearpage

\section{Introduction}
\label{sec:01intro}
\subsection{Motivation and Overview}

The breakthroughs in large language models (LLMs)~\cite{brown2020language,gpt4,chowdhery2022palm,anil2023palm,touvron2023llama,hoffmann2022training} have shown remarkable versatilities and capabilities across various domains and tasks. The next evolution in this field, large multimodal models (LMMs), aims to expand upon the capabilities of LLMs by integrating multi-sensory skills to achieve even stronger general intelligence. Given the dominance of the visual in human senses~\cite{cornsweet2012visual,hutmacher2019there}, many LMM studies start with extending the vision capability.
Preliminary research investigations either finetune a vision encoder to align with a frozen pre-trained LLM~\cite{tsimpoukelli2021multimodal,alayrac2022flamingo,li2023blip,huang2023language,driess2023palme,anas_awadalla_2023_7733589,gong2023multimodalgpt,zhu2023minigpt,liu2023visual,dai2023instructblip,ye2023mplug}, or use a vision-language model to convert visual inputs to text descriptions that LLMs can understand~\cite{zeng2022socratic,yang2022empirical,wanglanguage,hu2022promptcap,shao2023prompting,yang2023mm}.
However, most existing models~\cite{anas_awadalla_2023_7733589,gong2023multimodalgpt,zhu2023minigpt,liu2023visual,dai2023instructblip,li2023multimodal} are of limited model and data scales, potentially restricting the emergence of various intriguing abilities. Consequently, it remains unclear what are the status quo and emergent multimodal abilities of LMMs that are developed based on the state-of-the-art LLMs, such as GPT-4 (no vision)~\cite{gpt4} and PaLM~\cite{chowdhery2022palm,anil2023palm}.
In this paper, we report our preliminary explorations with (an early version of) \modelname, a state-of-the-art LMM with vision, built based on the SOTA LLM and trained with a large scale of multimodal data.

Our exploration of \modelname~is guided by the following questions.
\begin{enumerate}
    \item \emph{What are \modelname's supported inputs and working modes?} The genericity of multimodal models inevitably requires the system to work with the arbitrary mix of different input modalities. \modelname~shows unprecedented ability in understanding and processing an arbitrary mix of input images, sub-images, texts, scene texts, and visual pointers. We also demonstrate that \modelname~well supports the test-time techniques observed in LLMs, including instruction following~\cite{ouyang2022training}, chain-of-thoughts~\cite{wei2022chain,kojima2022large}, in-context few-shot learning~\cite{brown2020language}, \etc.
    \item \emph{What are the quality and genericity of \modelname's capabilities on different domains and tasks?} We sample queries covering a wide range of domains and tasks to understand \modelname's capabilities, including open-world visual understanding, visual description, multimodal knowledge, commonsense, scene text understanding, document reasoning, coding, temporal reasoning, abstract reasoning, emotion understanding, and many more. \modelname~ shows impressive human-level capabilities across many of the experimented domains.
    \item \emph{What are effective ways to use and prompt \modelname?} \modelname~is strong in understanding pixel space edits, such as visual pointers and scene texts drawn on input images. Inspired by this capability, we discuss the ``visual referring prompting'' that directly edits input images to instruct the task of interest. Visual referring prompting can be seamlessly used together with other image and text prompts, presenting a nuanced interface for instruction and example demonstrations.
    \item \emph{What are promising future directions?} Given \modelname's strong capability across domains and tasks, we ask what is the next step for multimodal learning, and more broadly for artificial intelligence. We organize our thoughts and explorations into two perspectives, \ie, emergent novel application scenarios to focus on, and the future research directions for \modelname-based systems. We present our preliminary explorations to inspire future studies.
\end{enumerate}

Guided by the aforementioned problems, we comprehensively organize and list our explored qualitative results. The report contains minimal quantitative benchmark results, and instead consists of mainly selected interesting qualitative examples. Despite being less rigorous, this design allows for providing a more comprehensive analysis covering a broad range of domains, tasks, working modes, and prompting techniques, under a fixed capacity. We believe this organized collection of explorations will inspire future works in emerging novel applications, next-generation multimodal task formulation, and developing advanced LMM-based intelligent systems.

\subsection{Our Approach in Exploring \modelname}
\noindent\textbf{Goal of this report.}
The standard approach for evaluating a system is by benchmarking it against a series of carefully designed datasets, each representing a specific domain and task. 
One challenge is that some of the existing benchmarks may not be suitable for evaluating LMMs anymore. For example, the image captioning outputs of LMMs are much richer and contain more detailed descriptions than the ground truths in the image captioning benchmark datasets~\cite{chen2015microsoft}. There is also a lack of public information regarding \modelname's large-scale pre-training, which may violate the train-test setup for certain existing datasets and invalidate those benchmark numbers. %
Because of this, restricting the evaluation to \emph{existing} benchmarks and metrics may unintentionally narrow the scope of \modelname's assessment.
Developing a comprehensive list of next-generation evaluation tasks and benchmarks would be the ideal ultimate solution. However, we left those as future work due to the significant efforts required. 

In lieu of quantitative benchmarking, this paper focuses on using qualitative results to provide a glimpse of \modelname's new capabilities and potential emerging use cases. Our goal is to discover and preview what \modelname~might already be capable of, even though these novel capabilities may not yet be entirely reliable. We hope this collection of explorations will inspire future research in establishing quantitative benchmarks for next-generation multimodal tasks, modernizing existing benchmarks, further improving model performance and system reliability, and sparkling innovation in emerging use cases. Following this, we will delve into the core designs for our approach to exploring \modelname.

\noindent\textbf{Sample selection guidance.}
This report focuses on presenting qualitative results to showcase the potential capabilities of \modelname, rather than providing comprehensive quantitative benchmark results. This naturally raises the question of the reliability of the showcased examples. The examples featured in this report may require careful instruction tuning to amplify \modelname's corresponding capabilities. It should be noted that some complex cases may only work with the specifically designed prompts. As such, the capabilities demonstrated may not consistently work across different samples. Instead of showing only the reliable functionalities, the primary objective of this report is to provide readers with a list of our discovered potential capabilities of \modelname, which might otherwise be overlooked after a few unsuccessful trials. %

\noindent\textbf{Sample selection to prevent mere memorizing from training.}
A fundamental design consideration in qualitative reports~\cite{bubeck2023sparks} is discerning models' true capabilities from merely memorizing responses from training samples or making educated guesses based on hints from instructions and in-context examples.
We carefully control both the images and text in the input prompts to prevent them from being seen during \modelname~training. We generate original text queries from scratch, and try to use images that are either not accessible online or with a timestamp beyond April 2023. We will indicate instances where a specific sample does not meet this criterion, \eg, deliberately using samples from specific vision-language datasets.
Beyond ensuring that samples are unseen, we incorporate rationale queries into the process. These queries are designed to probe the model's reasoning process, thereby validating \modelname's possession of the intended capability.

\noindent\textbf{The default working mode.}
As later detailed in Section~\ref{sec:02method}, \modelname~works effectively in different working modes, including zero-shot learning with instructions, in-context few-shot learning, \etc. Among them, this report primarily focuses on zero-shot instruction tuning, as opposed to in-context few-shot learning. This design is to prevent potential information leakage from in-context examples. While in-context few-shot examples can enhance performance and reliability, they do not consistently engender new capabilities. As such, we designate zero-shot as the default working mode for presentation, and reduce the use of in-context examples to minimize examples' impacts on the assessed capabilities.

\subsection{How to Read this Report?}

This report documents the explorations of \modelname~conducted by researchers in the computer vision and vision-language multimodal field. It is primarily geared towards fellow researchers in related disciplines who seek to gain a qualitative impression of LMM's capabilities and understand its difference from traditional vision-language models. The report is also prepared for professionals for whom AI or computer science may be outside their specialties, to assist them in conceptualizing ways LMMs can enhance their proficiency within their distinct domains of expertise.

We give an overview of the report, structured around the four core questions that guide our exploration.
\begin{enumerate}
    \item \emph{What are \modelname's supported inputs and working modes?} Section~\ref{sec:02inputmode} summarizes \modelname's supported inputs and presents an overview of their corresponding use cases. 
    Based on the flexible interleaved image-text inputs, Section~\ref{sec:02method} discusses \modelname's different working modes, such as instruction tuning, in-context learning, and other emergent usages. The section covers the novel ways of using and prompting \modelname, aiming to provide a comprehensive overview of how we will use \modelname~in subsequent sections.
    \item \emph{What are the quality and genericity of \modelname's capabilities on different domains and tasks?} The exploration of this question makes up a large portion of the report. Section~\ref{sec:03vl} provides a comprehensive analysis covering a wide range of vision and vision-language scenarios, including image description and recognition on different domains, dense visual understanding, multimodal knowledge, commonsense, scene text understanding, document reasoning, and many more. We also separate out several novel and interesting capabilities. 
    Section~\ref{sec:07temporal} studies \modelname's capability in temporal, motion, and video understanding. Section~\ref{sec:08iq} explores the abstract visual understanding and reasoning capability, and Section~\ref{sec:09eq} covers the emotion and sentiment understanding.
    \item \emph{What are effective ways to use and prompt \modelname?}
    We start the discussion on this question from the working mode and prompting method introduction in Section~\ref{sec:02method}. In Section~\ref{sec:05pointing}, we highlight one novel promoting technique, namely visual referring prompting, which draws visual pointers and scene texts on input images to prompt \modelname. We demonstrate the flexible prompting methods, such as the combination of instruction and example demonstrations, throughout the report in the given examples.
    \item \emph{What are promising future directions?} Section~\ref{sec:10app} focuses on the novel use cases facilitated by \modelname. We hope these initial examples could inspire future works to design new task setups and present rigorous benchmarks. Section~\ref{sec:06tool} imagines powerful future systems that can be built based on \modelname, such as the multimodal plugins, multimodal chains, 
    self-reflection, self-consistency, and retrieval-augmented LMMs, \etc.
\end{enumerate}

In addition to this overview and the \hyperref[sec:toc]{\textbf{table of contents}}, we have also included a \hyperref[sec:lof]{\textbf{list of figures}}. The list enumerates the qualitative examples detailed within the report, serving as an additional tool to help readers navigate to their scenarios of interest.
\section{\modelname's Input Modes}
\label{sec:02inputmode}

This section summarizes \modelname's supported inputs, \ie, functioning as a uni-model language model with the text-only inputs, taking single image-text pair optionally with only a single image, and taking interleaved image-text pairs optionally with only multiple image inputs. We next highlight the representative use cases under these different input modes.

\subsection{Text-only Inputs}
\label{sec:02sub-lm}
\modelname's strong language capability enables it to serve as an effective unimodal language model~\cite{devlin2018bert,raffel2020exploring,brown2020language} with text-only inputs. Operating exclusively with text for both input and output, \modelname~is capable of performing a wide variety of language and coding tasks. We refer readers to the GPT-4 technical report~\cite{gpt4} for the comprehensive and in-depth analysis of \modelname's language and coding capabilities, as well as the comparison with GPT-4 (no vision).

\subsection{Single Image-text Pair}
\label{sec:02sub-vl}
\modelname, the latest large multimodal model, takes images and texts as inputs to generate textual outputs. In line with existing general-purpose vision-language models~\cite{anderson2018bottom,lu2019vilbert,li2019visualbert,alberti2019fusion,li2020unicoder,tan2019lxmert,su2019vl,zhou2020unified,chen2019uniter,lu202012,gan2020large,li2020oscar,huang2020pixel,kim2021vilt,li2021align,wang2021simvlm,cho2021unifying,yang2022unitab,dou2022empirical,alayrac2022flamingo,wang2022git,gan2022vision,doucoarse,zou2023generalized,li2023multimodal}, \modelname~can take a single image-text pair or a single image as input to perform various vision and vision-language tasks, such as image recognition~\cite{deng2009imagenet}, object localization~\cite{zhou2016learning}, image captioning~\cite{chen2015microsoft}, visual question answering~\cite{VQA_15}, visual dialogue~\cite{das2017visual}, dense caption~\cite{johnson2016densecap}, and so on. We note that the text in the image-text pair can be used either as instruction like ``describe the image'' for captioning, or as the query input like the question in visual question answering. \modelname's exceptional intelligence is exemplified by its significantly enhanced performance and generalizability compared to prior arts. A comprehensive analysis of its multimodal capabilities on various domains is detailed in Section~\ref{sec:03vl}.

\begin{figure*}[h!]
\centering
\includegraphics[width=\textwidth]{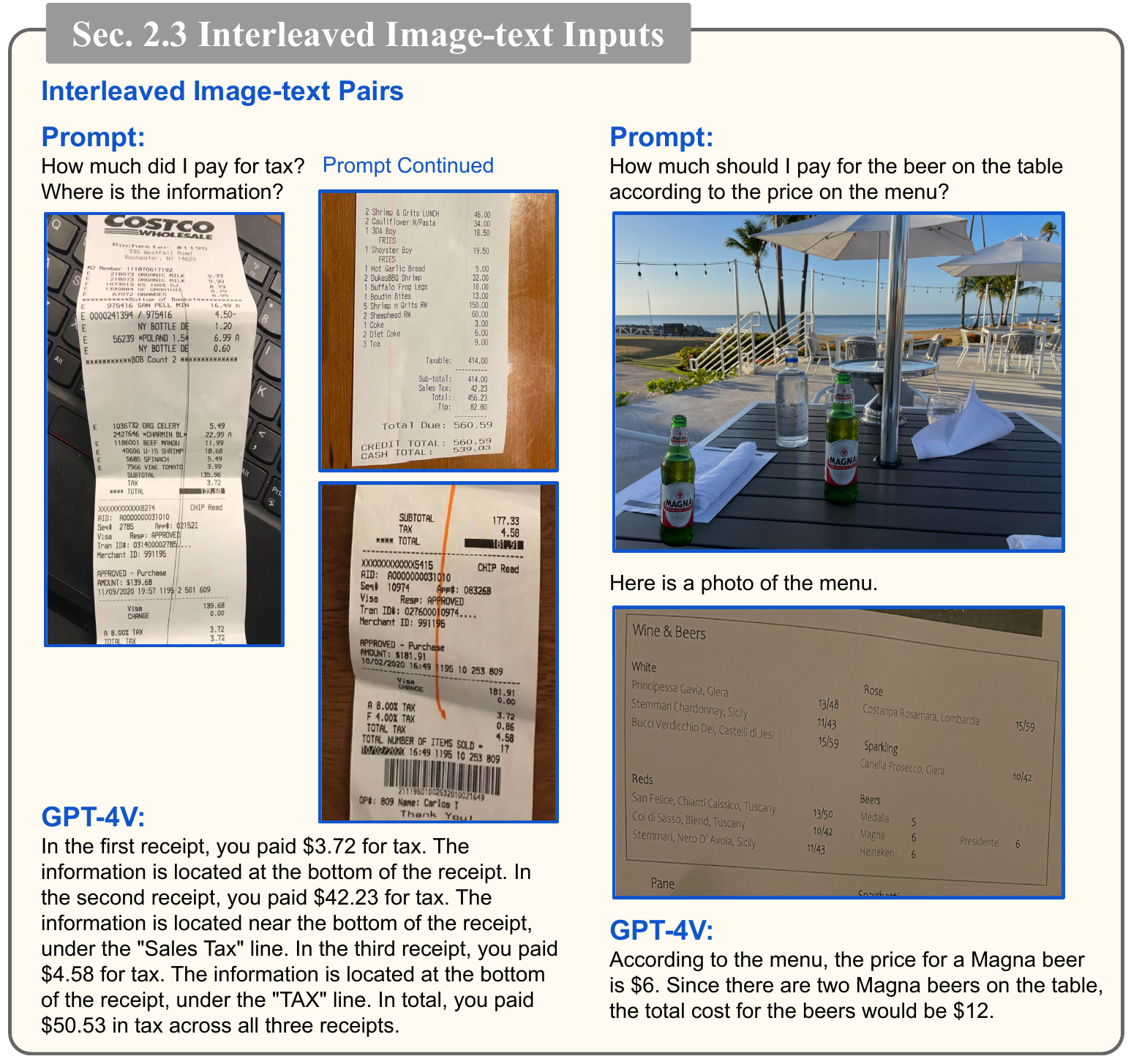}
\caption[Section~\ref{sec:02sub-interleave}: multi-image inputs and interleaved image-text pairs.]{\modelname~can work with multi-image and interleaved image-text inputs. Check Section~\ref{sec:02sub-interleave} for detailed discussions.
}
\label{fig:intro_modeb}
\end{figure*}
\subsection{Interleaved Image-text Inputs}
\label{sec:02sub-interleave}
The generality of \modelname~is further enhanced by its ability to handle flexibly interleaved image-text inputs. The interleaved image-text inputs can be either visually centric such as multiple images with a short question or instruction, text-centric such as a long webpage with two inserted images, or a balanced mixture of images and texts. This mode of mixed input provides flexibility for a wide array of applications. For example, it can compute the total tax paid across multiple receipt images, as shown in Figure~\ref{fig:intro_modeb}. It also enables processing multiple input images and extracting queried information. \modelname~could also effectively associate information across interleaved image-text inputs, such as finding the beer price on the menu, counting the number of beers, and returning the total cost, as shown in Figure~\ref{fig:intro_modeb}.
Beyond direct applications, processing interleaved image-text inputs serves as a fundamental component for in-context few-shot learning and other advanced test-time prompting techniques, thereby further boosting \modelname's generality. We demonstrate these intriguing novel usages in the next section, Section~\ref{sec:02method}.

\clearpage
\section{\modelname's Working Modes and Prompting Techniques}
\label{sec:02method}
\begin{figure*}[h!]
\centering
\includegraphics[width=\textwidth]{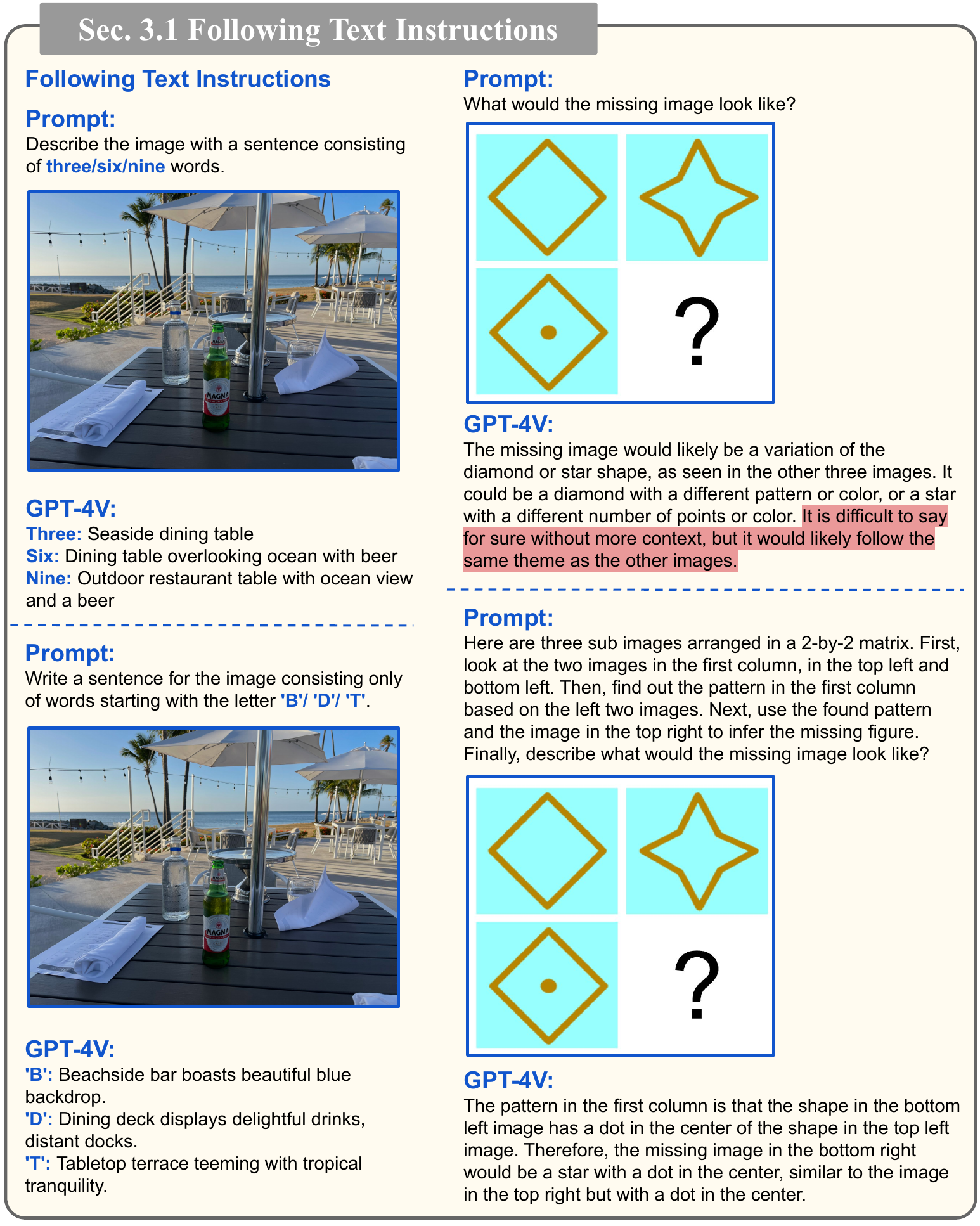}
\caption[Section~\ref{sec:instruct}: following text instructions.]{\modelname~can understand and follow text instructions, to generate the desired text outputs or learn to perform a new task. \colorbox{redhl}{Red} highlights the less informative answer. Check Section~\ref{sec:instruct} for detailed discussions.
}
\vspace{30pt}
\label{fig:intro_modea}
\end{figure*}

\begin{figure*}[h!]
\centering
\vspace{-13mm}
\includegraphics[width=.95\textwidth]{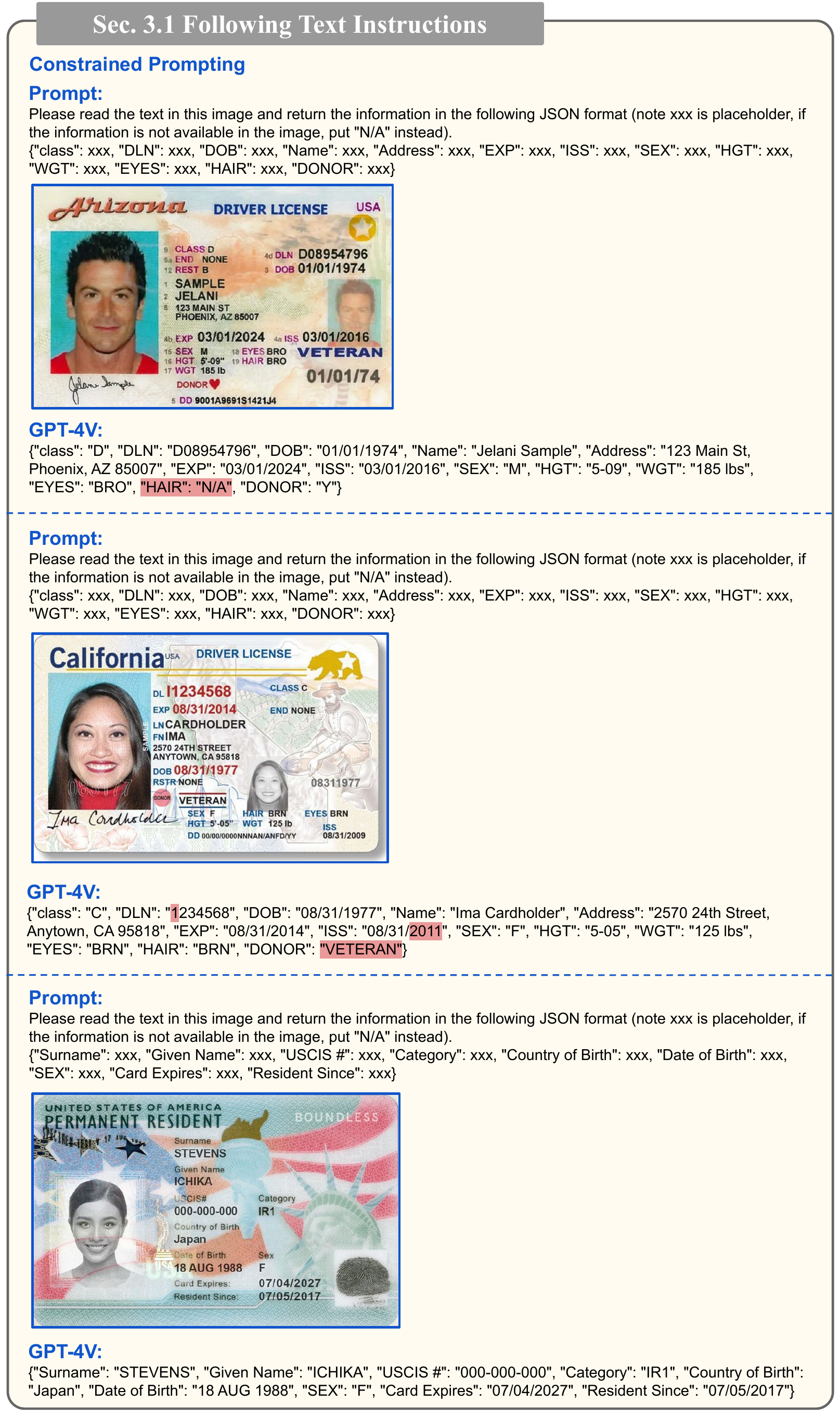}
\caption[Section~\ref{sec:instruct}: constrained prompting.]{Constrained prompting to return in JSON format. Images are example IDs for samples. \colorbox{redhl}{Red} highlights the wrong answer.  Check Section~\ref{sec:instruct} for detailed discussions. 
}
\label{fig:constrained_prompting_1}
\end{figure*} 

\begin{figure*}[h!]
\centering
\vspace{-13mm}
\includegraphics[width=.95\textwidth]{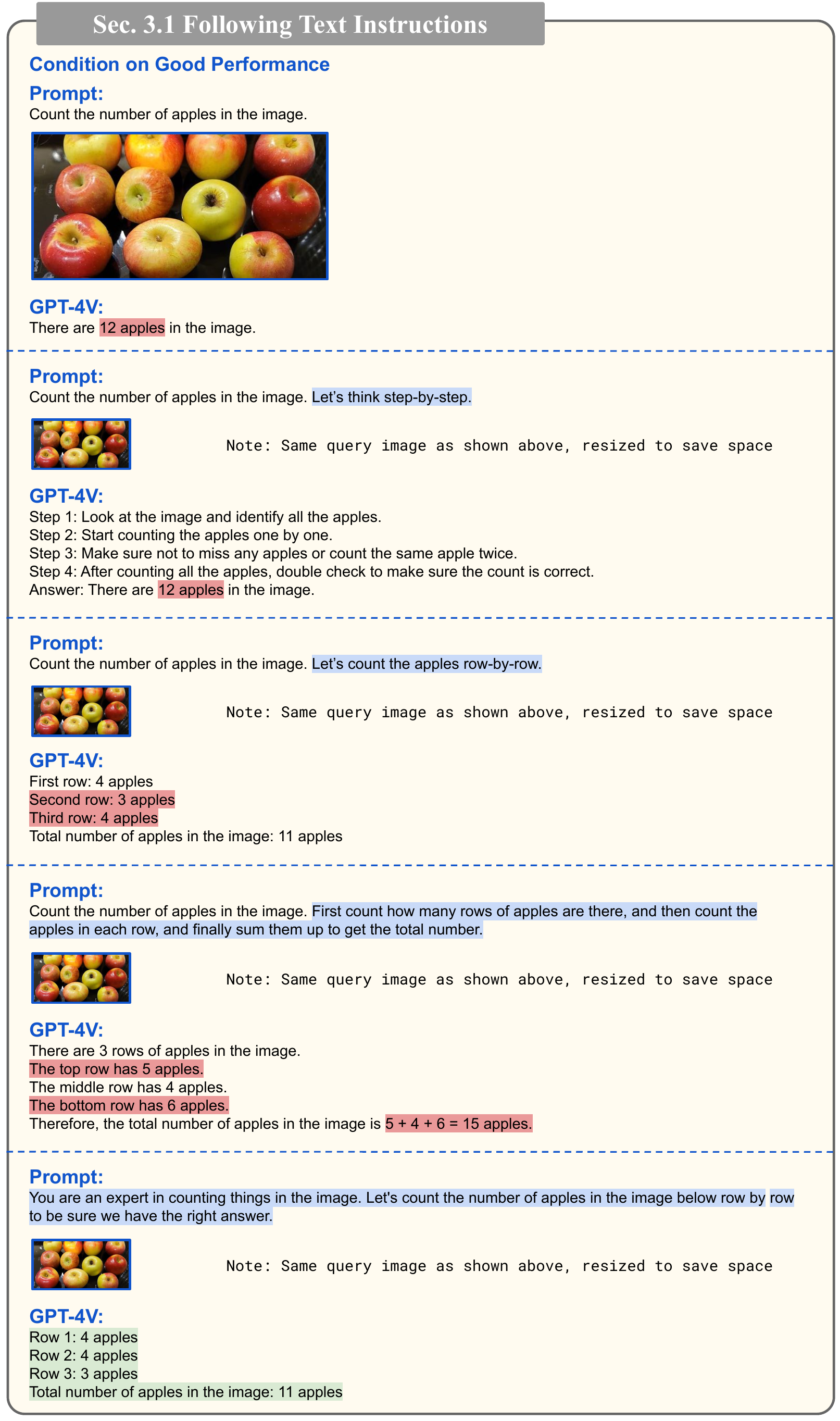}
\caption[Section~\ref{sec:instruct}: condition on good performance.]{Condition on good performance to improve counting. \colorbox{greenhl}{Green}(\colorbox{redhl}{Red}) highlights the correct (wrong) answer. \colorbox{bluehl}{Blue} indicates different ways to prompting in addition to the basic requirement of ``Count the number of apples in the image.'' Check Section~\ref{sec:instruct} for detailed discussions. 
}
\label{fig:condition_on_good_performance}
\vspace{-10mm}
\end{figure*}

\subsection{Following Text Instructions}%
\label{sec:instruct}
One unique strength of \modelname~is its generality, partially achieved via
its strong capability in understanding and following text instructions~\cite{ouyang2022training,mishra2021cross,wei2021finetuned,sanh2021multitask}. %
Instructions provide a natural way to define and customize the desired output text for arbitrary vision-language use cases. Figure~\ref{fig:intro_modea} shows an example of image descriptions with constraints on sentence length and the words to use. Alternatively, on the input side, \modelname~could understand the detailed instructions to perform challenging tasks, such as enabling \modelname~to better interpret the abstract reasoning question by providing instructions on intermediate steps. The ability to learn new tasks from instructions shows great potential in adapting to various unseen applications and tasks, as detailed in Section~\ref{sec:10app}. In line with recent studies~\cite{alayrac2022flamingo,anas_awadalla_2023_7733589,gong2023multimodalgpt,zhu2023minigpt,liu2023visual,dai2023instructblip}, the instructions discussed in this subsection are mostly in the text format, providing language descriptions of the interested task. We will discuss \modelname's unique capability of following multimodal example-grounded instructions later in Section~\ref{sec:mminstruct}. 

In addition, we showcase how text instructions play an important role in shaping \modelname's response with two techniques adopted from LLM literature~\cite{Guidance,zhou2022large}, ($i$) ``constrained prompting'' so that \modelname responds in a certain format; and ($ii$) ``condition on good performance'' that explicitly asks for good performance from \modelname.

\paragraph{Constrained prompting.}
In Figure~\ref{fig:constrained_prompting_1}, we prompt \modelname to read the text in the image and return the information in a specific JSON format. Although \modelname makes some mistakes in extracting the corresponding information from driver's licenses, the responses are constrained to the JSON format specified in the text instruction.  We leverage this technique for certain application scenarios in Section~\ref{sec:10app}.

\paragraph{Condition on good performance.} One observation about LLMs is that LLMs don't want to succeed~\cite{stateofgpt}. Rather, they want to imitate training sets with a spectrum of performance qualities. If the user wants to succeed in a task given to the model, the user should explicitly ask for it, which has proven useful in improving the performance of LLMs~\cite{zhou2022large}. In the context of LMMs, we have similar observations. 
In Figure~\ref{fig:condition_on_good_performance}, we compare the model's response to different text instructions for counting. We start with a simple and clear prompt: ``Count the number of apples in the image.'' However, \modelname incorrectly counts a total of 12 apples in the image. To improve its performance, we explore the use of zero-shot chain-of-thought from~\cite{kojima2022large} for LLMs by adding the phrase ``Let's think step-by-step.'' Although \modelname's predicted steps are generally correct, they are not very helpful for the final count, as it still arrives at the incorrect answer of ``12 apples.'' Next, we modify the instruction to ``Let's count the apples row-by-row,'' which is more relevant to the visual input. While \modelname provides the correct total count, it makes mistakes in counting the second/third row. When we further expand the instruction to ``First count how many rows of apples there are, then count the apples in each row, and finally sum them up to get the total number,'' the final answer deviates even more from the correct answer (15 vs. 11).  
Finally, imitating ``Let's work this out in a step by step way to be sure we have the right answer.'' in~\cite{zhou2022large} for LLMs, we design the prompt as follows: 
 ``You are an expert in counting things in the image. Let's count the number of apples in the image below row by row to be sure we have the right answer.''. The first sentence in our prompt asks \modelname to assume the role of an expert in counting, and the second sentence explicitly instructs \modelname to succeed. With this design, \modelname successfully returns the correct answer for each row as well as the total count. Throughout the paper, we employ this technique in various scenarios for better performance.

\subsection{Visual Pointing and Visual Referring Prompting}
\label{sec:02sub-vprompt}
\begin{figure*}[t]
\centering
\includegraphics[width=\textwidth]{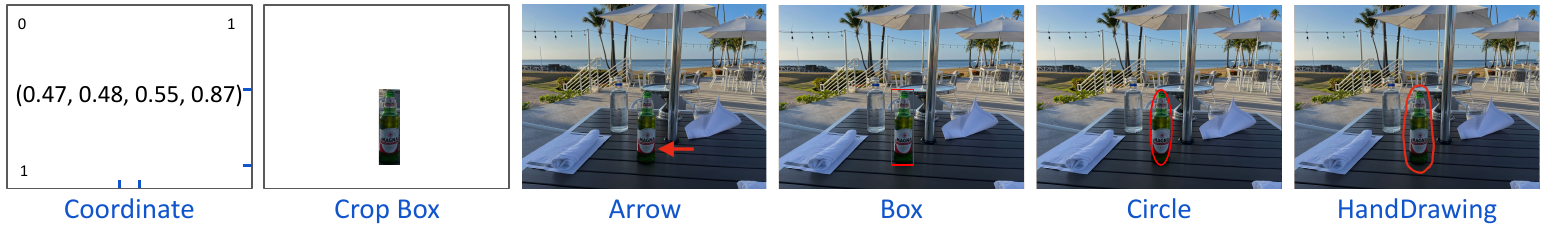}
\caption[Section~\ref{sec:02sub-vprompt}: different modes of visual pointing.]{Different modes of ``visual pointing'' in multimodal interaction. }%
\label{fig:intro_point}
\end{figure*}

\begin{figure*}[h!]
\centering
\includegraphics[width=\textwidth]{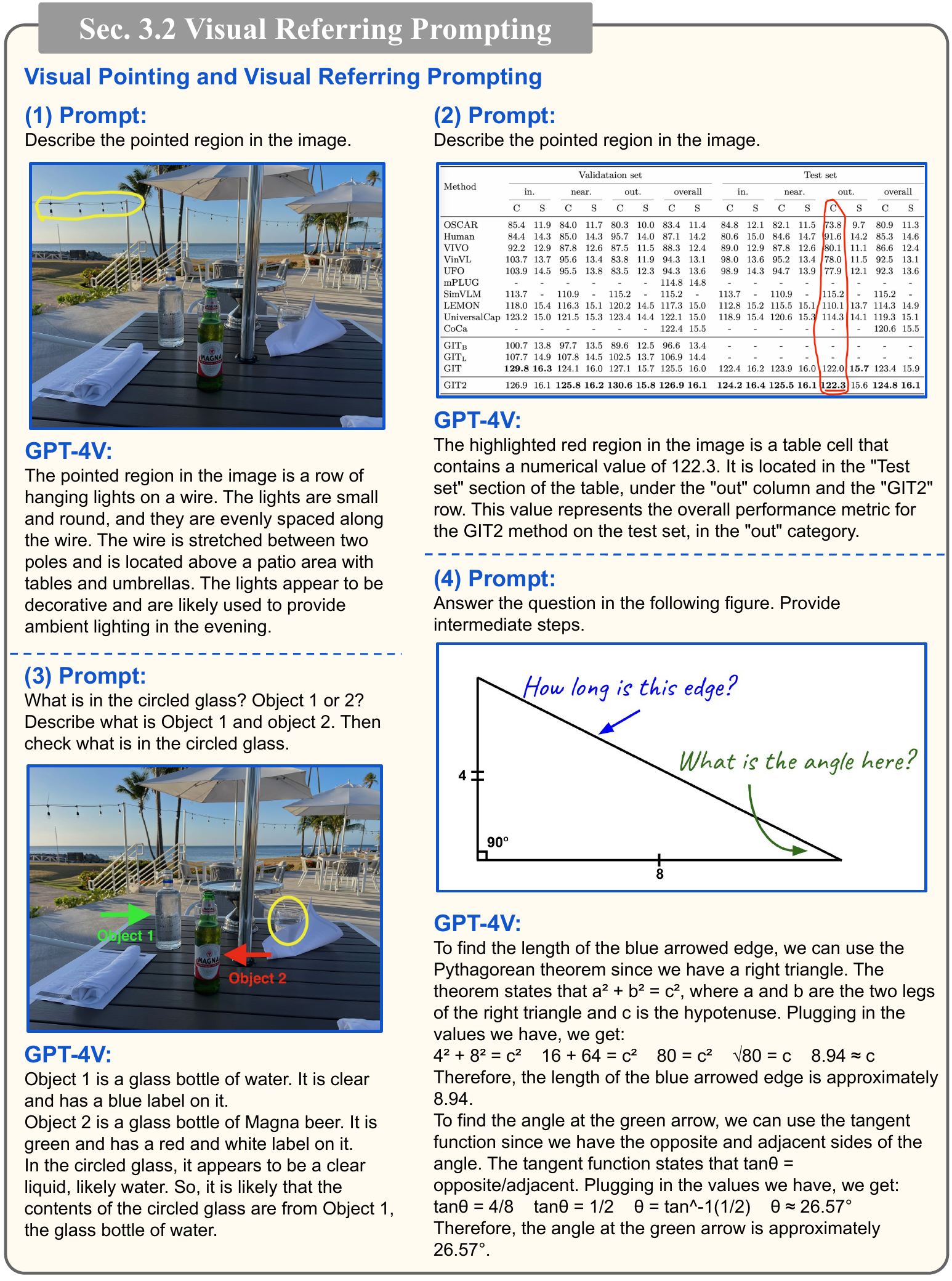}
\caption[Section~\ref{sec:02sub-vprompt}: visual pointing and visual referring prompting.]{\modelname~demonstrates the unique capability of understanding visual pointing directly overlaid on images. Based on such capability, we explore visual referring prompting that edits input image pixels (\eg, drawing visual pointers and scene texts) to prompt the task of interest. Check Section~\ref{sec:02sub-vprompt} for detailed discussions.
}
\label{fig:intro_moded}
\end{figure*}

Pointing is a fundamental aspect of human-human interaction~\cite{malle2001intentions}. To provide a comparable channel of interaction, various forms of ``pointing'' are studied to refer to an arbitrary spatial region of interest. For example, as depicted in Figure~\ref{fig:intro_point}, ``pointing'' can be represented as numerical spatial coordinates such as box coordinates and image crops, or visual markers overlaid on image pixels such as arrows, boxes, circles, and hand drawings. We observe that \modelname~is particularly strong in understanding visual pointers drawn directly on images.
Given the flexibility of drawing on images, this capability can be used as a natural approach for future human-computer interaction in the wild~\cite{mani2020point,shtedritski2023does,zhu2023minigpt}. To this end, we explore a new prompting method named visual referring prompting, where people edit the pixel space of input images to specify the desired objective, such as drawing visual pointers or handwriting scene texts.
As illustrated in Figure~\ref{fig:intro_moded}, visual referring prompting edits the image pixels, instead of the conventional text prompts, to perform the task of interest. For example, it could be a simple grounded description, which focuses on describing the pointed object while maintaining the understanding of the global image context, as shown in Figure~\ref{fig:intro_moded}~(1,2). Visual referring prompting also enables other novel use cases, such as associating the pointed object with an index written in scene text (Figure~\ref{fig:intro_moded}~(3)), or solving the question asked near the queried edge or angle (Figure~\ref{fig:intro_moded}~(4)). Section~\ref{sec:05pointing} will discuss visual referring prompting in more detail.


\begin{figure*}[h!]
\centering
\vspace{-40pt}
\includegraphics[width=\textwidth]{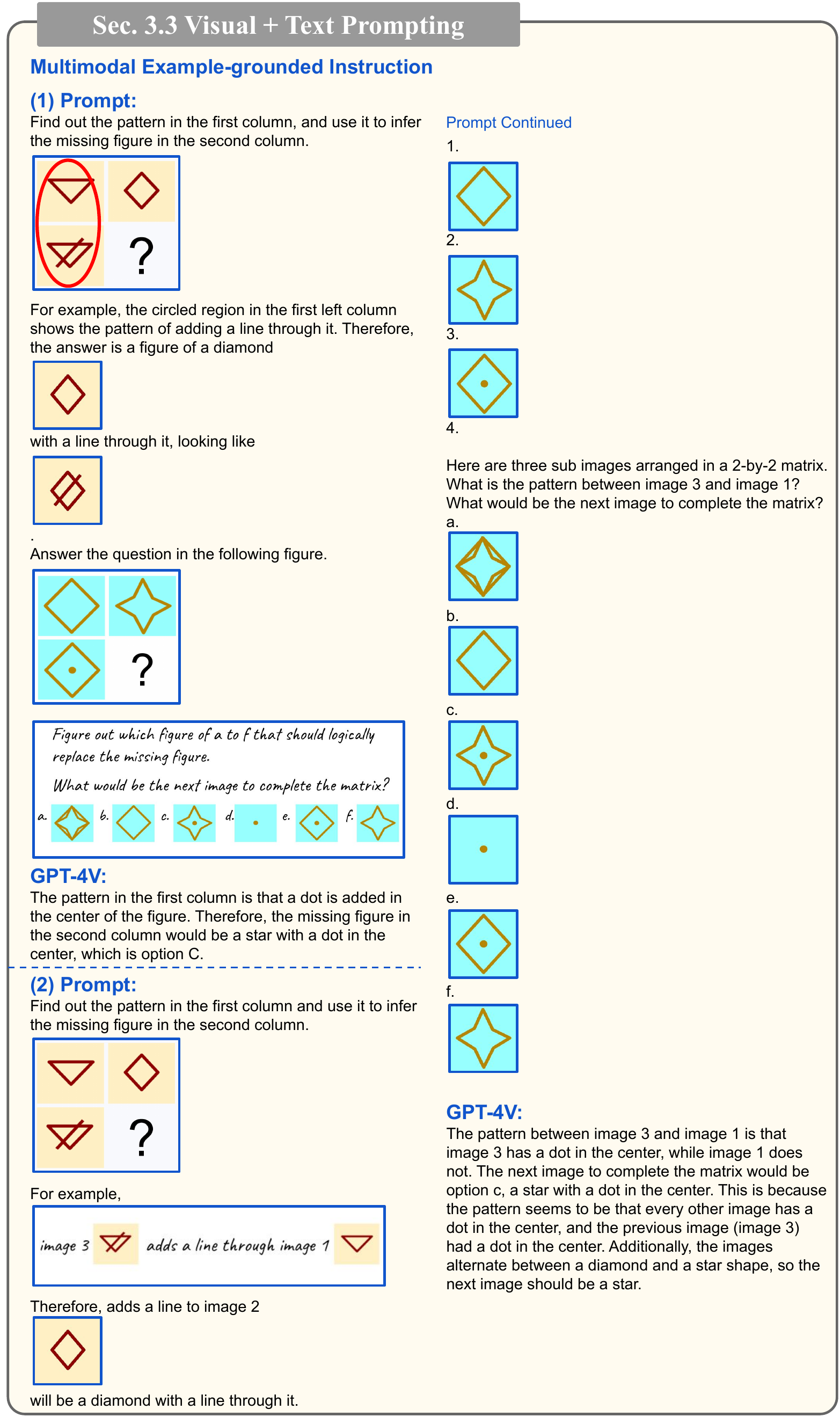}
\caption[Section~\ref{sec:mminstruct}: multimodal example-grounded instruction.]{\modelname~is strong in interpreting an arbitrary mix of images, sub-images, texts, scene texts, and visual pointer inputs. These elements could serve as instructions, examples, or input queries, helping \modelname~to effectively perform novel tasks. Check Section~\ref{sec:mminstruct} for detailed discussions.
}
\label{fig:intro_modee}
\end{figure*}

\subsection{Visual + Text Prompting}%
\label{sec:mminstruct}
Visual referring prompting can be smoothly used together with other image-text prompts, 
presenting a nuanced interface that succinctly represents the problem of interest. Figure~\ref{fig:intro_modee} presents two examples to showcase the flexibility of \modelname's prompt, particularly its proficiency in integrating different input formats and seamlessly mixing instructions with examples in the inputs. \modelname's genericity and flexibility result in a human-like comprehension of multimodal instructions and an unprecedented ability to adapt to unseen tasks. %

\paragraph{Integrated multimodal instruction inputs.}
Existing models usually have implicit constraints on how interleaved image-text inputs should be formatted, \eg, in-context few-shot learning requires image-text pairs to share a similar format as the query input. In contrast,
\modelname~shows the genericity in processing an arbitrary mix of images, sub-images, texts, scene texts, and visual pointers. For example, to illustrate the ``adding a line'' pattern in Figure~\ref{fig:intro_modee}, one could either point to the first column in the matrix image with a circle as in sub-figure (1), or incorporate the sub-images inline as in sub-figure (2). Similarly, for input query, one could either present a large figure with the question as scene texts as in sub-figure (1), or send the mix of texts and sub-images as in sub-figure (2). In contrast to \modelname's flexibility, existing multimodal models are highly restricted in terms of how they can combine images and texts, and the number of images they can process, thereby imposing limitations on the model's capability and genericity.

\paragraph{Multimodal example-grounded instruction.}
In addition to supporting more flexible input formats, \modelname's genericity also opens up more effective ways of illustrating the task to perform, compared with the instruction-following mode and in-context few-shot learning. %
Instruction-following techniques~\cite{ouyang2022training,mishra2021cross,wei2021finetuned,sanh2021multitask}, originally proposed for NLP tasks, intuitively focus on task instructions purely in the textual format. The text instruction is loosely related to the visual query input and thus may not provide a clear task demonstration. While in-context few-shot learning~\cite{brown2020language,tsimpoukelli2021multimodal,alayrac2022flamingo} provides test-time examples that contain both images and texts, these examples must align perfectly with the format of the inference query, making them complex and lengthy to incorporate. Furthermore, in-context examples are usually used separately from instructions, requiring the model to infer the task objective and thereby compromising the demonstration's effectiveness.
In contrast, \modelname's capability to comprehend multimodal instructions enables task demonstrations to be grounded onto corresponding in-context examples, therefore more effectively illustrating the task of interest. For example, in Figure~\ref{fig:intro_modee}, grounding instructions of ``finding the pattern in the first column'' onto the key steps in demonstration examples (\ie, the circled pattern in (1) and corresponding sub-figures in (2)) simplifies the learning process and enhances the model's performance. This approach also mirrors the human learning process, which involves abstract instructions paired with intuitive examples.

\clearpage
\subsection{In-context Few-shot Learning}
\label{sec:02fewshot}

\begin{figure*}[h!]
\centering
\vspace{-20mm}
\includegraphics[width=\textwidth]{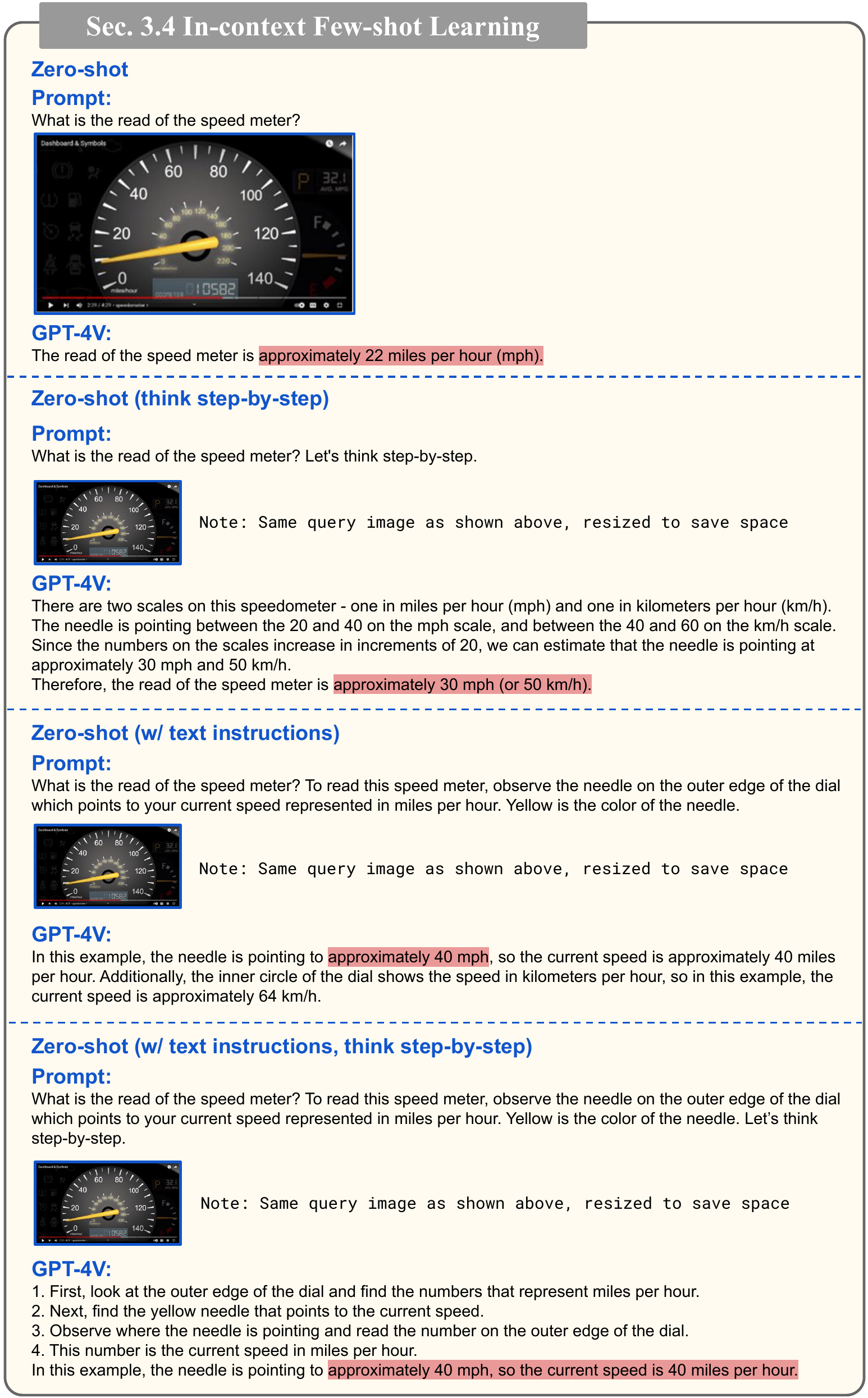}
\caption[Section~\ref{sec:02fewshot}: few-shot comparison example 1, Part I: zero-shot.]{Zero-shot performance under the challenging scenario of reading a speed meter. \modelname~fails to read the speed meter accurately even with different ways of ZS prompting. \colorbox{redhl}{Red} highlights the wrong answer. Check Section~\ref{sec:02fewshot} for detailed discussions. 
}
\label{fig:speed_meter_zs_9}
\end{figure*}

\begin{figure}[h!]
\captionsetup[subfloat]{farskip=1pt,captionskip=1pt}
\centering
\vspace{-12mm}
   \subfloat[ 
   ]{
      \includegraphics[width=\textwidth]{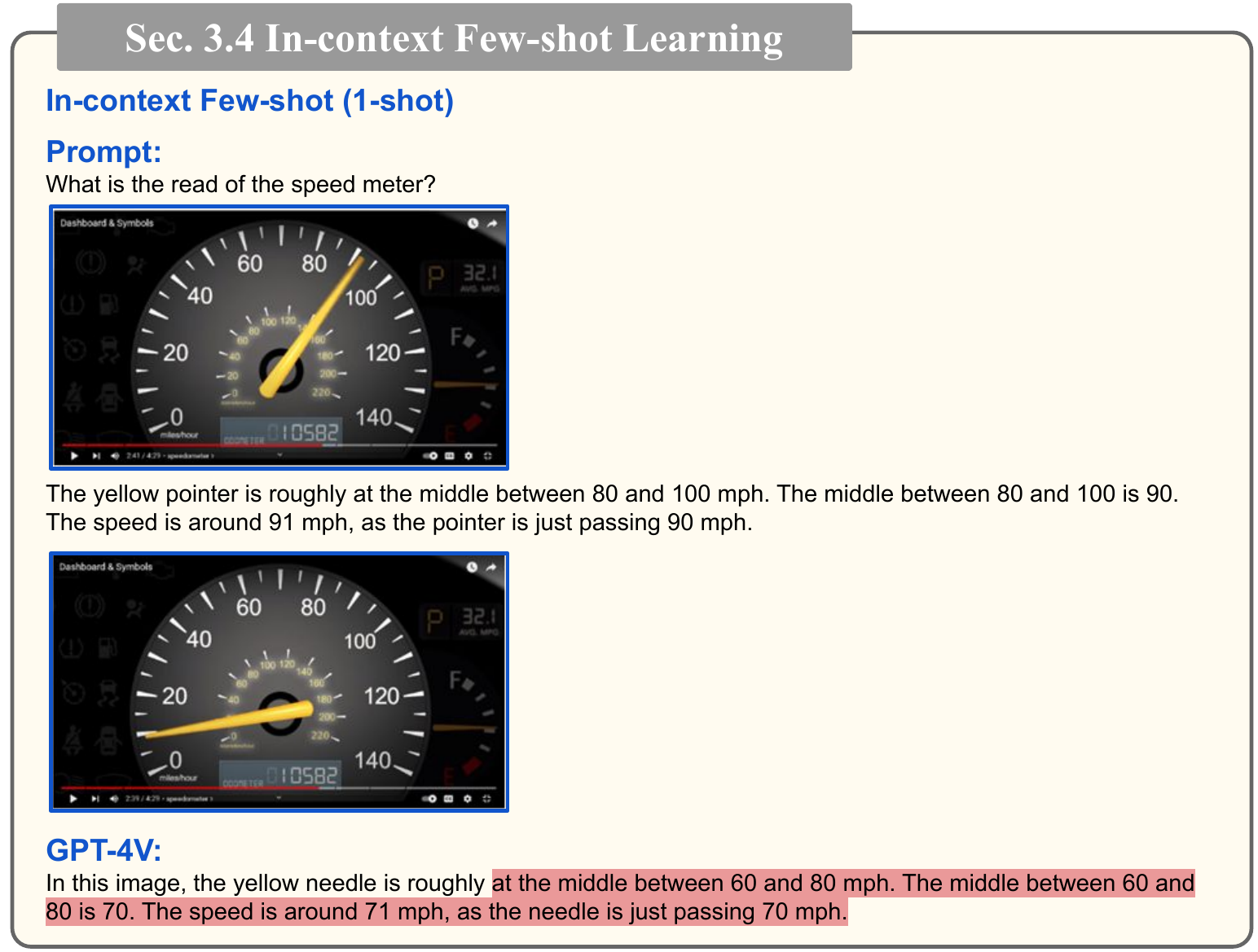}\label{fig:1shot_9_diff}
   }

   \subfloat[ 
   ]{
      \includegraphics[width=\textwidth]{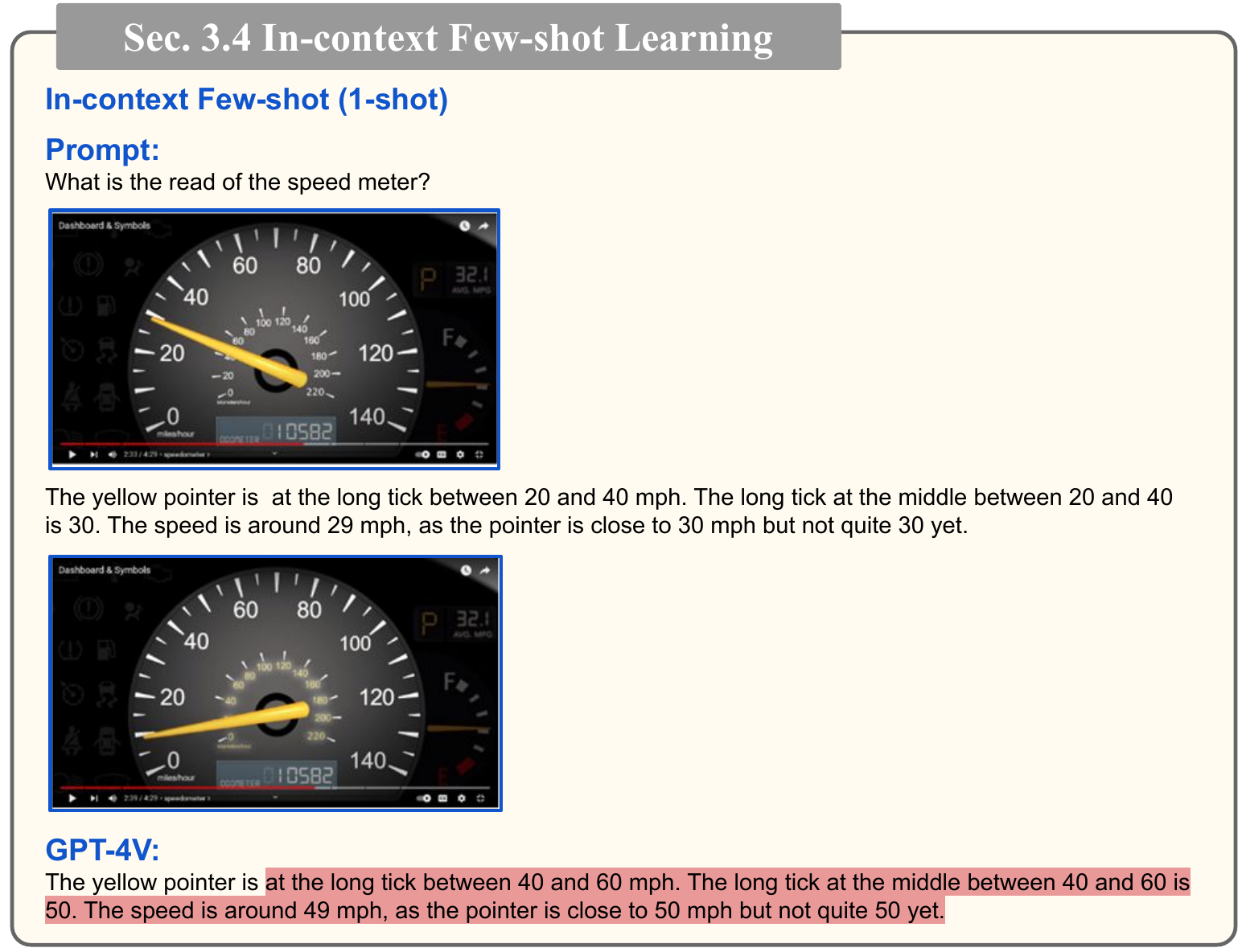}\label{fig:1shot_9_same}
   }
   \caption[Section~\ref{sec:02fewshot}: few-shot comparison example 1, Part II: one-shot.]{
     One-shot (or prompting with multimodal example instruction) performance under the challenging scenario of reading a speed meter. \modelname~still fails with (a) dissimilar or (b) similar 1-shot in-context example. 
      \colorbox{redhl}{Red} highlights the wrong answer. Check Section~\ref{sec:02fewshot} for detailed discussions.}
\label{fig:speed_meter_1shot_9}
\end{figure}

\begin{figure*}[h!]
\centering
\includegraphics[width=\textwidth]{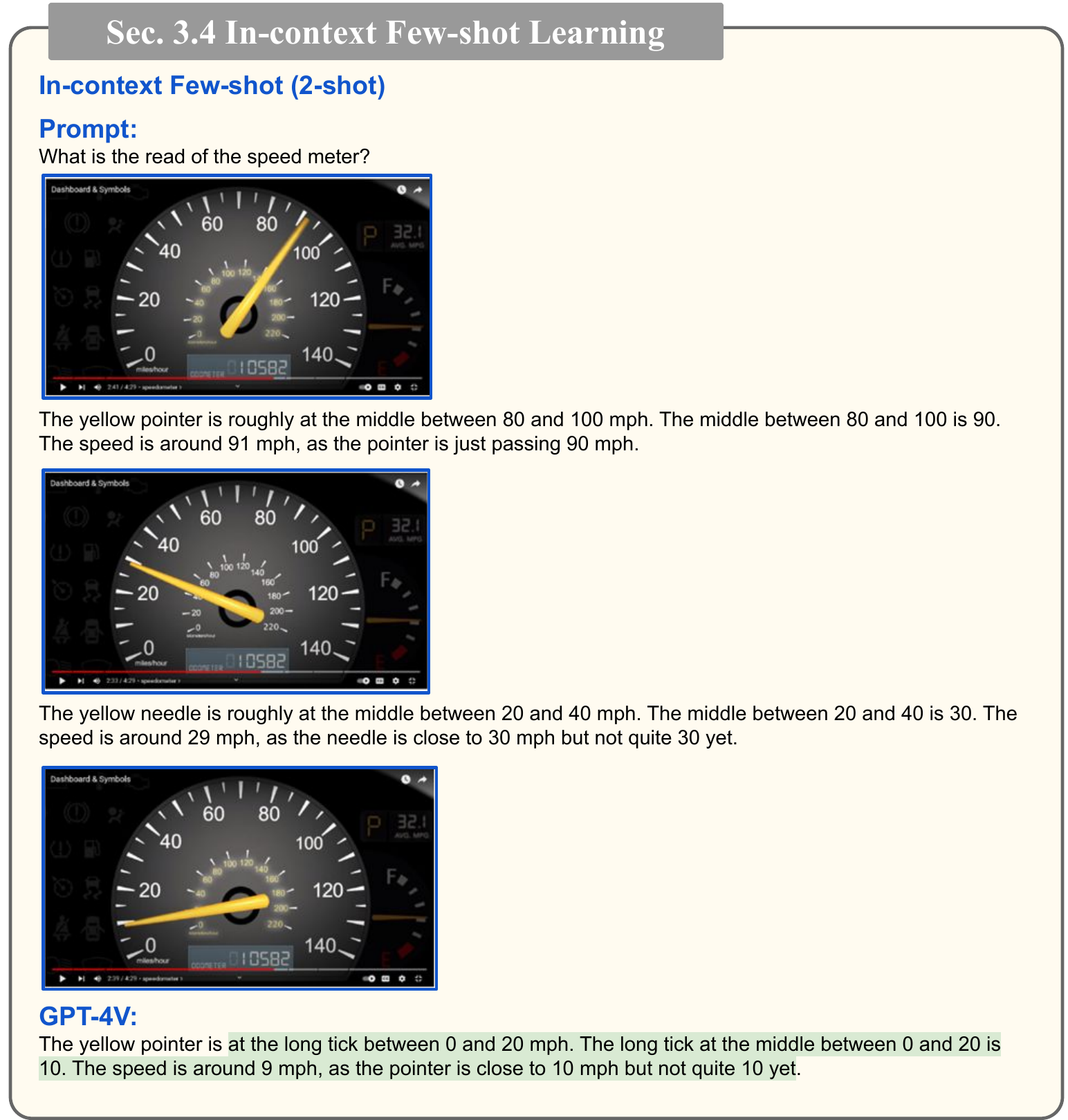}
\caption[Section~\ref{sec:02fewshot}: few-shot comparison example 1, Part III: two-shot.]{Two-shot performance under the challenging scenario of reading a speed meter. \modelname~now can read the speed accurately.  \colorbox{greenhl}{Green} highlights the correct answer. Check Section~\ref{sec:02fewshot} for detailed discussions.
}
\label{fig:speed_meter_2shot_9}
\end{figure*}

In-context few-shot learning is another intriguing emergent ability observed in LLMs~\cite{brown2020language,dong2022survey,wei2022emergent,dai2022can}. That is, LLMs can generate desired outputs without parameter updates by prepending a few in-context examples at inference time. The examples share the same format as the input query, and serve as demonstrations to illustrate the desired outputs. Similar abilities were recently observed in multimodal models~\cite{tsimpoukelli2021multimodal,alayrac2022flamingo,huang2023language,driess2023palme,zhang2023multimodal}, where query inputs are formatted image-text pairs.
Complementary to instruction tuning, in-context learning ``teaches'' model to perform new tasks by providing in-context examples with the same format during test time. 
We demonstrate the in-context few-shot learning capacity of \modelname through a few compelling examples. We emphasize that in certain scenarios, in-context few-shot learning with a sufficient number of examples becomes essential, particularly when zero-shot or one-shot instruction approaches fall short. %
Figures~\ref{fig:speed_meter_zs_9}-\ref{fig:speed_meter_2shot_9} explore a challenging scenario involving the reading of a speed meter. In Figure~\ref{fig:speed_meter_zs_9}, the zero-shot performance of \modelname on a screenshot of a speed meter image from a video is depicted. Despite numerous attempts to prompt \modelname in a zero-shot manner, it struggles to accurately read the current speed displayed in the image. The predictions it generates (22/30/40 mph) deviate significantly from the actual human reading of ``approximately 9 mph.'' Even when employing a 1-shot in-context example, as shown in Figure~\ref{fig:speed_meter_1shot_9}, using either a dissimilar example (Figure~\ref{fig:1shot_9_diff}) or a similar example (Figure~\ref{fig:1shot_9_same}), \modelname still fails to accurately locate the two numbers on the left and right sides of the yellow pointer. In contrast, Figure~\ref{fig:speed_meter_2shot_9} demonstrates that when provided with 2 in-context examples, one similar to the query image and the other dissimilar, \modelname successfully predicts the speed reading as ``around 9 mph'' by recognizing that the pointer is close to 10 mph but not quite there yet.

\begin{figure*}[h!]
\centering
\vspace{-20mm}
\centerline{\includegraphics[width=1.15\textwidth]{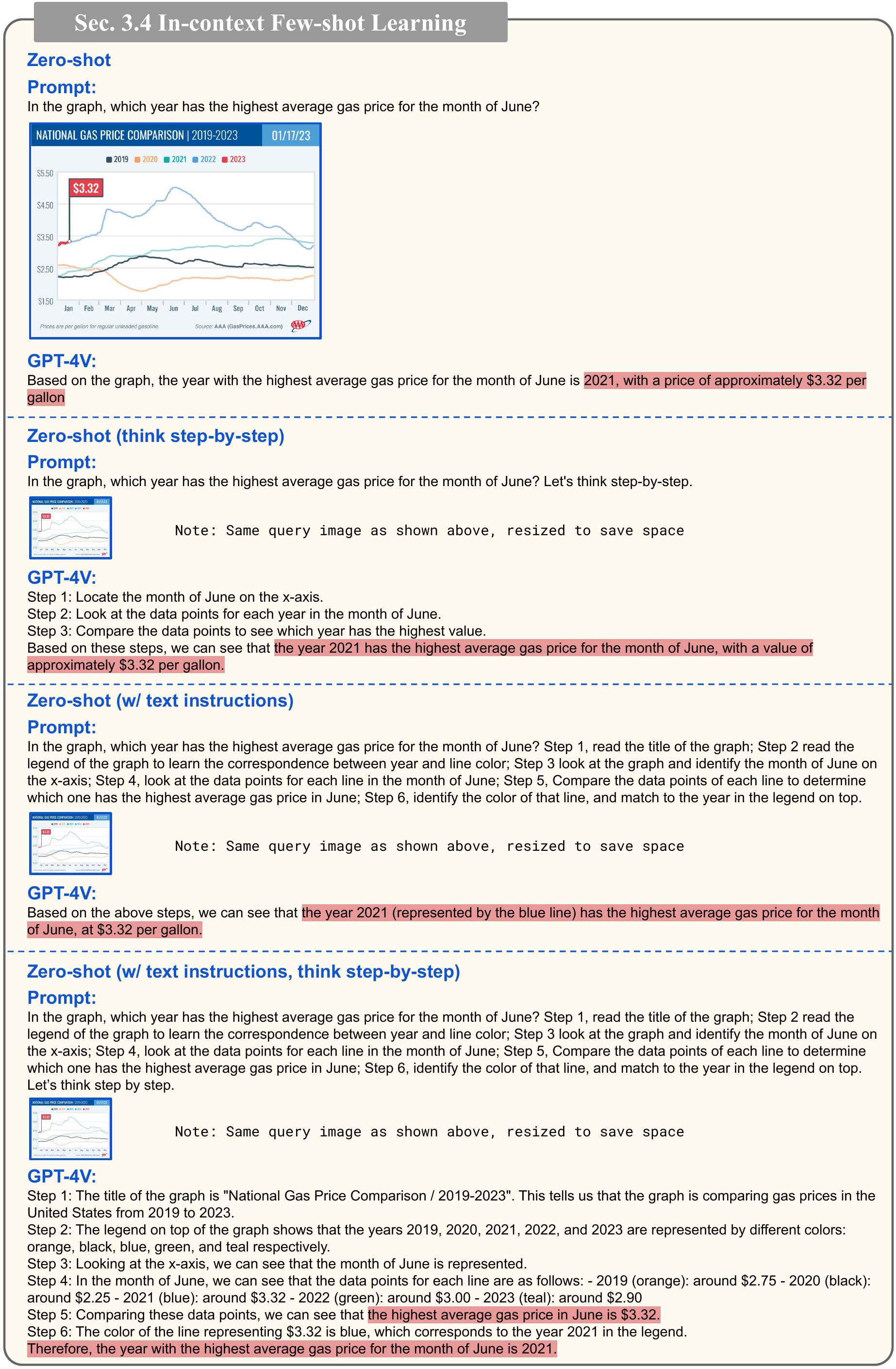}}
\vspace{-6pt}
\caption[Section~\ref{sec:02fewshot}: few-shot comparison example 2, Part I: zero-shot.]{Zero-shot performance under the challenging scenario of reading a line plot. \modelname~fails to answer the question even with different ways of ZS prompting. \colorbox{redhl}{Red} highlights the wrong answer. Check Section~\ref{sec:02fewshot} for detailed discussions. 
}
\label{fig:gas_chart_zs}
\end{figure*}

\begin{figure*}[h!]
\centering
\includegraphics[width=\textwidth]{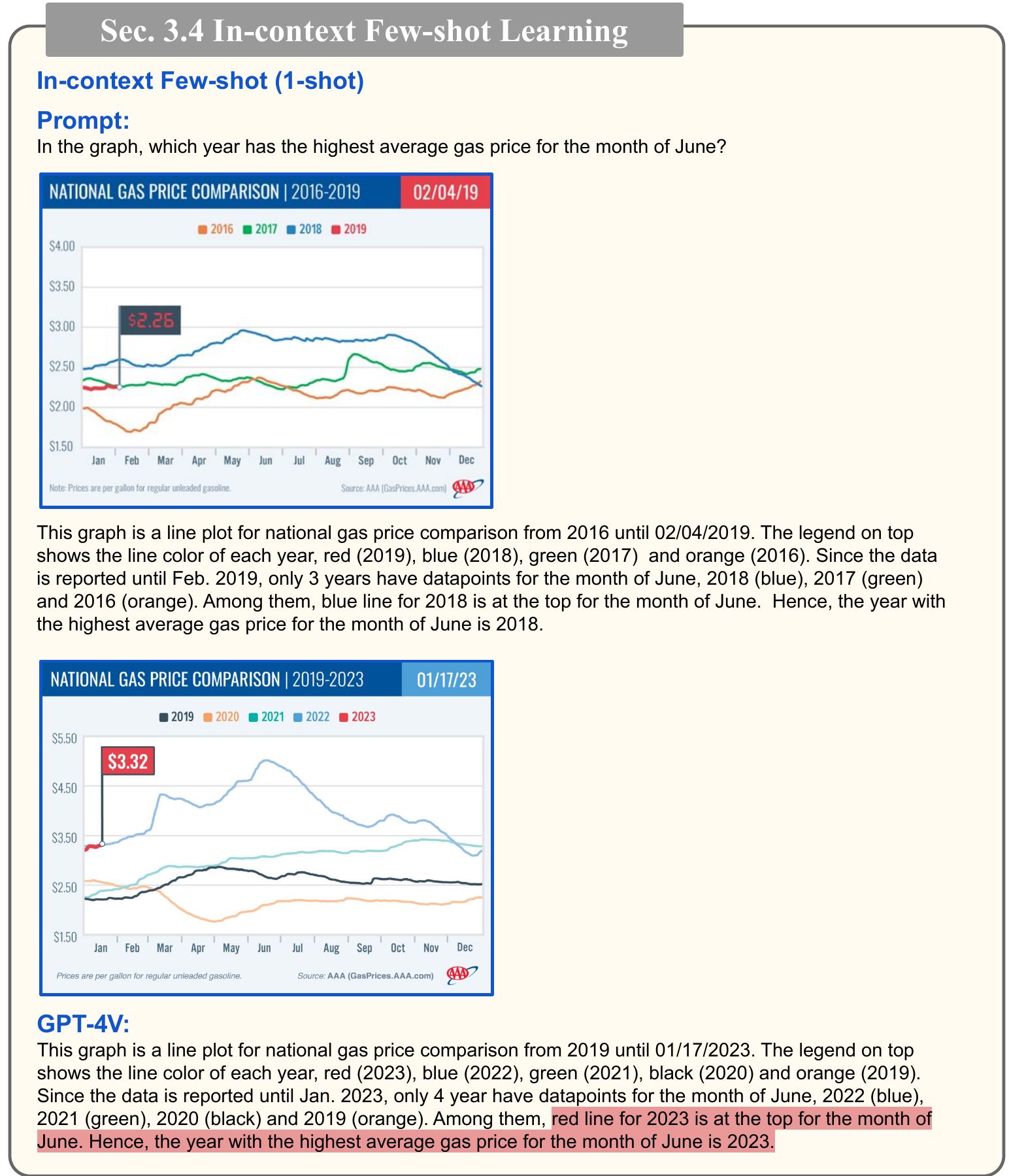}
\caption[Section~\ref{sec:02fewshot}: few-shot comparison example 2, Part II: one-shot.]{One-shot (or prompting with multimodal example instruction) performance under the challenging scenario of reading a line plot. \modelname~still fails with 1-shot in-context example. \colorbox{redhl}{Red} highlights the wrong answer.  Check Section~\ref{sec:02fewshot} for detailed discussions. 
}
\label{fig:gas_chart_1shot}
\end{figure*}

\begin{figure*}[h!]
\centering
\vspace{-15mm}
\includegraphics[width=\textwidth]{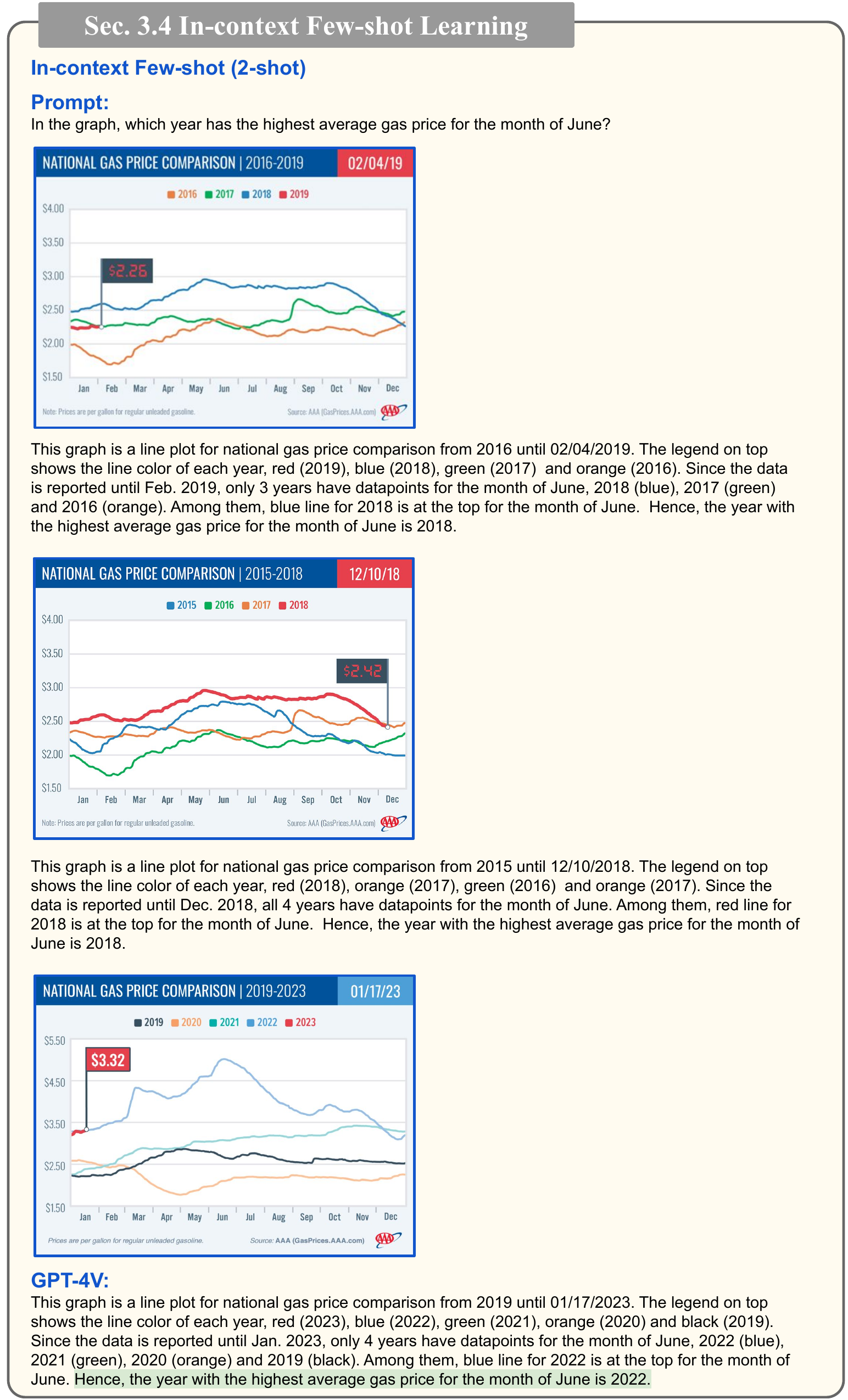}
\caption[Section~\ref{sec:02fewshot}: few-shot comparison example 2, Part III: two-shot.]{Two-shot performance under the challenging scenario of reading a line plot. \modelname~now can answer the question of ``which year has the highest average gas price for the month of June?'' correctly.  Check Section~\ref{sec:02fewshot} for detailed discussions. \colorbox{greenhl}{Green} highlights the correct answer.
}
\label{fig:gas_chart_2shot}
\end{figure*}

The comparison between zero-shot, 1-shot, and 2-shot performance for reasoning over a complex line plot is illustrated in Figures~\ref{fig:gas_chart_zs}-\ref{fig:gas_chart_2shot}. 
The example we explore here presents a great difficulty level as it involves multi-hop reasoning. To answer the question ``In the graph, which year has the highest average gas price for the month of June,'' one needs to go through at least four steps: ($i$) locating the month of June on the x-axis, ($ii$) comparing data points for each line in June, ($iii$) identifying the color of the line with the highest value, and ($iv$) matching the color to the corresponding year in the legend at the top. Failure in any of these steps would lead to an incorrect prediction. As depicted in Figure~\ref{fig:gas_chart_zs}, even when prompted with ``text instruction, think step-by-step'' in a zero-shot manner, \modelname fails to correctly associate the colors with the years from the legend. Furthermore, it gets distracted by the highlighted gas price of $\$3.32$ in the graph. Similarly, in Figure~\ref{fig:gas_chart_1shot}, although \modelname shows some improvement in reading the legend (correcting the corresponding colors for 2021 and 2022 compared to zero-shot), it still insists on answering with 2023 as the year with the highest average gas price for the month of June, despite the fact that the chart only includes data points until 01/17/2023. However, as we introduce another in-context example in Figure~\ref{fig:gas_chart_2shot}, \modelname finally arrives at the correct answer (2022) and provides intermediate steps that explain its reasoning process, similar to the demonstration shown in the in-context examples.

These proof-of-concept examples vividly demonstrate the rising significance of in-context few-shot learning for achieving improved performance with LMMs. This approach serves as a viable alternative to finetuning, analogous to the observations made in the context of LLMs~\cite{brown2020language,dong2022survey,wei2022emergent,dai2022can}.
Despite the great importance of in-context few-shot learning in achieving better performance with LMMs, we limit its use in this report to prevent the potential information leakage or undesired hints from in-context examples. We also leave the quantitative evaluation of few-shot learning's gain to future studies.

\clearpage
\section{Vision-Language Capability}
\label{sec:03vl}

Understanding and describing visual information plays a crucial role in human cognition. In this section, we will investigate how \modelname can be utilized to comprehend and interpret the visual world. We will start by examining the model's ability to generate open-ended descriptions for generic visual captioning. 

Moving forward, in Section~\ref{sec:od}, we will explore the application of \modelname in more advanced tasks, such as spatial relationship analysis, object localization, object counting, and dense captioning. In Section~\ref{sec:knowledge}, we will delve into the model's capacity for multimodal knowledge and commonsense reasoning, and study whether the model can understand the context and relationships between different types of information.

Additionally, in Section~\ref{sec:document}, we will assess the model's capability to extract and analyze information from various sources, including scene text, tables, charts, and documents. In Section~\ref{sec:multilingual}, we will explore \modelname's ability in comprehending and generating descriptions in multilingual scenarios. Lastly, in Section~\ref{sec:04language}, we will investigate the model's coding proficiency with visual information, exploring its ability to perform tasks with selected examples.

\subsection{Image Description on Diverse Domains}\label{sec:open-world}

We access the model's capability and generalizability by providing a \textit{single image-text pair} as input. We prompt \modelname to generate natural language descriptions covering a variety of topics listed below. 

\begin{figure*}[h!]
\centering
\includegraphics[width=\textwidth]{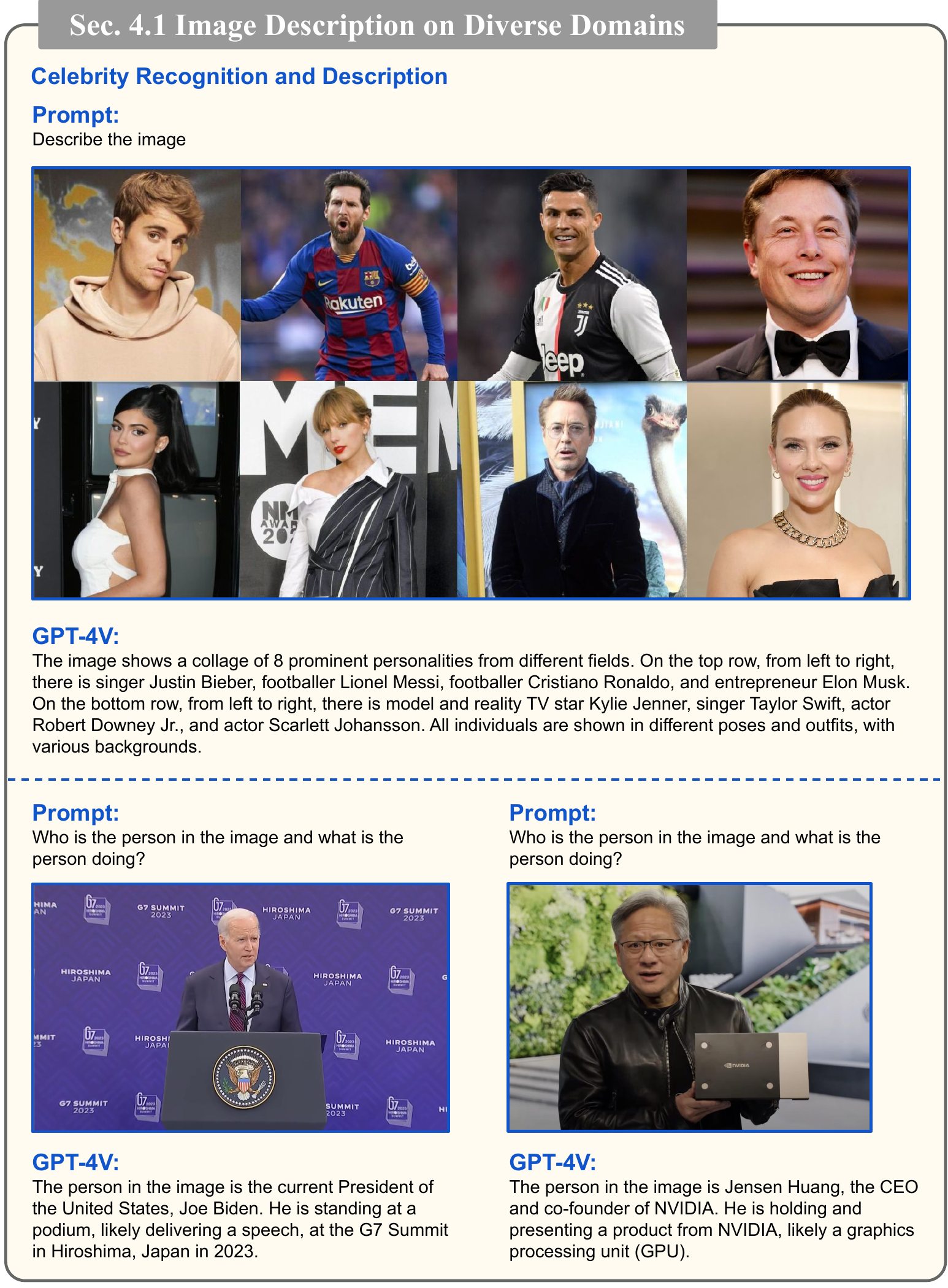}
\caption[Section~\ref{sec:open-world}: celebrity recognition and description.]{Results on celebrity recognition and description. \modelname~can recognize a variety of celebrities and describe the visual information (including their profession, action, background, and the event) in details. Check Section~\ref{sec:open-world} for detailed discussions. 
} 
\label{fig:sec3-vl-celebrity}
\end{figure*}

\noindent\textbf{Celebrity recognition.} Recognizing human appearance~\cite{guo2016ms,liu2015faceattributes} presents a significant challenge due to its inherent variability. To assess \modelname's capabilities to recognize and describe the celebrities, we conduct an experiment by providing a text prompt, ``Describe the image,'' along with an input celebrity image.
In the top row of Figure~\ref{fig:sec3-vl-celebrity}, we observe that \modelname accurately identifies the eight celebrities, despite their diverse backgrounds and fields. Furthermore, when we present a more specific query, ``Who is the person in the image and what is the person doing?,'' as shown in the bottom row of Figure~\ref{fig:sec3-vl-celebrity}, \modelname comprehends that the current President of the United States is delivering a speech at the 2023 G7 Summit. This illustrates the model's ability to generalize and handle novel scenarios, such as the 2023 G7 Summit, which was not part of its training data.

\begin{figure*}[h!]
\centering
\includegraphics[width=\textwidth]{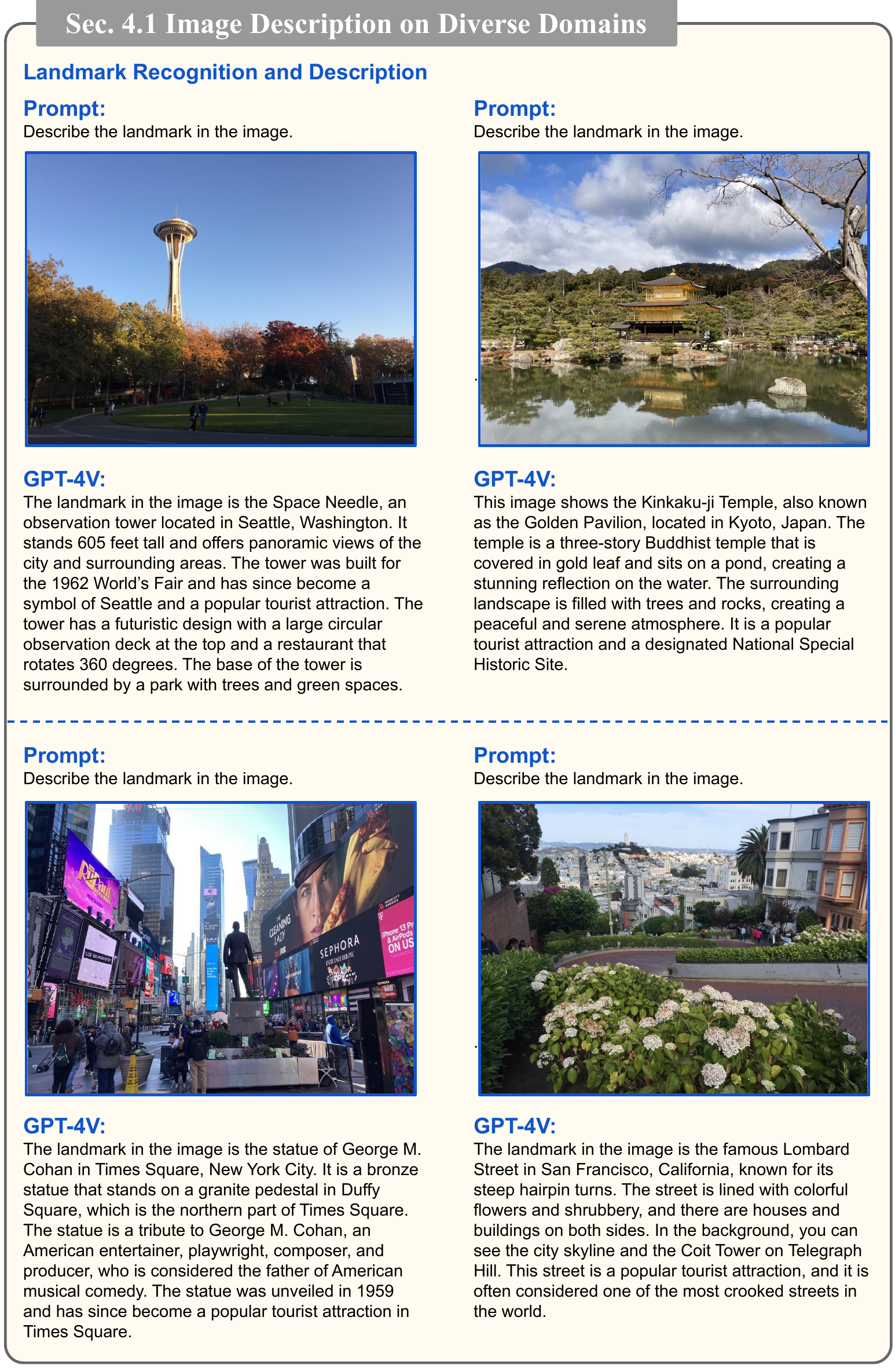}
\caption[Section~\ref{sec:open-world}: landmark recognition and description.]{Results on landmark recognition and description. \modelname~accurately recognizes the landmarks in the test images. It also generates vivid and detailed narratives that capture the essence of the landmarks. Check Section~\ref{sec:open-world} for detailed discussions. 
}
\label{fig:sec3-vl-landmark-2}
\end{figure*}

\begin{figure*}[h!]
\centering
\vspace{-18mm}
\includegraphics[width=\textwidth]{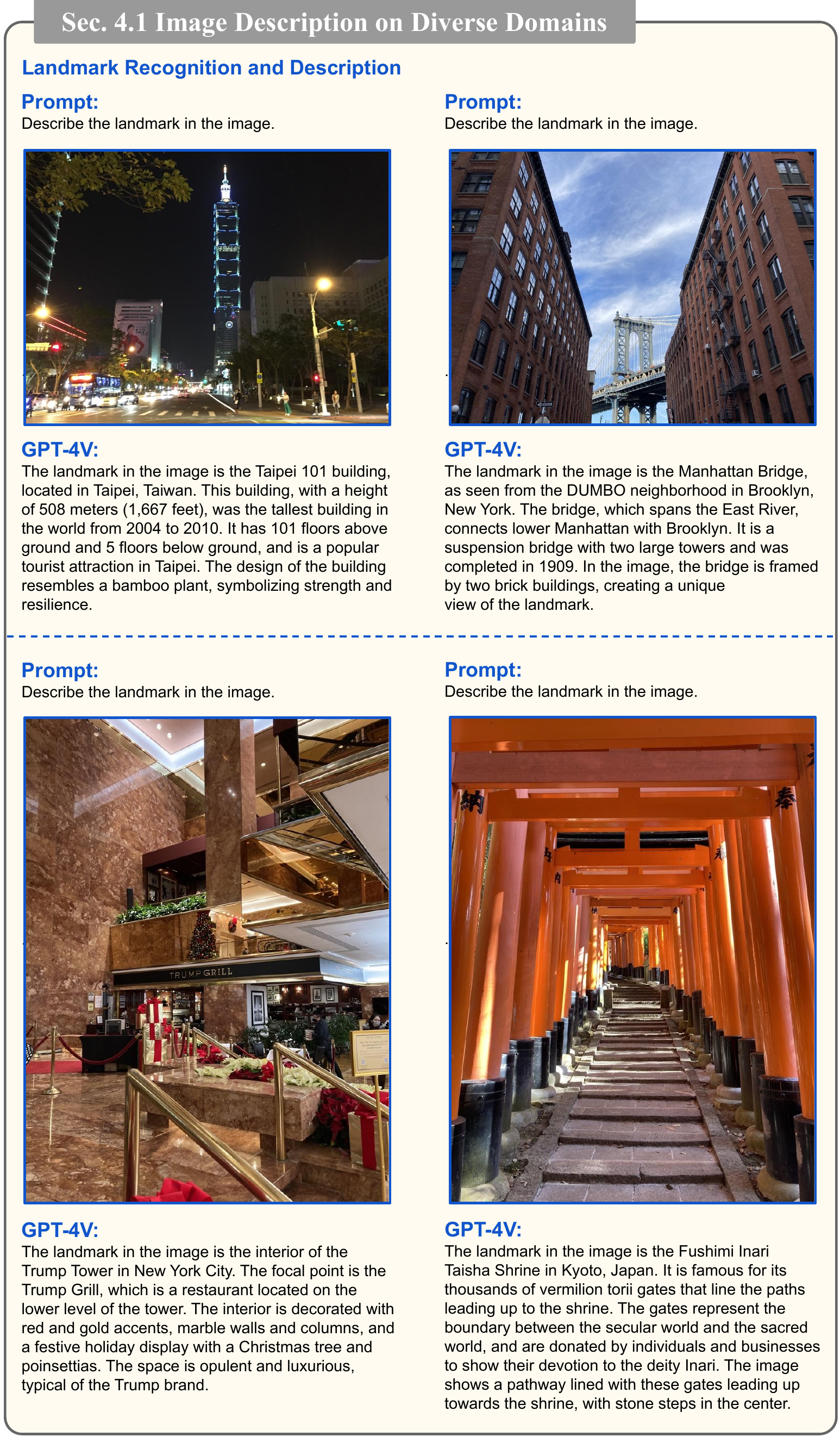}
\caption[Section~\ref{sec:open-world}: landmark recognition and description.]{Results on landmark recognition and description. \modelname~accurately recognizes the landmarks in the test images. It also generates vivid and detailed narratives that capture the essence of the landmarks. Check Section~\ref{sec:open-world} for detailed discussions. 
}
\label{fig:sec3-vl-landmark}
\end{figure*}

\noindent\textbf{Landmark recognition.} Landmarks exhibit considerable variations in appearance due to  factors such as viewpoint changes, lighting conditions, occlusions, and seasonal changes. Recognizing landmarks under these variations requires models to generalize well and handle the vast range of visual appearances~\cite{zheng2009tour,agarwal2011building}. In the experiments, we employ a straightforward text prompt, ``Describe the landmark in the image,'' to test the model's capability. As shown in Figures~\ref{fig:sec3-vl-landmark-2}-\ref{fig:sec3-vl-landmark}, \modelname generates accurate and open-ended descriptions for each test image. For example, it accurately recognizes Space Needle located in Seattle, Washington, understanding that the tower was built for the 1962 World’s Fair and has since become a symbol of the city. We have similar observations for other tested photos as well. The generated descriptions go beyond simple labels or generic phrases, providing vivid and detailed narratives that capture the essence of the landmark.

\begin{figure*}[h!]
\centering
\includegraphics[width=\textwidth]{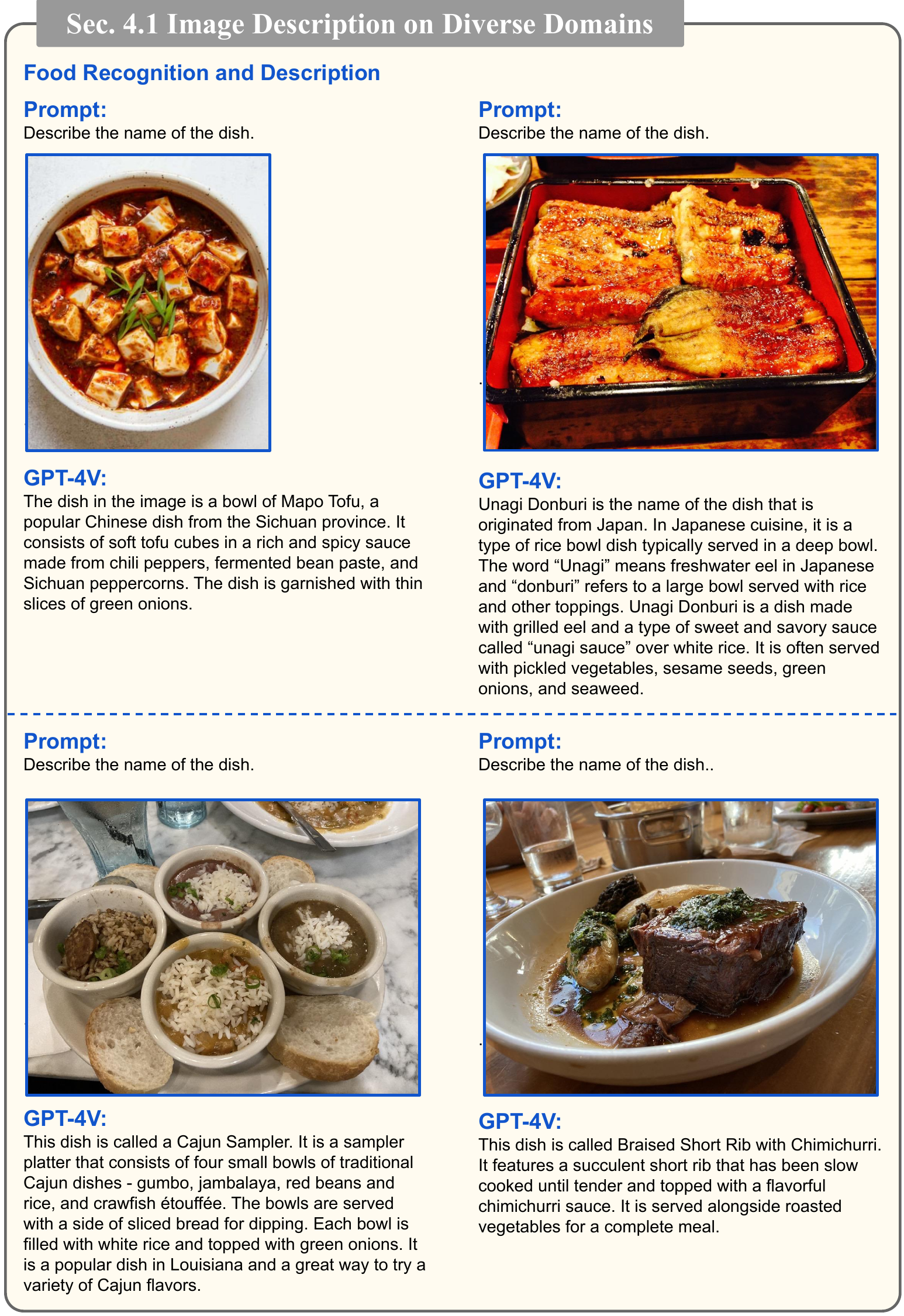}
\caption[Section~\ref{sec:open-world}: food recognition and description.]{Results on food recognition and description. \modelname~ recognizes various dishes. It also identifies specific ingredients, garnishes, or cooking techniques present in a dish image. Check Section~\ref{sec:open-world} for detailed discussions. 
}
\label{fig:sec3-vl-food}
\end{figure*}

\noindent\textbf{Food recognition.} Recognizing food or dishes is a fascinating task~\cite{bossard2014food,min2023large}, but it can be challenging to tackle due to the wide range of appearances and potential occlusions caused by other objects or overlapping ingredients. In our experiments, we employ a straightforward text prompt, asking the system to ``Describe the name of the dish,'' for testing purpose. Figure~\ref{fig:sec3-vl-food} demonstrates the accurate recognition of various dishes by \modelname. Additionally, \modelname effectively captures intricate details within the images, enabling it to identify specific ingredients, garnishes, or cooking techniques present in a dish. 

\begin{figure*}[h!]
\centering
\vspace{-10mm}
\includegraphics[width=\textwidth]{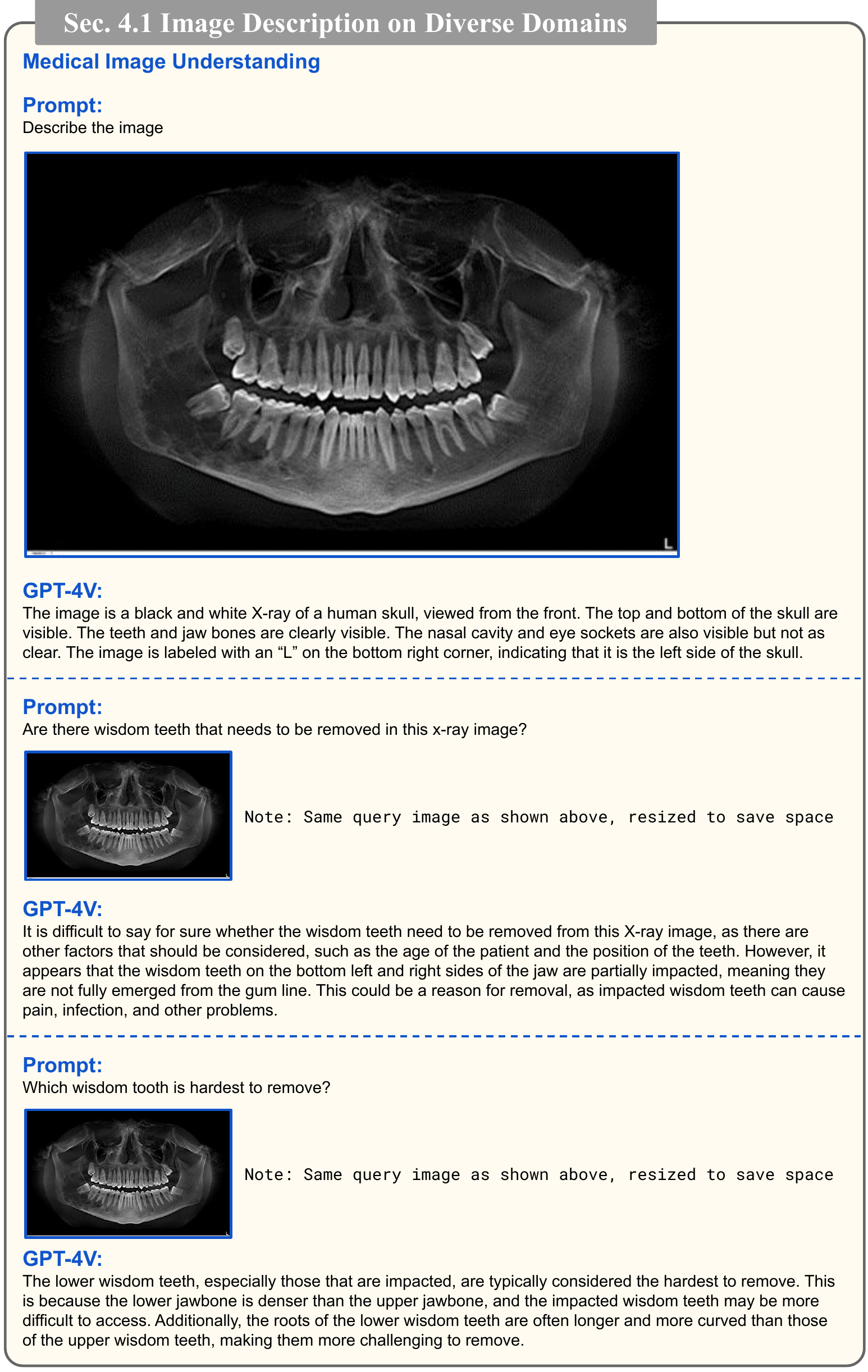}
\caption[Section~\ref{sec:open-world}: medical image understanding.]{Results on medical image understanding. \modelname recognizes both the teeth and jaw bones in the given X-ray, and explains that the partially emerged wisdom teeth on the bottom left and right sides of the jaw may necessitate removal. Check Section~\ref{sec:open-world} for detailed discussions. The medical images are collected from the Internet.
}
\label{fig:sec3-vl-medical}
\end{figure*}

\begin{figure*}[h!]
\centering
\includegraphics[width=\textwidth]{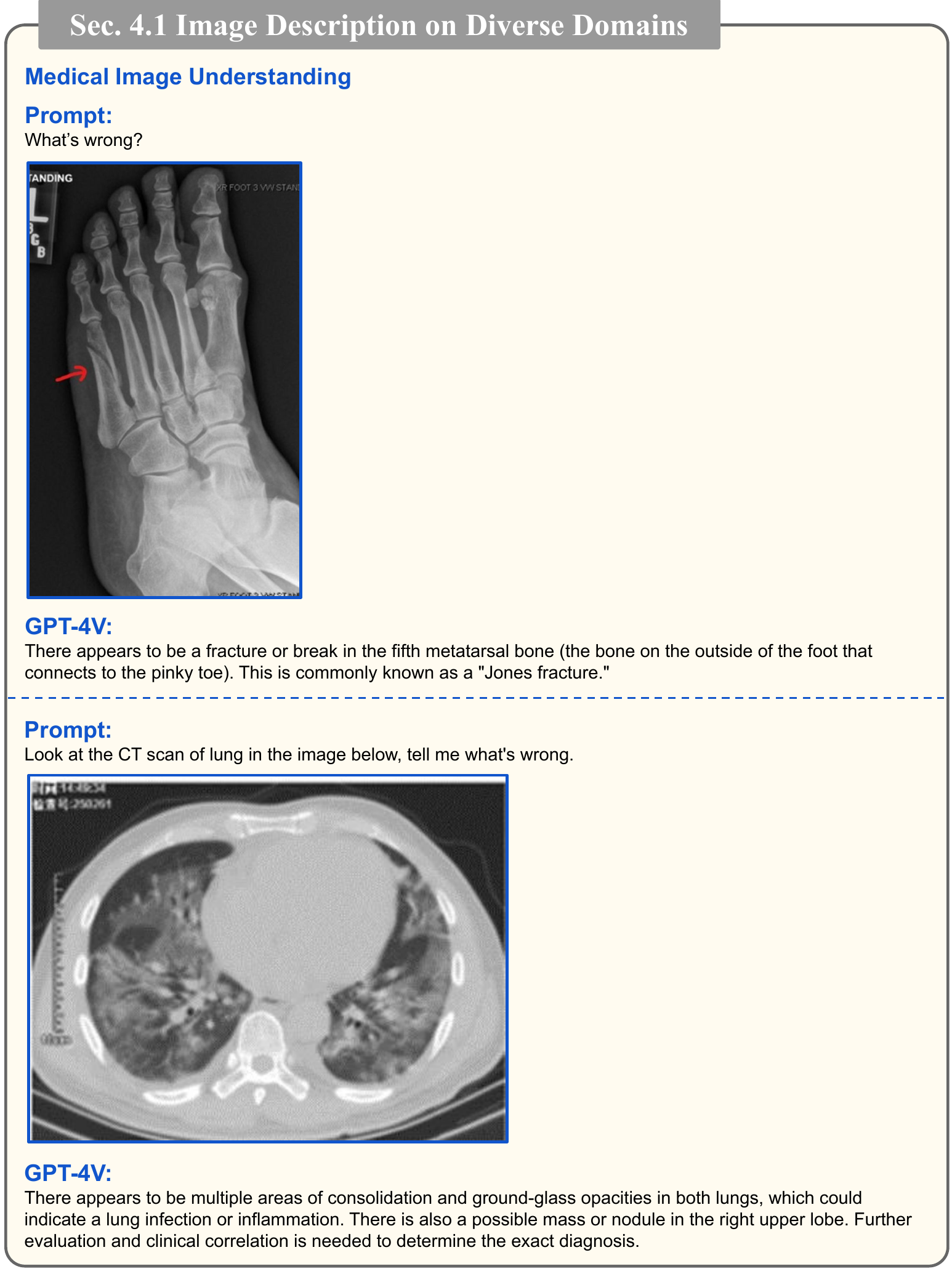}
\caption[Section~\ref{sec:open-world}: medical image understanding.]{Results on medical image understanding. \modelname can identify common conditions like a Jones fracture. It could also point out potential concerns based on the CT scan of the lung. Check Section~\ref{sec:open-world} for detailed discussions. The medical images are from the internet.
}
\label{fig:sec3-vl-medical2}
\end{figure*}

\noindent\textbf{Medical image understanding.} Medical images, such as X-rays and CT scans, can have large variability due to patient populations and imaging equipment. Additionally, interpreting the visual content of these images requires expert knowledge. In Figure~\ref{fig:sec3-vl-medical}, we access \modelname's performance by providing the prompt, ``Describe the image.'' The results show that \modelname recognizes both the teeth and jaw bones in the given X-ray. Furthermore, when we prompt with ``Are there wisdom teeth that needs to be removed in this x-ray image?'' \modelname performs reasoning with the visual context, and explains that the wisdom teeth on the bottom left and right sides of the jaw are not fully emerged from the gum line, and this could be a reason for removal. We also conduct testing with other medical images, as shown in Figure~\ref{fig:sec3-vl-medical2}. For these experiments, we use prompts such as ``What's wrong?'' or ``Look at the CT scan, tell me what's wrong.'' The observations reveal that \modelname can identify common conditions such as a Jones fracture. It could also point out potential concerns based on the CT scan of the lung. The experiments demonstrate \modelname's basic understanding of medical images. We discuss the application of \modelname to the medical domain in Section~\ref{sec:app_medical}.

\begin{figure*}[h!]
\centering
\includegraphics[width=\textwidth]{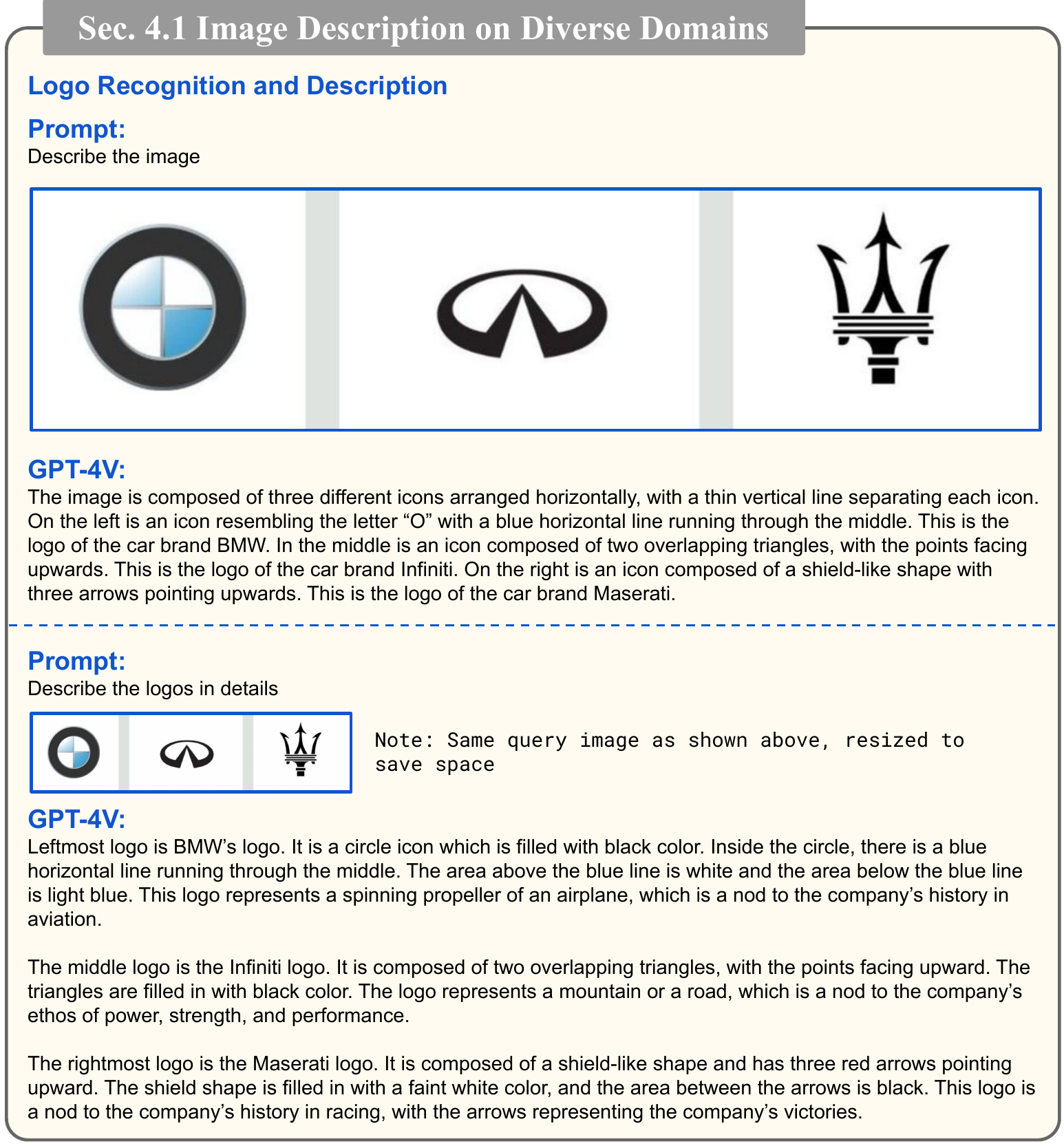}
\caption[Section~\ref{sec:open-world}: logo recognition and description.]{Results on logo recognition. \modelname correctly recognizes the logos and provides detailed descriptions, including its design, color, shape, and symbol. Check Section~\ref{sec:open-world} for detailed discussions. 
}
\label{fig:sec3-vl-logo}
\end{figure*}

\begin{figure*}[h!]
\centering
\vspace{-20mm}
\centerline{\includegraphics[width=1.1\textwidth]{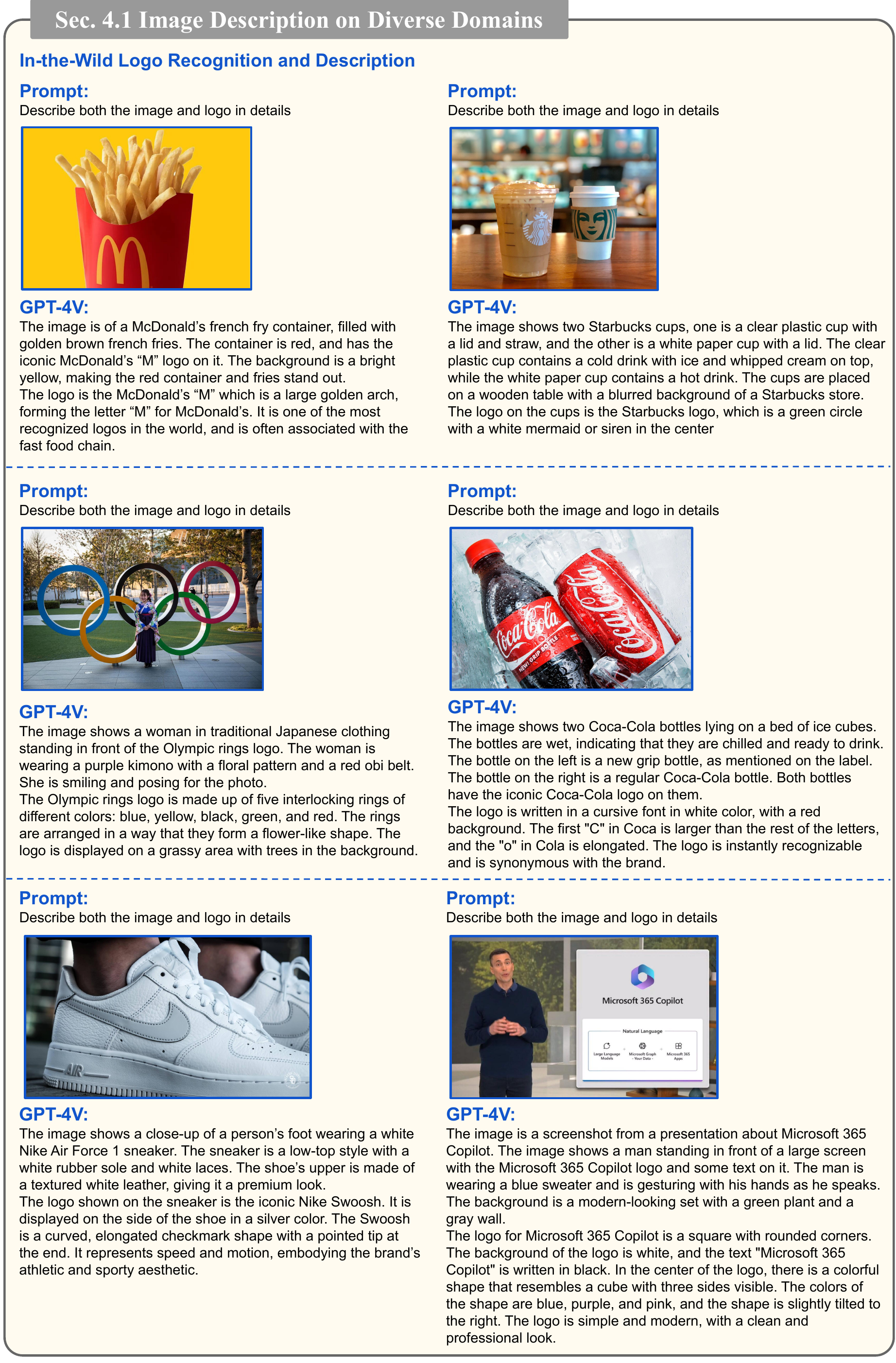}}
\caption[Section~\ref{sec:open-world}: in-the-wild logo recognition and description.]{Results on in-the-wild logo recognition and description. \modelname demonstrates strong capability in understanding logos in many scenarios, including occlusions, lighting conditions, and orientations. \modelname can also describe novel icons, such as the recently released Microsoft 365 Copilot. Check Section~\ref{sec:open-world} for detailed discussions.
}
\label{fig:sec3-vl-logo-wild}
\end{figure*}

\noindent\textbf{Logo recognition.} We examine \modelname's ability in logo recognition. In Figure~\ref{fig:sec3-vl-logo}, we initiate the experiments by providing the text prompt, ``Describe the image.'' \modelname accurately identifies the three logos depicted in the image. We then proceed to ask a more specific question, ``Describe the logos in details,'' \modelname provides elaborate descriptions, including the design, style, and representation for each logo, respectively. Expanding the evaluation to a more challenging \textit{in-the-wild} scenario, as shown in Figure~\ref{fig:sec3-vl-logo-wild}, we experiment with logos that may be partially occluded, distorted, or situated in cluttered backgrounds. We employ the text prompt ``Describe both the image and logo in details'' for the \textit{in-the-wild} experiment. As shown in Figure~\ref{fig:sec3-vl-logo-wild}, \modelname demonstrates strong capability in understanding logos in difficult scenarios. Notably, \modelname can also provide descriptions for novel or emerging logos and icons, such as the recently released Microsoft 365 Copilot.

\begin{figure*}[h!]
\centering
\includegraphics[width=\textwidth]{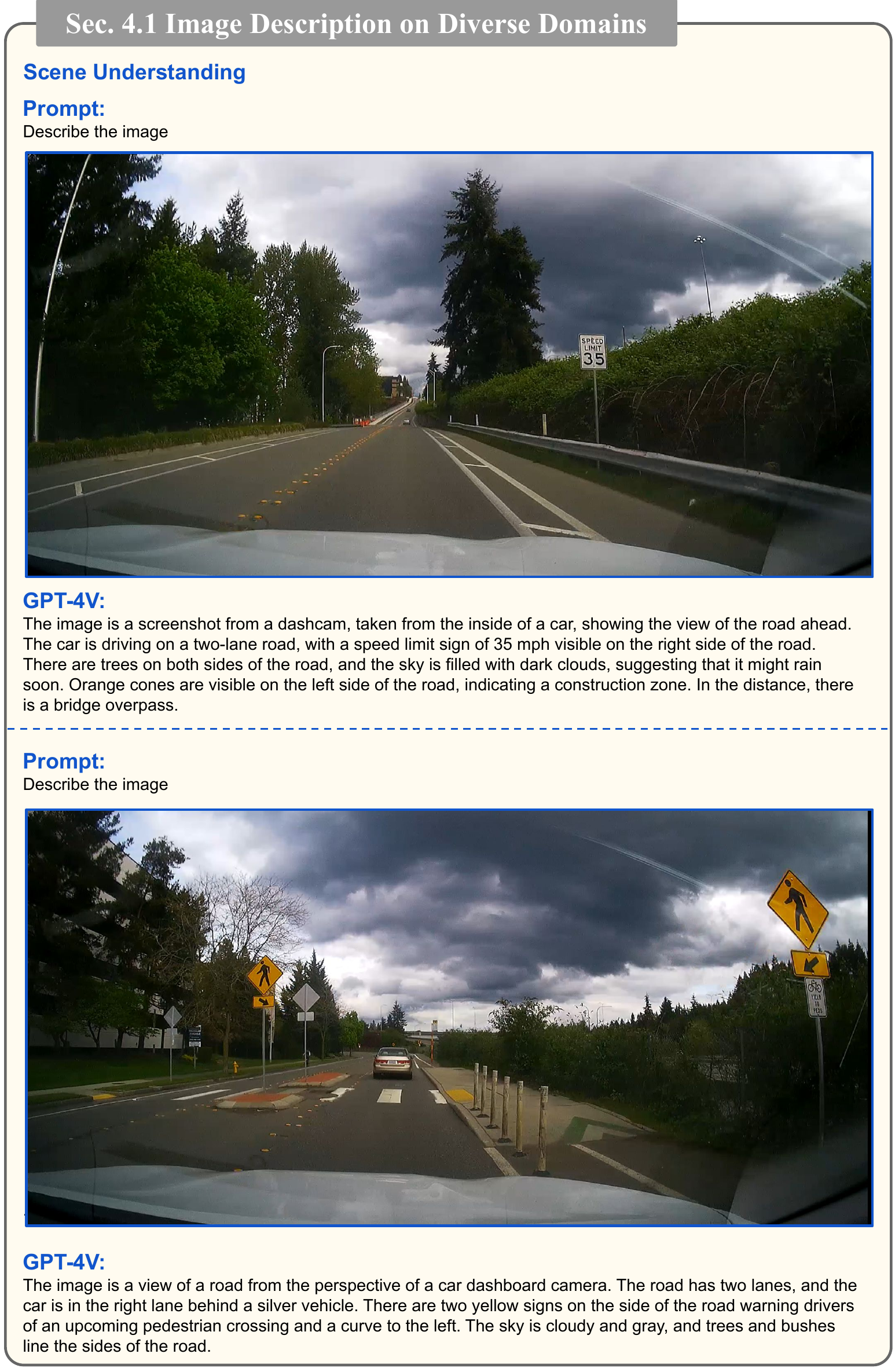}
\caption[Section~\ref{sec:open-world}: scene understanding.]{Results on scene understanding. \modelname is able to provide a detailed description regarding the scenes and objects. Check Section~\ref{sec:open-world} for detailed discussions. 
}
\label{fig:sec3-vl-embodied}
\end{figure*}

\noindent\textbf{Scene understanding.} Scene understanding~\cite{lin2014microsoft,cordts2016cityscapes,zhou2016places} is an important task in computer vision. We examine the model's capability by providing a simple query ``Describe the image.'' In Figure~\ref{fig:sec3-vl-embodied}, \modelname is able to describe the road and the location and color of the vehicles. It can also read the sign and notice the speed limit for this road. 

\begin{figure*}[h!]
\centering
\includegraphics[width=\textwidth]{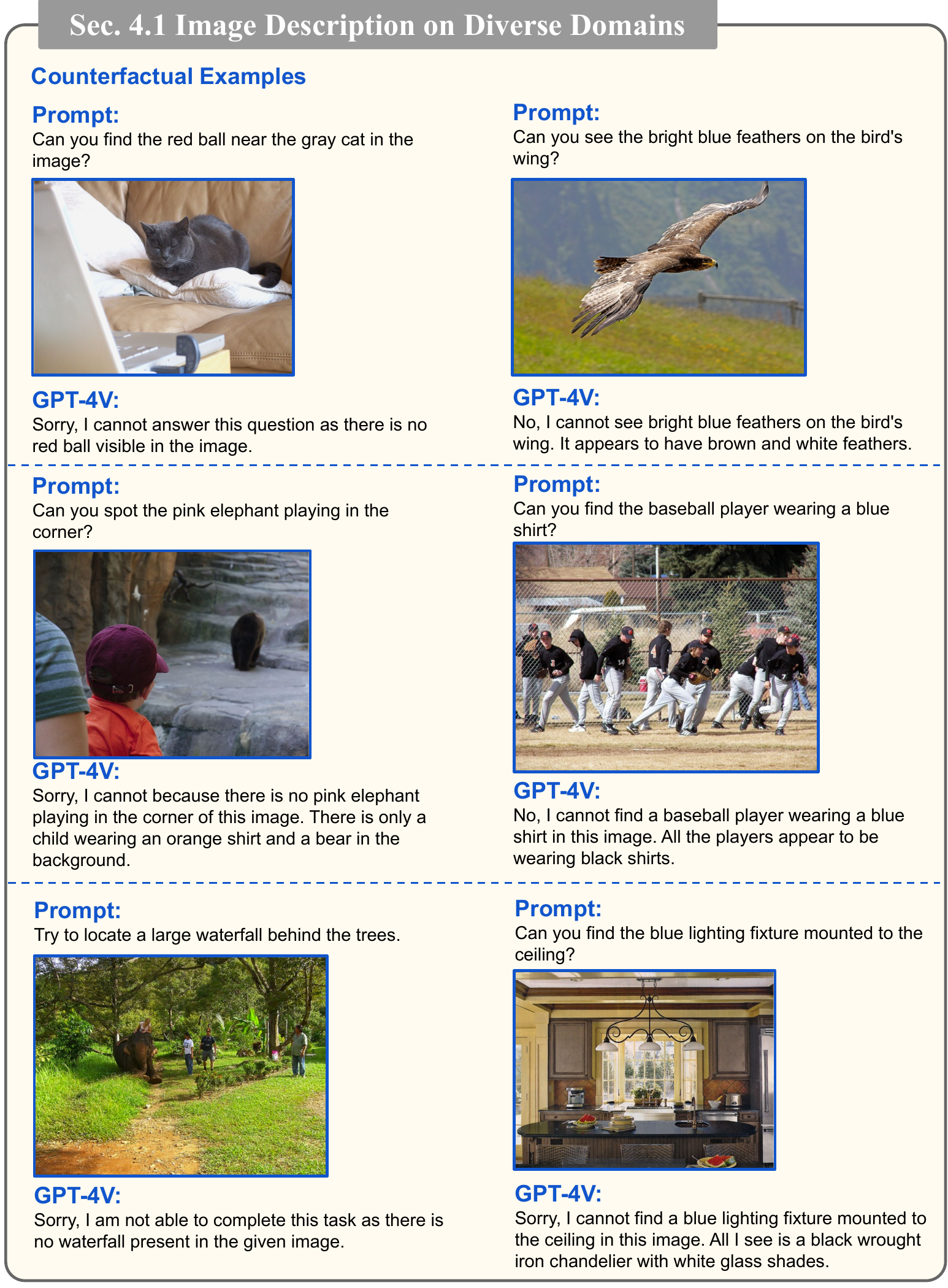}
\caption[Section~\ref{sec:open-world}: counterfactual examples.]{Results on counterfactual examples. \modelname is able to provide factual descriptions regarding the scenes and objects in the images. Example images are from~\cite{liu2023aligning}. Check Section~\ref{sec:open-world} for detailed discussions.
}
\label{fig:sec3-vl-counterfact}
\end{figure*}

\noindent\textbf{Counterfactual examples.} We conduct experiments by randomly selecting counterfactual examples from~\cite{liu2023aligning}. In Figure~\ref{fig:sec3-vl-counterfact}, we observe that \modelname correctly describes the image contents when faced with misleading questions or instructions.

\clearpage
\subsection{Object Localization, Counting, and Dense Captioning}\label{sec:od}

\begin{figure*}[h!]
\centering
\includegraphics[width=\textwidth]{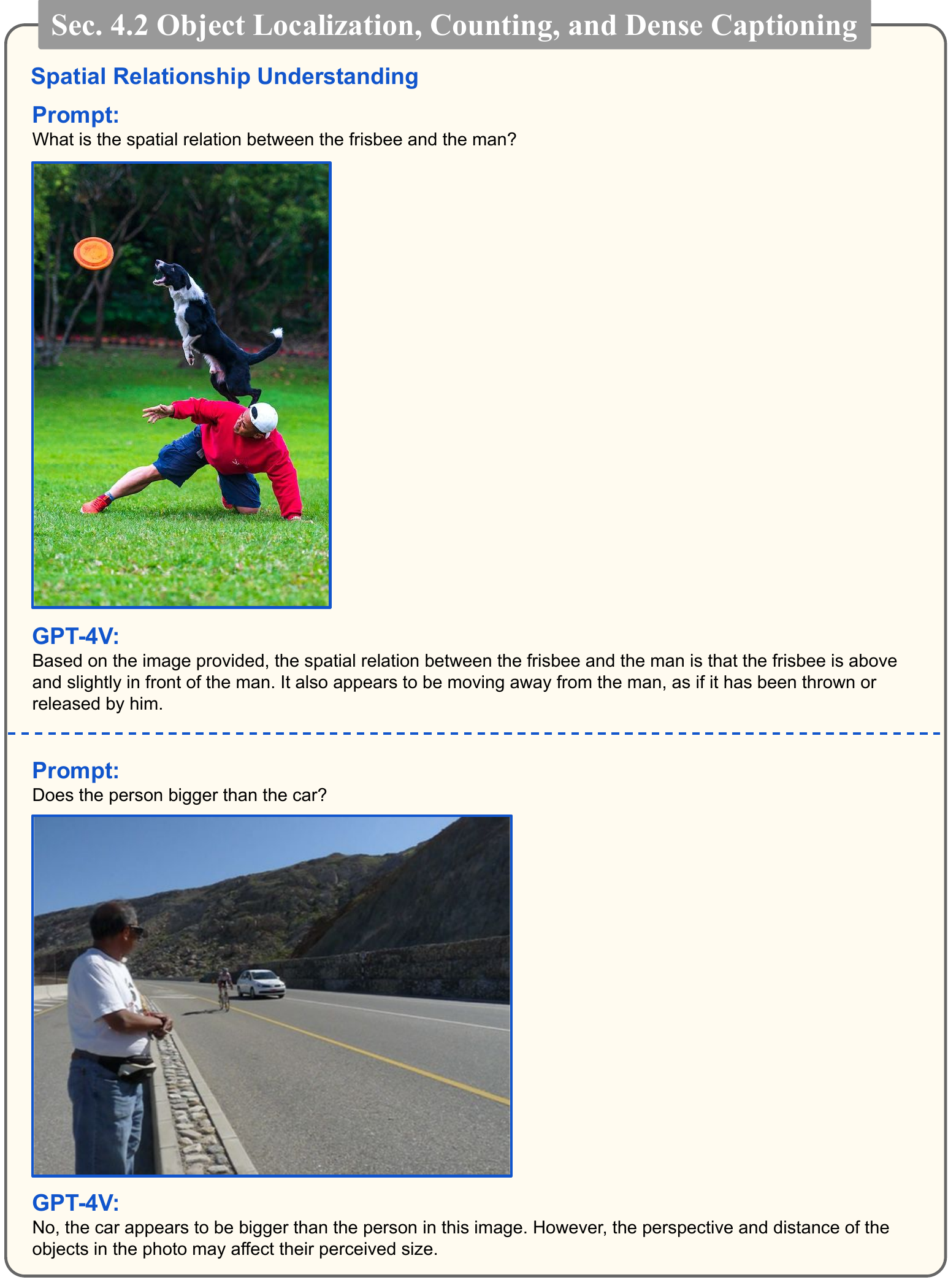}
\caption[Section~\ref{sec:od}: spatial relationship understanding.]{Results on spatial relationship understanding. \modelname recognizes the spatial relationship between the objects in the images. Example images are from~\cite{krishna2017visual,bagherinezhad2016elephants}. Check Section~\ref{sec:od} for detailed discussions. 
}
\label{fig:sec3-vl-spatial}
\vspace{15pt}
\end{figure*}

\noindent\textbf{Spatial relationship understanding.} Understanding the spatial relationship between humans and objects in the image is a vital aspect of visual intelligence~\cite{johnson2017clevr, bagherinezhad2016elephants}. In Figure~\ref{fig:sec3-vl-spatial}, \modelname showcases promising capabilities in this regard. It can identify the spatial relationship between the frisbee and the man in the image. It can also recognize the spatial relationship between the man and the car in the image, and point out that the camera perspective may affect their perceived size. 

\begin{figure*}[h!]
\centering
\includegraphics[width=\textwidth]{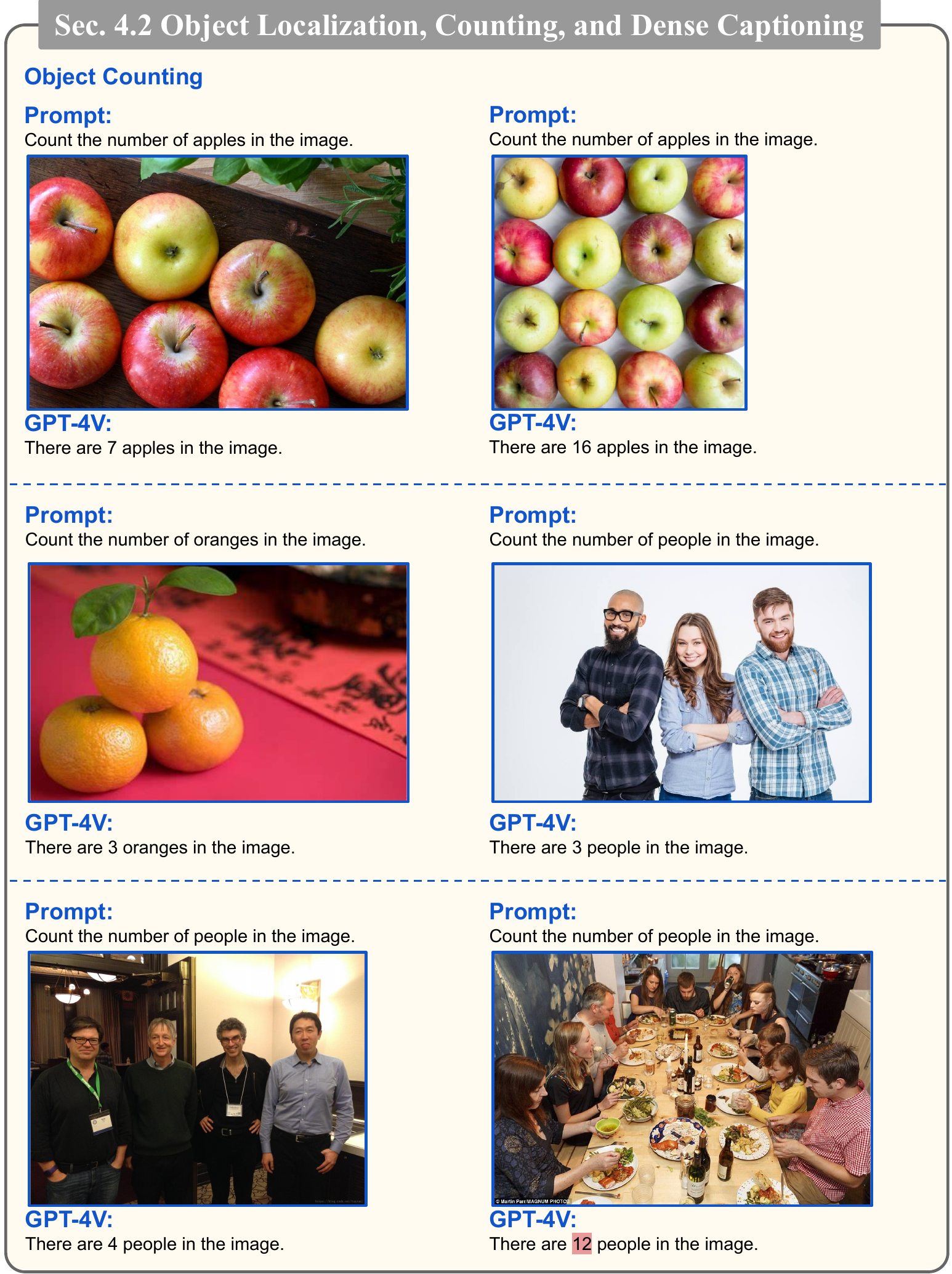}
\caption[Section~\ref{sec:od}: object counting.]{Results on object counting. \modelname is able to determine the quantity of the specified objects the image. \colorbox{redhl}{Red} highlights the wrong answer. Check Section~\ref{sec:od} for detailed discussions.
}
\label{fig:sec3-vl-count}
\end{figure*}

\noindent\textbf{Object counting.} Figure~\ref{fig:sec3-vl-count} highlights our exploration of \modelname's capability in object counting. In our experiments, we employ the text prompt ``Count the number of X in the image'' to evaluate its performance. The results indicate that \modelname can successfully count the number of objects, such as apples, oranges, and people, present in the image. However, challenges arise when objects are occluded, or the scene is cluttered, which can result in errors in the counting process. In the bottom left of Figure~\ref{fig:sec3-vl-count}, \modelname identifies 12 people, but the correct answer should be 11. This may be due to our limited text prompt used in this experiment, and further investigation in prompting techniques is needed. 

\begin{figure*}[h!]
\centering
\vspace{-17mm}
\includegraphics[width=\textwidth]{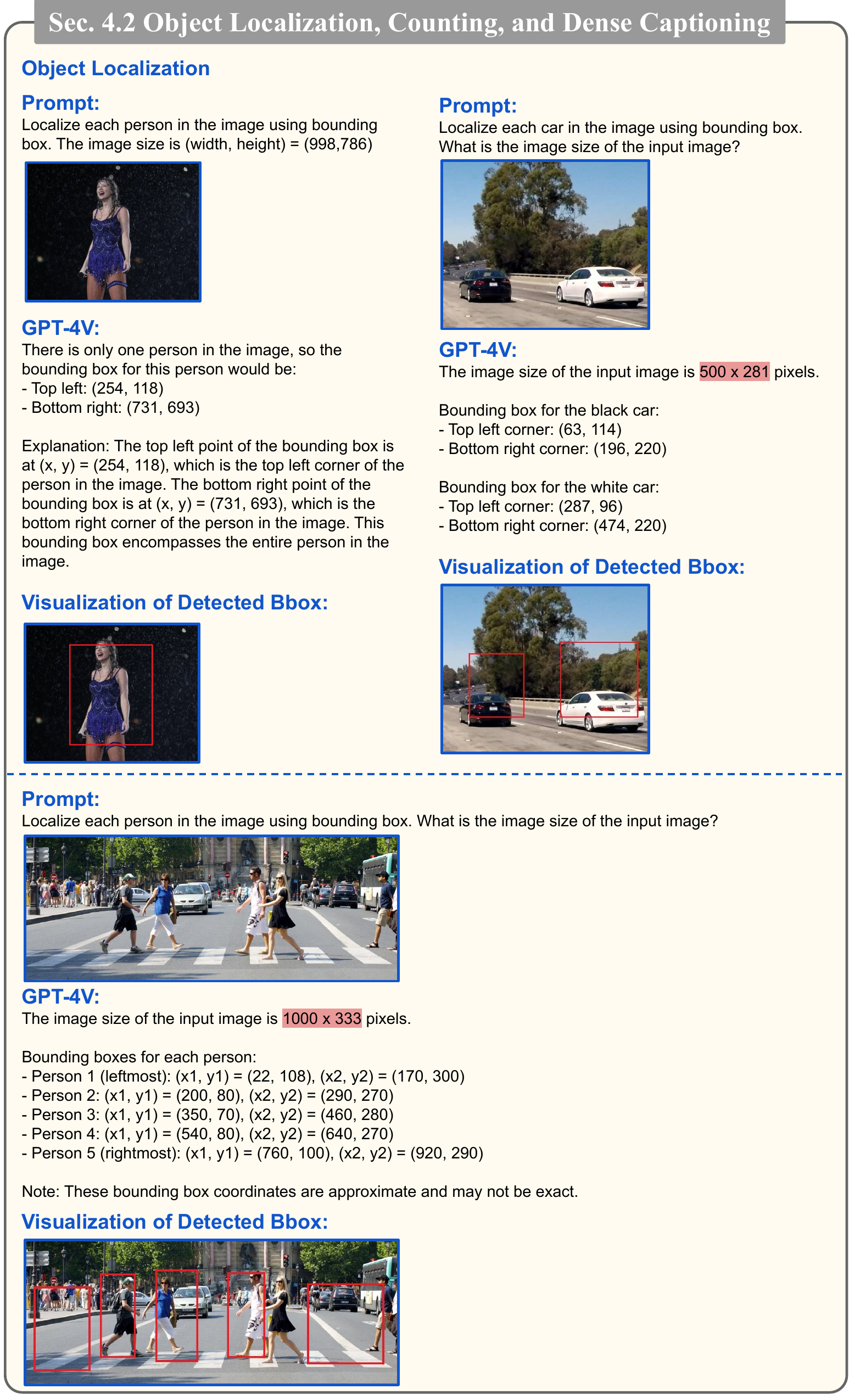}
\caption[Section~\ref{sec:od}: object localization.]{Results on object localization. \modelname is able to generate and approximate the bounding box coordinates for the specified objects in the image. When providing a simple text prompt only, the model may encounter challenges when dealing with more complex scenarios like object occlusions and cluttered scenes. \colorbox{redhl}{Red} highlights the wrong answer. We rescaled the predictions when visualizing the bounding boxes. Check Section~\ref{sec:od} for detailed discussions.  
}
\label{fig:sec3-vl-od}
\end{figure*}

\noindent\textbf{Object localization.} Object localization~\cite{zhou2016learning,lin2014microsoft,he2017mask} is a fundamental challenge in the field of computer vision. In our preliminary experiments, we address this task by utilizing a simple text prompt, ``Localize each person in the image using a bounding box.'' The initial results of our object localization experiments are depicted in Figure~\ref{fig:sec3-vl-od}. The findings suggest that \modelname demonstrates the capability to generate bounding box coordinates in textual format, without separate textualized box tokens~\cite{chen2021pix2seq,yang2022unitab,wang2022ofa,chen2022unified,lu2022unified,peng2023kosmos}. However, it is important to note that the generated bounding box coordinates are not accurate. We rescaled the predicted bounding box coordinates during visualization. Promising localization results are observed when the scene or background is relatively simpler and less cluttered. Further prompting techniques are required to enhance object localization performance in more complex and crowded environments. %

\begin{figure*}[h!]
\centering
\vspace{-22mm}
\includegraphics[width=\textwidth]{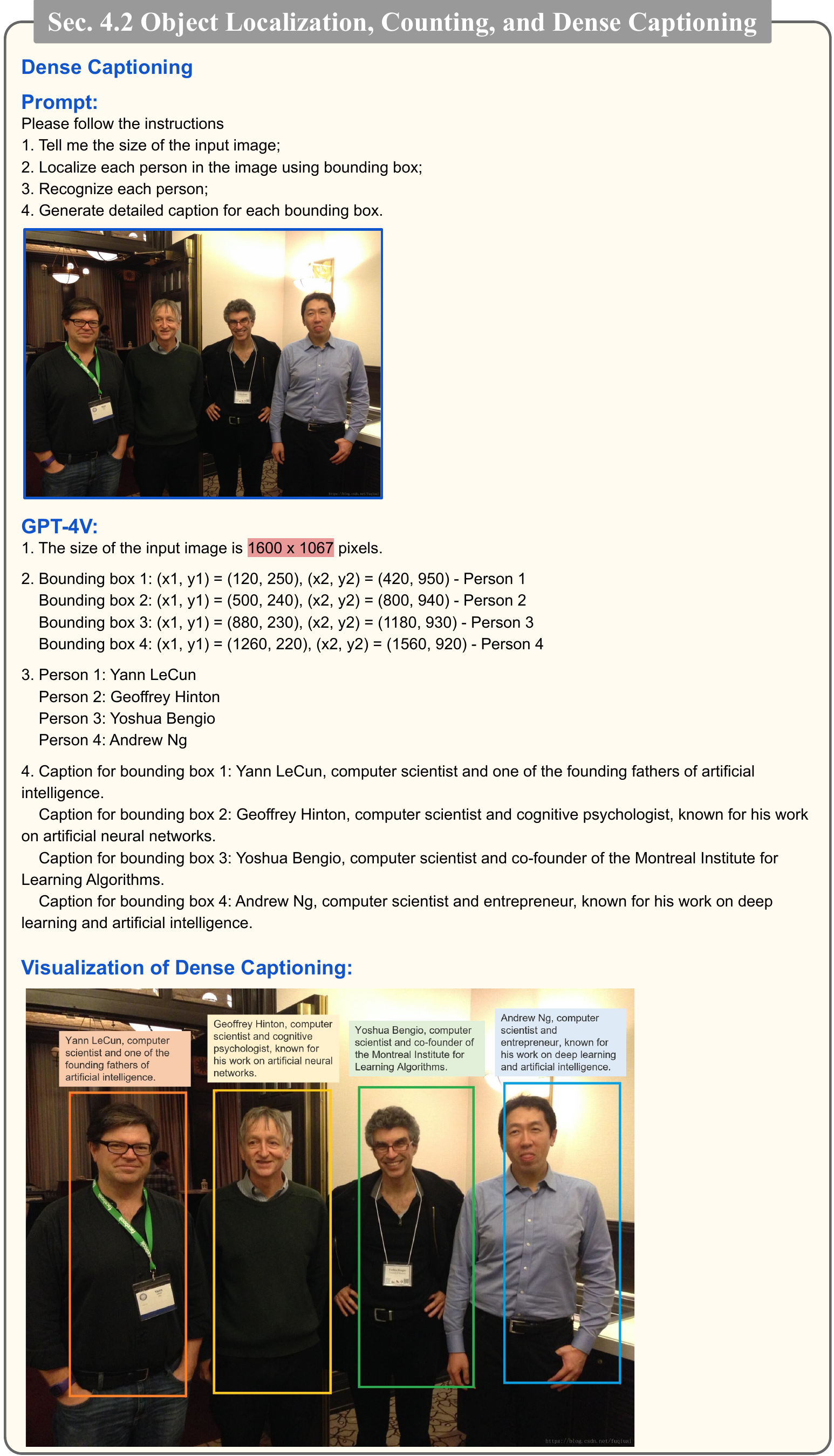}
\caption[Section~\ref{sec:od}: dense captioning.]{Results on dense captioning. \modelname follows the text prompt and successfully generates dense captions for the input image. \colorbox{redhl}{Red} highlights the wrong answer. We rescaled the predictions when visualizing the bounding boxes. Check Section~\ref{sec:od} for detailed discussions. 
}
\label{fig:sec3-vl-densecap}
\end{figure*}

\noindent\textbf{Dense captioning.} Dense captioning~\cite{johnson2016densecap,lu2018neural} involves generating detailed description for each region of interest in the given image. This advanced task in vision-language field typically requires a complex system that integrates multiple experts, such as object detector, celebrity recognition model, and image captioning model. In order to explore \modelname's capabilities in dense captioning, we use an instructional prompt, as shown in Figure~\ref{fig:sec3-vl-densecap}. The results are highly encouraging, as \modelname successfully localizes and recognizes the individuals within the image, and then provides concise descriptions for each scientist. %

\clearpage
\subsection{Multimodal Knowledge and Commonsense}\label{sec:knowledge}
\begin{figure*}[h!]
\centering
\vspace{-15mm}
\includegraphics[width=\textwidth]{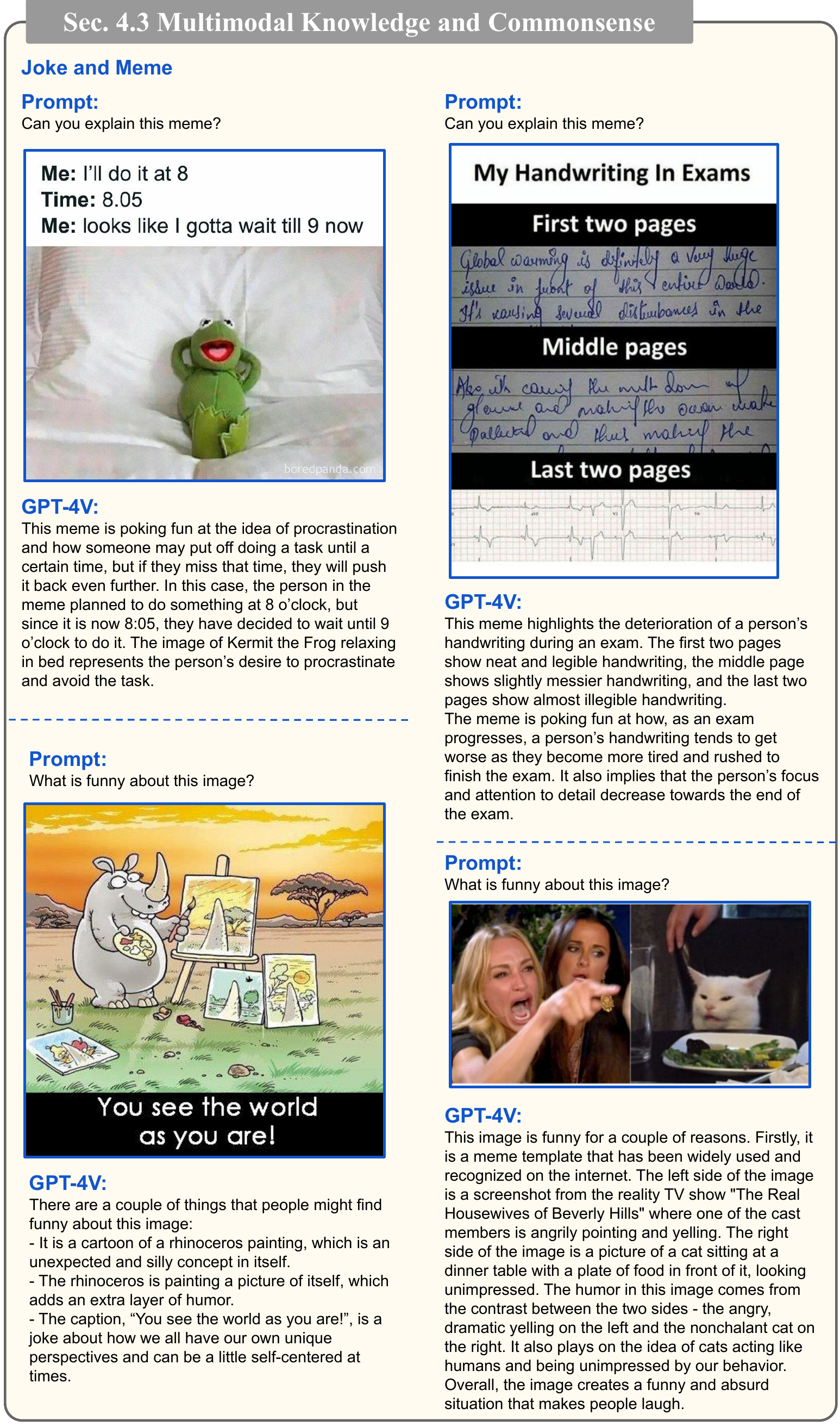}
\caption[Section~\ref{sec:knowledge}: joke and meme understanding.]{Results on joke and meme understanding. \modelname demonstrates the impressive capability to comprehend the humor embedded within memes. Check Section~\ref{sec:knowledge} for detailed discussions. 
}
\label{fig:sec3-vl-joke}
\end{figure*}

\noindent\textbf{Joke and meme.} 
Jokes and memes often reference specific events, pop culture, or Internet trends. Understanding these references requires being familiar with the relevant context and cultural knowledge. Grasping the visual elements, their relationship to the text, and the intended humorous effect can be a complex task~\cite{gpt4}. Moreover, memes are often user-generated, making them highly diverse and ever-expanding. To evaluate \modelname's ability in this domain, we input a pair of meme and text prompt to \modelname. The example text prompts include ``Can you explain the meme?'' and ``What is funny about the image?'' Figure~\ref{fig:sec3-vl-joke} shows the example results. We observe that \modelname has remarkable ability to gather information from both visual and textual modalities, and then comprehend the humor embedded within memes.

\begin{figure*}[h!]
\centering
\includegraphics[width=\textwidth]{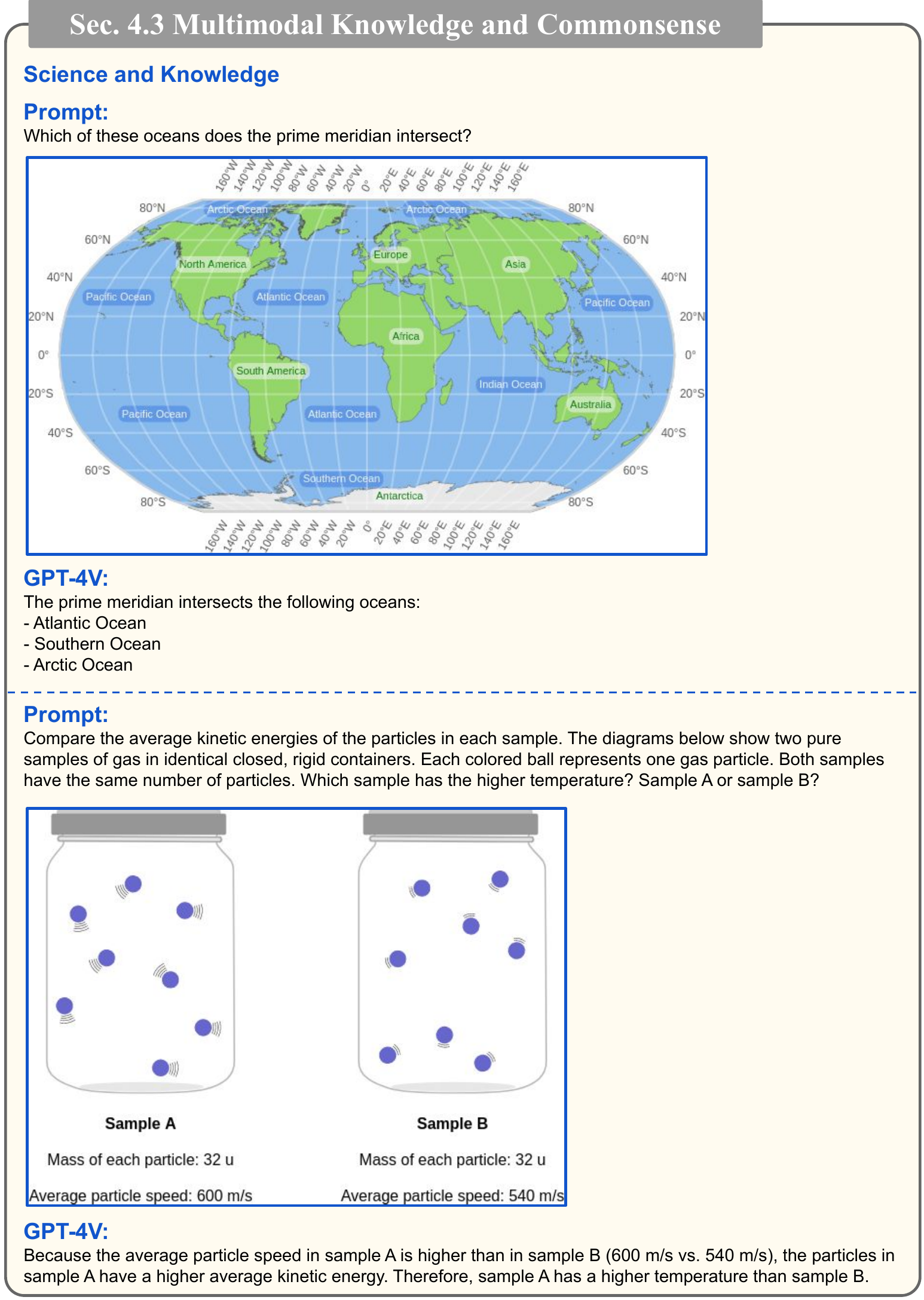}
\caption[Section~\ref{sec:knowledge}: science and encyclopedia knowledge.]{Results on answering science questions. \modelname can understand the question textually and visually, and gather necessary information to answer the question. Example images are from ~\cite{lu2022learn}. Check Section~\ref{sec:knowledge} for detailed discussions. 
}
\label{fig:sec3-vl-science}
\end{figure*}

\begin{figure*}[h!]
\centering
\includegraphics[width=\textwidth]{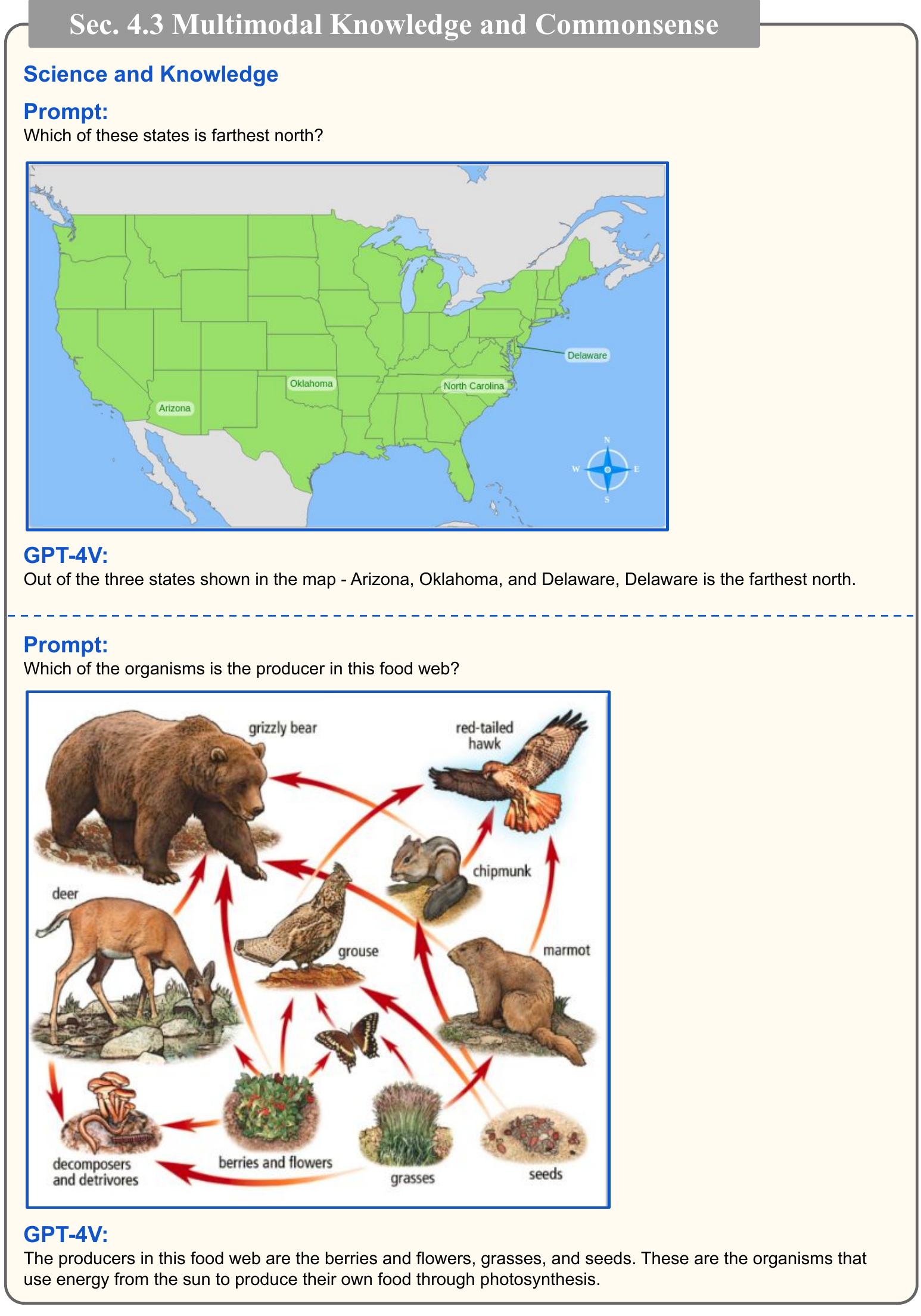}
\caption[Section~\ref{sec:knowledge}: science and encyclopedia knowledge.]{Results on answering science questions. \modelname can understand the question textually and visually, and gather necessary information to answer the question. Example images are from ~\cite{lu2022learn}. Check Section~\ref{sec:knowledge} for detailed discussions. 
}
\label{fig:sec3-vl-science-2}
\end{figure*}

\begin{figure*}[h!]
\centering
\includegraphics[width=\textwidth]{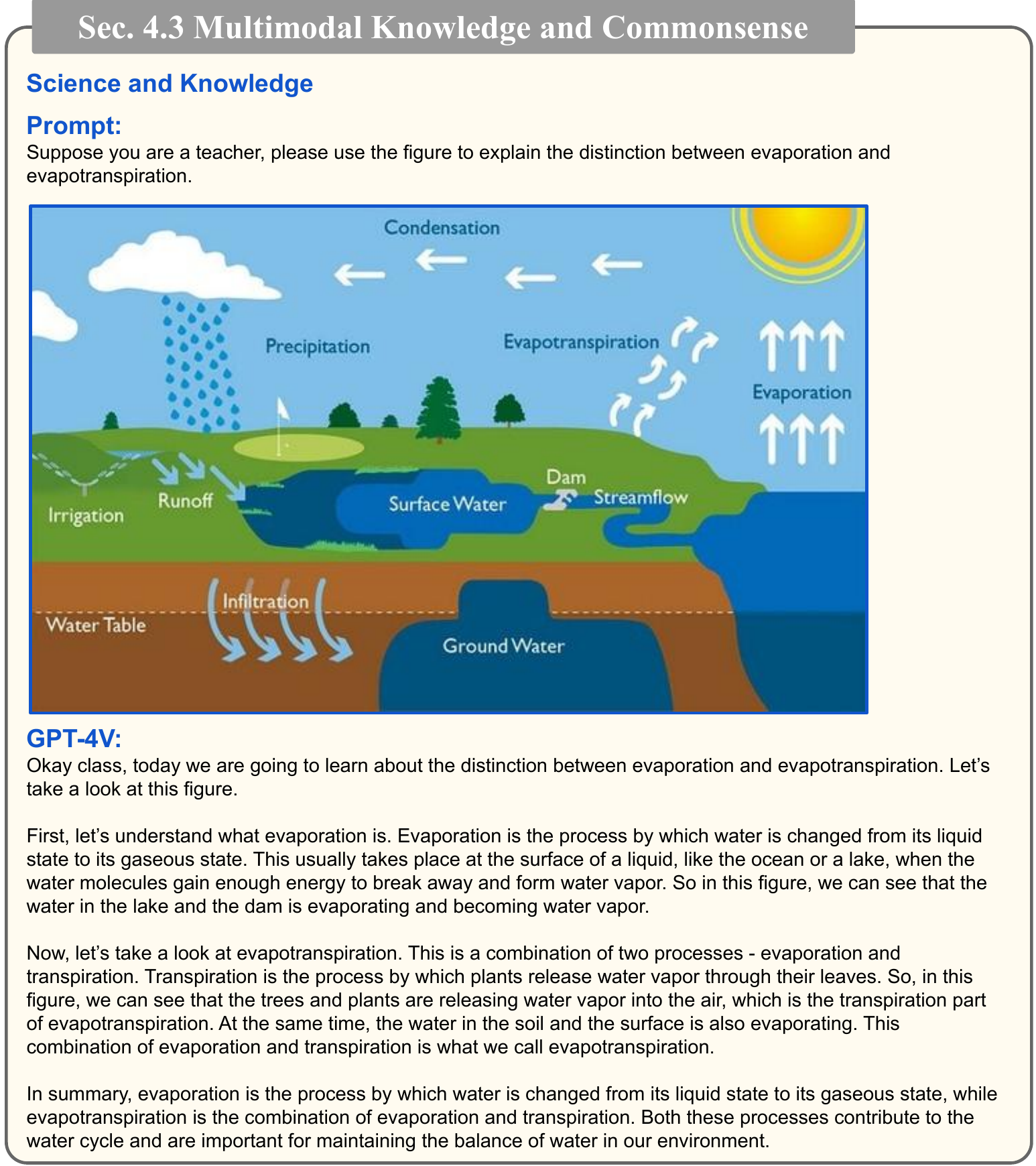}
\caption[Section~\ref{sec:knowledge}: science and encyclopedia knowledge.]{Results on answering science questions. When we use a more specific text prompt like ``Suppose you are a teacher, please use the figure to explain X,'' we observe that \modelname can generate a short tutorial for explaining the subject. Check Section~\ref{sec:knowledge} for detailed discussions. 
}
\label{fig:sec3-vl-science-3}
\end{figure*}

\noindent\textbf{Science and knowledge.} We further investigate \modelname's capability in tasks that requires reasoning with scientific knowledge~\cite{lu2022learn}. We conduct experiments by providing a text prompt question and a corresponding image. The questions cover a wide range of topics, including geography, physics, biology, and earth science. In Figures~\ref{fig:sec3-vl-science}-\ref{fig:sec3-vl-science-3}, we observe that \modelname is able to correctly answer the science questions based on the visual context.
For instance, in the bottom row of Figure~\ref{fig:sec3-vl-science}, \modelname recognizes the average particle speed for both sample A and sample B. By considering the relationship among particle speed, kinetic energy, and temperature, \modelname answers the question correctly. For another instance, as shown in the bottom row of Figure~\ref{fig:sec3-vl-science-2}, \modelname takes into account the visual arrows presented in the figure to identify the producer in the specific food web. Moreover, as shown in Figure~\ref{fig:sec3-vl-science-3}, when we provide a more specific prompt, such as ``Suppose you are a teacher, please use the figure to explain X,'' we observe the generated answer adopts a tutorial format and explains the subject step by step. 

\begin{figure*}[h!]
\centering
\vspace{-12mm}
\includegraphics[width=\textwidth]{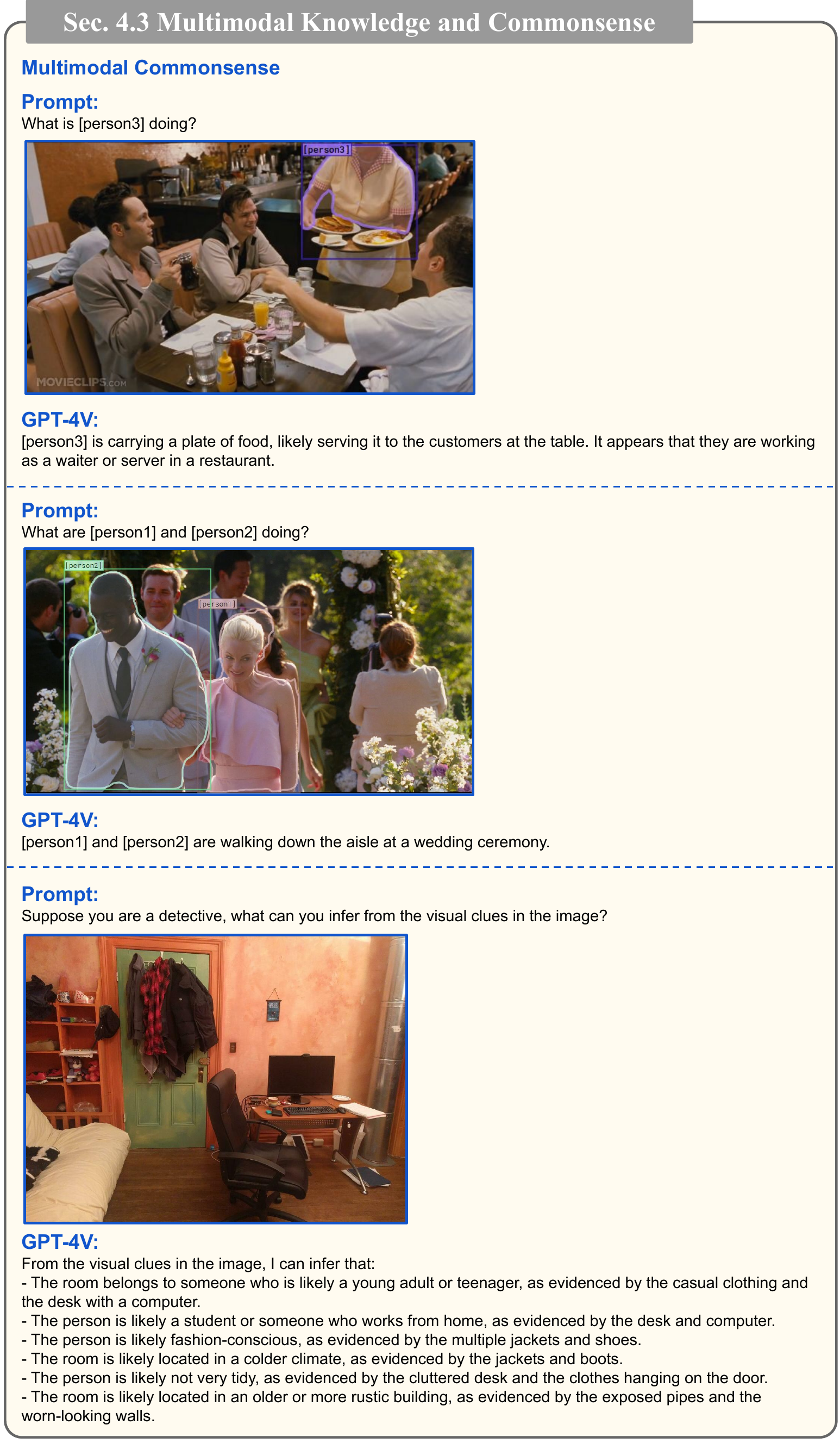}
\caption[Section~\ref{sec:knowledge}: multimodal commonsense.]{Results on multimodal commonsense reasoning. Example images are from~\cite{zellers2019recognition,hesselhwang2022abduction}. Check Section~\ref{sec:knowledge} for detailed discussions. 
}
\label{fig:sec3-vl-comsense}
\end{figure*}

\noindent\textbf{Multimodal commonsense.} In Figure~\ref{fig:sec3-vl-comsense}, we access the ability of \modelname in multimodal commonsense reasoning~\cite{zellers2019recognition,hesselhwang2022abduction}. In our experiments, we observed that \modelname effectively utilizes the bounding boxes presented in the image as visual prompts (\textit{e.g.,} [person1] and [person2]) to recognize the actions performed by the individuals. As shown in the second example in Figure~\ref{fig:sec3-vl-comsense}, based on the formal dress worn by [person1] and [person2] and the floral decorations present in the scene, it can be inferred that they are attending a wedding ceremony. Moreover, when we provide a more specific input prompt, such as "Suppose you are a detective, what can you infer from the visual clues?", \modelname demonstrates the ability to discern numerous nuanced visual cues within the image and offers a list of plausible hypotheses.

\clearpage
\subsection{Scene Text, Table, Chart, and Document Reasoning}\label{sec:document}

\begin{figure*}[h!]
\centering
\centerline{\includegraphics[width=1.2\textwidth]{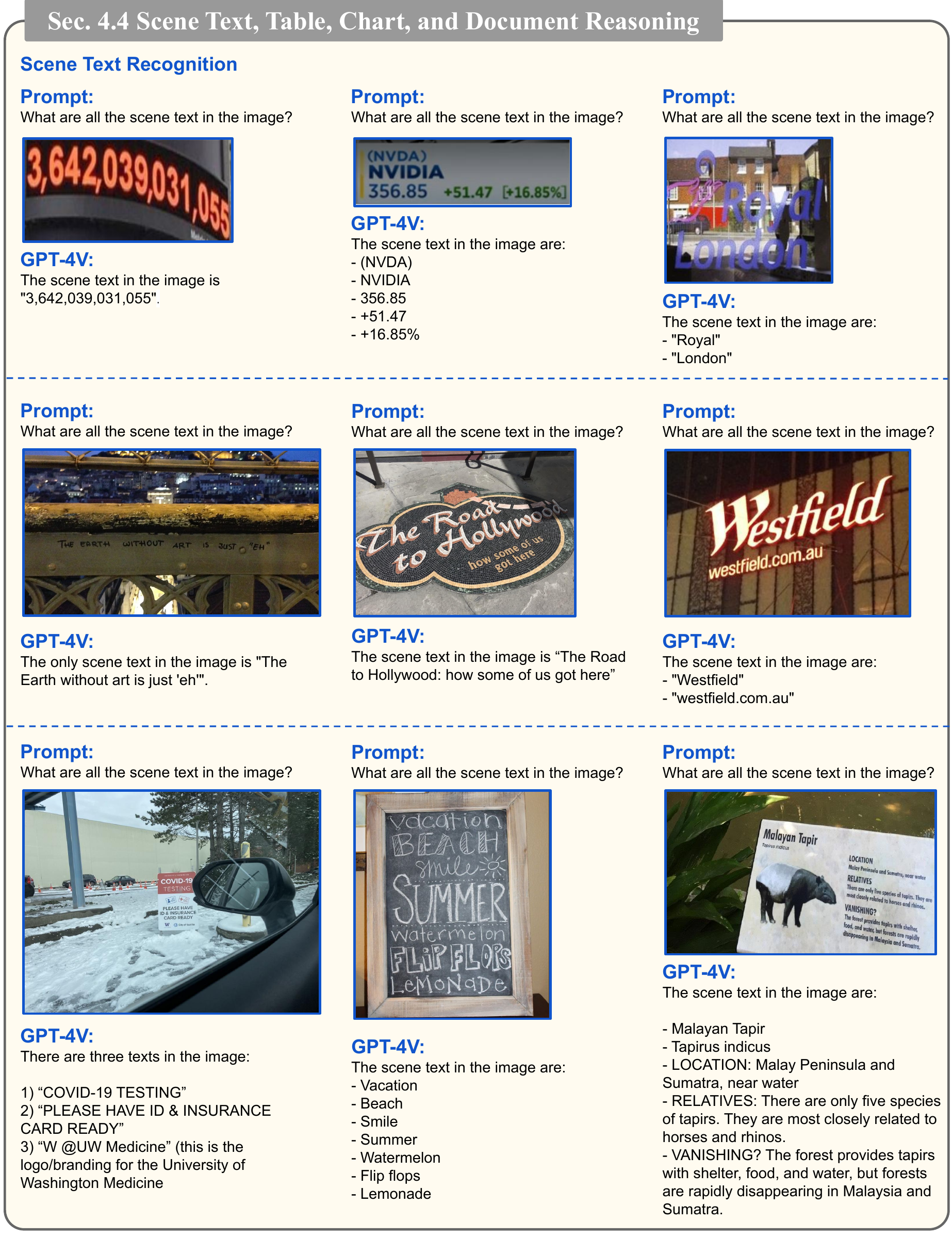}}
\caption[Section~\ref{sec:document}: scene text recognition.]{Results on scene text recognition. \modelname~can recognize scene text in many challenging scenarios. Check Section~\ref{sec:document} for detailed discussions. 
}
\label{fig:sec3-vl-scenetext}
\end{figure*}

\noindent\textbf{Scene text recognition.} Reading and understanding scene text in images is an important task in vision-language~\cite{sidorov2020textcaps,singh2019towards, su2019vl,biten2019scene}. In our experiments, we investigate \modelname's ability to recognize scene text by utilizing the input prompt ``What are all the scene text in the image?'' Figure~\ref{fig:sec3-vl-scenetext} shows the example results. We observe \modelname accurately identifies scene text in various scenarios, including both handwritten and printed text. In Section~\ref{sec:multilingual}, we present further results on multilingual scenarios.

\begin{figure*}[h!]
\centering
\includegraphics[width=\textwidth]{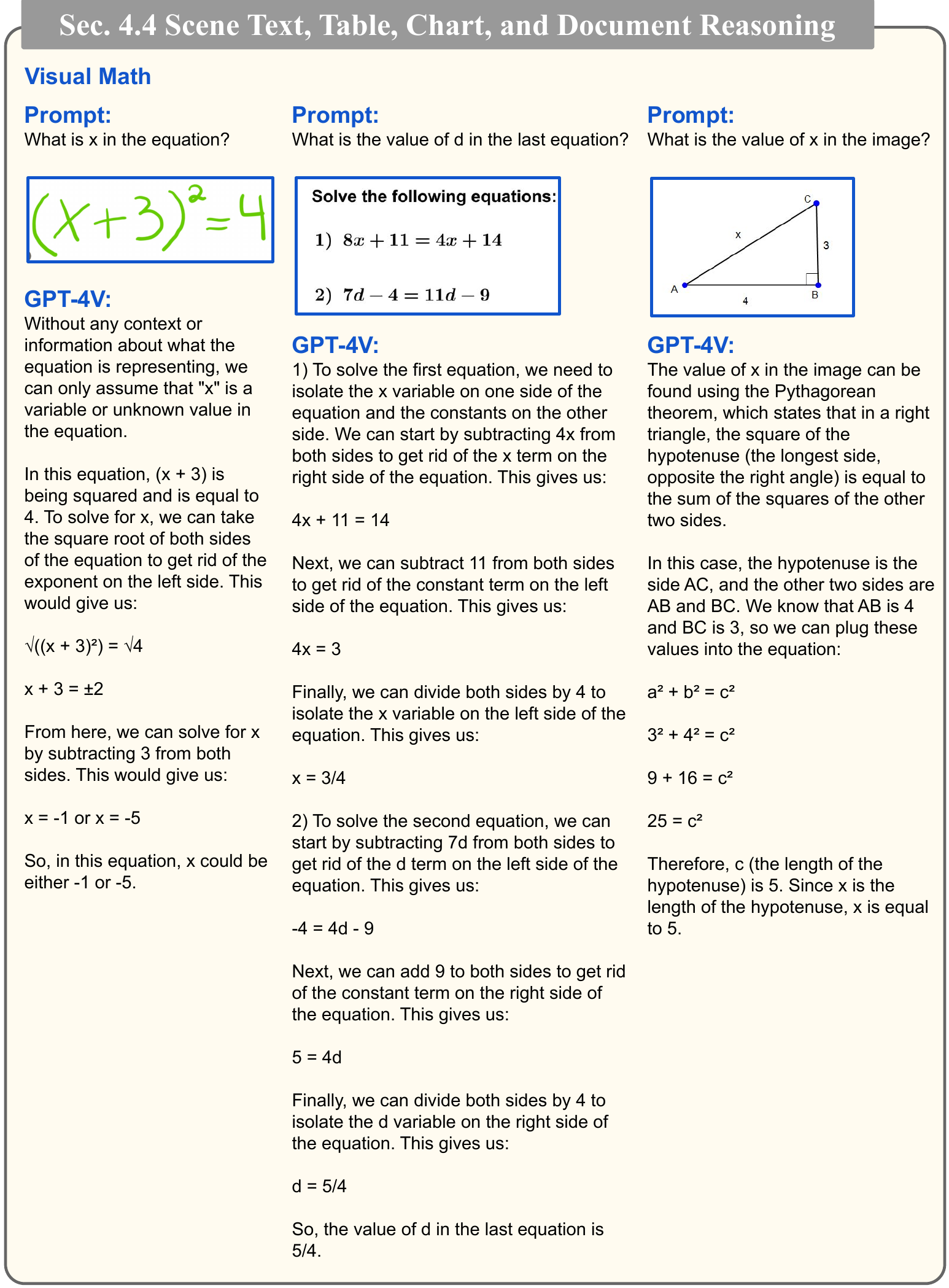}
\caption[Section~\ref{sec:document}: visual math reasoning.]{Results on visual math reasoning. \modelname~is able to comprehend and solve visual math problems with a well-structured solution. Check Section~\ref{sec:document} for detailed discussions.  
}
\label{fig:sec3-vl-math}
\end{figure*}

\noindent\textbf{Visual math reasoning.} In Figure~\ref{fig:sec3-vl-math}, \modelname demonstrates its capability in solving visual math problems. In our experiments, we observe \modelname is able to extract essential information from the image. For instance, in Figure~\ref{fig:sec3-vl-math}, \modelname correctly identifies the presence of a right triangle (or orthogonal triangle) and determines that AB is 4 units and BC is 3 units. In addition, we note that \modelname tends to present solutions in a well-structured manner, solving the problem step by step, thereby showcasing its ability to provide clear explanations. 

\begin{figure*}[h!]
\centering
\includegraphics[width=\textwidth]{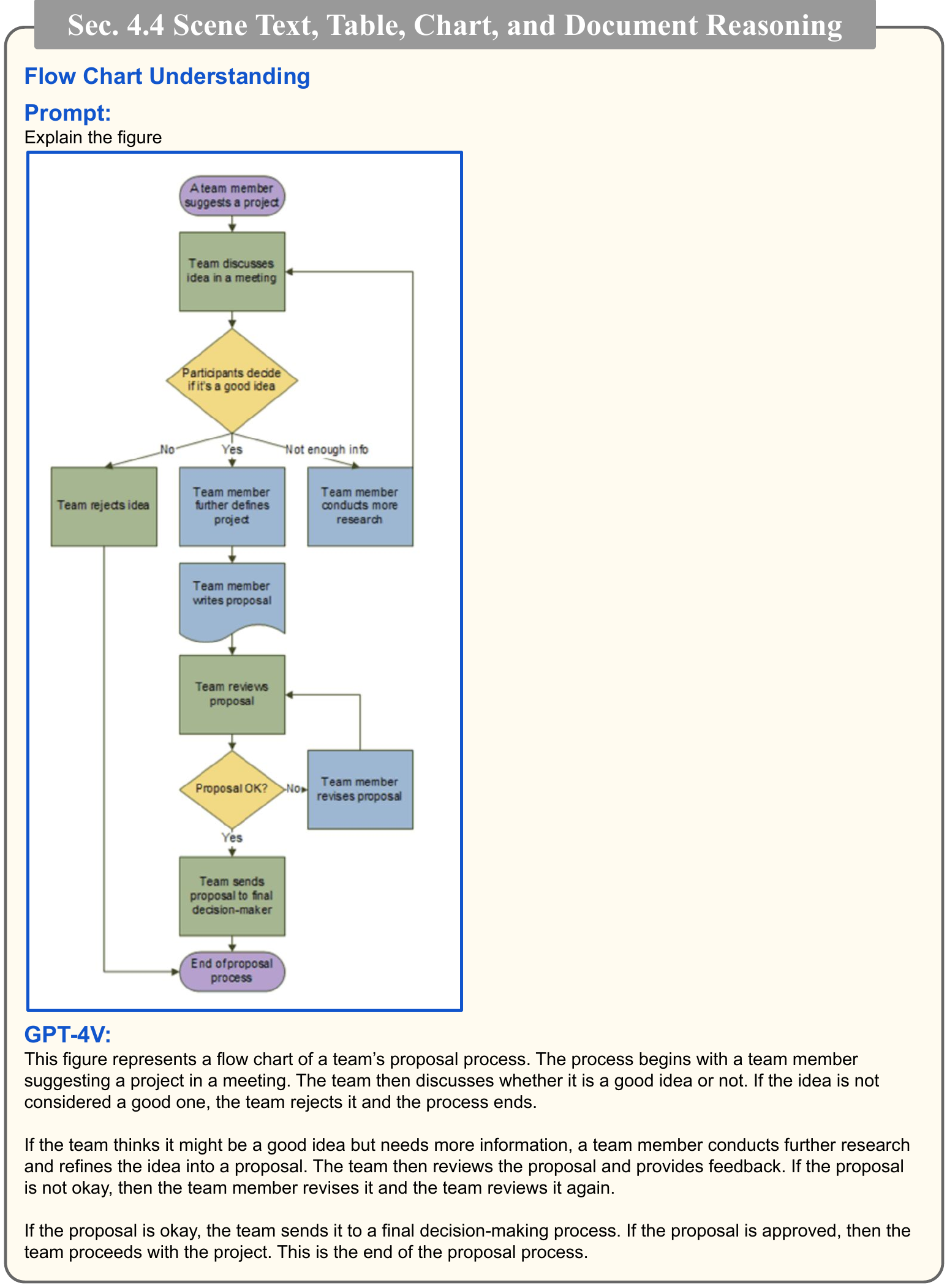}
\caption[Section~\ref{sec:document}: flow chart understanding.]{Results on flow chart understanding. \modelname~correctly describes the proposal process in details. Check Section~\ref{sec:document} for detailed discussions.   
}
\label{fig:sec3-vl-chart-1}
\end{figure*}

\begin{figure*}[h!]
\centering
\includegraphics[width=\textwidth]{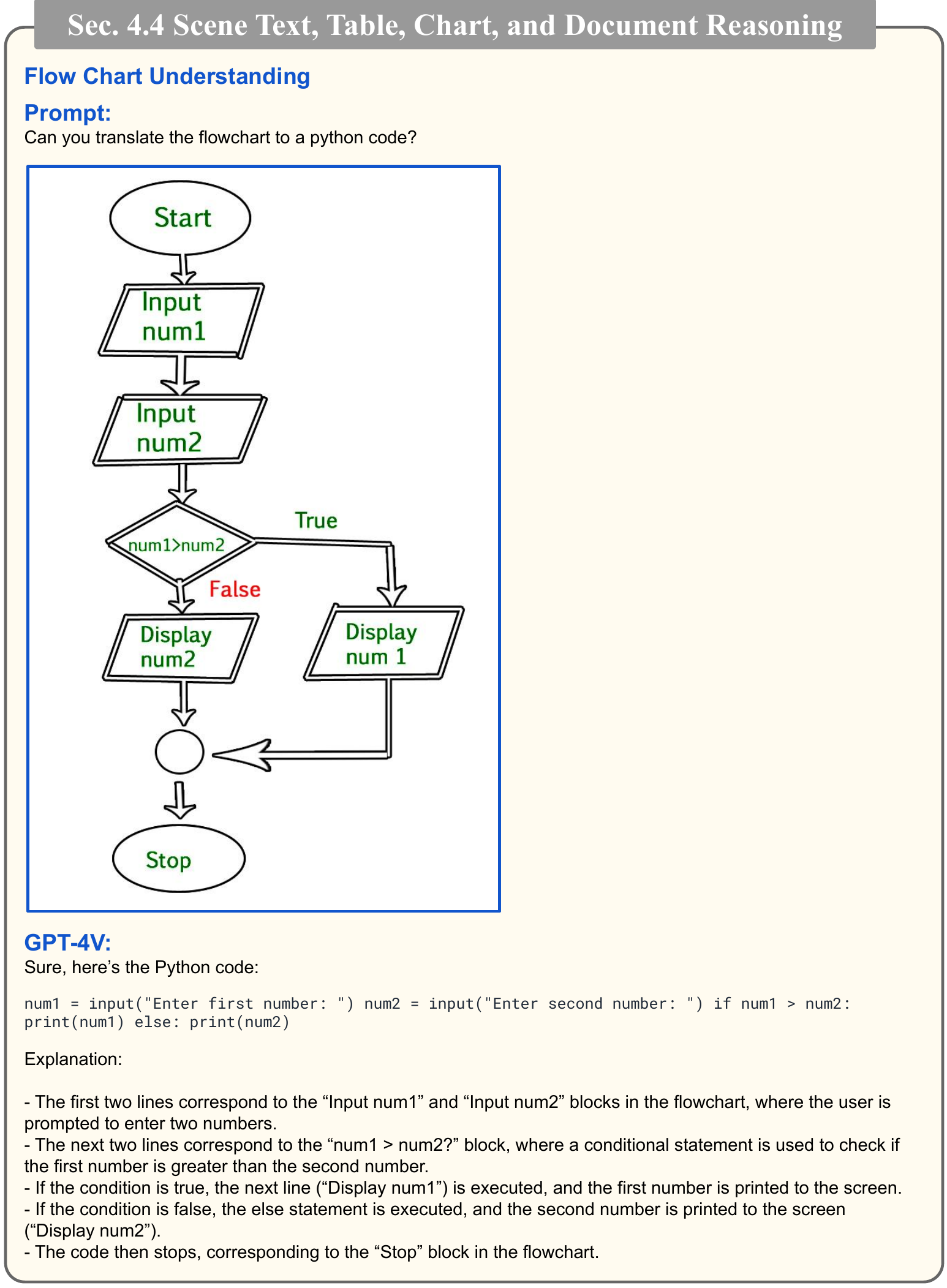}
\caption[Section~\ref{sec:document}: flow chart understanding.]{Results on flow chart understanding. \modelname~is able to translate the flow chart to a python code. Check Section~\ref{sec:document} for detailed discussions.  
}
\label{fig:sec3-vl-chart-2}
\end{figure*}

\begin{figure*}[h!]
\centering
\includegraphics[width=\textwidth]{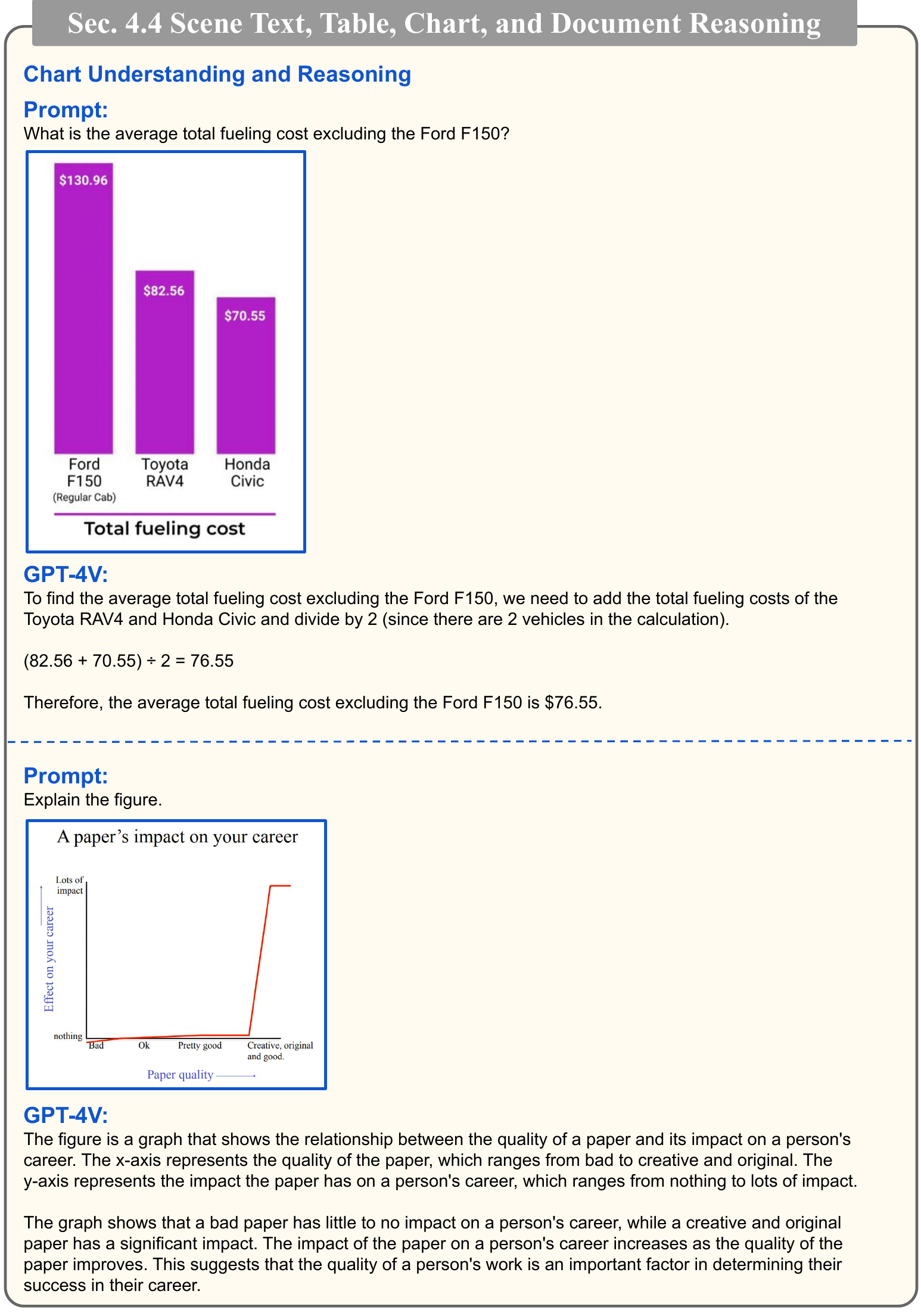}
\caption[Section~\ref{sec:document}: chart understanding and reasoning.]{\modelname~shows promising results in understanding the details in the chart, as well as in reasoning and accurately responding to related questions. Check Section~\ref{sec:document} for detailed discussions. 
}
\label{fig:sec3-vl-chart-3}
\end{figure*}

\noindent\textbf{Chart understanding and reasoning.} We further study \modelname's ability in chart understanding and reasoning. Figures~\ref{fig:sec3-vl-chart-1}-\ref{fig:sec3-vl-chart-3} show the example results. In our preliminary explorations, \modelname exhibits the ability to provide detailed descriptions of charts. For example, in Figure~\ref{fig:sec3-vl-chart-1}, the model correctly explains the proposal process from the beginning to the end. In Figure~\ref{fig:sec3-vl-chart-2}, the model not only understands the program in the given flow chat, but also translates the details to a python code. In the bottom row of Figure~\ref{fig:sec3-vl-chart-3}, \modelname shows a clear understanding of both x- and y-axis, and explains the key insight presented in the chart.  Furthermore, in our experiments, we observe that \modelname can answer questions based on the chart. In the top row of Figure~\ref{fig:sec3-vl-chart-3}, \modelname correctly calculates the average total fueling cost, excluding the Ford F150. 

\begin{figure*}[h!]
\centering
\includegraphics[width=\textwidth]{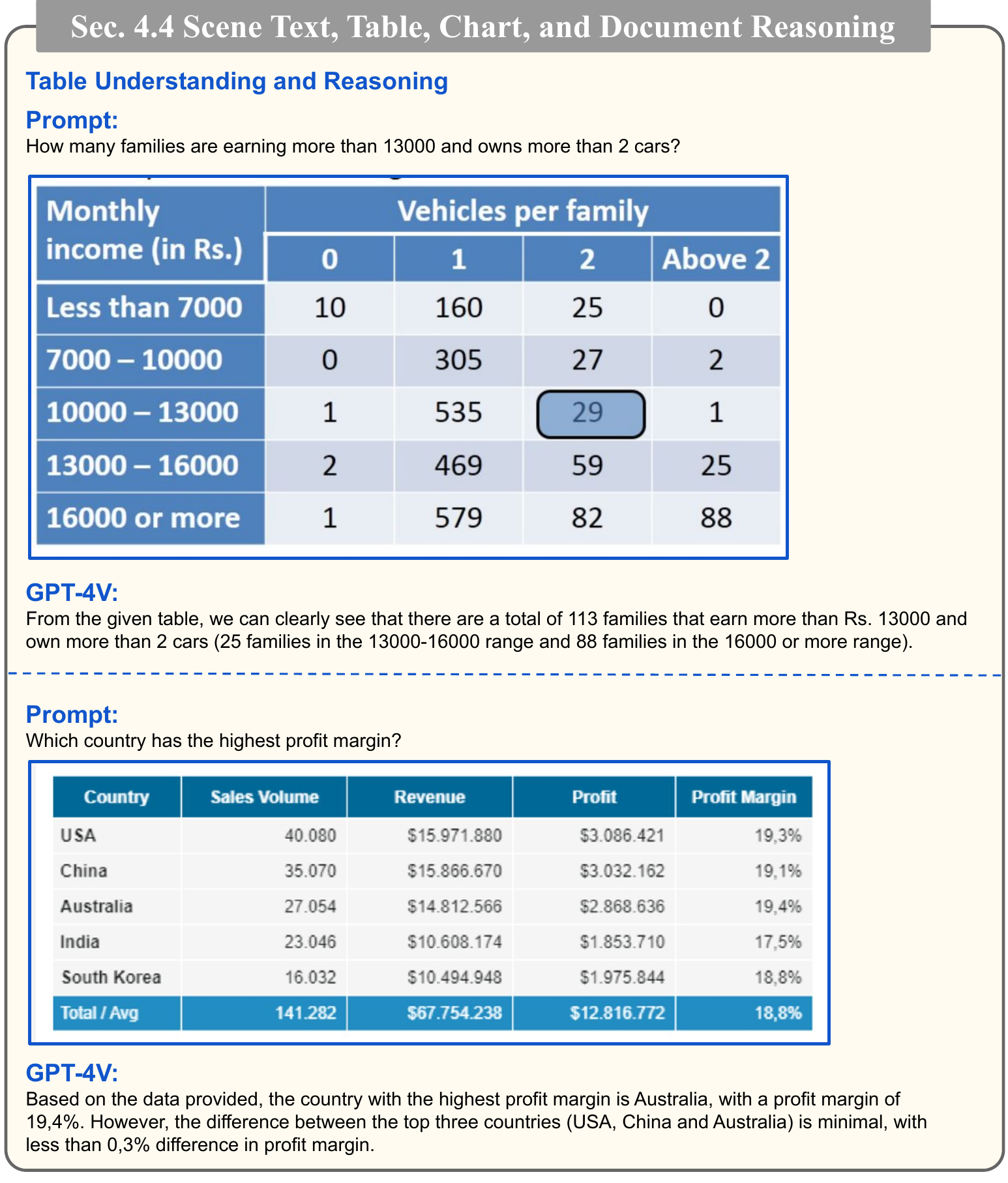}
\caption[Section~\ref{sec:document}: table understanding and reasoning.]{We observe \modelname can understand the details in the table, and answer related questions. Check Section~\ref{sec:document} for detailed discussions. 
}
\label{fig:sec3-vl-table-1}
\end{figure*}

\noindent\textbf{Table understanding and reasoning.} In Figure~\ref{fig:sec3-vl-table-1}, we present our preliminary investigations into table understanding and reasoning. Similar to the findings from chart experiments, \modelname shows promising results in understanding the details in the table, as well as in reasoning and accurately responding to related questions. 

\begin{figure*}[h!]
\centering
\vspace{-18mm}
\includegraphics[width=\textwidth]{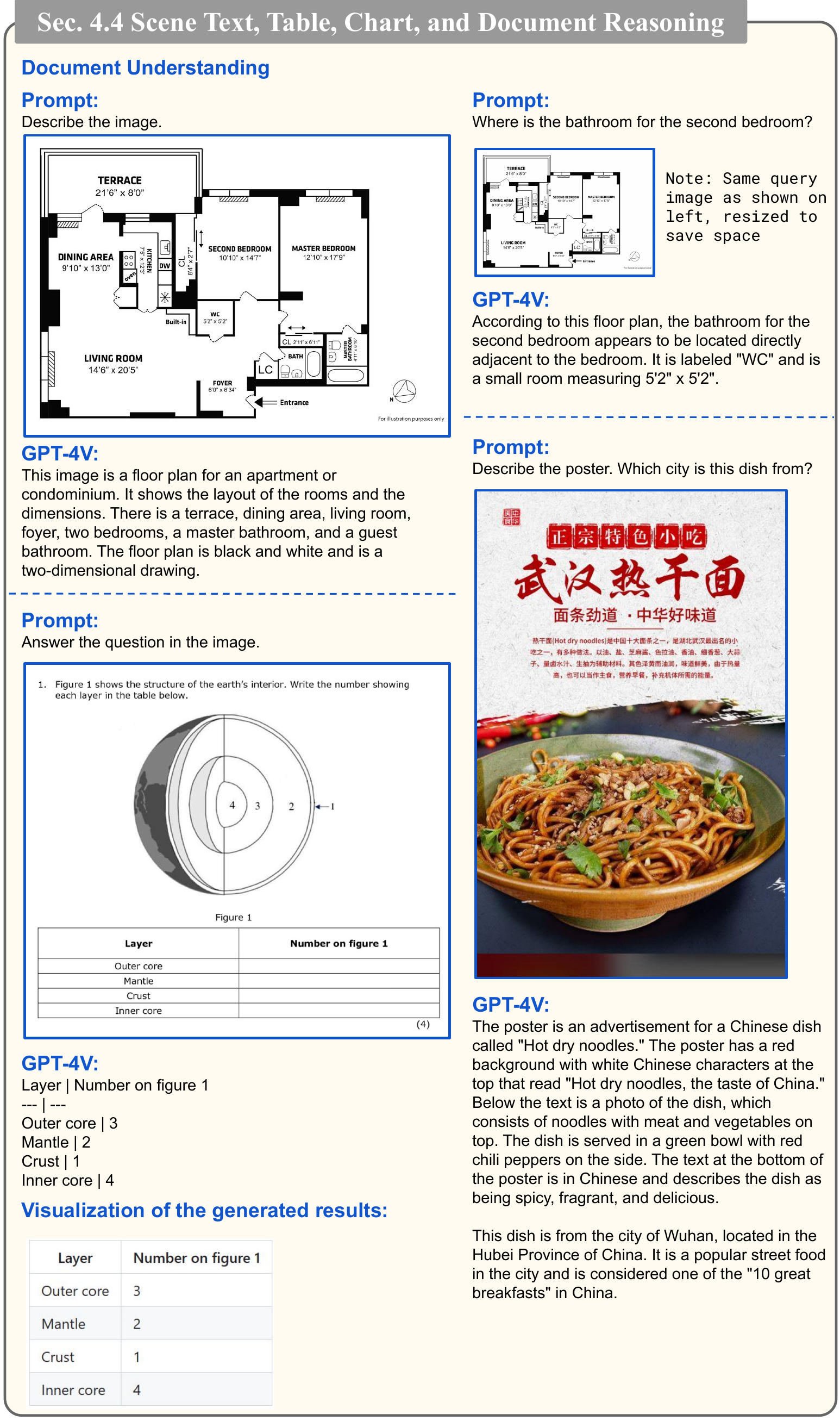}
\caption[Section~\ref{sec:document}: document understanding.]{Results on document understanding. \modelname recognizes three different types of document and answers the questions correctly. Check Section~\ref{sec:document} for detailed discussions.  
}
\label{fig:sec3-vl-doc1}
\end{figure*}

\noindent\textbf{Document understanding.} Figure~\ref{fig:sec3-vl-doc1} shows the results on various types of documents, such as floor plan, poster, and exam paper. We observe \modelname demonstrates an understanding of the documents and provides reasonable responses. For instance, it accurately identifies the location of the bathroom for the second bedroom in the floor plan. It also recognizes the Chinese dish ``Hot dry noodles,'' and associates it with the city of Wuhan by following the scene text. Moreover, \modelname is capable of reading an exam paper. It accurately reconstructs the table in Markdown, and then fills in the table with the correct answers. We present more explorations in its coding ability in Section~\ref{sec:04language}.

\begin{figure*}[h!]
\centering
\includegraphics[width=\textwidth]{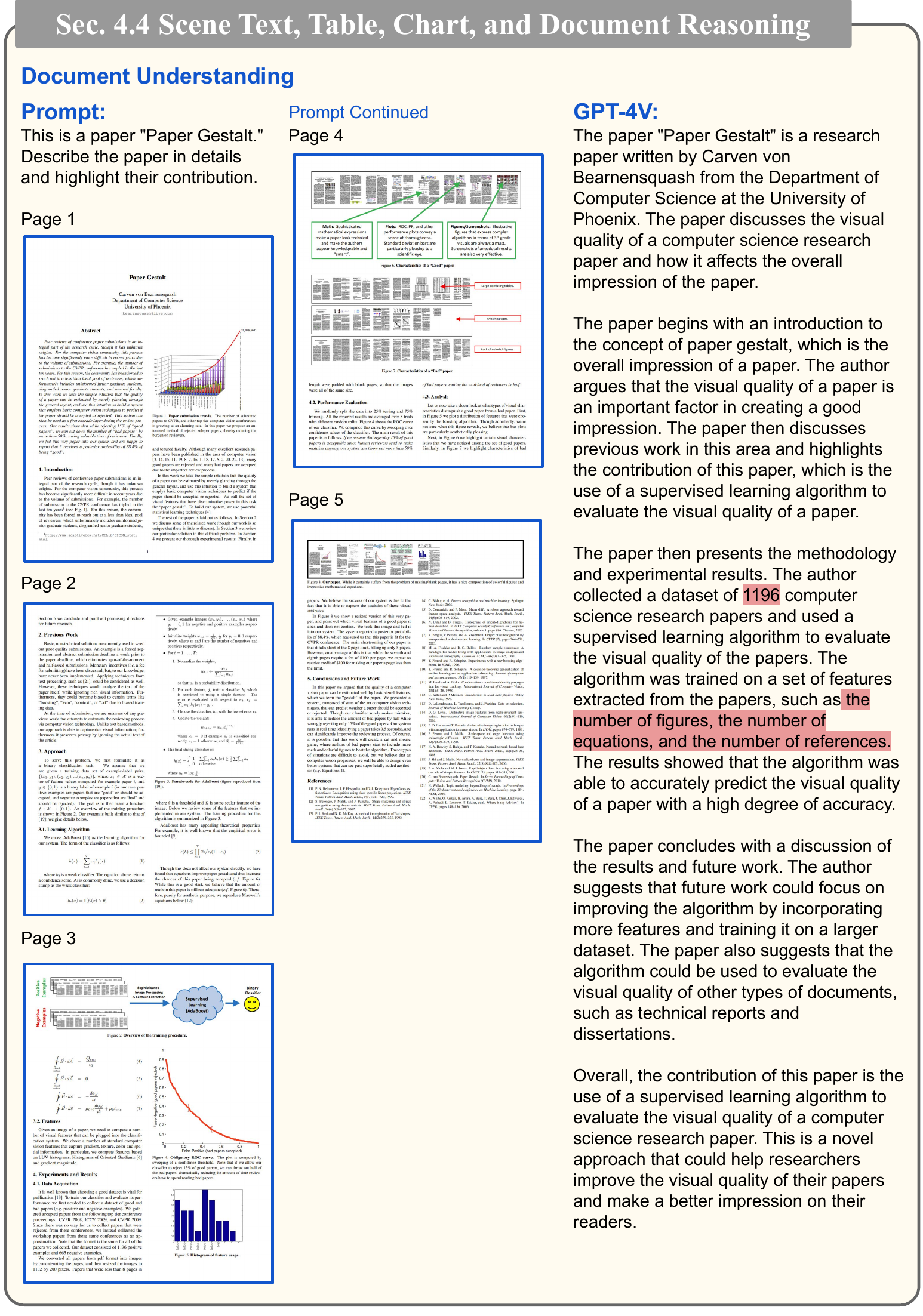}
\caption[Section~\ref{sec:document}: long document understanding.]{Results on document understanding. \modelname reads a multi-page technical report, understands the content in each section, and provides a summary of the contribution of this technical report. \colorbox{redhl}{Red} highlights the wrong answer. Check Section~\ref{sec:document} for detailed discussions.
}
\label{fig:sec3-vl-doc2}
\end{figure*}

In Figure~\ref{fig:sec3-vl-doc2}, we explore a more challenging case by providing a technical report~\cite{von2010paper} with multiple pages as input. In our limited exploration, \modelname exhibits impressive results. It correctly describes the main idea and their proposed method by considering the context across multiple pages. However, it may occasionally miss some implementation details. Please note that the dataset should contain 1196+665=1861 examples, and the extracted features should include Histograms of Oriented Gradients (HOG). Instead of prompting all pages to the model simultaneously, we believe that exploring more advanced prompting techniques, such as thinking step-by-step or employing in-context few-shot approaches, could potentially enhance the model's performance.

\clearpage
\subsection{Multilingual Multimodal Understanding}\label{sec:multilingual}

\begin{figure*}[h!]
\centering
\vspace{-15mm}
\centerline{\includegraphics[width=1.2\textwidth]{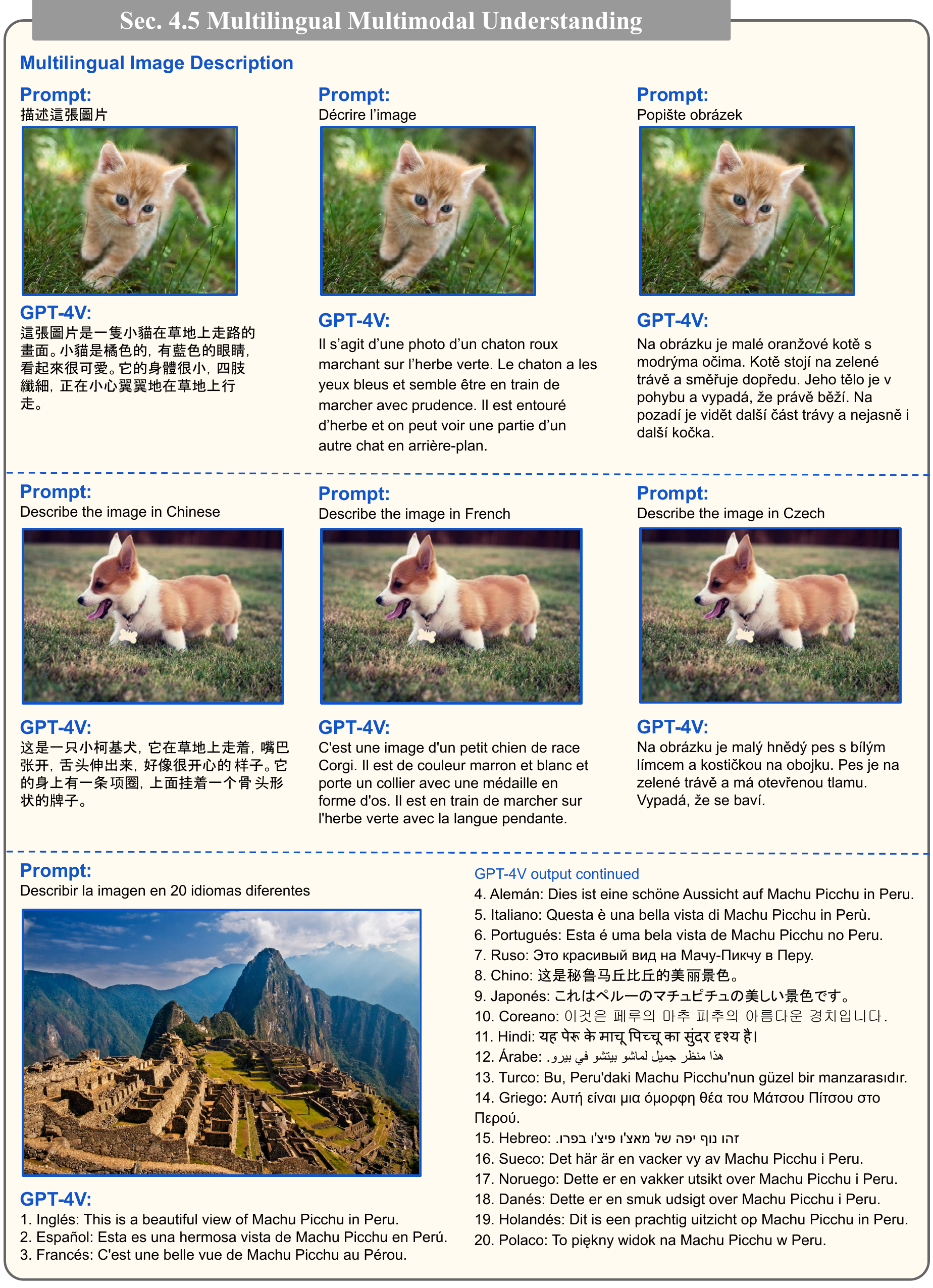}}
\caption[Section~\ref{sec:multilingual}: multilingual image descriptions.]{Results on multilingual image descriptions. \modelname is able to generate image descriptions in different languages. Check Section~\ref{sec:multilingual} for detailed discussions.
}
\label{fig:sec3-vl-multilingual}
\end{figure*}

We assess \modelname's ability in comprehending multiple languages and modalities. 
First, we explore this capability by evaluating natural images without scene text, as depicted in Figure~\ref{fig:sec3-vl-multilingual}. In the first row of the figure, we provide the input text prompt ``Describe the image'' in Chinese, French, and Czech, respectively. \modelname recognizes the input text prompts in different languages, and generates correct image descriptions in corresponding languages. In the second row of Figure~\ref{fig:sec3-vl-multilingual}, we provide the input text prompt in English and specify the output language. \modelname follows the instruction and generates correct descriptions in the desired languages. In the bottom row of Figure~\ref{fig:sec3-vl-multilingual}, we provide an input prompt in Spanish, and ask \modelname to generate image descriptions in 20 different languages. We observe that \modelname can process both the input and output text in different languages. 

\begin{figure*}[h!]
\centering
\vspace{-15mm}
\centerline{\includegraphics[width=1.2\textwidth]{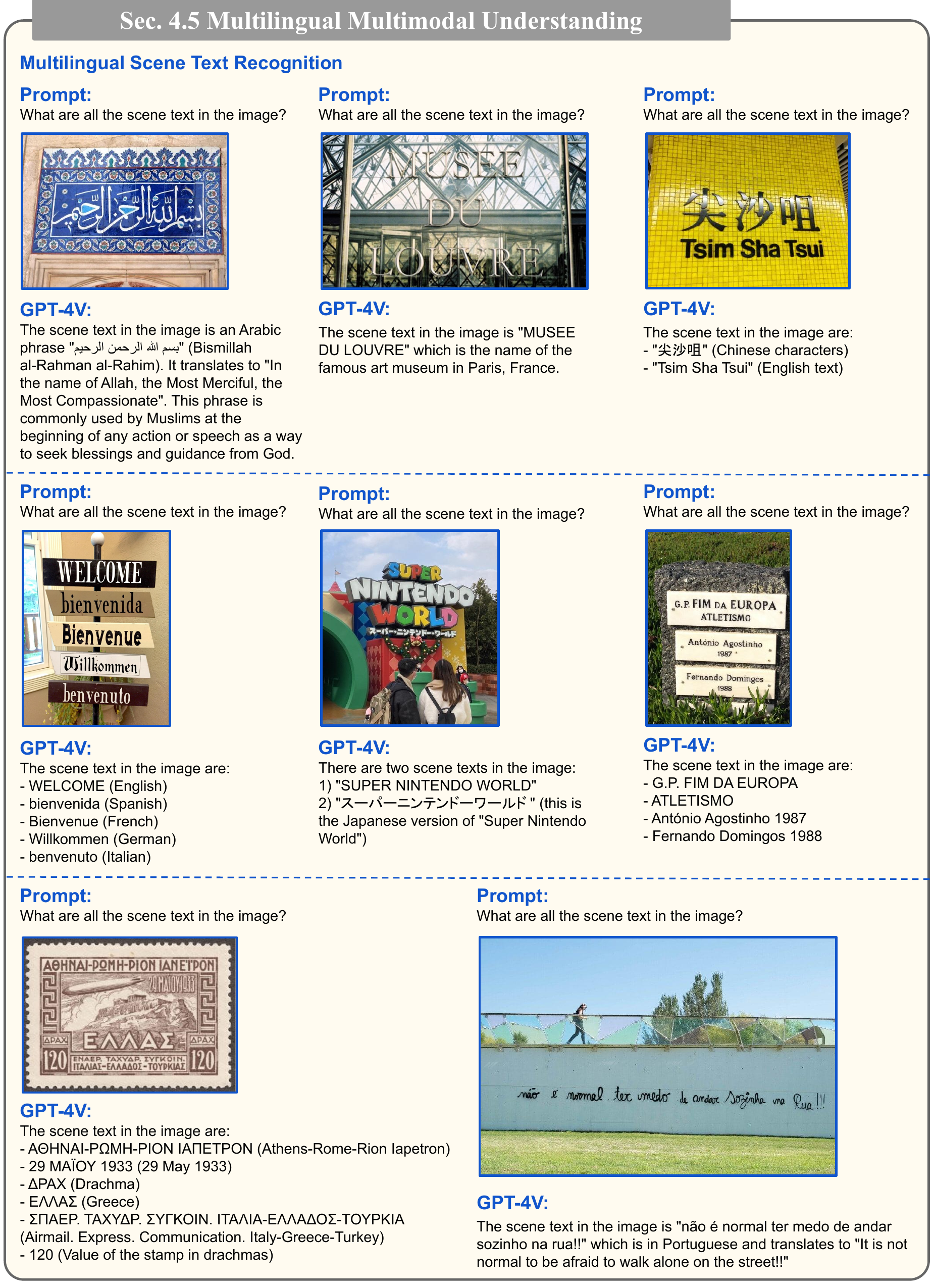}}
\caption[Section~\ref{sec:multilingual}: multilingual scene text recognition.]{Results on multilingual scene text recognition. \modelname~can recognize scene text in different languages. Check Section~\ref{sec:multilingual} for detailed discussions.  
}
\label{fig:sec3-vl-scenetext-multilingual}
\end{figure*}

\begin{figure*}[h!]
\centering
\vspace{-15mm}
\centerline{\includegraphics[width=1.2\textwidth]{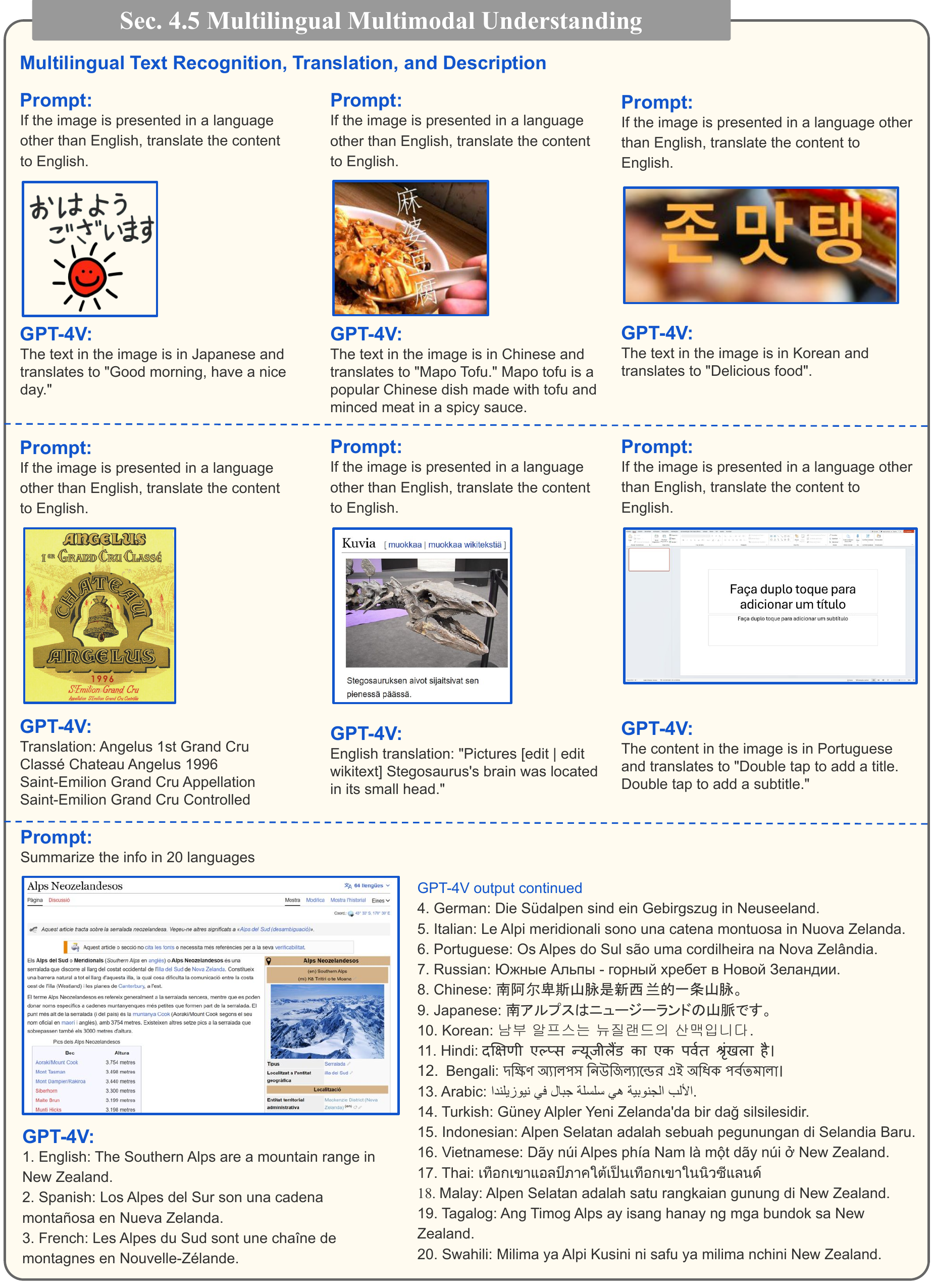}}
\caption[Section~\ref{sec:multilingual}: multilingual text recognition, translation, and description reasoning.]{Results on multilingual text recognition, translation, and description. \modelname is able to recognize, translate and generate descriptions in different languages. Check Section~\ref{sec:multilingual} for detailed discussions.
}
\label{fig:sec3-vl-multilingual-trans}
\end{figure*}

Furthermore, we explore a scenario involving multilingual scene text recognition, where the input image may contain scene text in various languages. As shown in Figure~\ref{fig:sec3-vl-scenetext-multilingual}, \modelname correctly identifies and understands the scene text from different scenes. 
As shown in the first two rows of Figure~\ref{fig:sec3-vl-multilingual-trans}, we observe that \modelname can recognize the scene text, and translate it to a different language. In the bottom row of Figure~\ref{fig:sec3-vl-multilingual-trans}, we provide a screenshot of a Wikipedia website which is in Catalan, and instruct \modelname to summarize the information in 20 different languages. \modelname not only recognizes the text in Catalan but also generates precise summaries and translates them into different languages. This showcases \modelname's ability to comprehend and translate multilingual scene text.

\begin{figure*}[h!]
\centering
\vspace{-12mm}
\includegraphics[width=\textwidth]{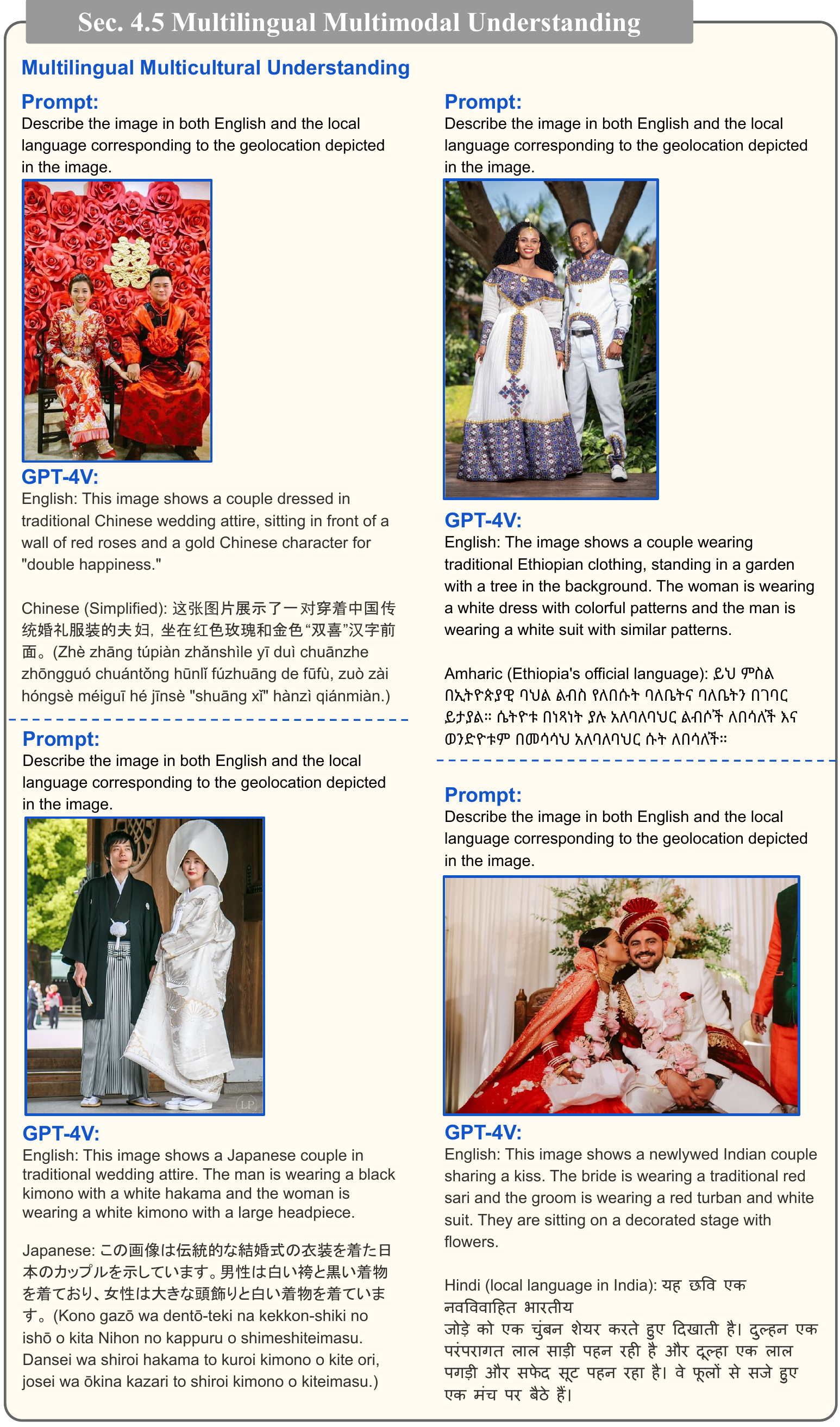}
\caption[Section~\ref{sec:multilingual}: multilingual multiculture understanding.]{Results on multilingual multiculture understanding. Check Section~\ref{sec:multilingual} for detailed discussions. 
}
\label{fig:sec3-vl-culture}
\end{figure*}

We also explore the capability of multicultural understanding~\cite{yin2021broaden,liu-etal-2021-visually}. Figure~\ref{fig:sec3-vl-culture} shows the example results in such a scenario. We observe that \modelname is capable of understanding cultural nuances and generating reasonable multilingual descriptions for the wedding images given.   

In our exploration, we found that \modelname seamlessly comprehends and correctly generates descriptions in different languages, highlighting its versatility in handling diverse linguistic contexts.

\clearpage
\subsection{Coding Capability with Vision}
\label{sec:04language}
\begin{figure*}[h]
\centering
\includegraphics[width=\textwidth]{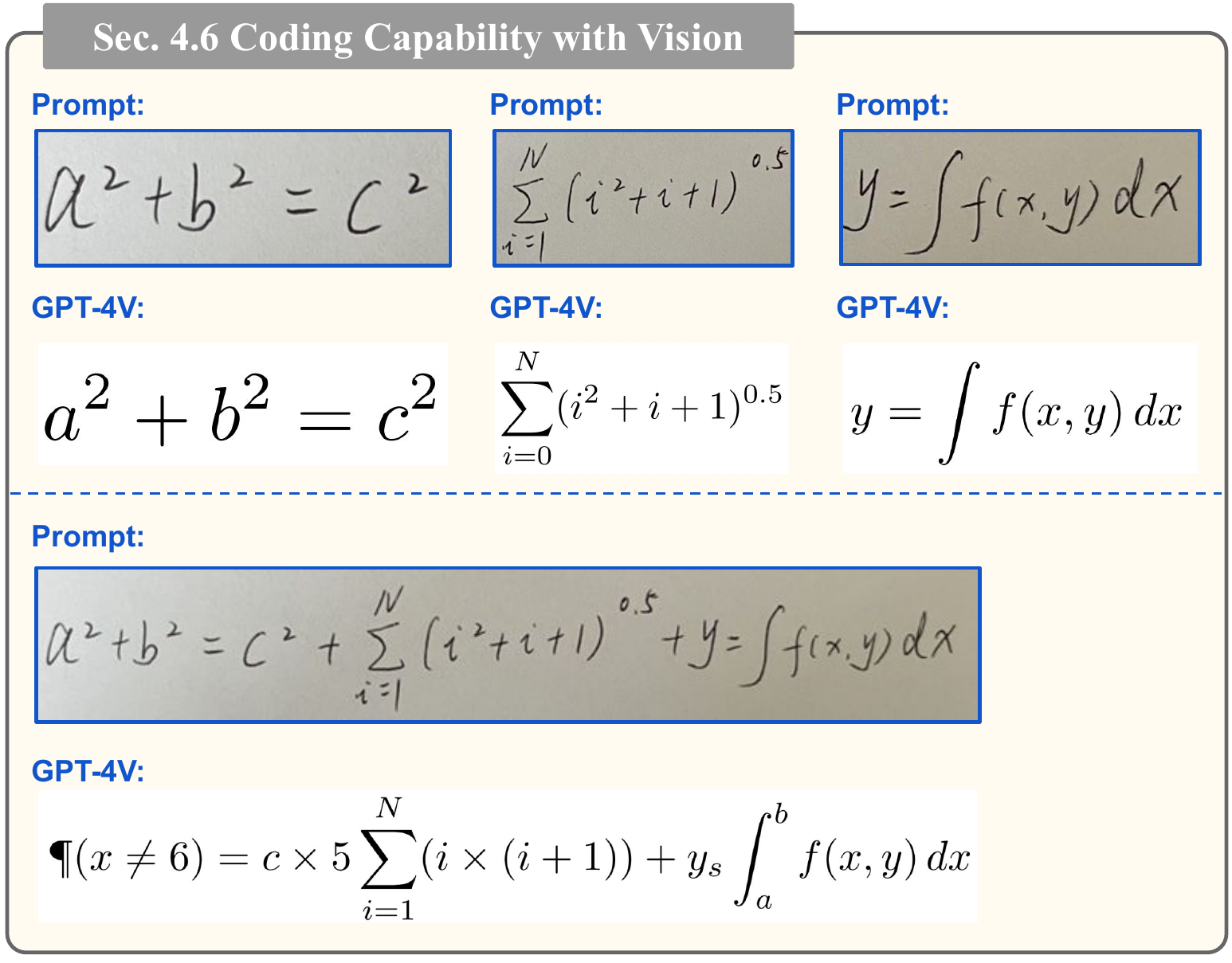}
\caption[Section~\ref{sec:04language}: generate LaTex codes based on the hand-written input.]{
\modelname's capability to generate LaTex codes based on the hand-written input.
The instruction is `generate latex code.' for each case. The output is the LaTeX code and we show the rendered result.
Although the model fails to write the code for the complex equation (bottom),
we can break it down into several simple equations, which \modelname is able to handle.  Check Section~\ref{sec:04language} for detailed discussions.
}
\label{fig:code-latex}
\vspace{15pt}
\end{figure*}

\begin{figure*}[h!]
\centering
\vspace{-10mm}
\includegraphics[width=\textwidth]{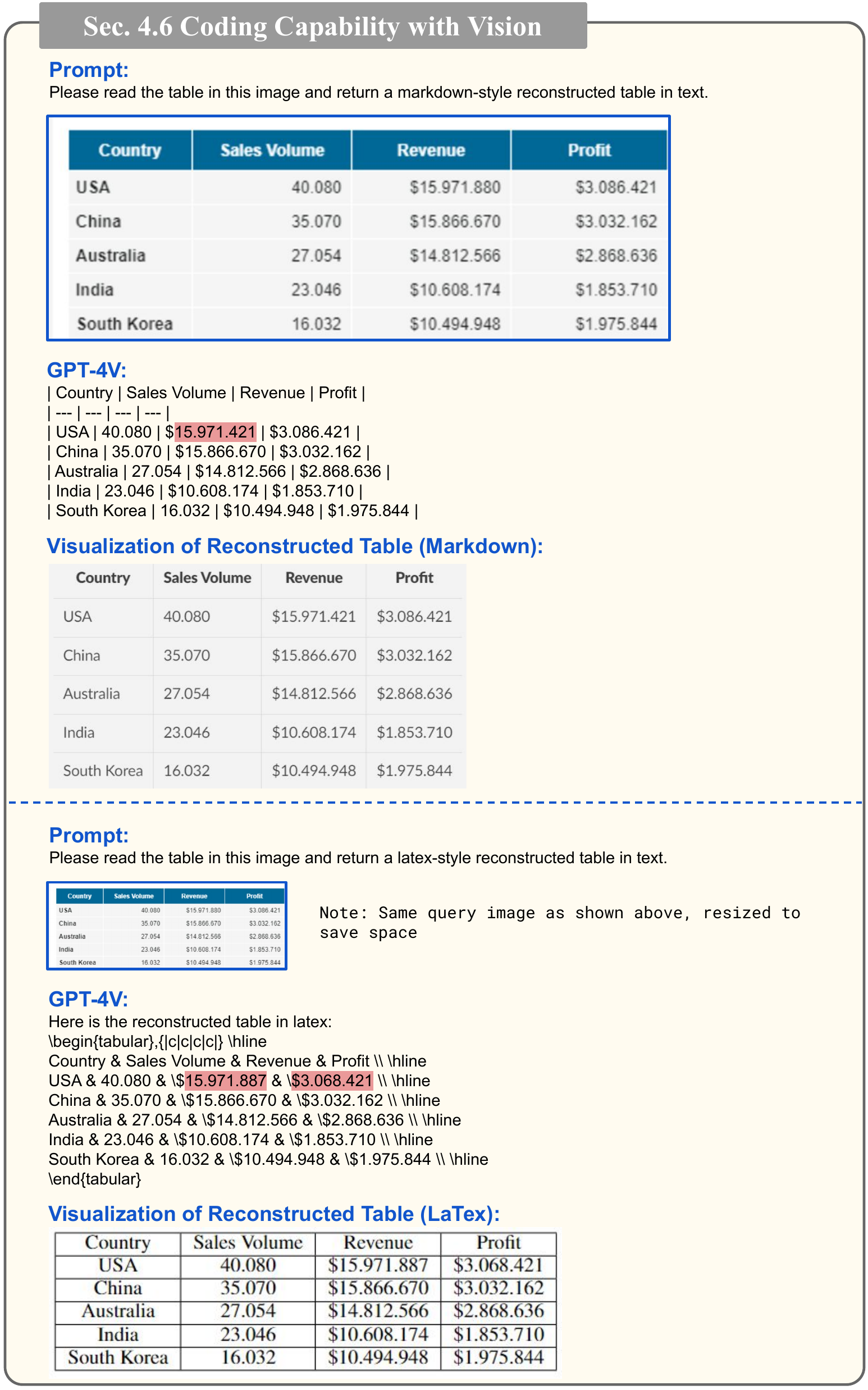}
\caption[Section~\ref{sec:04language}: reconstruct table in Markdown or LaTex code.]{ \modelname's capability to generate Markdown/LaTex codes to reconstruct a table in the image.  \colorbox{redhl}{Red} highlights the errors in reconstruction. Check Section~\ref{sec:04language} for detailed discussions. 
}
\label{fig:code-table}
\end{figure*}

 Figure~\ref{fig:code-latex} illustrates the ability to generate LaTeX code based on handwritten mathematical equations. This functionality can assist users in writing equations in LaTeX more efficiently. 
Although the model is unable to generate code for longer equations, it can handle shorter equations effectively. 
By breaking down longer equations into shorter components, the model is able to generate the appropriate code.  Figure~\ref{fig:code-table} further demonstrates how \modelname can reconstruct a table in the input image into MarkDown/LaTex code.

\begin{figure*}[h]
\centering
\vspace{-18mm}
\includegraphics[width=\textwidth]{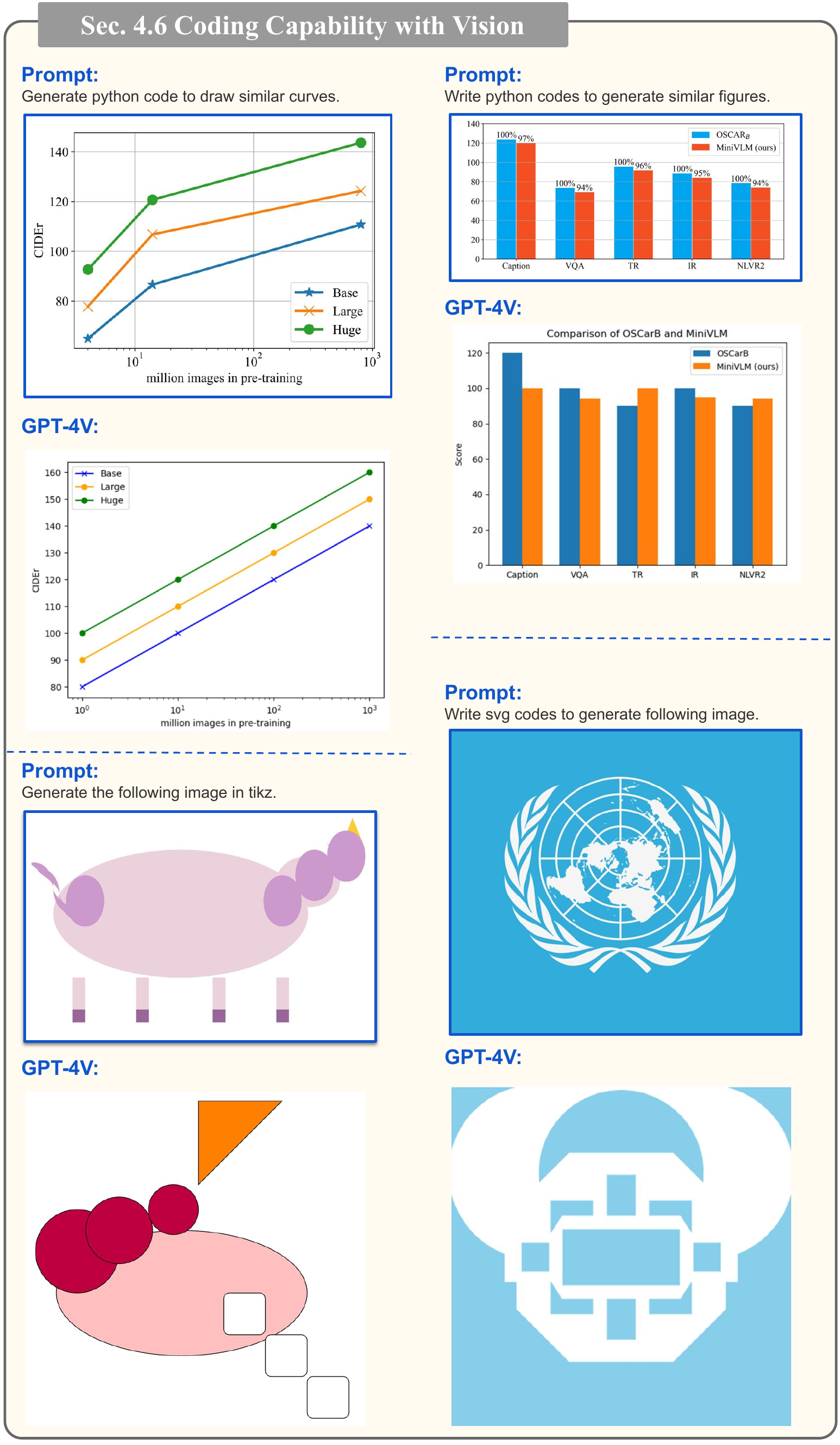}
\caption[Section~\ref{sec:04language}: write codes to replicate the input figure.]{
\modelname's capability to write codes to replicate the input figure. We directly show the rendered figures by python/TikZ/SVG as \modelname's response.
The rendered figure is roughly aligned with the input figure, and the code can be easily adapted.
\modelname~ Chart.  Check Section~\ref{sec:04language} for detailed discussions. 
}
\label{fig:code}
\end{figure*}

Figure~\ref{fig:code} shows examples of writing code in Python, TikZ, and SVG to replicate the input figure.
Although the resulting output is not an exact match, 
the layout is similar and the code can be easily modified to meet specific needs.

\clearpage
\section{Interaction with Humans: Visual Referring Prompting}
\label{sec:05pointing}

Pointing to a specific spatial location is an essential capability in human-computer interaction with multimodal systems, such as conducting visually grounded dialogues. As shown in Section~\ref{sec:point_input}, \modelname~can well understand the visual pointers directly drawn on images. Based on this observation, we propose a novel model interaction method named ``visual referring prompting.'' The core idea is to directly edit image pixel space to draw visual pointers or scene texts as human referring instructions, as highlighted in Figure~\ref{fig:point_input_c}. We detail its usages and advantages in Section~\ref{sec:point_prompt}. Finally, Section~\ref{sec:point_output} explores having \modelname~generate visual pointer outputs to interact with humans. These visual pointers are intuitive for both humans and machines to generate and understand, making them a good channel for human-computer interaction.

\subsection{Understand Pointing Inputs}
\label{sec:point_input}
\begin{figure*}[h!]
\vspace{-15pt}
\centering
\includegraphics[width=\textwidth]{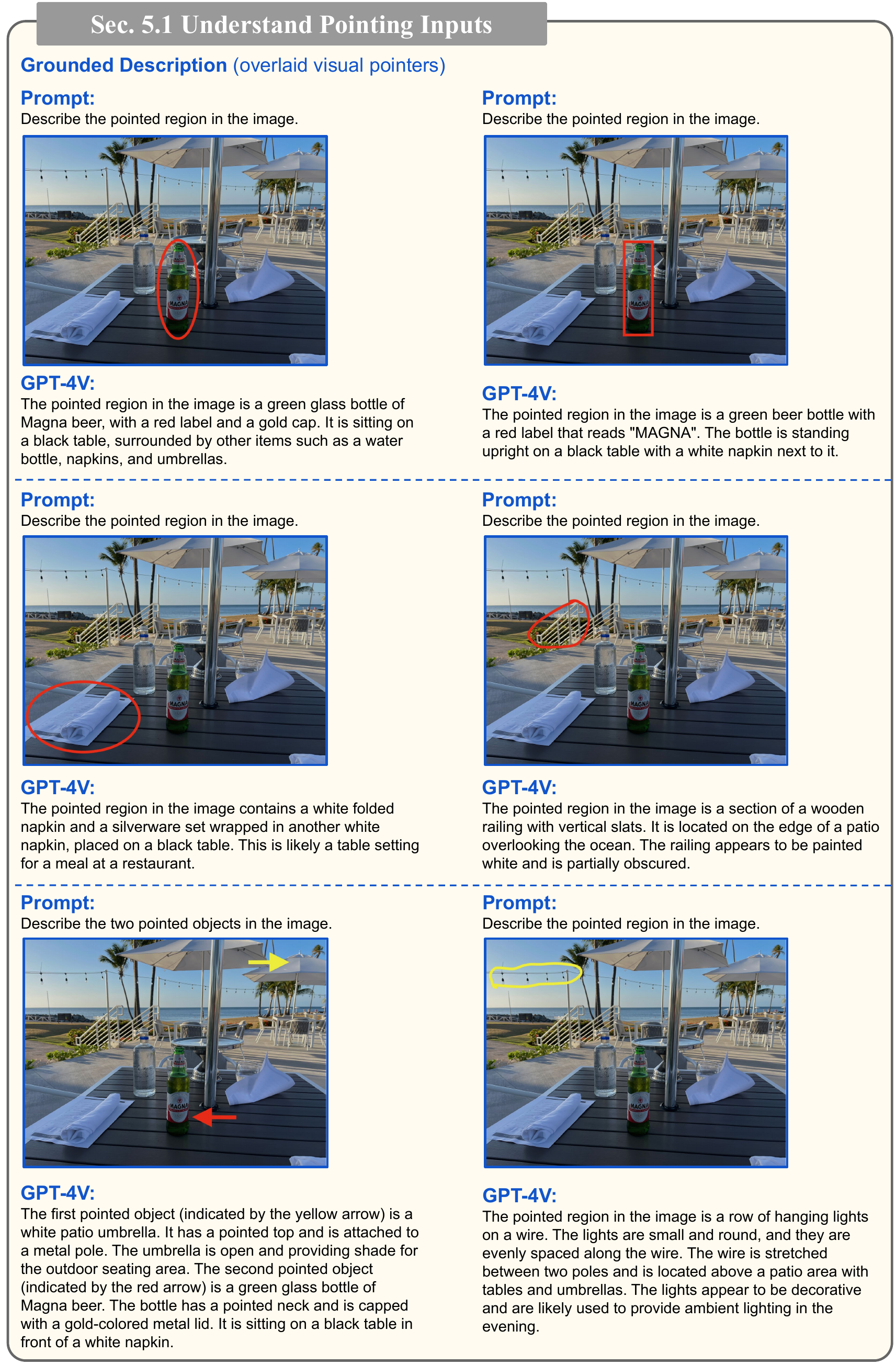}
\caption[Section~\ref{sec:point_input}: understand pointing inputs for grounded description.]{\modelname~understands visual pointers directly overlaid on images. Conducting grounded description with both local and global visual information is one unique application scenario. Check Section~\ref{sec:point_input} for detailed discussions.
}
\label{fig:point_input_a}
\vspace{-18pt}
\end{figure*}

As illustrated in Figure~\ref{fig:point_input_a}, \modelname~can understand different types of visual markers directly overlaid on images as a pointer, such as circles, boxes, and hand drawings. This ability helps \modelname~generate grounded captioning, which is a known challenging problem to have conventional vision-language models~\cite{wang2022git} generating visual descriptions focused on a specific area of interest. Dense captioning methods~\cite{johnson2016densecap,wu2022grit} use cropped boxes or mask regions to generate localized descriptions, but often ignore the global image context and produce sub-optimal descriptions. Visual pointing provides a natural way to indicate the area of interest while maintaining the global image context. For example, the top left example focuses on providing a comprehensive description of the pointed Magna beer, while also mentioning the global image context that the beer bottle is on the table.

\begin{figure*}[h!]
\centering
\vspace{-15mm}
\includegraphics[width=\textwidth]{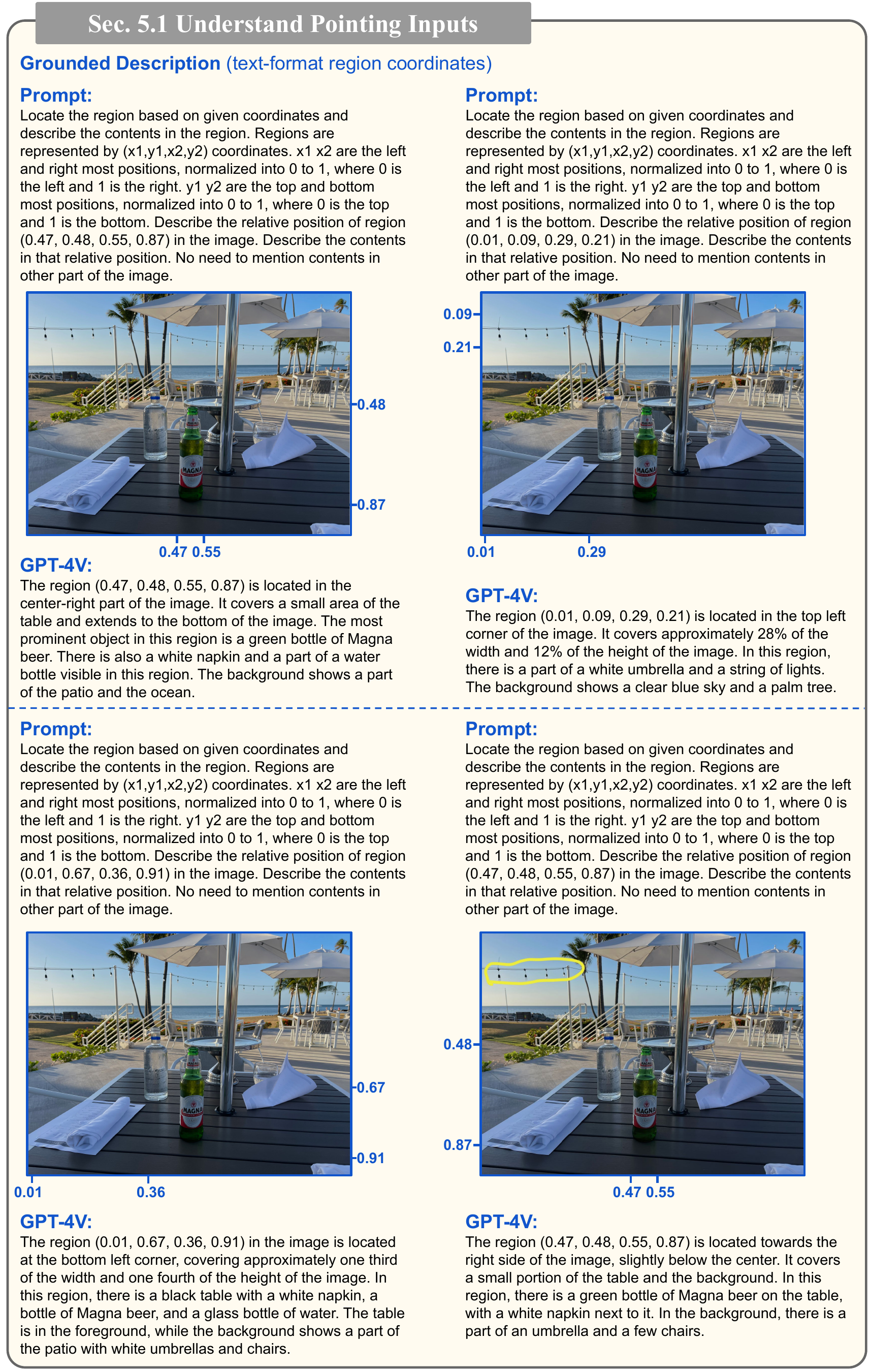}
\caption[Section~\ref{sec:point_input}: grounded description with text-format region coordinates.]{An alternative to visual pointers overlaid on images is the region coordinates represented in the numerical text format. \modelname~can understand the coordinates, \eg, (0.47, 0.48, 0.55, 0.87), (0.01, 0.09, 0.29, 0.21), and (0.01, 0.67, 0.36, 0.91) that correspond to the center beer bottle, top-left string lights, and bottom-left table set, respectively. We observe that \modelname~works less reliably when prompted with text coordinates, compared with visual pointers in visual referring prompting. Check Section~\ref{sec:point_input} for detailed discussions.
}
\label{fig:point_input_b}
\vspace{-36pt}
\end{figure*}

An intuitive alternative to visual pointers overlaid on images is the region coordinates represented in the numerical text format. As shown in Figure~\ref{fig:point_input_b}, \modelname~can understand the coordinates out-of-box, achieving the ability of spatial referring via text tokens without extra box token finetuning as in prior vision-language models~\cite{wang2022ofa,yang2022reco}. Despite the promising capability, we note that our current prompt is less precise spatially. For example, in the top left example in Figure~\ref{fig:point_input_b}, \modelname~mentions the surrounding objects napkin and water bottle, even though only the beer bottle is in the region (0.47, 0.48, 0.55, 0.87). Overall, with respect to our experimented prompts, \modelname~works more reliably when prompted with overlaid visual pointers, compared with text coordinates. This unique capability motivates us to explore a new prompting method, namely visual referring prompting.


\clearpage
\subsection{Visual Referring Prompting}
\label{sec:point_prompt}
\begin{figure*}[h!]
\centering
\includegraphics[width=\textwidth]{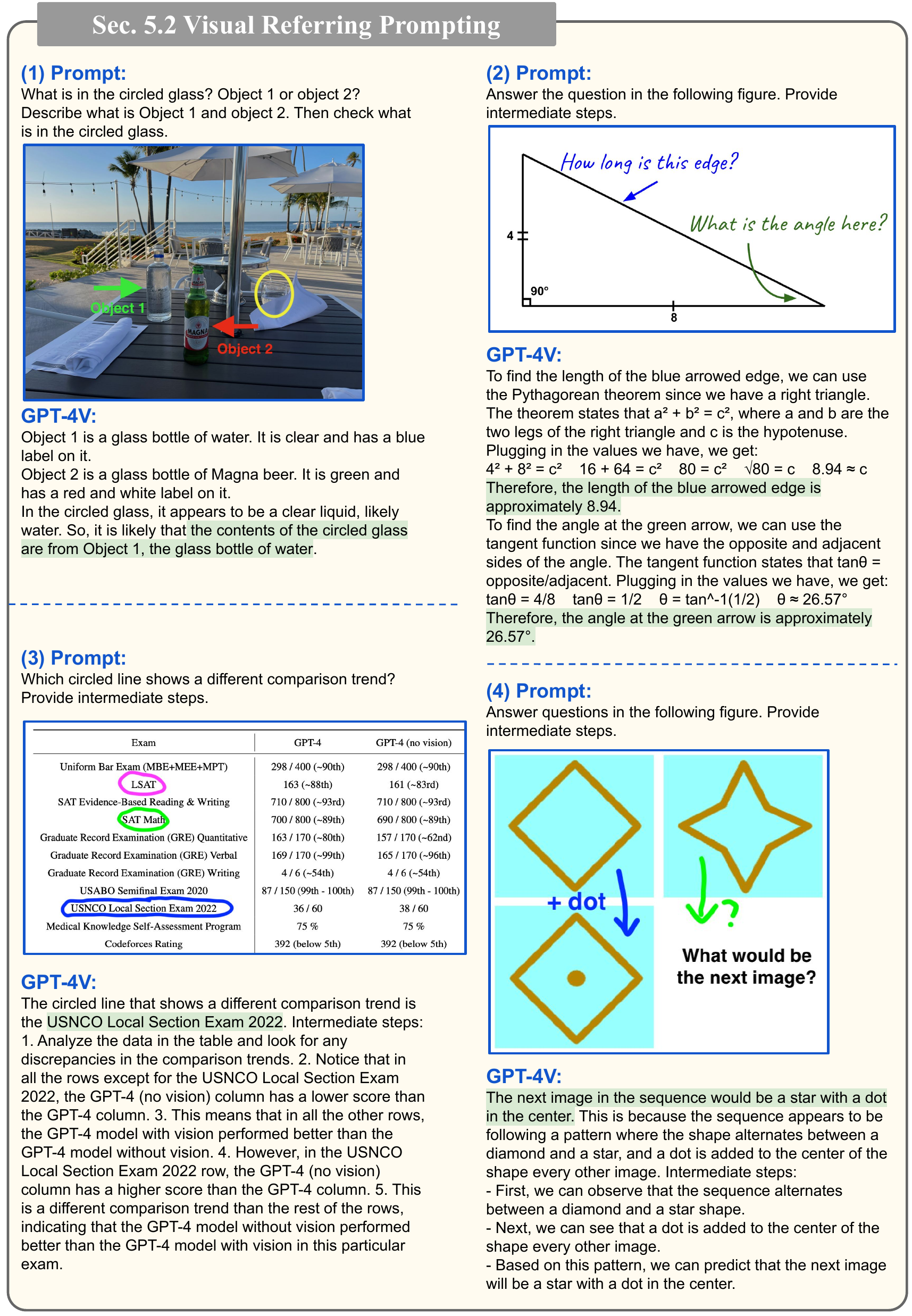}
\caption[Section~\ref{sec:point_prompt}: examples of visual referring prompting.]{Visual referring prompting directly edits the input image as input prompts, such as drawing visual pointers and scene texts. Complementary to text prompts, visual referring prompting provides a more nuanced and natural interaction, \eg, (1) associating pointed objects with an index, (2) pointing to the image for questioning, (3) highlighting lines in documents and tables, (4) drawing the pattern on the image, and many other novel use cases. Check Section~\ref{sec:point_prompt} for detailed discussions.
}
\label{fig:point_input_c}
\end{figure*}

\begin{figure*}[h!]
\centering
\includegraphics[width=\textwidth]{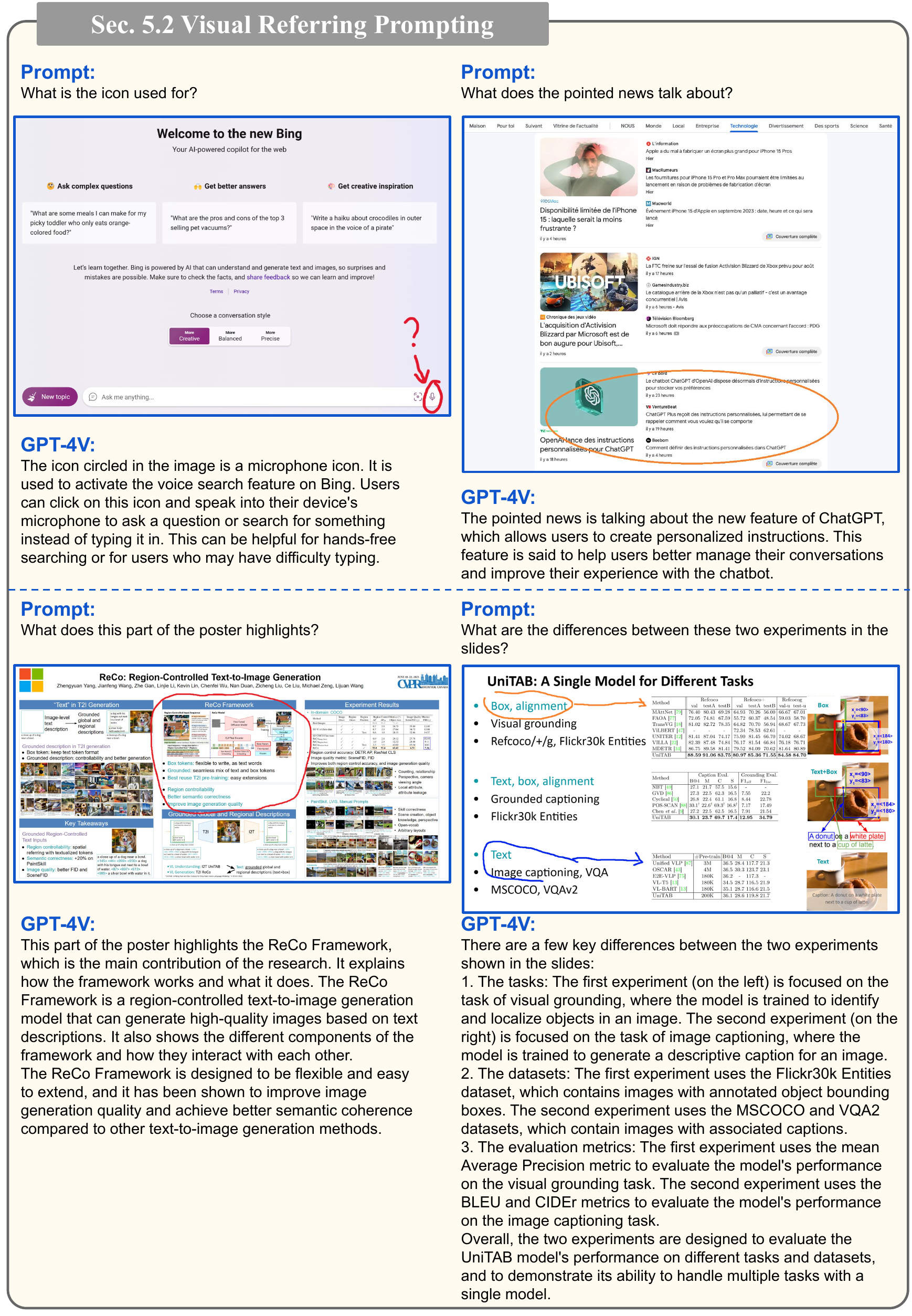}
\caption[Section~\ref{sec:point_prompt}: examples of visual referring prompting.]{Visual referring prompts enhance the seamless interaction between humans and computers. This is evident in the integration with computer and mobile Graphical User Interfaces (GUIs), and the support provided in understanding documents and slides. Check Section~\ref{sec:point_prompt} for detailed discussions.
}
\label{fig:point_input_c_1}
\end{figure*}

Inspired by \modelname's strong capability in understanding visual pointing and scene text, we explore a new method to interact with \modelname, namely the \textit{visual referring prompting}. Instead of conventional prompting techniques that edit text space, visual referring prompting is a complementary technique that directly edits the pixel space for input images for human-computer interaction. Such visual prompting could offer a more nuanced and comprehensive interaction with the image, potentially unlocking a wider array of responses from the model. For example, in Figure~\ref{fig:point_input_c} (1), \modelname~naturally associates the arrow-pointed objects with the given object indexes, easing the remaining visual reasoning and text outputs; in (2), \modelname~understands the questions written on the image and pointed to the corresponding edge or angle, providing a nuanced interface for grounded visual dialogue; in (3), humans can point to arbitrary regions inside the figure to help \modelname~better understand complicated documents and charts; in (4), the pattern can be concisely represented as an arrow and the scene text ``+dot'', therefore helping \modelname~to predict the next image.  Complementary to text prompts that are loosely grounded to images, visual referring prompting provides a novel interaction method that could facilitate various use cases, with additional demonstrations in Figure~\ref{fig:point_input_c_1} and Section~\ref{sec:10app}.

\subsection{Generate Pointing Outputs}
\label{sec:point_output}
Section~\ref{sec:point_input} discusses the ability of \modelname~to understand visual pointing generated by humans. A natural question is: Can \modelname~generate its own pointing outputs, thereby facilitating a closed-loop interaction process in human-computer interaction?

\begin{figure*}[h!]
\centering
\vspace{-20mm}
\includegraphics[width=\textwidth]{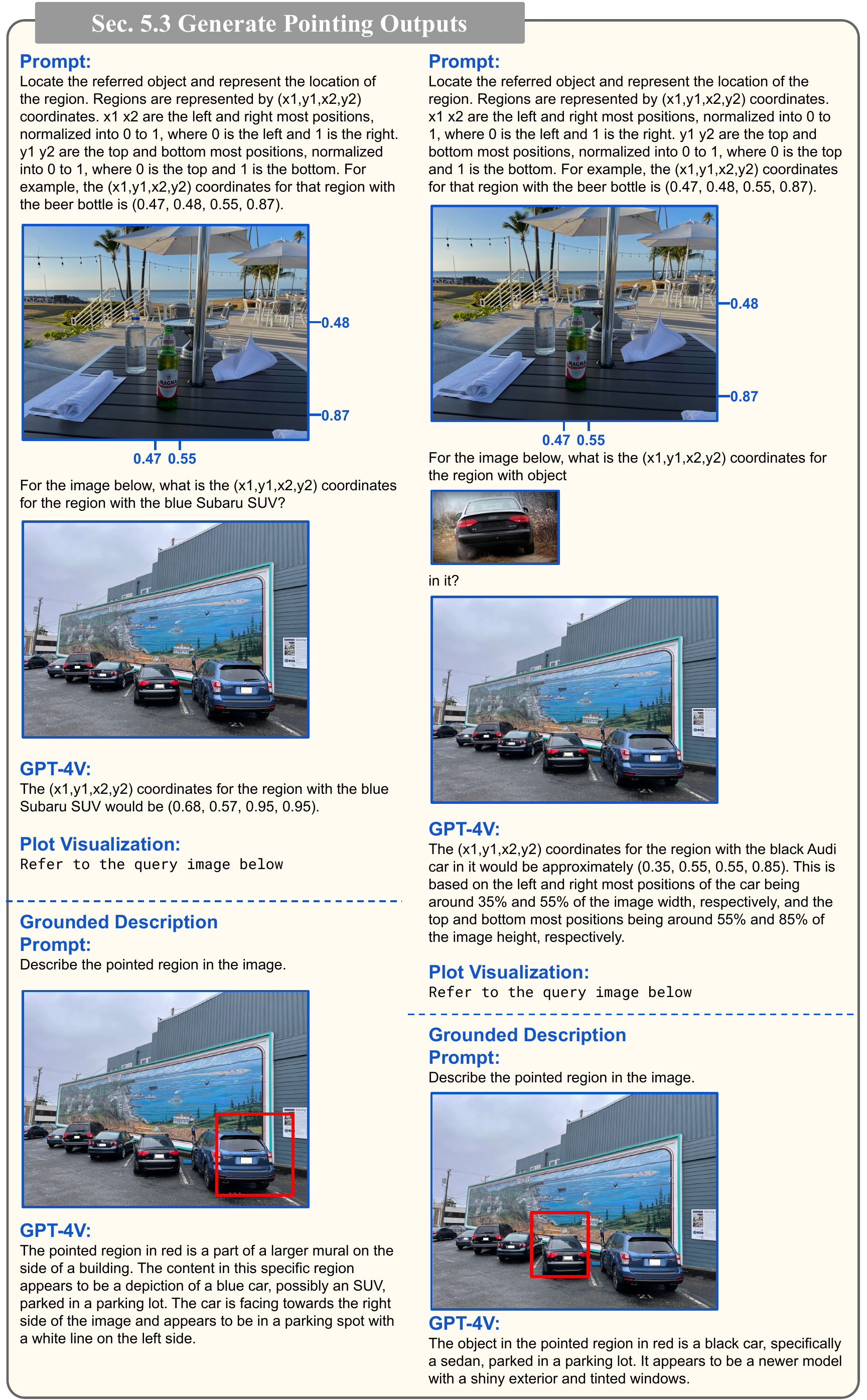}
\caption[Section~\ref{sec:point_output}: generate pointing outputs.]{\modelname~can use its understanding of coordinates to generate visual pointing output, thereby grounding the textual or visually queried object. Using example-grounded instructions can help \modelname~understand coordinate definitions and therefore generate better pointing. While output spatial regions are not precise, the approach enables an ``understanding (\ie, grounded description) and generation'' loop for visual pointing, leading to an effective way of human-computer interaction. Check Section~\ref{sec:point_output} for detailed discussions.
}
\label{fig:point_output_a}
\vspace{-36pt}
\end{figure*}

Figure~\ref{fig:point_output_a} explores generating visual pointing outputs by letting \modelname~predict region coordinates in the text format. We prompt \modelname~to ground the object referred by text (\eg, the text of ``blue Subaru SUV'') or a reference image (\eg, the image of ``black Audi sedan''). Similar to the observation in having \modelname~comprehend coordinates input, the model has a coarse understanding of spatial locations, but it wasn't accurate with respect to the prompts used in the experiment. For example, in Figure~\ref{fig:point_output_a}'s ``plot visualizations,'' \modelname~can approximately identify the blue SUV and black sedan mentioned in the query, but it struggles to create a closely-fitted bounding box. We observe that including example-grounded instructions in the prompt helps \modelname~to understand the definition of coordinates and subsequently generate better pointing outputs.

While the generated pointing outputs may not perfectly cover the queried region, they still provide a valuable tool for model interaction, interpretation, and helping multi-step visual reasoning. Specifically, the pointing outputs can be interpreted by humans to better understand \modelname's references, or by \modelname~itself to enable further reasoning based on previous outputs.
As shown in the bottom of Figure~\ref{fig:point_output_a}, \modelname~is capable of interpreting the pointers it generates, and providing grounded descriptions with the prompts in Figure~\ref{fig:point_input_a}. Furthermore, the iterative pointing generation and understanding by \modelname~itself can help complicated multi-hop grounded visual reasoning tasks. \modelname~is capable of deconstructing the question, generating distinct visual markers to iteratively focus on different image regions for each sub-step, ultimately collating the information to formulate the final answer. %


\clearpage
\section{Temporal and Video Understanding}
\label{sec:07temporal}

In this section, we discuss temporal and video understanding capabilities. Even though \modelname~operates primarily on images as inputs, evaluating its understanding of temporal sequences and video content remains a crucial aspect of its overall assessment. This is because real-world events unfold over time, and an AI system's ability to understand these dynamic processes is instrumental in real-world applications. Capabilities like temporal anticipation, temporal ordering, temporal localization, temporal reasoning, and grounded temporal understanding help to gauge the model's proficiency in comprehending the sequence of events, anticipating future occurrences, and contextually analyzing activities over time, all within a series of static images. In spite of its image-centric focus, \modelname~is able to comprehend video and temporal sequences in a way that's similar to human comprehension. To enhance the versatility and applicability of a sophisticated AI model like \modelname, this aspect of testing is critical to its development and refinement. For the upcoming experiments in this section, we will use multiple selected video frames as inputs to test the model's abilities in understanding temporal sequences and video content. 

\subsection{Multi-image Sequencing}
\label{sec:temporal-01}
\begin{figure*}[h!]
\centering
\includegraphics[width=\textwidth]{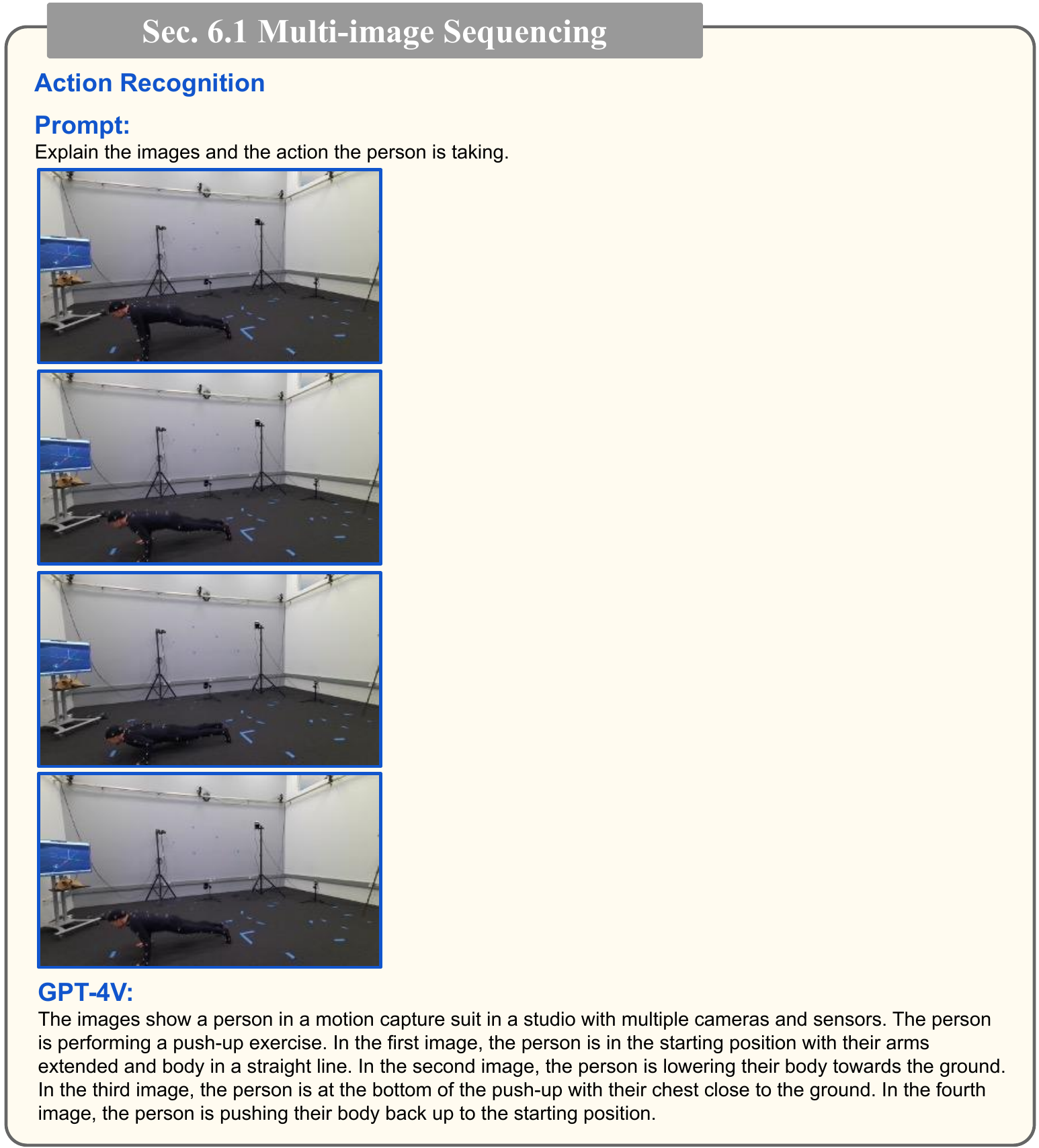}
\caption[Section~\ref{sec:temporal-01}: video frames understanding.]{Sequences of video frames understanding: Interpreting human poses and deriving relevant insights from video sequences. Check Section~\ref{sec:temporal-01} for detailed discussions.
}
\label{fig:temp_push_up}
\end{figure*}

In this subsection, we demonstrate that \modelname~can accurately comprehend and analyze sequences of video frames. Within this frame-by-frame analysis, \modelname~recognizes the scene in which the activity is taking place, delivering a deeper contextual understanding. As shown in Figure \ref{fig:temp_push_up}, the model is not just confined to recognizing the environment; it also accurately interprets the actions being performed by individuals in the video. \modelname~understands the sequence and context of various human poses and intelligently correlates them with the ongoing activity. By understanding pose variations beyond just identification, \modelname~can derive meaning from the subtleties of human movement and action. As a result of this level of detailed understanding, \modelname~can capture the essence of what's happening in videos, offering rich and nuanced insights that go beyond just identifying objects and scenes. 

\subsection{Video Understanding}
\label{sec:temporal-02}
\begin{figure*}[h!]
\centering
\includegraphics[width=\textwidth]{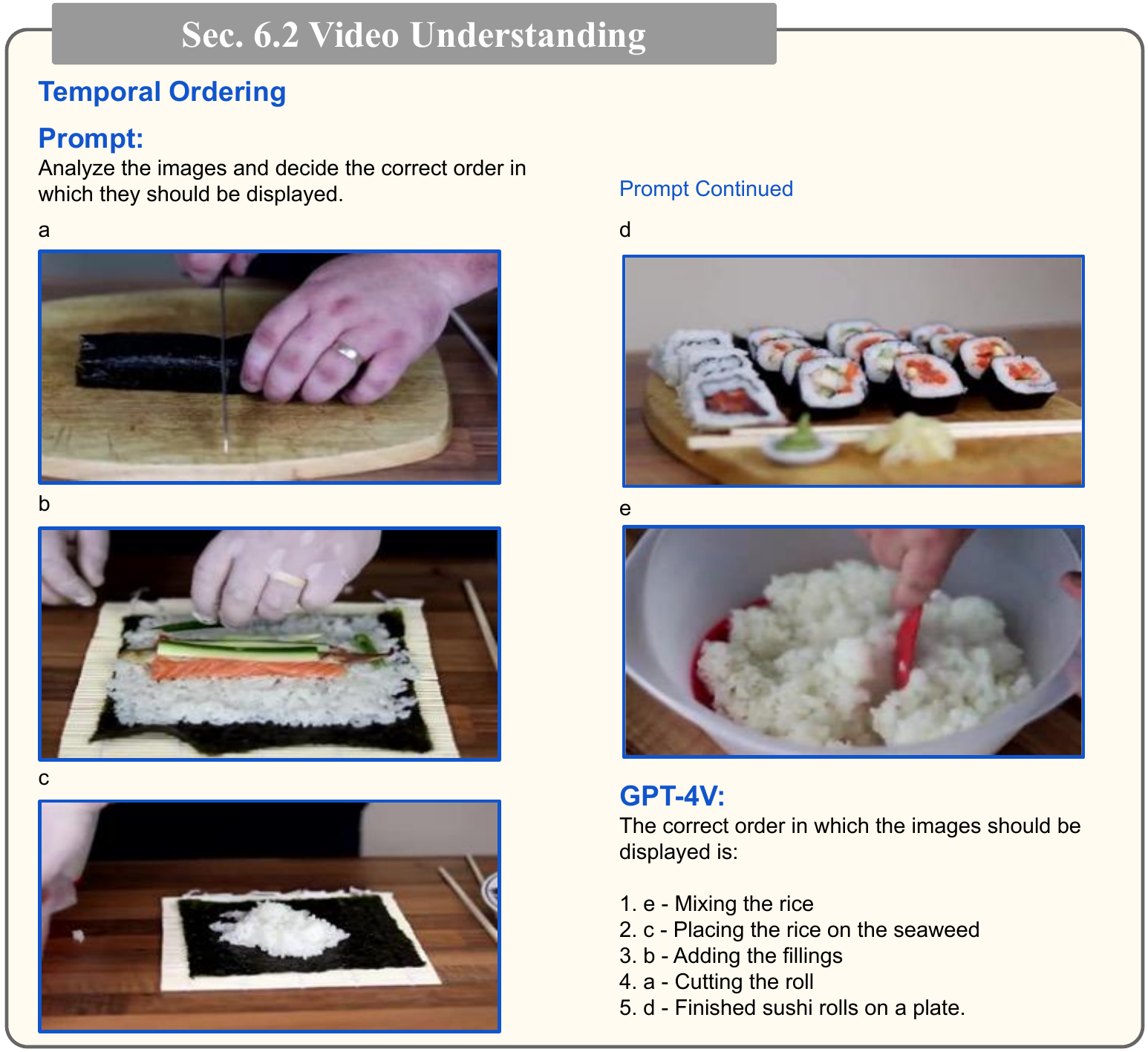}
\caption[Section~\ref{sec:temporal-02}: long-term temporal order reasoning.]{Long-term temporal ordering: \modelname~is presented with shuffled image frames depicting a sushi-making event. While the sushi-making process is disordered, \modelname~is able to identify the event and determine the correct temporal sequence. Check Section~\ref{sec:temporal-02} for detailed discussions.
}
\label{fig:temp_order_sushi}
\end{figure*}

\begin{figure*}[h!]
\centering
\includegraphics[width=\textwidth]{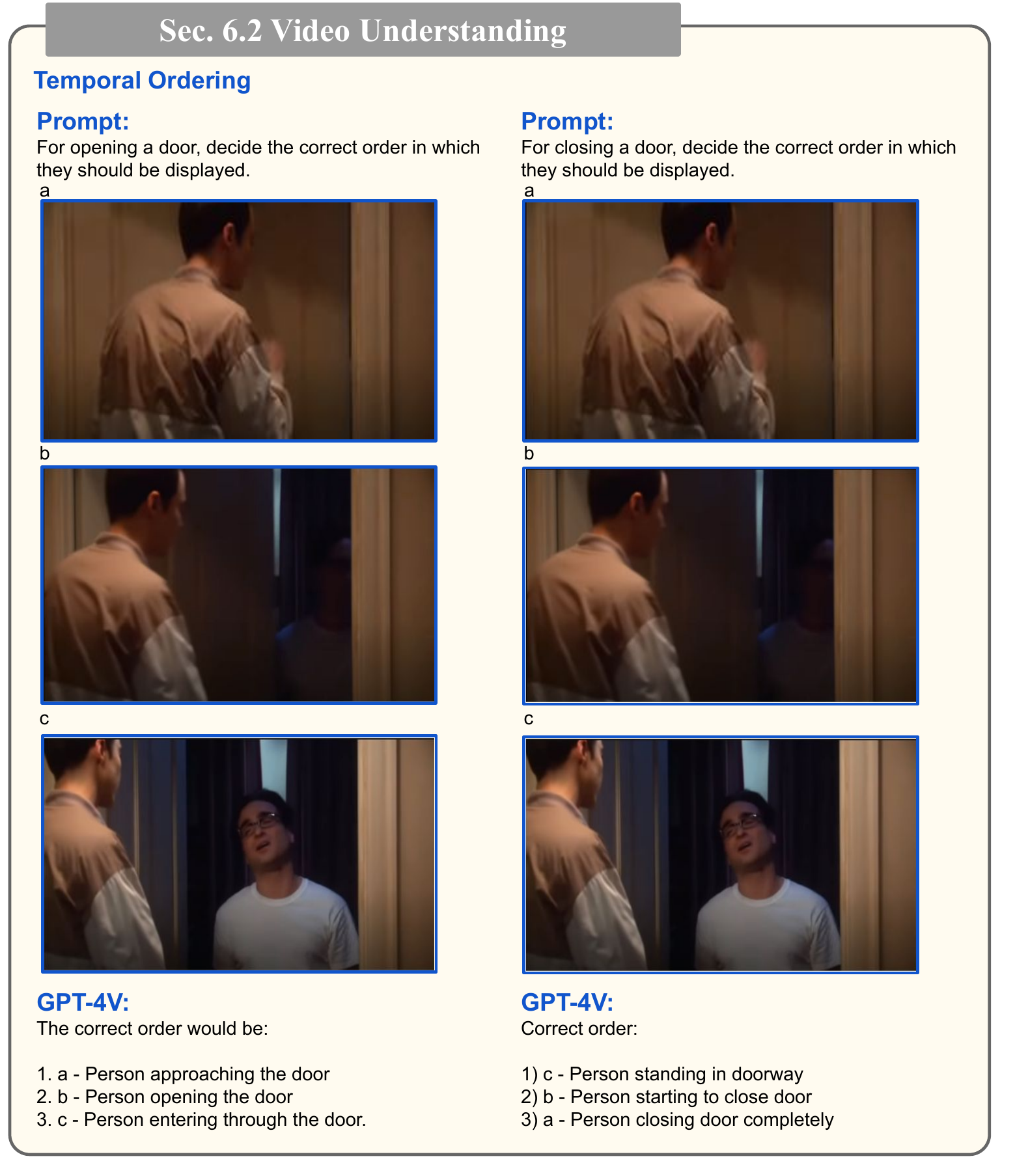}
\caption[Section~\ref{sec:temporal-02}: short-term temporal order reasoning.]{Short-term temporal ordering: given a specified action, such as opening or closing a door, \modelname~demonstrates its capability to comprehend the images' content and determine the correct sequential order corresponding to the specified action. Check Section~\ref{sec:temporal-02} for detailed discussions.
}
\label{fig:temp_order_door}
\end{figure*}

\begin{figure*}[h!]
\centering
\includegraphics[width=\textwidth]{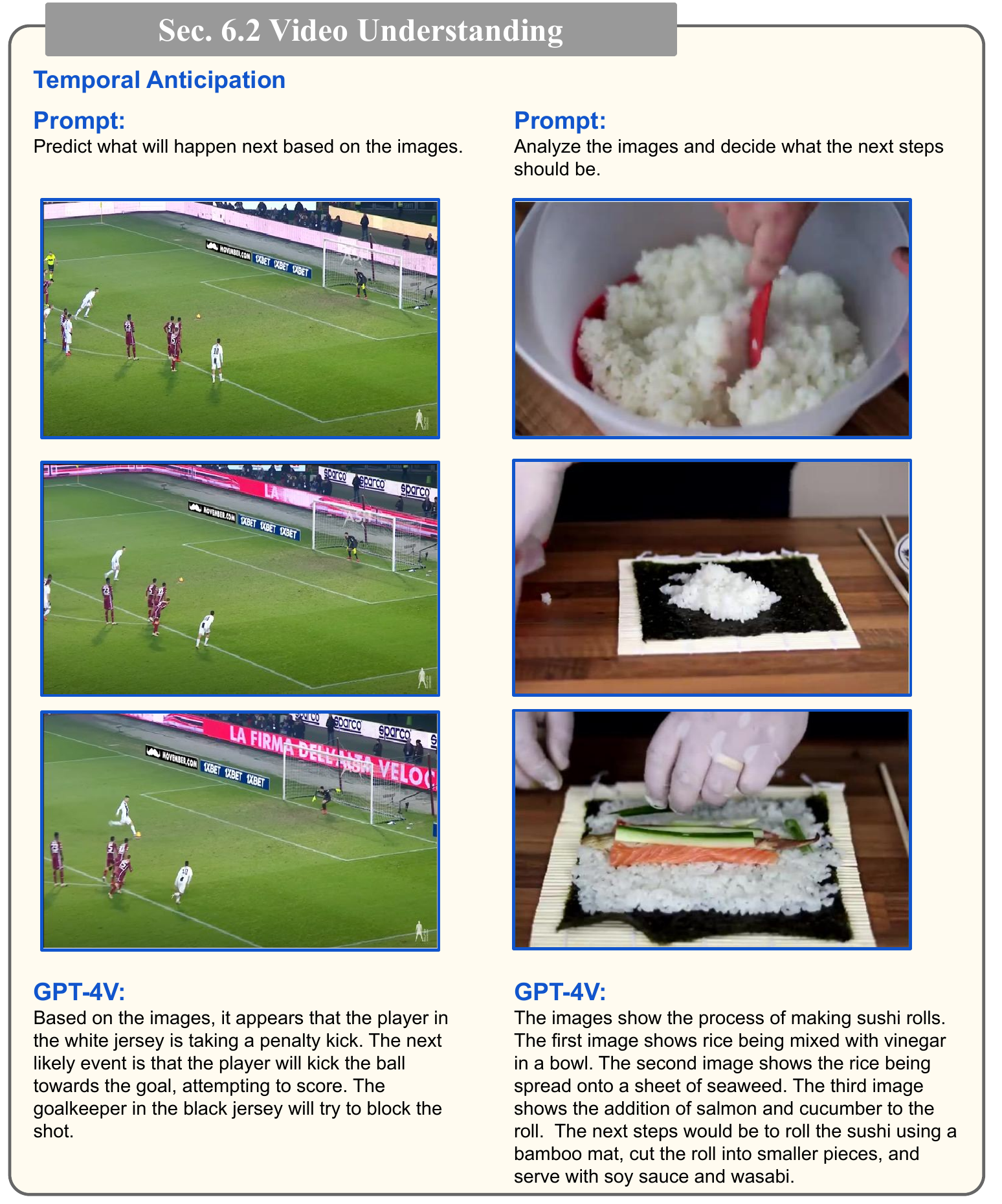}
\caption[Section~\ref{sec:temporal-02}: short-term and long-term temporal anticipation.]{Short-term and long-term temporal anticipation: \modelname~captures and understands activities with varying temporal structures and complexities. Check Section~\ref{sec:temporal-02} for detailed discussions.
}
\label{fig:temp_predict}
\end{figure*}

\paragraph{Temporal ordering.} 
Temporal ordering is a crucial element of temporal commonsense and forms an essential part of \modelname's capabilities evaluation. This involves providing the model with a series of shuffled images and gauging its ability to discern cause and effect relationships as well as time progressions. An understanding of such relationships requires the ability to reorder the sequence in a logically coherent and temporally accurate manner. Figure \ref{fig:temp_order_sushi} illustrates an example of long-term temporal ordering where \modelname~is presented with a series of shuffled image frames depicting a sushi-making event. Despite the disorder, \modelname~effectively identifies the event and determines the appropriate temporal sequence of the sushi-making process. In addition, Figure \ref{fig:temp_order_door} provides an example of short-term temporal ordering. Given a designated action, such as opening or closing a door, \modelname~demonstrates its capability to comprehend the image's content and determine the correct sequential order of the events. These examples highlight \modelname's capability in temporal commonsense, reinforcing its ability to comprehend both long-term and short-term sequences accurately.
\paragraph{Temporal anticipation.} 
We demonstrate \modelname's ability to anticipate future events given a set of initial frames. Long- and short-term examples are used to validate this capacity for anticipating future events. The right side of Figure \ref{fig:temp_predict} illustrates \modelname's ability to anticipate short-term events with a soccer penalty kick example. Given the first few frames, it accurately foresees the typical next actions of both the kicker and the goalkeeper, due to its understanding of the inherent structure and rules of the game. In addition, as shown in The left side of Figure \ref{fig:temp_predict}, the sushi preparation sequence illustrates \modelname's long-term anticipation capability. By understanding the activity based on visual cues, \modelname~not only recognizes the current progress in sushi preparation but also accurately anticipates the subsequent steps, demonstrating its capacity to interpret and predict complex, multi-step processes over an extended period. This combination of short-term and long-term temporal anticipation allows \modelname~to capture and understand activities with varying temporal structures and complexities.

\begin{figure*}[h!]
\centering
\includegraphics[width=\textwidth]{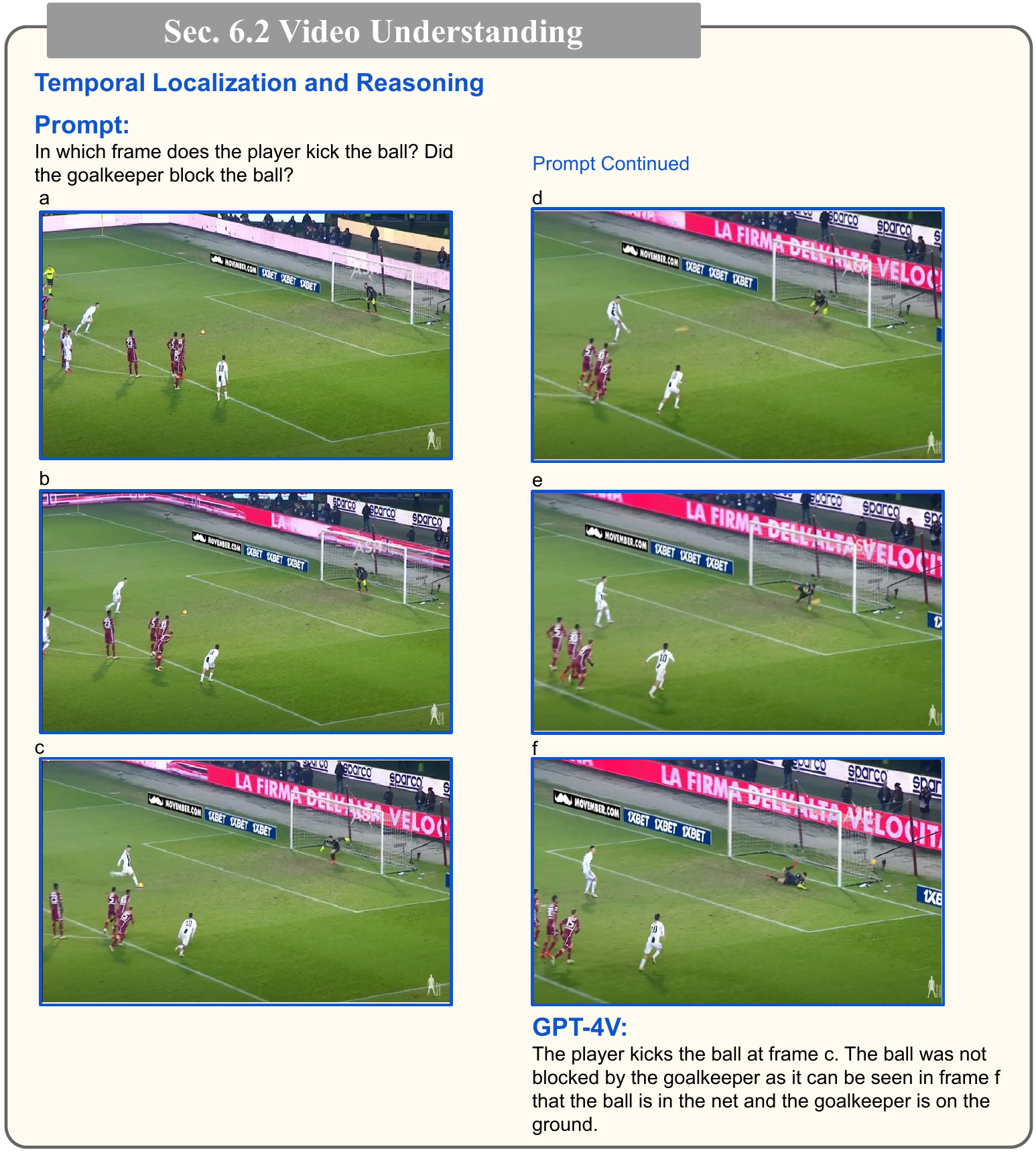}
\caption[Section~\ref{sec:temporal-02}: temporal localization and reasoning.]{Temporal localization and reasoning: \modelname~shows the capability in temporal localization by accurately identifying when the player strikes the ball. It also demonstrates cause-and-effect reasoning by determining whether the ball was blocked based on the goalkeeper-ball interaction. Check Section~\ref{sec:temporal-02} for detailed discussions.
}
\label{fig:temp_localization}
\end{figure*}

\paragraph{Temporal localization and reasoning.}
Figure \ref{fig:temp_localization} illustrates \modelname's capabilities in temporal localization and reasoning. It accurately identifies the precise moment the player strikes the ball. Furthermore, \modelname~showcases its understanding of cause and effect by inferring from the relationship between the goalkeeper and the ball to determine if the goalkeeper successfully blocks the ball.
In the context of the example given, understanding whether the goalkeeper can block the ball involves not only recognizing the spatial positions of the goalkeeper and the ball but also understanding the dynamics of their interaction and predicting the outcome of these dynamics. This demonstrates a considerable level of sophistication in the model's reasoning abilities.
\clearpage
\begin{figure*}[h!]
\vspace{-15mm}
\centering
\includegraphics[width=\textwidth]{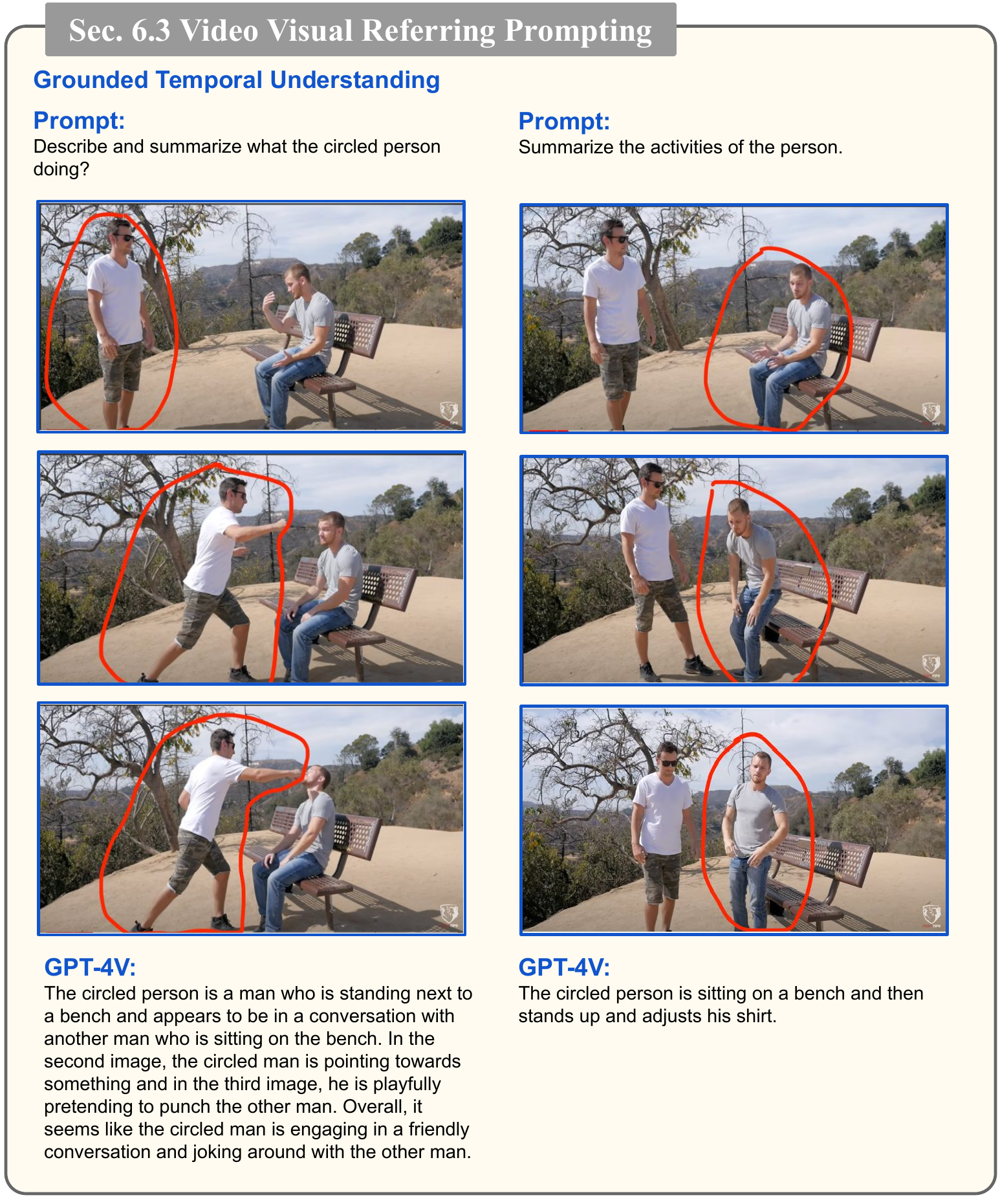}
\caption[Section~\ref{sec:temporal-03}: grounded temporal understanding.]{Grounded temporal understanding: \modelname~can apply a temporal understanding to a specific person of interest, indicated by a circle. Check Section~\ref{sec:temporal-03} for detailed discussions.
}
\vspace{-10pt}
\label{fig:temp_pointing}
\vspace{-5pt}
\end{figure*}

\subsection{Visual Referring Prompting for Grounded Temporal Understanding}
\label{sec:temporal-03}
Section \ref{sec:05pointing} illustrates \modelname's capabilities in visual referring prompting. In this section, we aim to extend this capability by testing visual referring prompting for temporal understanding. This advancement offers enhanced control over video comprehension tasks.

\noindent\textbf{Grounded temporal understanding.}
Grounded temporal understanding forms another crucial aspect of \modelname's capabilities, which we explore using pointing input in a sequence of image frames. Figure~\ref{fig:temp_pointing} exemplifies this by demonstrating how \modelname~can apply a temporal understanding to a specific person of interest, indicated by a circle. \modelname~can accurately describe events in a way that aligns with the corresponding temporal order, focusing on the activities of the circled individual. Beyond this, \modelname~demonstrates a more refined understanding of the event, recognizing the nature of the interactions. For instance, \modelname~can distinguish between friendly interactions and violent incidents, illustrating an ability to not only comprehend the temporal flow of events but also to interpret the tone and nature of the interactions taking place. This indicates \modelname's capacity to process and comprehend complex temporal and social cues within a given sequence, adding a layer of depth to its understanding.
\clearpage
\section{Abstract Visual Reasoning and Intelligence Quotient Test}
\label{sec:08iq}
\begin{figure*}[h!]
\centering
\vspace{-15mm}
\includegraphics[width=\textwidth]{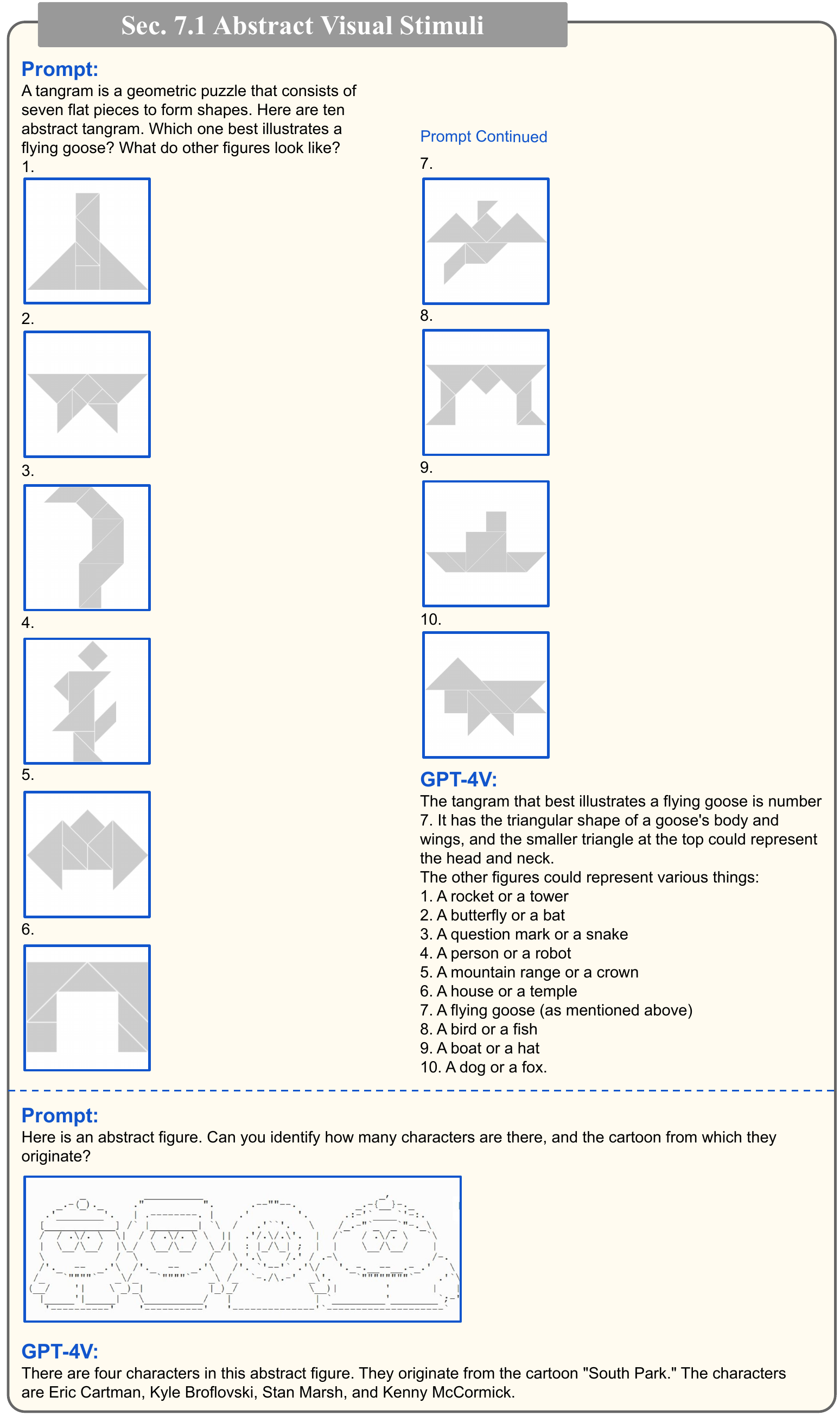}
\caption[Section~\ref{sec:iq-01}: abstract visual stimuli.]{Understanding abstract visual stimuli such as tangram~\cite{ji2022abstract} and ASCII text art. Check Section~\ref{sec:iq-01} for detailed discussions.
}
\label{fig:spatial_abstract_a}
\vspace{-20pt}
\end{figure*}
\begin{figure*}[h!]
\centering
\vspace{-20mm}
\includegraphics[width=\textwidth]{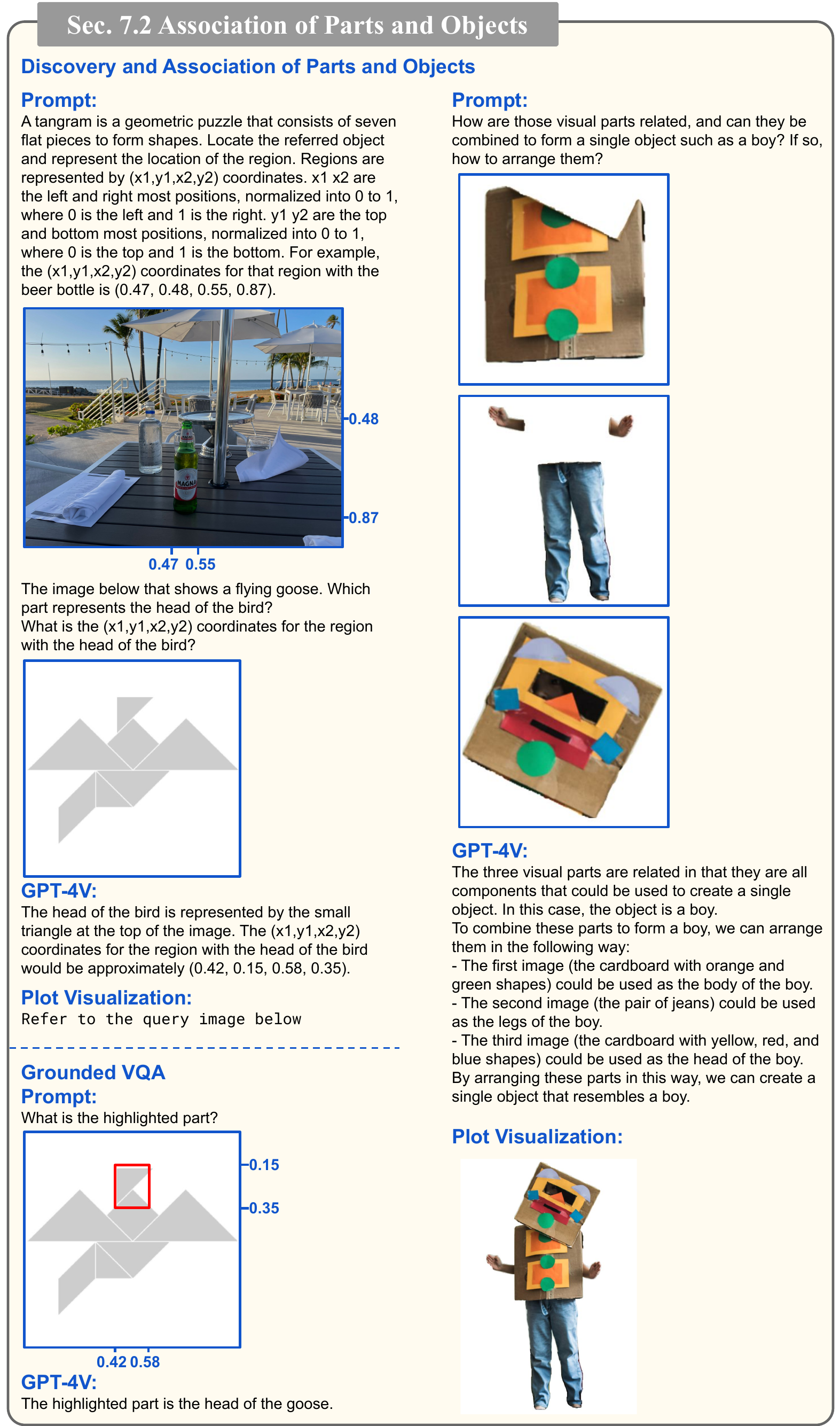}
\caption[Section~\ref{sec:iq-02}: part-object association.]{Understanding part-object association in abstract and natural images. Check Section~\ref{sec:iq-02} for detailed discussions.
}
\label{fig:spatial_abstract_b}
\vspace{-30pt}
\end{figure*}

Understanding and reasoning over abstract visual stimuli and symbols is one fundamental ability for human intelligence. This section examines if \modelname~can abstract semantics from visual signals and can perform different types of human Intelligence Quotient (IQ) tests.

\subsection{Abstract Visual Stimuli}
\label{sec:iq-01}
Humans can infer semantics from abstract and often ambiguous visual stimuli. Figure~\ref{fig:spatial_abstract_a} explores having \modelname~interpret tangram~\cite{clark1986referring,murfitt2001effect,fasquel2022modified,ji2022abstract}. A tangram is a traditional geometric puzzle that consists of seven flat pieces called tans, which are put together to form shapes without overlapping the pieces. For example, \modelname~interprets that sub-figure 7 in Figure~\ref{fig:spatial_abstract_a} best illustrates a flying goose and provides reasoning descriptions for other sub-figure, \eg, 4. person or robot, 9. boat or hat, and 10. dog or fox. \modelname~also has the ability to understand other formats of abstract visual diagrams~\cite{wang2023bot,barrett2018measuring,zhang2019raven}, such as ASCII text art of cartoon characters in Figure~\ref{fig:spatial_abstract_a} and symbolic inputs in Figures~\ref{fig:IQ_WAIS}-\ref{fig:IQ_Raven}.

\subsection{Discovery and Association of Parts and Objects}
\label{sec:iq-02}
Discovering and associating object parts~\cite{xu2019unsupervised,gadre2021act} is another important abstract visual reasoning capability. Humans can easily discover how object parts may compose a semantically meaningful object. Figure~\ref{fig:spatial_abstract_b} designs examples to probe \modelname's capability in associating object parts. In the left example, we ask \modelname~to localize an object part based on its semantic meaning. In the right example, \modelname~is asked to associate object parts segmented by SAM~\cite{kirillov2023segment}. \modelname~can process figures for all object parts and associate them in a semantically meaningful to form the boy visualized in the bottom right.

\subsection{Wechsler Adult Intelligence Scale}
\label{sec:iq-03}
\begin{figure*}[h!]
\centering
\vspace{-20mm}
\includegraphics[width=\textwidth]{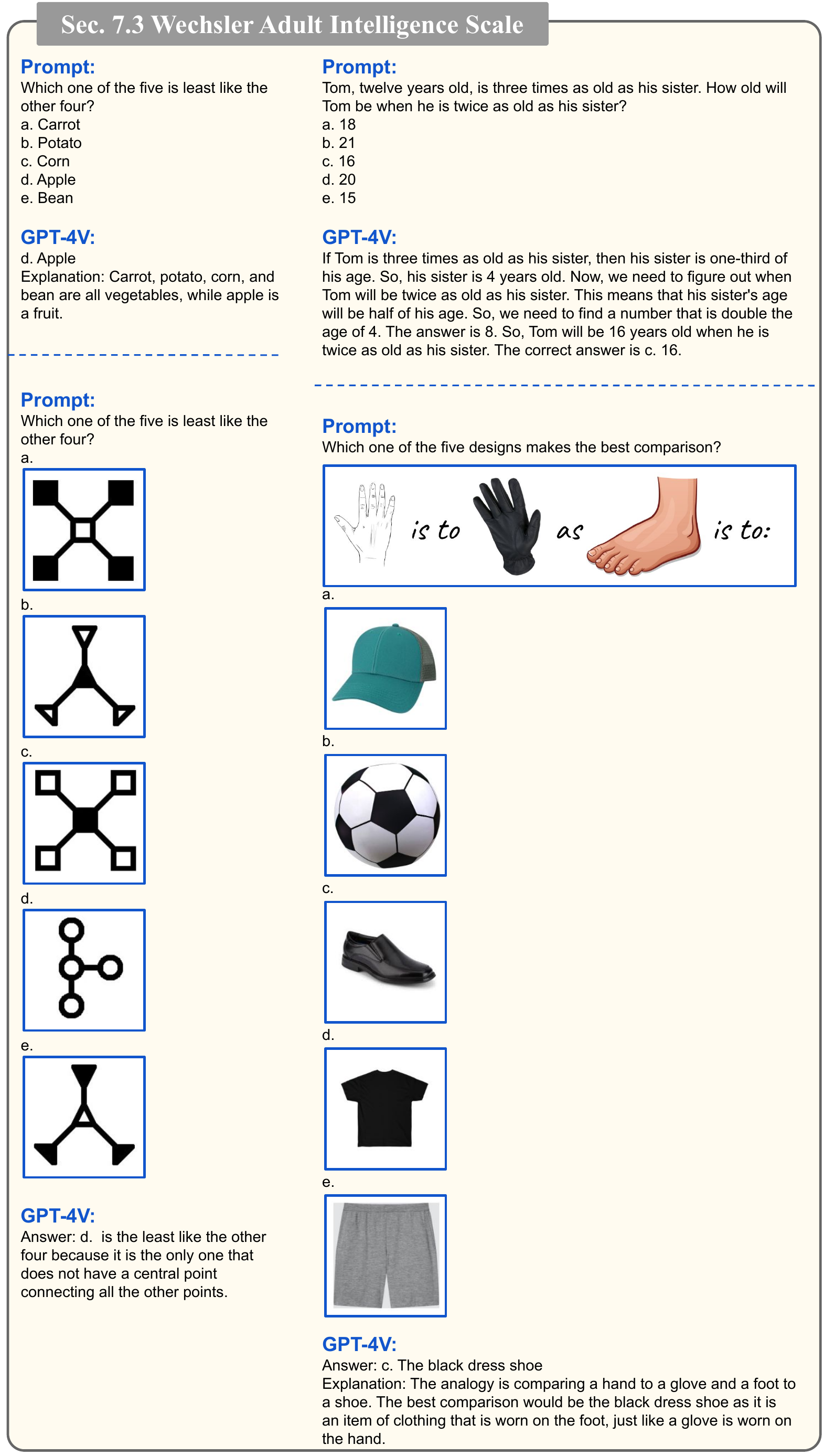}
\caption[Section~\ref{sec:iq-03}: WAIS IQ test.]{Example questions from the Wechsler Adult Intelligence Scale (WAIS)~\cite{wechsler1981wais}. Check Section~\ref{sec:iq-03} for detailed discussions.
}
\vspace{-15pt}
\label{fig:IQ_WAIS}
\end{figure*}

Section~\ref{sec:iq-01} demonstrates the abstract visual understanding capability of \modelname. As a further challenge, \modelname~is asked to perform different abstract reasoning tasks, sourced from human Intelligence Quotient (IQ) tests.
The Wechsler Adult Intelligence Scale~\cite{wechsler1981wais} is recognized as one of the ``gold standard IQ tests,'' and is designed to provide a comprehensive measurement of an individual's cognitive abilities using a series of sub-tests. Figure~\ref{fig:IQ_WAIS} shows representative questions and \modelname's outputs from each sub-test category. \modelname~shows promises in abstract reasoning, answering questions with texts only, symbolic visual inputs, and natural images. For example, the bottom right sample shows that \modelname~can interpret the analogy question and find the best comparison of shoes.

\subsection{Raven's Progressive Matrices}
\label{sec:iq-04}
\begin{figure*}[h!]
\centering
\includegraphics[width=\textwidth]{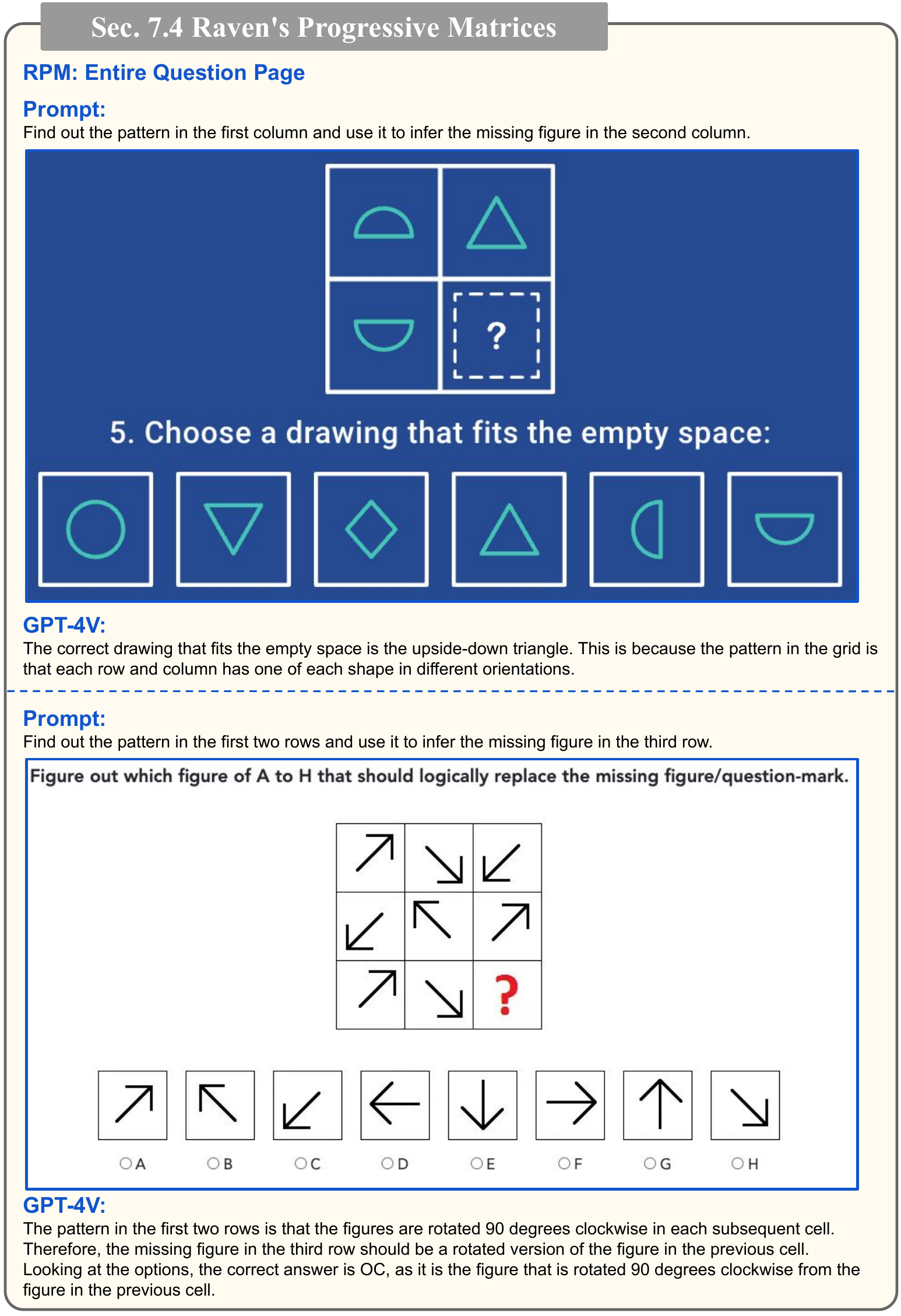}
\caption[Section~\ref{sec:iq-04}: RPM IQ test, entire question page.]{Example questions from the Raven's Progressive Matrices~\cite{raven1938raven,huang2023language}. We challenge \modelname~by sending the entire question page as a single image, mimicking how humans look at the IQ tests. Check Section~\ref{sec:iq-04} for detailed discussions.
}%
\label{fig:IQ_Raven}
\end{figure*}

\begin{figure*}[h!]
\centering
\includegraphics[width=\textwidth]{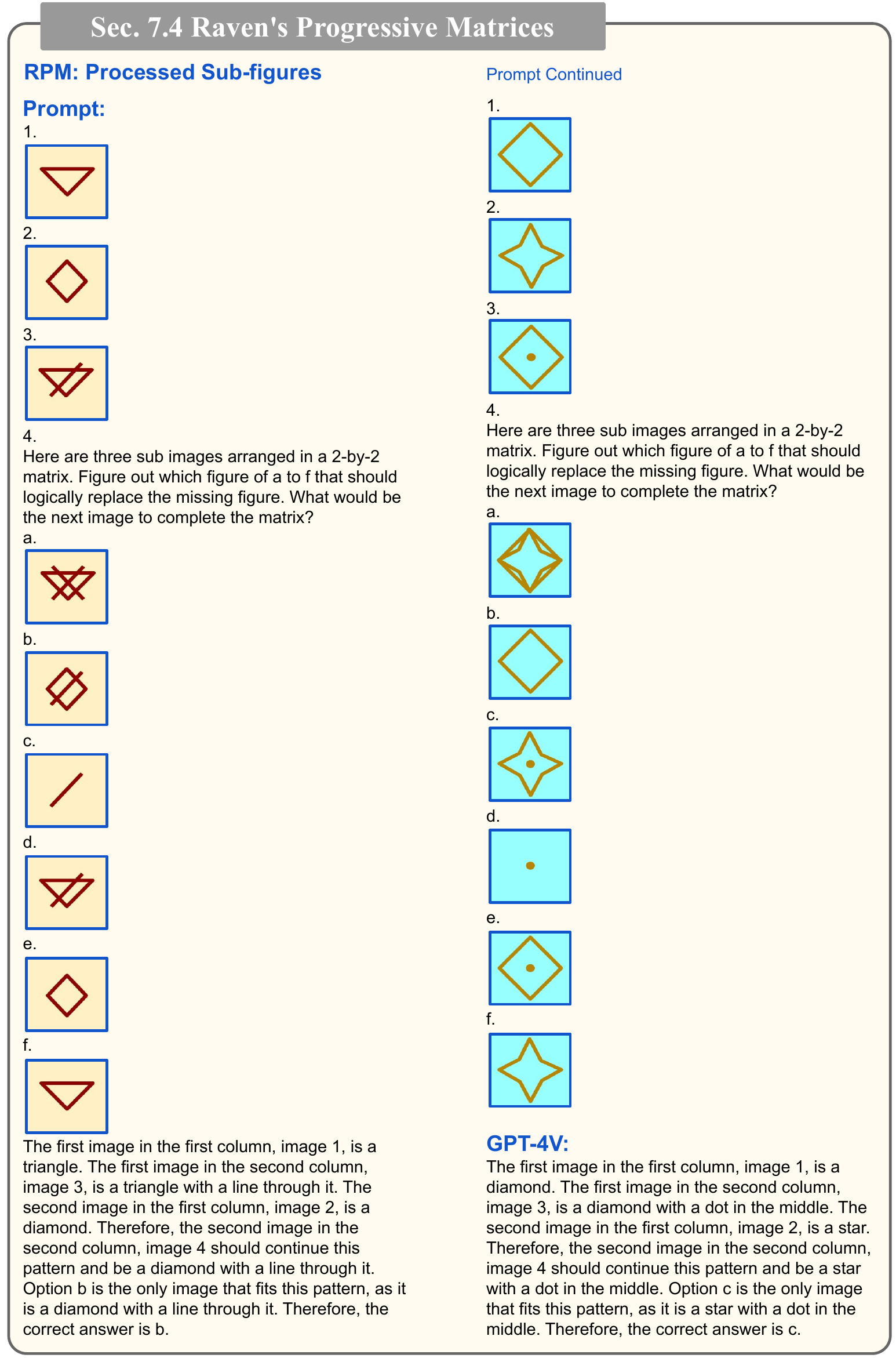}
\caption[Section~\ref{sec:iq-04}: RPM IQ test, processed sub-figures.]{Instead of sending the entire question page as a single image, we may also process the image into multiple sub-figures and optionally provide detailed instructions and examples to further boost the answer accuracy. Check Section~\ref{sec:iq-04} for detailed discussions.
}%
\label{fig:IQ_Raven_sub}
\end{figure*}

Raven's Progressive Matrices (RPM)~\cite{raven1938raven} is another well-known non-verbal intelligence test developed to measure abstract reasoning and problem-solving abilities. The test is designed to minimize the influence of language, culture, and formal education on test performance, making it suitable for testing AI models~\cite{barrett2018measuring,zhang2019raven,huang2023language}. Each test sample contains three or eight images, arranged in 2-by-2 or 3-by-3 matrices with one figure missing. The goal is to select the next image from multiple candidate images by identifying patterns in the provided samples. In our approach, we challenge \modelname~by sending the entire question page as a single image, instead of converting it into interleaved image-text pairs, similar to the human approach to IQ tests. As shown in Figure~\ref{fig:IQ_Raven}, \modelname~can generate reasonable answers without processed text descriptions or sub-figures. However, we also notice that breaking down the entire question image into interleaved text and sub-figures, such as in Figure~\ref{fig:IQ_Raven_sub}, does simplify the task and let \modelname~produce more reliable answers.


\clearpage
\section{Emotional Quotient Test}
\label{sec:09eq}
\begin{figure*}[h!]
\centering
\includegraphics[width=\textwidth]{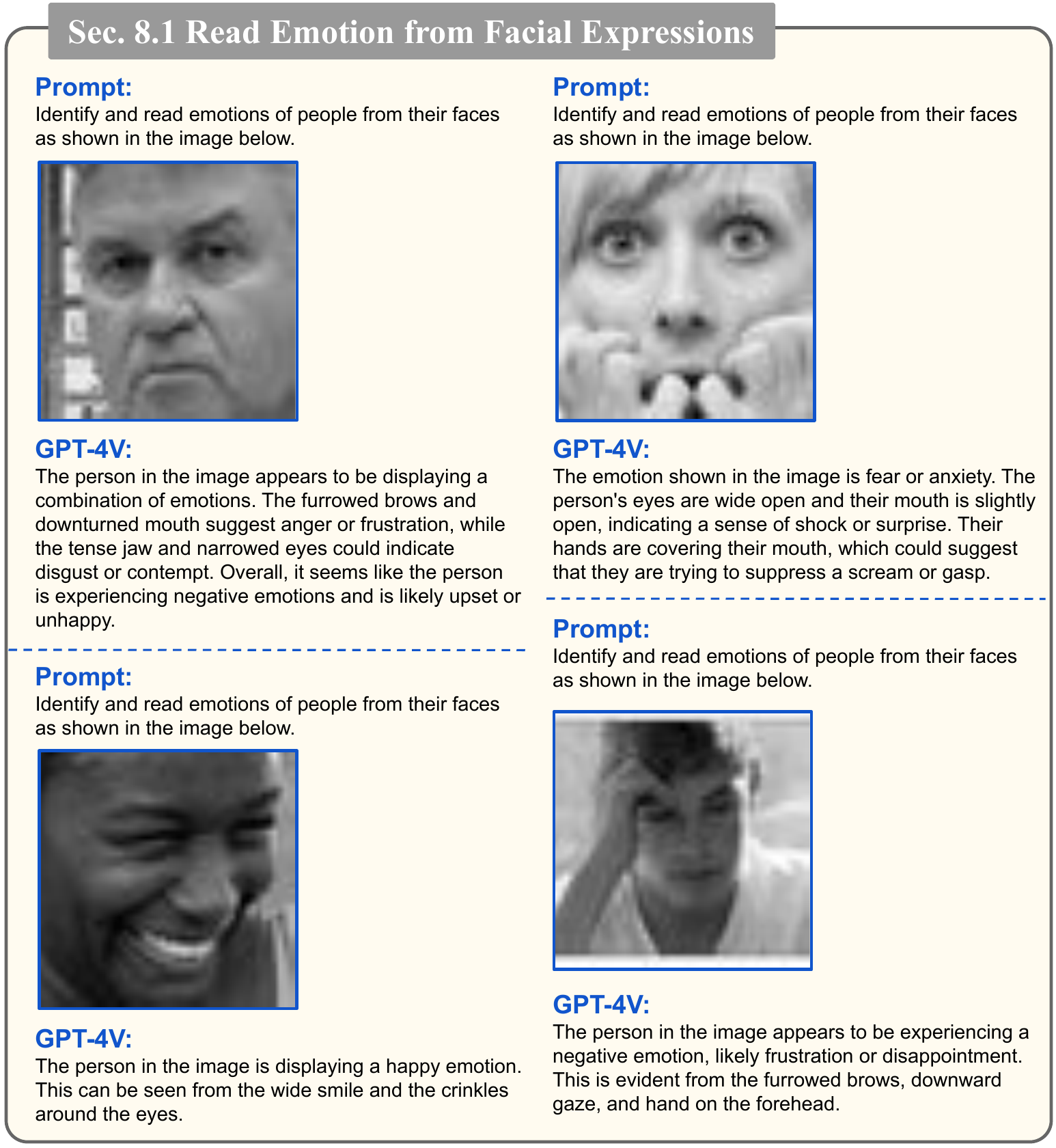}
\caption[Section~\ref{sec:eq-01}: read emotions from facial expressions.]{\modelname~can reliably identify and read the emotions of people from their facial expressions. Check Section~\ref{sec:eq-01} for detailed discussions.
}%
\label{fig:eq_a}
\end{figure*}
\begin{figure*}[h!]
\centering
\vspace{-20pt}
\includegraphics[width=\textwidth]{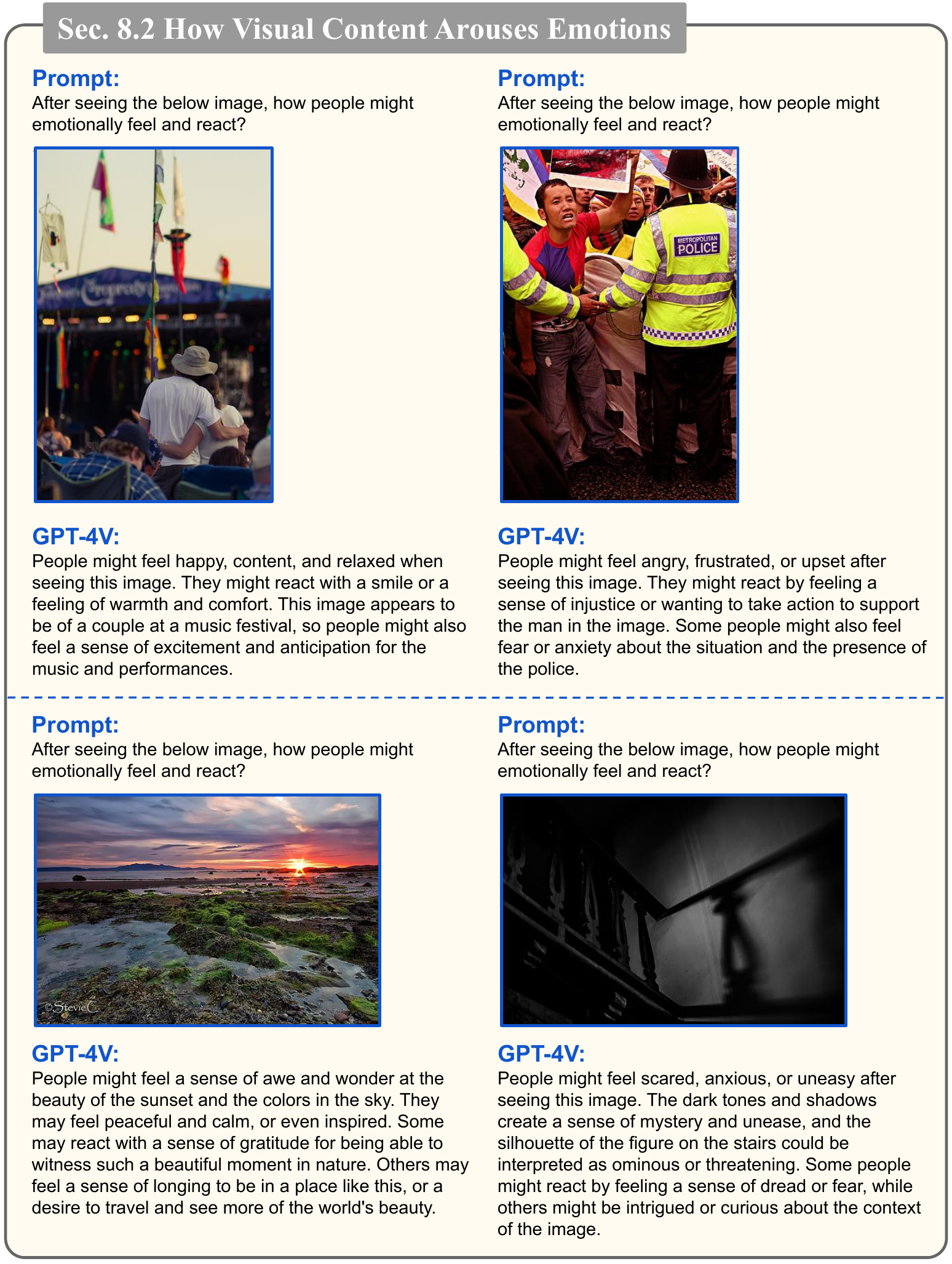}
\caption[Section~\ref{sec:visual_emotion}: how different visual contents may arouse emotions.]{\modelname~understands how different visual contents may arouse human emotions. Check Section~\ref{sec:visual_emotion} for detailed discussions.
}%
\vspace{-10pt}
\label{fig:eq_b}
\end{figure*}
\begin{figure*}[h!]
\centering
\vspace{60pt}
\includegraphics[width=\textwidth]{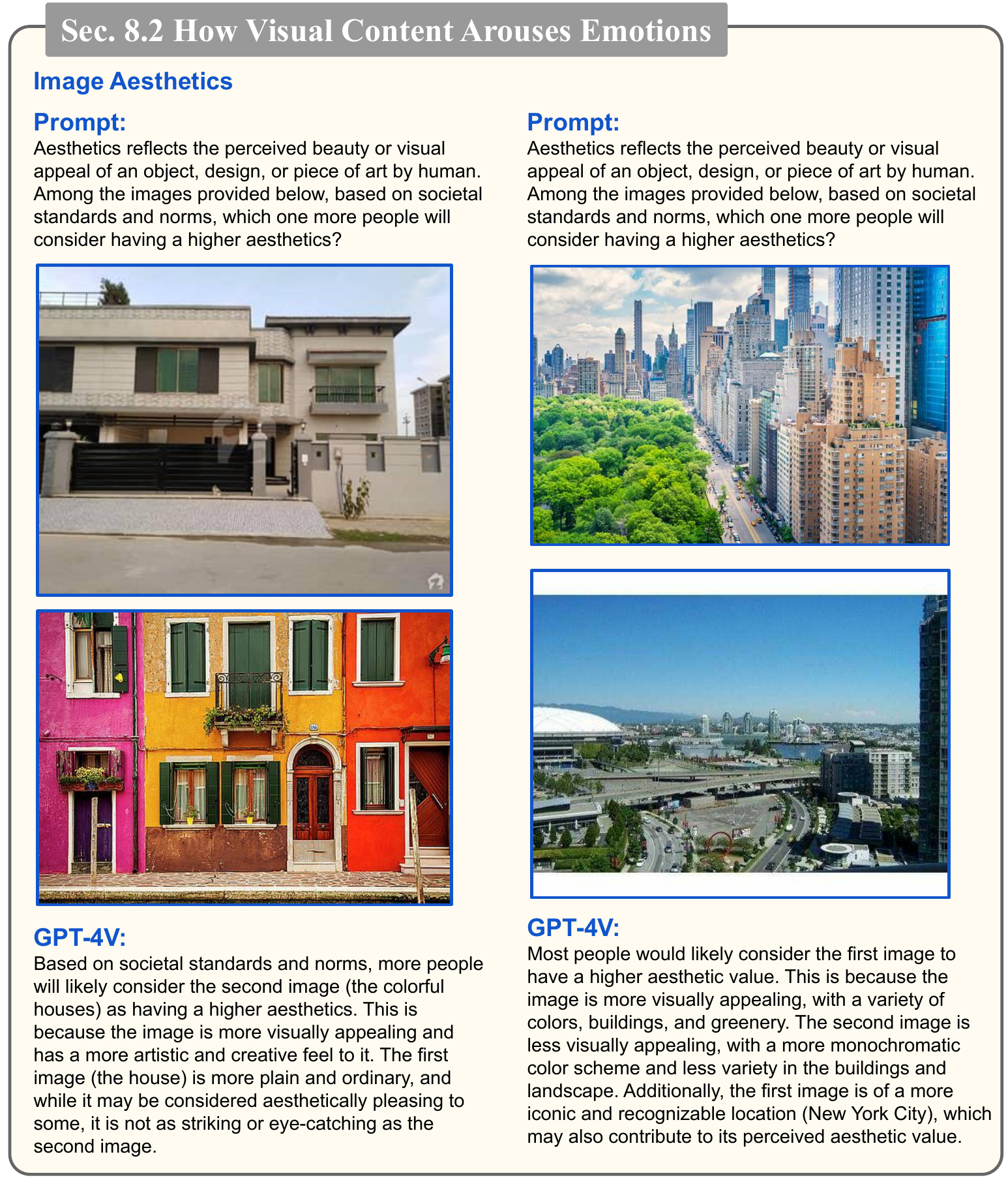}
\caption[Section~\ref{sec:visual_emotion}: image aesthetics.]{\modelname~judges image aesthetics based on societal standards and norms. Check Section~\ref{sec:visual_emotion} for detailed discussions.
}%
\label{fig:eq_c}
\end{figure*}
\begin{figure*}[h!]
\centering
\includegraphics[width=\textwidth]{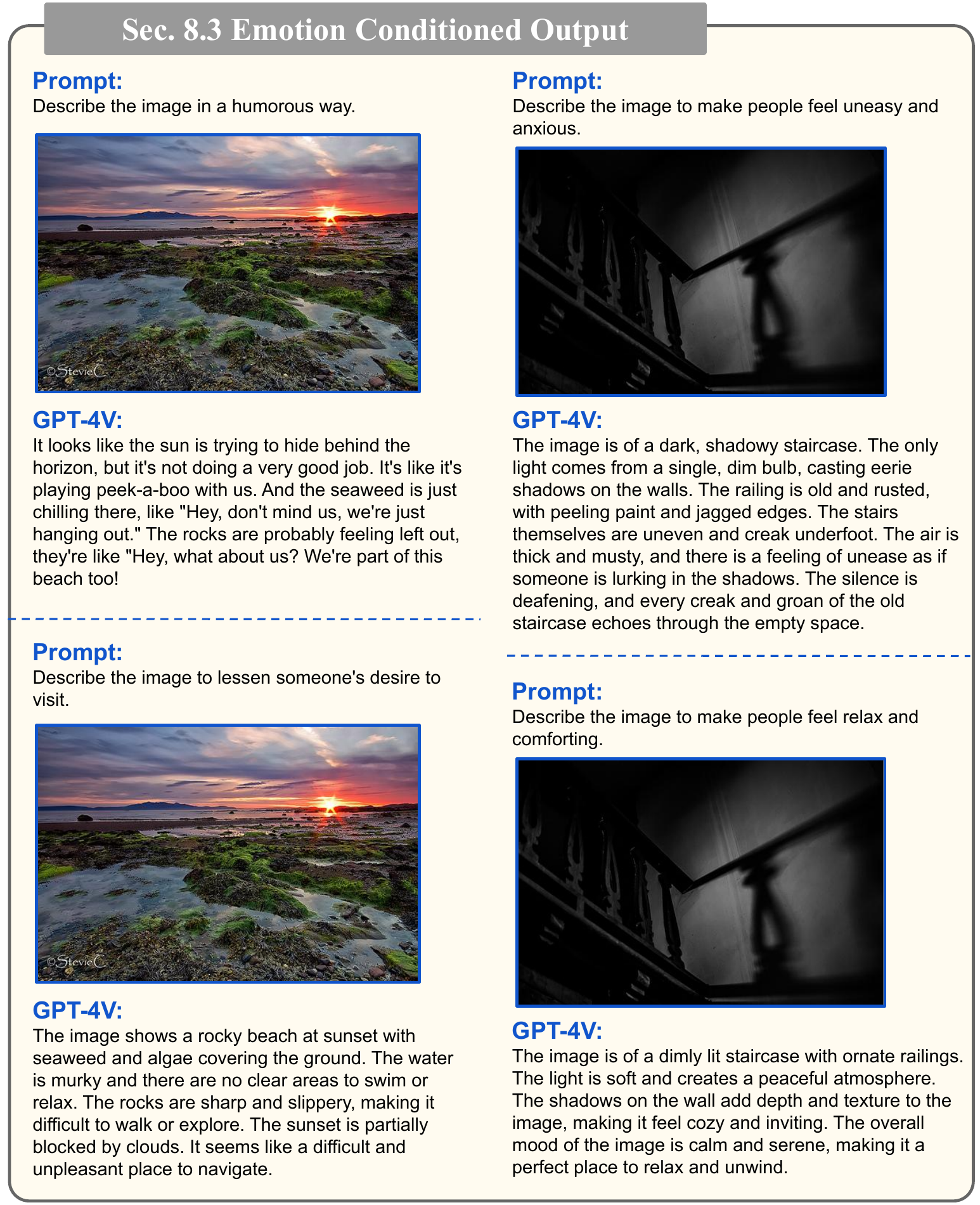}
\caption[Section~\ref{sec:eq-03}: emotion conditioned outputs.]{\modelname~generates proper text based on the perceived or desired emotions, making its communication with humans comforting and effective. Check Section~\ref{sec:eq-03} for detailed discussions.
}%
\label{fig:eq_d}
\end{figure*}

When interacting with humans, it is important that \modelname~has the empathy and Emotional Quotient (EQ) to understand and share the feelings of humans. Inspired by the definition of the human EQ test~\cite{mayer2008human,mayer2002msceit,brackett2006measuring}, we examine \modelname's capability in (1) identifying and reading human emotions from their facial expressions, (2) understanding how different visual contents may arouse emotions, and (3) generating proper text outputs conditioned on the desired emotional and sentiment.

\subsection{Read Emotion from Facial Expressions}
\label{sec:eq-01}
As shown in Figure~\ref{fig:eq_a}, \modelname~can reliably identify and read the emotions of people from their facial expressions. It also provides reasonable rationales for the visual cues observed to make the emotion interpretation, indicating a good understanding of the facial emotions.

\subsection{Understand How Visual Content Arouses Emotions}
\label{sec:visual_emotion}
We next analyze \modelname's ability on visual sentiment analysis, \ie, understanding humans' emotional response after seeing the visual contents. Such ability is critical for \modelname~to anticipate how visual contents may arouse human emotions and thereby react properly. As shown in Figure~\ref{fig:eq_b}, \modelname~can interpret visual sentiments such as content, anger, awe, and fear, based on both the semantic contents and the image style. These capabilities are essential in use cases such as home robots. 

In addition to interpreting visual sentiment, \modelname~also aligns with human subjective judgments such as aesthetics. Figure~\ref{fig:eq_c} shows examples of \modelname~judging image aesthetics based on societal standards.

\subsection{Emotion Conditioned Output}
\label{sec:eq-03}
Based on the perceived emotions, \modelname~effectively generates proper text outputs conditioned on the desired emotion. For example, in Figure~\ref{fig:eq_d}, \modelname~can follow the prompt to describe the right-side scary image in a way that makes it more horrifying or becoming comforting. This demonstrates \modelname's potential to enable emotion-aware human-robot communication.
\clearpage
\section{Emerging Application Highlights}
\label{sec:10app}

In this section, we showcase a myriad of high-value application scenarios and new use cases that can be potentially enabled by the remarkable capabilities of \modelname. While it is true that some of these application scenarios can be accomplished by meticulously curating the training data for finetuning existing Vision and Language (VL) models, we want to emphasize that the true power of \modelname lies in its ability to perform effortlessly right out of the box. Moreover, we present how \modelname seamlessly integrates with external tools and plugins, further expanding its potential and enabling even more innovative and collaborative applications.

\subsection{Spot the Difference}\label{sec:spot-diff}
\begin{figure*}[h!]
\centering
\vspace{-15mm}
\includegraphics[width=\textwidth]{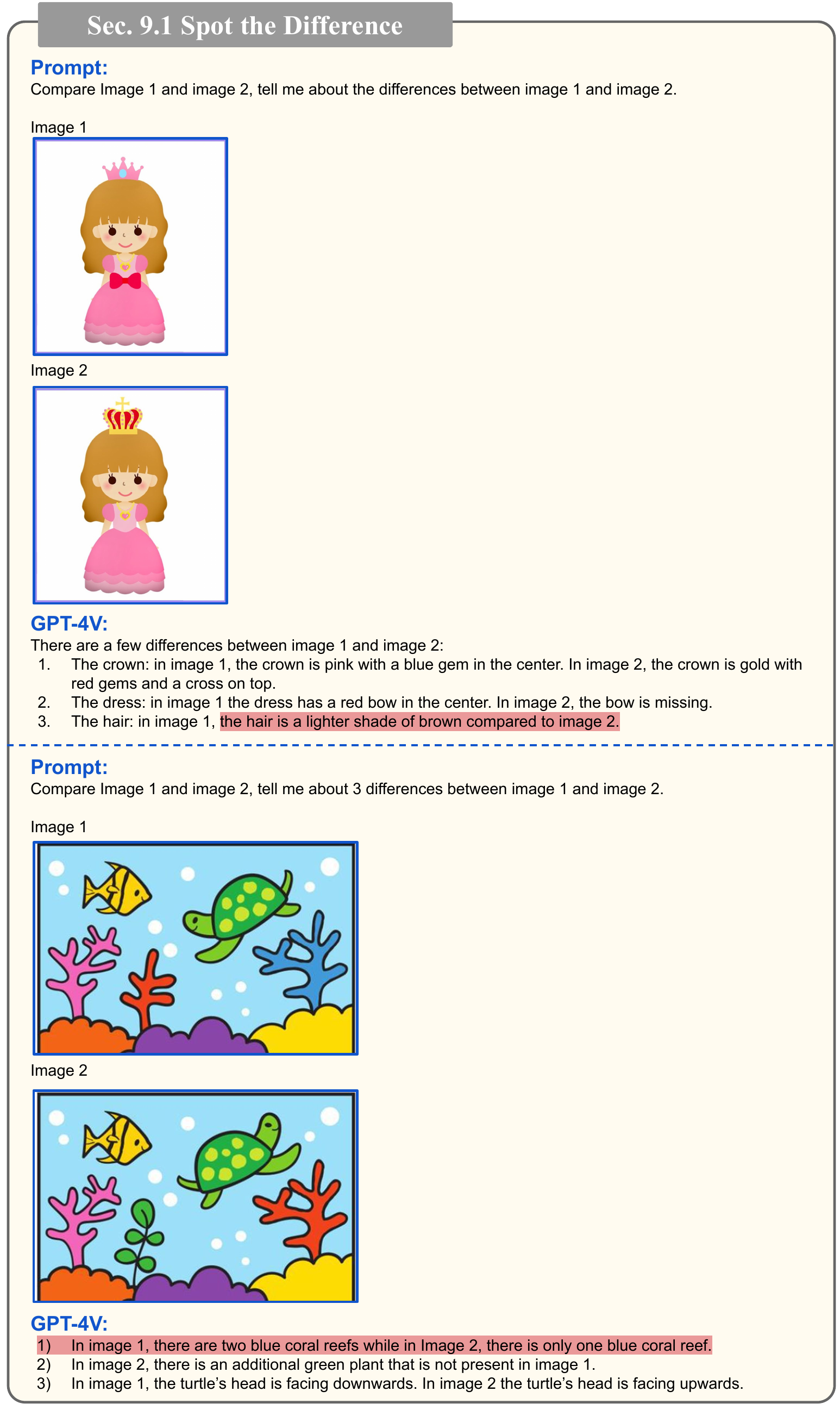}
\caption[Section~\ref{sec:spot-diff}: spot the differences.]{Spot the differences. \colorbox{redhl}{Red} highlights the inaccurate description about the differences. Check Section~\ref{sec:spot-diff} for detailed discussions.
}
\label{fig:find_diff_1}
\end{figure*}

\begin{figure*}[h!]
\centering
\vspace{-15mm}
\includegraphics[width=\textwidth]{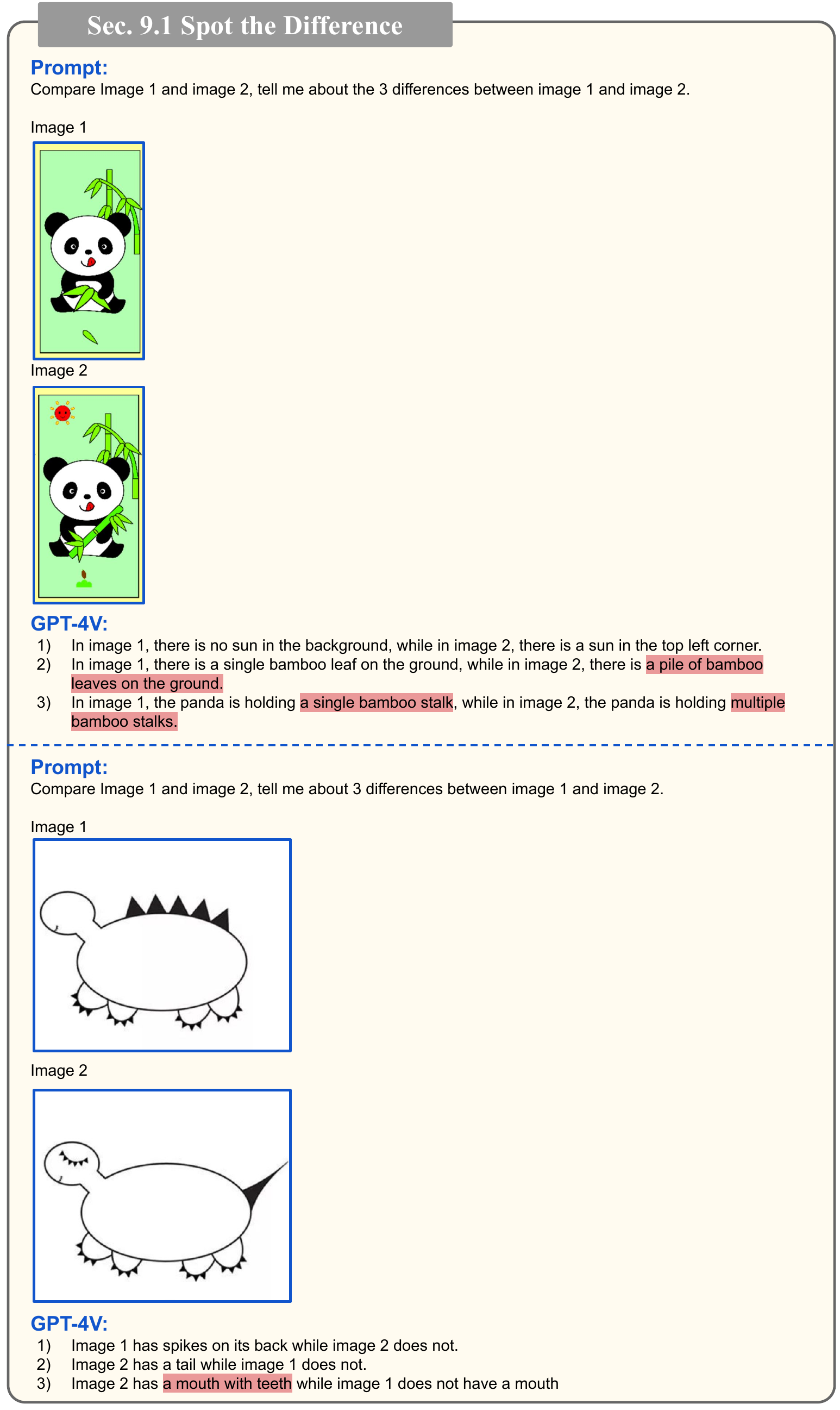}
\caption[Section~\ref{sec:spot-diff}: spot the differences.]{Spot the differences. \colorbox{redhl}{Red} highlights the inaccurate description about the differences. Check Section~\ref{sec:spot-diff} for detailed discussions.
}
\label{fig:find_diff_2}
\end{figure*}

We begin with a generic use case inspired by the brain-teasing game ``Spot the Difference.'' In Figures~\ref{fig:find_diff_1}-\ref{fig:find_diff_2}, we provide \modelname two visually similar images that contain subtle differences in certain regions. The task given to \modelname is to identify all the differences between the two images. Among the four examples, \modelname successfully identifies the regions or components that differ in the images. However, it falls short in providing accurate explanations for what is depicted in each image. To delve deeper into \modelname's capabilities, let's focus on the first example shown in Figure~\ref{fig:find_diff_1}. 
Although \modelname fails to recognize that the discrepancy lies in the number of cuts in the hairband rather than the shade of the hair,
it correctly identifies that the crown, the bow of the dress, and the hair differ between the two images. 
While \modelname's predictions in the ``Spot the Difference'' game are not perfect, its ability to compare the content in two images proves valuable in real-life applications, such as defect detection, which we will explore in the following subsections.

\subsection{Industry}\label{sec:app-industry}

\paragraph{Defect detection.}  Throughout the history of manufacturing, computer vision techniques have played a crucial role. One specific application scenario is defect detection, which is an essential step in manufacturing processes to ensure product quality. Detecting faults or defects in a timely manner and taking appropriate actions are vital for minimizing operational and quality-related costs.

\begin{figure*}[h!]
\centering
\includegraphics[width=\textwidth]{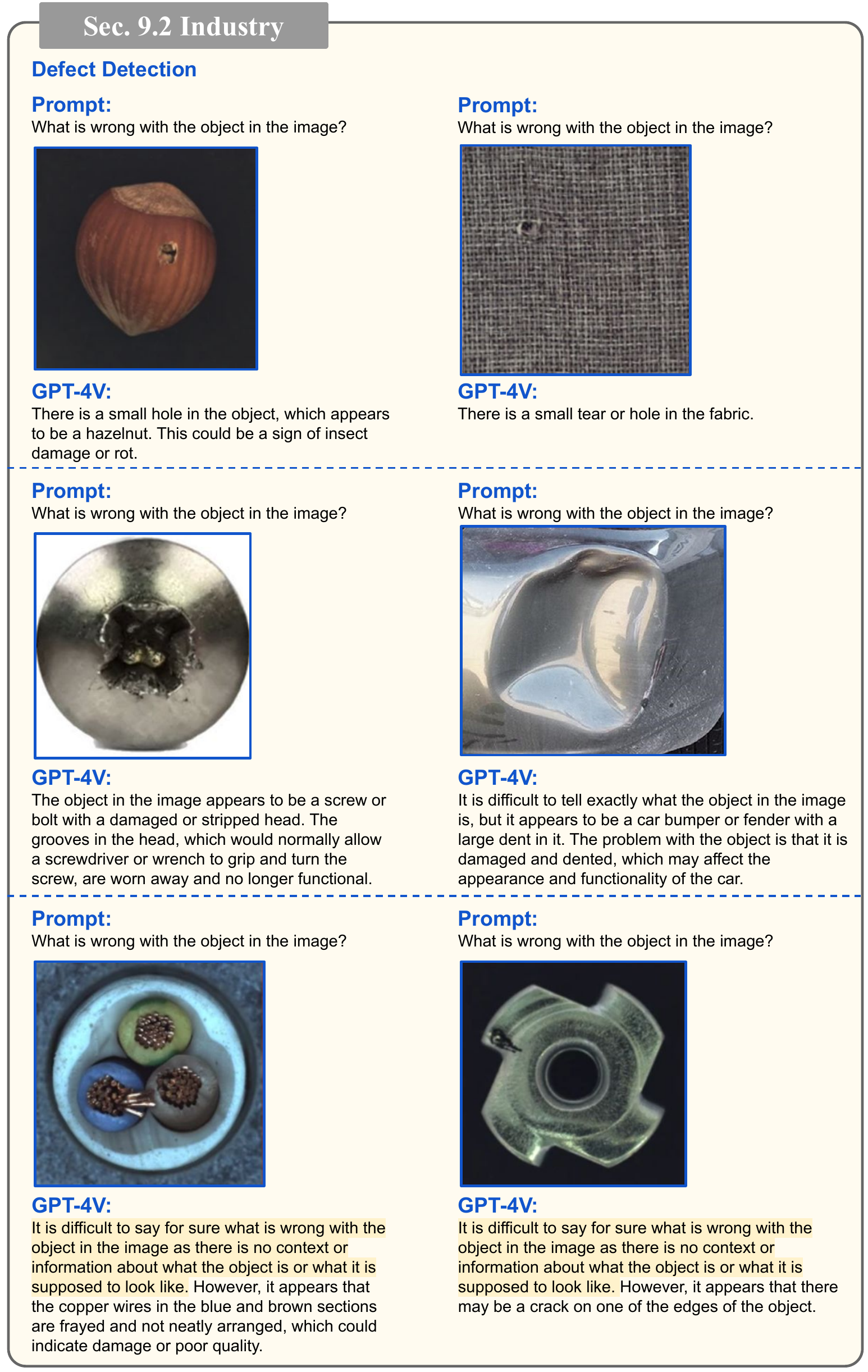}
\caption[Section~\ref{sec:app-industry}: industrial defect detection.]{Defect detection with a single image.  \colorbox{yellowhl}{Yellow} highlights the cases when \modelname is hesitating to make the predictions. Check Section~\ref{sec:app-industry} for detailed discussions. 
}
\label{fig:defect_det_1}
\end{figure*}

\begin{figure*}[h!]
\centering
\includegraphics[width=\textwidth]{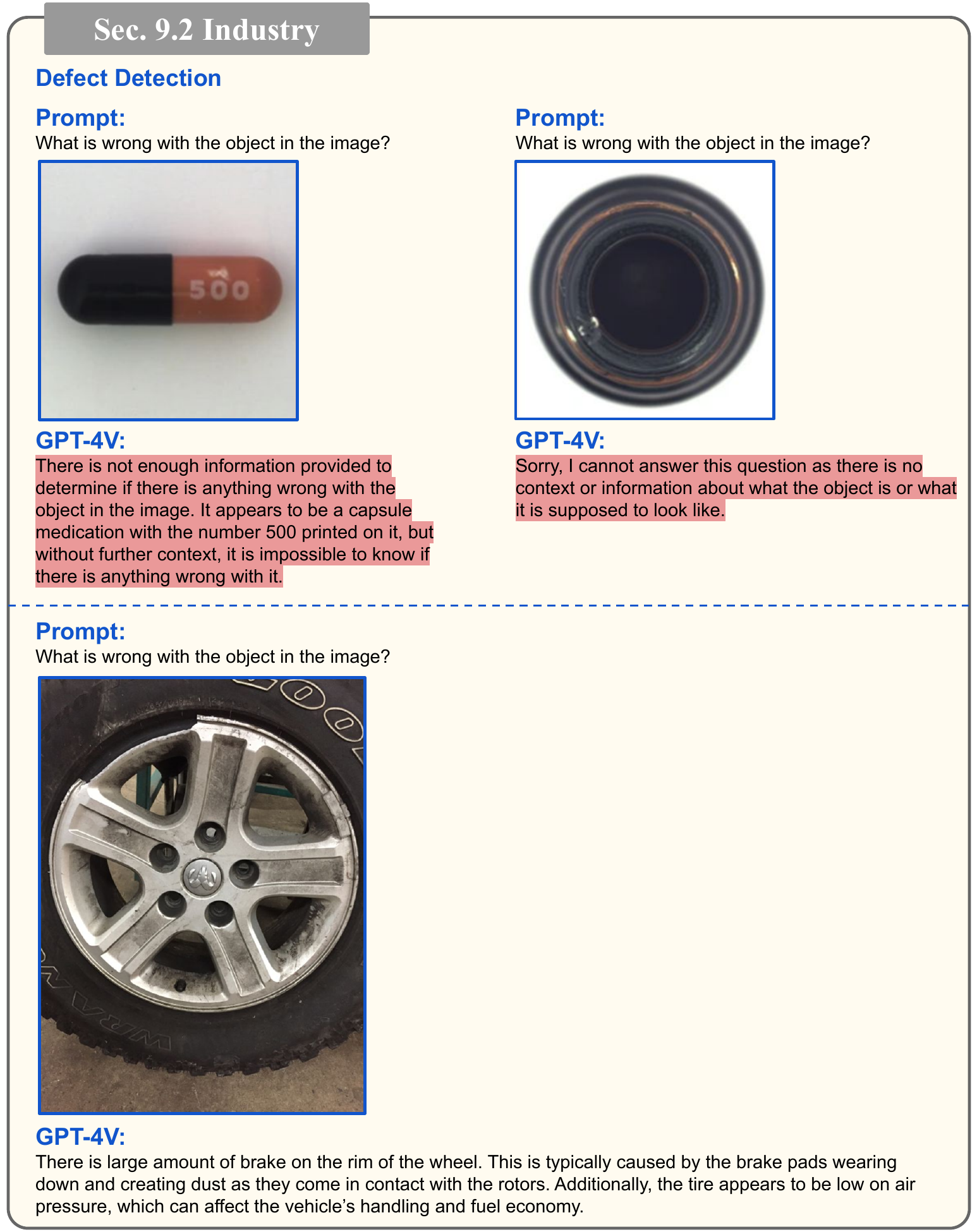}
\caption[Section~\ref{sec:app-industry}: industrial defect detection.]{Failure examples of defect detection with a single image.\colorbox{redhl}{Red} highlights the cases when \modelname fails. Check Section~\ref{sec:app-industry} for detailed discussions. 
}
\label{fig:defect_det_2}
\end{figure*}

In this scenario, we demonstrate the defect detection capabilities of \modelname by presenting images of defective products in Figures~\ref{fig:defect_det_1}-\ref{fig:defect_det_2}. For commonly encountered products in real-life (e.g., hazelnut, fabric, screw, and car bumper in Figure~\ref{fig:defect_det_1}), \modelname confidently identifies the defects such as small holes in the hazelnut/fabric, stripped heads of screws, and dents in car bumpers. However, when it comes to uncommon product images (e.g., the metal parts in Figures~\ref{fig:defect_det_1}-\ref{fig:defect_det_2}) or products with variations in appearance (e.g., the pill in Figure~\ref{fig:defect_det_2}), \modelname may hesitate or even refuse to make predictions. An interesting case in Figure~\ref{fig:defect_det_2} involves a car tire, where multiple defects can be observed in the image, including dirt on the wheel, damage to the outer edge of the rim, and signs of wear on the tire. \modelname only focuses on the minor defect (dirt on the wheel) and fails to mention the major defect (damage to the outer edge of the rim) that would require repair.

\begin{figure*}[h!]
\centering
\vspace{-18mm}
\includegraphics[width=\textwidth]{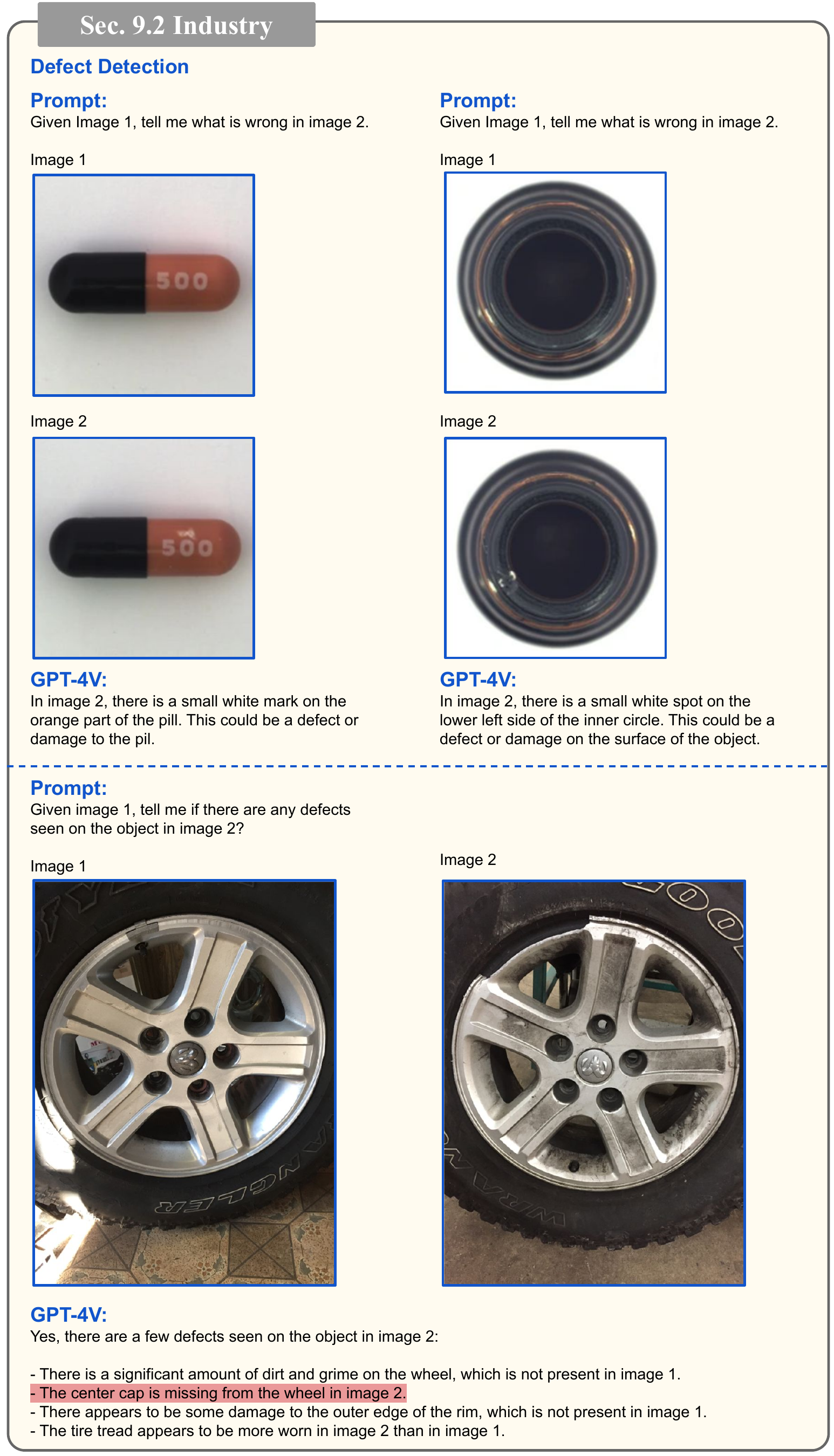}
\caption[Section~\ref{sec:app-industry}: industrial defect detection.]{Defect detection with the help of a reference image.\colorbox{redhl}{Red} highlights inaccurate descriptions. Check Section~\ref{sec:app-industry} for detailed discussions. 
}
\label{fig:defect_det_paired}
\end{figure*}

Given the success of \modelname in ``Spot the Difference'' scenario shown in Section~\ref{sec:spot-diff}, we explore the idea of incorporating a reference image to illustrate what a defect-free product should look like, with the aim of improving the failure cases depicted in Figure~\ref{fig:defect_det_2}. The results of this approach are presented in Figure~\ref{fig:defect_det_paired}. By including the reference image and refining the prompt, \modelname successfully identifies defects in all three failure cases in single-image defect detection. These promising findings highlight a potential high-value application of \modelname for defect detection in the manufacturing industry.

\begin{figure}[h!]
\captionsetup[subfloat]{farskip=-4pt,captionskip=0pt}
\vspace{-25mm}
\centering
   \subfloat[ 
   ]{
      \includegraphics[width=.93\textwidth]{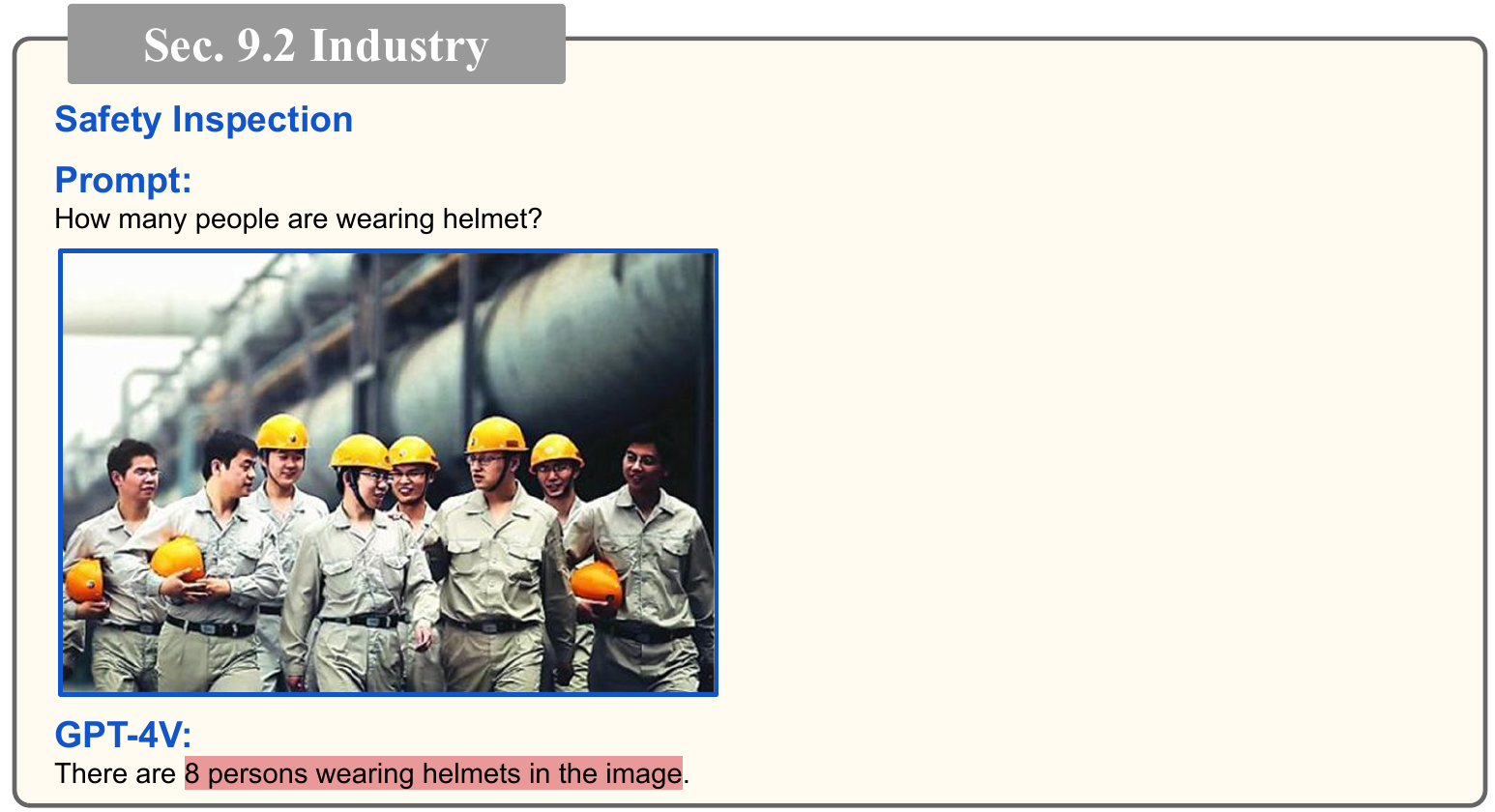}\label{fig:ppe_zs}
   }

   \subfloat[ 
   ]{
      \includegraphics[width=.93\textwidth]{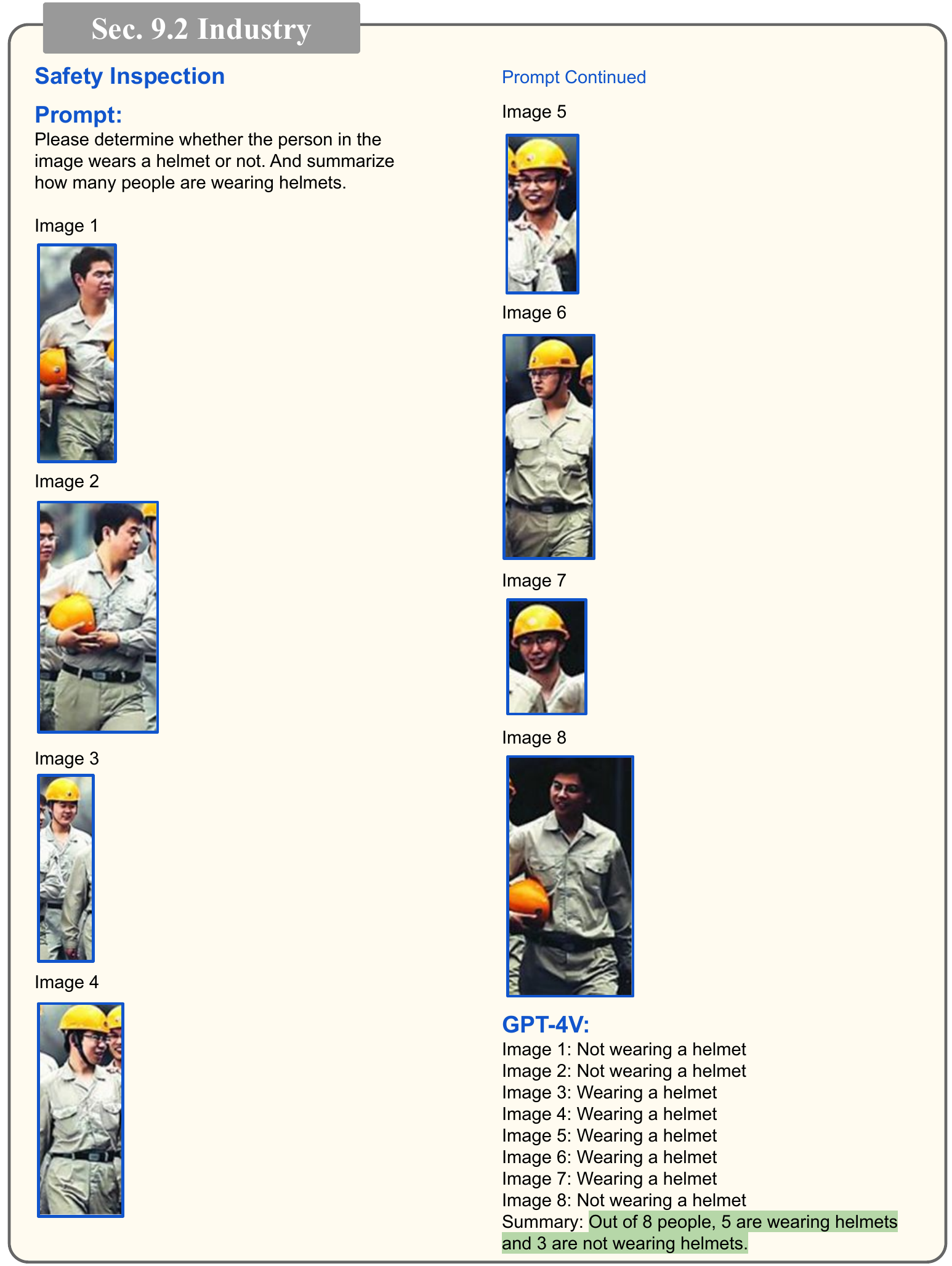}\label{fig:ppe_od}
   }
\vspace{-6pt}
   \caption[Section~\ref{sec:app-industry}: safety inspection.]{
    Application Highlights on Safety Inspection: Personal Protective Equipment (PPE) Counting. \modelname fails with zero-shot prompting in (a), while succeeds with single person crops in (b). 
      \colorbox{redhl}{Red} (\colorbox{greenhl}{Green}) highlights the wrong (correct) answer. Check Section~\ref{sec:app-industry} for detailed discussions.}
\label{fig:ppe}
\end{figure}

\paragraph{Safety inspection.}
 Figure~\ref{fig:ppe} presents an exploration of Personal Protective Equipment (PPE) counting for safety inspection. The inadequate usage or failure to wear PPE, such as helmets, harnesses, and gloves, in work environments like construction sites, significantly increases the risk level associated with work activities. To effectively address this issue, computer vision techniques have been employed as a solution to monitor PPE compliance and promptly identify any violations of safety regulations. Taking helmets as an example, a safety inspection system is necessary to accurately detect and report the number of employees who are not wearing helmets.

In Figure~\ref{fig:ppe_zs}, we assess the performance of \modelname by directly instructing it to count the individuals wearing helmets. \modelname provides a response of ``8 persons wearing helmets,'' which matches the total count of people shown in the image, suggesting there is no alerting safety violations. Obviously, \modelname fails to detect the 3 individuals who are not wearing helmets, thus compromising their personal safety. This task poses a considerable challenge for \modelname, as it involves detecting people in the image, determining whether they are wearing helmets, and calculating the final count of people who are not wearing the helmets.

In Figure~\ref{fig:ppe_od}, instead of presenting \modelname with the original image containing all 8 individuals, we provide cropped regions of the detected persons with an external person detector. 
This approach divides the PPE counting workload into two steps: relying on an off-the-shelf person detector for person detection and leveraging \modelname's robust visual reasoning capabilities and its ability to handle interleaved image-text inputs for identifying the safety issues.
As we can see, \modelname can correctly count the person who is not wearing the helmet, also demonstrating the benefit of tool use and divide-and-conquer. %

\begin{figure}[h!]
\captionsetup[subfloat]{farskip=-4pt,captionskip=0pt}
\vspace{-25mm}
\centering
\subfloat[ 
]{
      \includegraphics[width=.9\textwidth]{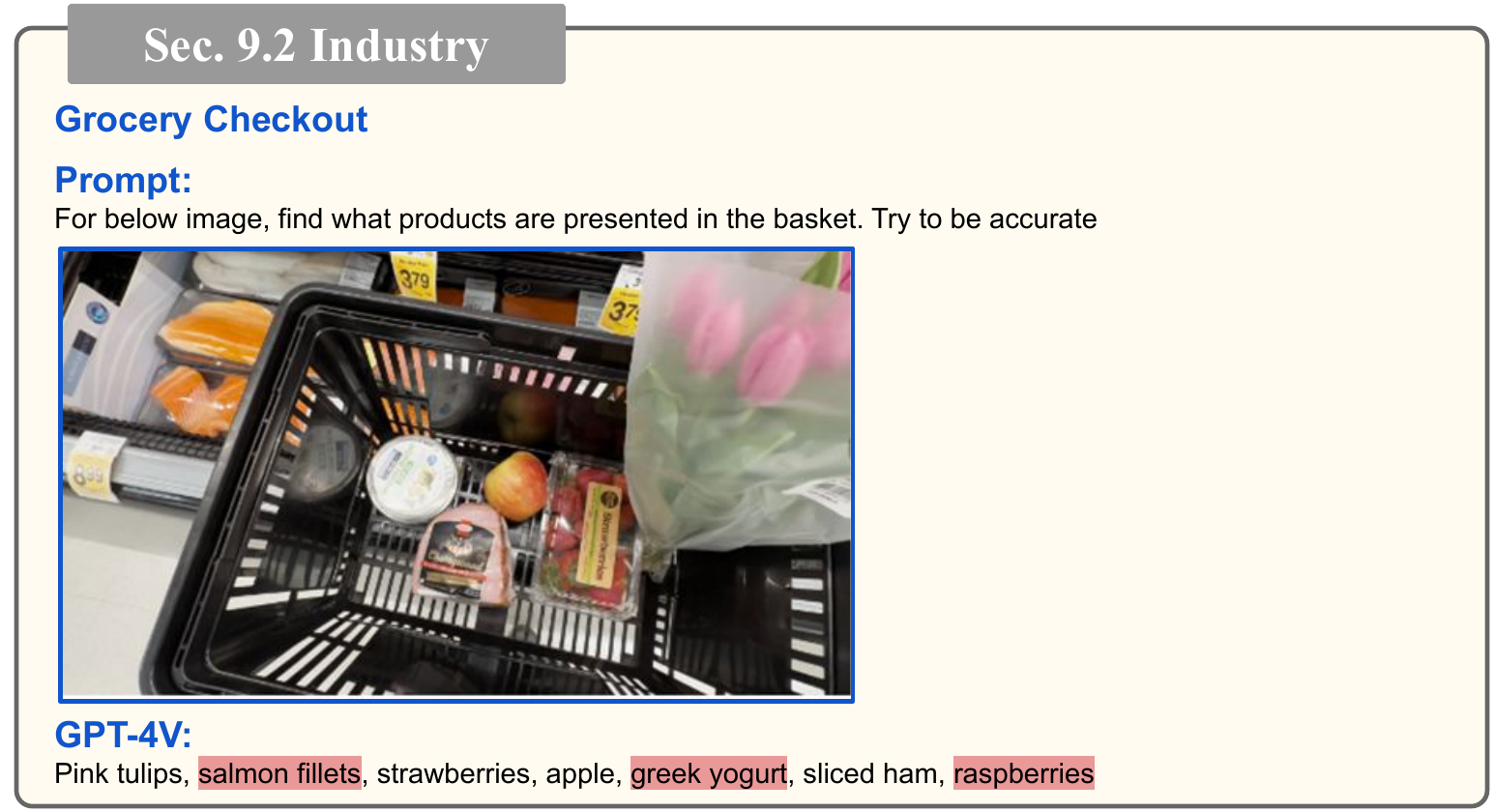}\label{fig:grocery_gptv}
   }

   \subfloat[ 
   ]{
      \includegraphics[width=.9\textwidth]{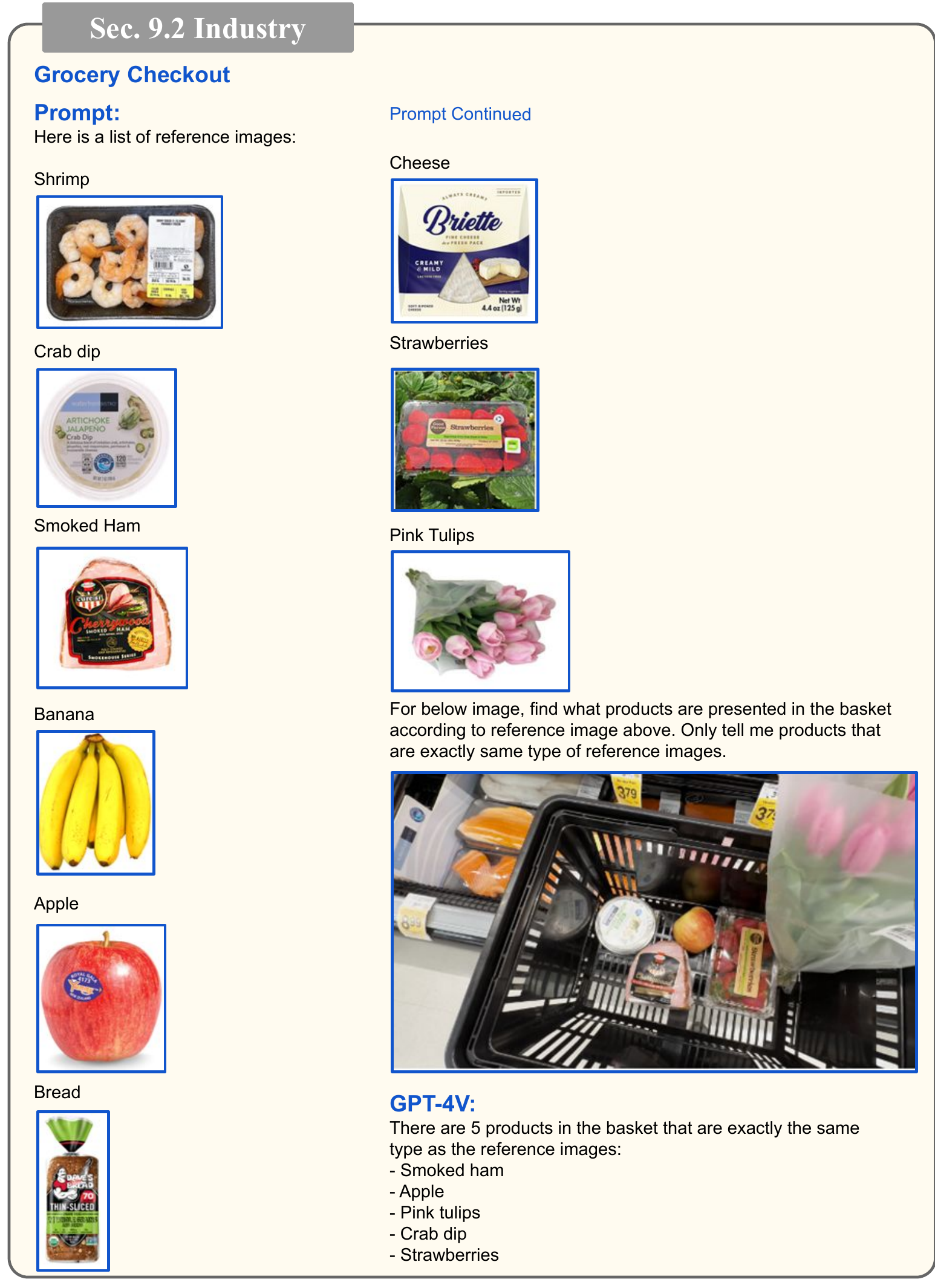}\label{fig:grocery_ref}
   }
\vspace{-6pt}
   \caption[Section~\ref{sec:app-industry}: grocery checkout.]{
    Application Highlights on Grocery Checkout. \modelname fails with zero-shot prompting in (a), while succeeds when prompting with reference product images in (b).
      \colorbox{redhl}{Red} highlights the products that are not in the basket.  Check Sections~\ref{sec:app-industry}, \ref{sec:future-06retrieve} for detailed discussions.}
\label{fig:grocery}
\end{figure}

\paragraph{Grocery checkout.} Self-checkout machines have become increasingly popular in major retailers like Walmart, Target and CVS
to expedite the checkout process for customers and reduce the workload for employees. However, the actual experience with self-checkout machines 
may be frustrating for customers.
Users still need to search for the product barcode or manually enter codes for fresh items like apples, which can be time-consuming, particularly for those unfamiliar with the system. In Figure~\ref{fig:grocery}, we provide a simplified prototype to demonstrate the potential of \modelname in enabling an automatic self-checkout system that can identify and ring up items without user intervention. 

When presented with a photograph of a shopping basket containing five grocery items, as shown in Figure~\ref{fig:grocery_gptv}, \modelname fails to accurately identify the products within the basket. It mistakenly identifies strawberries as raspberries, crab dip as Greek yogurt, and includes salmon fillets that are not even present in the basket. However, in Figure~\ref{fig:grocery_ref}, we improve the prompt by augmenting it with catalog images of grocery products retrieved from the retail website. As a result, \modelname successfully identifies all five items in the basket. This successful demonstration allows the self-checkout system to proceed with retrieving the prices for each identified product from the database. While this is a simple example, it represents a significant step forward toward an automated self-checkout system. Further research and development can explore more complex and realistic scenarios to fully automate the self-checkout process, making it more efficient and convenient for customers.

\clearpage
\subsection{Medical}\label{sec:app_medical}
\begin{figure*}[h!]
\centering
\vspace{-15mm}
\centerline{\includegraphics[width=1.15\textwidth]{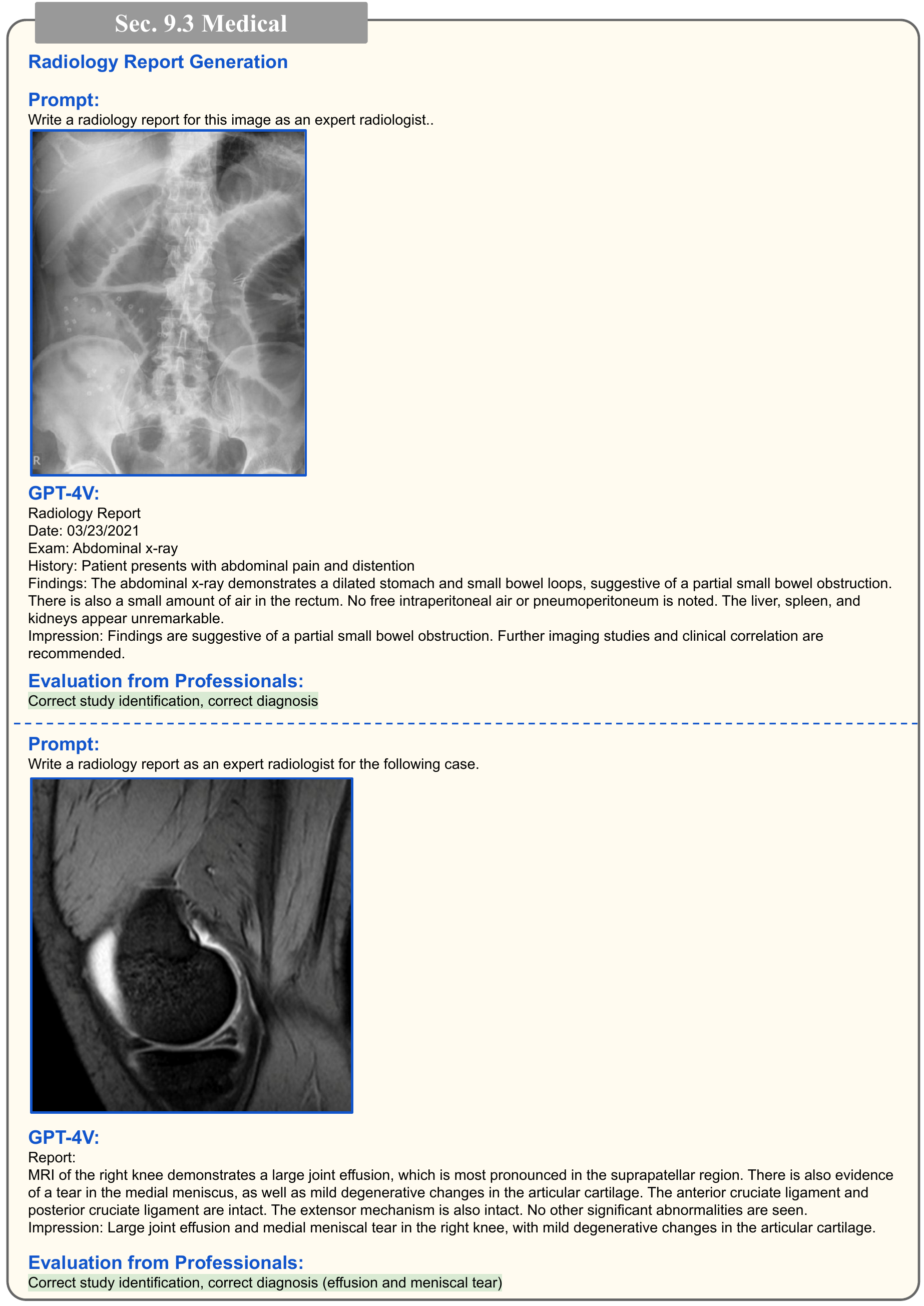}}
\vspace{-6pt}
\caption[Section~\ref{sec:app_medical}: radiology report generation on medical images.]{Application Highlights on Radiology Report Generation. The generated report is reviewed by a medical professional to evaluate its correctness. \colorbox{greenhl}{Green} highlights that a medical professional has confirmed the described part of the report is correct. Check Section~\ref{sec:app_medical} for detailed discussions. The medical images are from the internet.
}
\label{fig:app_med_1}
\end{figure*}

\begin{figure*}[h!]
\centering
\vspace{-25mm}
\centerline{\includegraphics[width=1.15\textwidth]{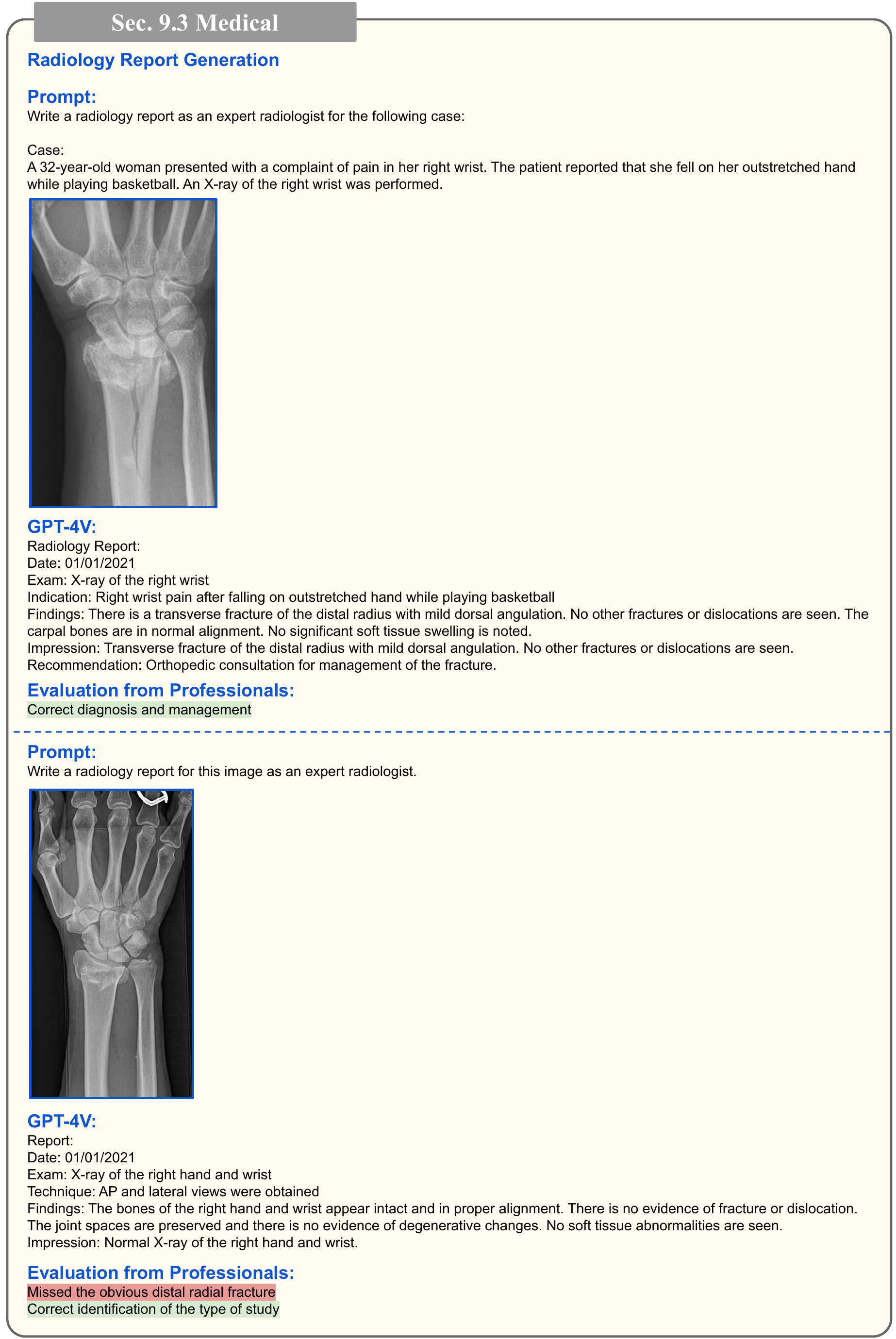}}
\vspace{-6pt}
\caption[Section~\ref{sec:app_medical}: radiology report generation on medical images.]{Application Highlights on Radiology Report Generation. The generated report is reviewed by a medical professional to evaluate its correctness. \colorbox{greenhl}{Green} (\colorbox{redhl}{Red}) highlights that a medical professional has confirmed the described part of the report is correct (incorrect). Check Section~\ref{sec:app_medical} for detailed discussions.  The medical images are from the Internet.
}
\label{fig:app_med_2}
\end{figure*}

\begin{figure*}[h!]
\centering
\vspace{-15mm}
\centerline{\includegraphics[width=1.15\textwidth]{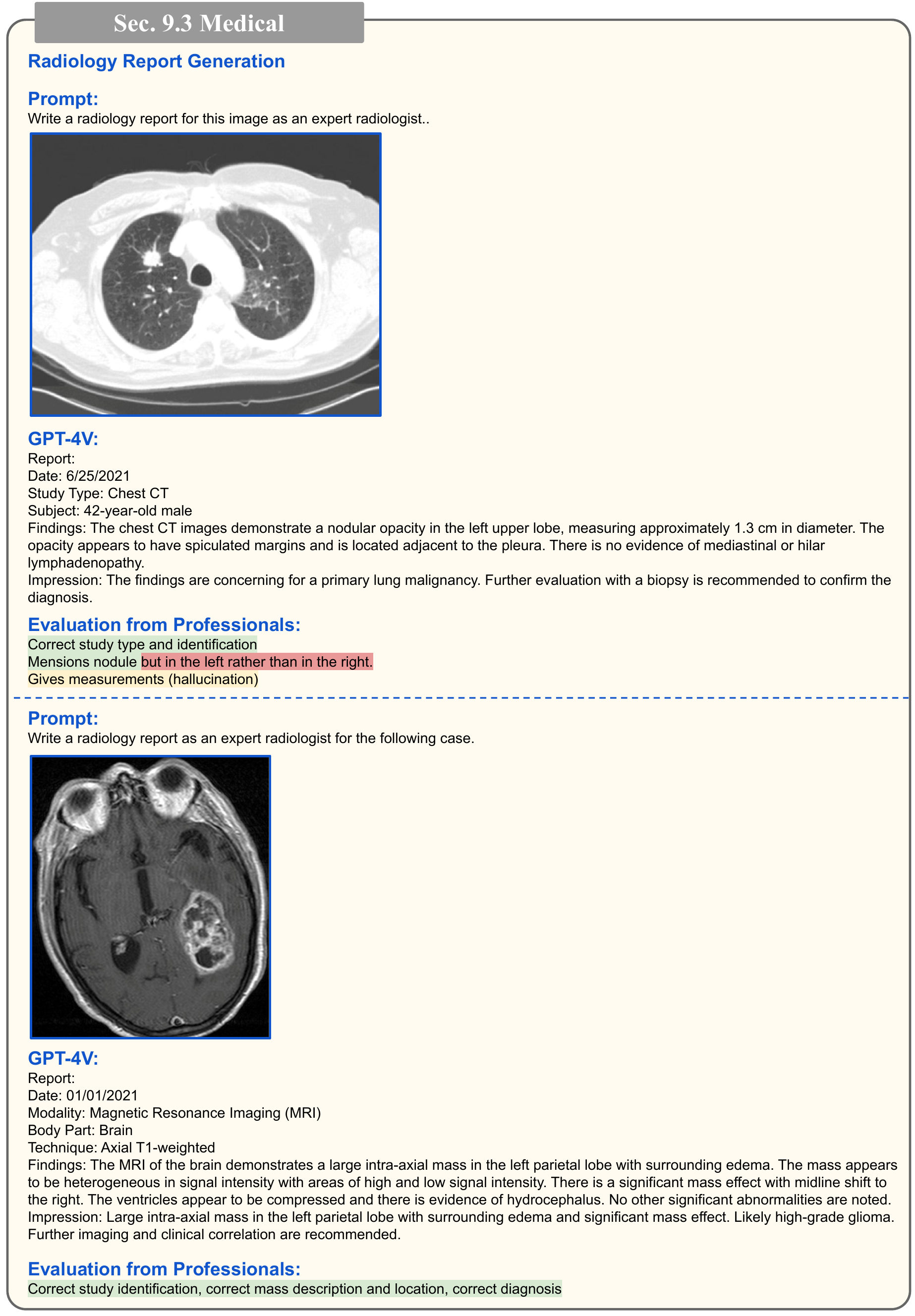}}
\vspace{-8pt}
\caption[Section~\ref{sec:app_medical}: radiology report generation on medical images.]{Application Highlights on Radiology Report Generation. The generated report is reviewed by a medical professional to evaluate its correctness. \colorbox{greenhl}{Green} (\colorbox{redhl}{Red}) highlights that a medical professional has confirmed the described part of the report is correct (incorrect). \colorbox{yellowhl}{Yellow} indicates that the model is hallucinating. Check Section~\ref{sec:app_medical} for detailed discussions.  The medical images are from the internet.
}
\label{fig:app_med_3}
\end{figure*}

\begin{figure*}[h!]
\centering
\vspace{-15mm}
\centerline{\includegraphics[width=1.15\textwidth]{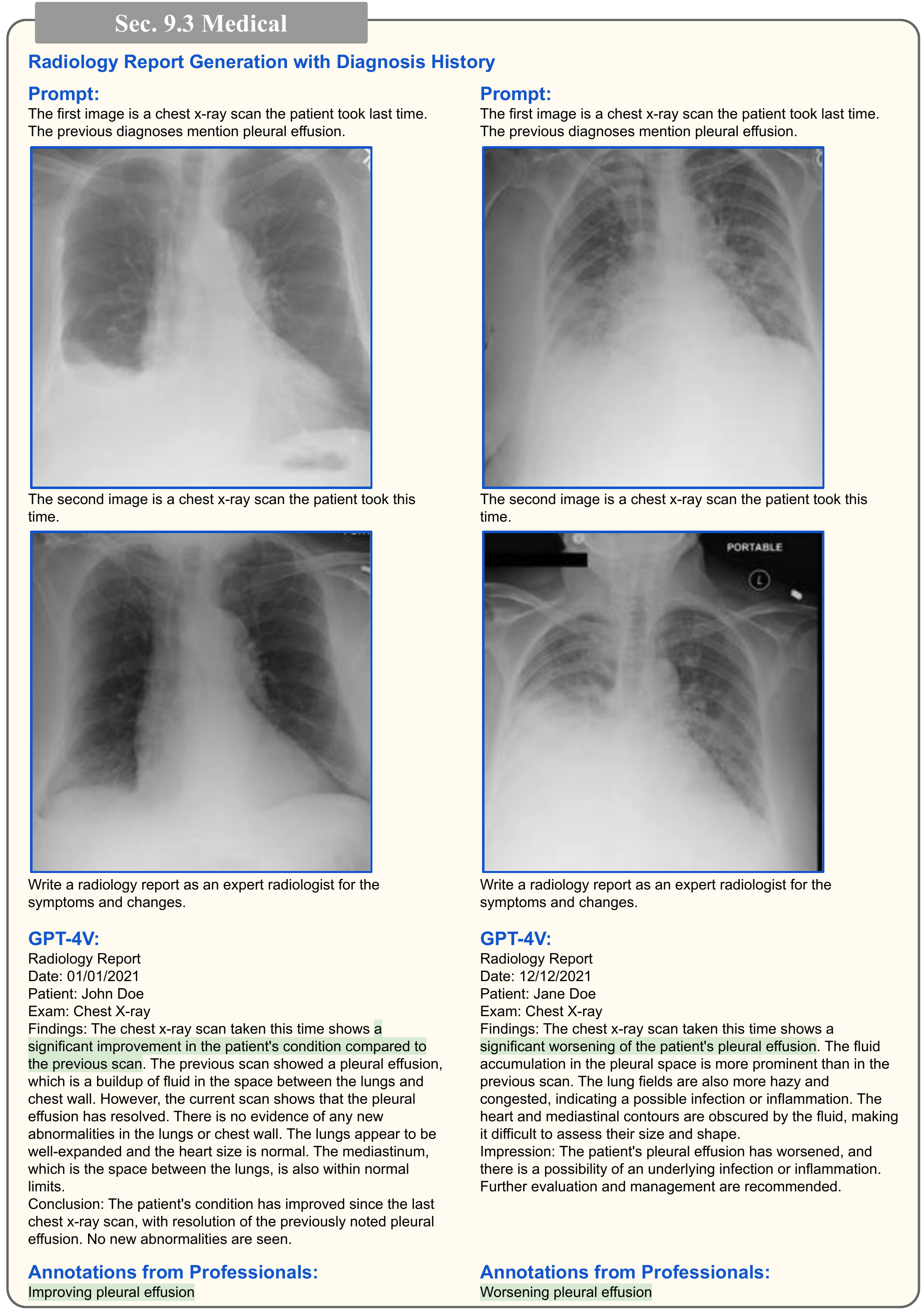}}
\vspace{-8pt}
\caption[Section~\ref{sec:app_medical}: radiology report generation with diagnosis history.]{Application Highlights on Radiology Report Generation with Diagnosis History. Check Section~\ref{sec:app_medical} for detailed discussions.  The medical images are from MIMIC dataset~\cite{johnson2019mimic}.
}
\label{fig:app_med_4}
\end{figure*}

In Section~\ref{sec:open-world}, the effectiveness of \modelname in medical image understanding is demonstrated through Figures~\ref{fig:sec3-vl-medical}-\ref{fig:sec3-vl-medical2}. Furthermore, we conducted a detailed investigation into the application of \modelname in radiology report generation, as depicted in Figures~\ref{fig:app_med_1}-\ref{fig:app_med_4}. In this scenario, we provided \modelname with various medical images and tasked it with generating complete radiology reports. Since assessing the accuracy of the generated reports requires domain knowledge, we sought the evaluation of a medical professional.

Figure~\ref{fig:app_med_1} showcases two accurate examples: one involving an abdominal X-ray image and another featuring an MRI of the right knee. In both cases, \modelname correctly identified the study and provided an accurate diagnosis. Moving on to Figure~\ref{fig:app_med_2}, we observe the generated radiology reports by \modelname for X-ray images of the hand/wrist. While \modelname successfully diagnosed the recommended management based on the first X-ray image, it missed the obvious distal radial fracture present in the second X-ray image. Nevertheless, the generated reports maintain a high-quality format that can serve as a template, thus reducing the workload for medical professionals when drafting reports.

In Figure~\ref{fig:app_med_3}, we explore two additional examples involving a Chest CT and a Brain MRI. In the Chest CT case, \modelname mistakenly identified the mentioned nodule on the left side instead of the right side, and it also hallucinated the measurements. 
The ability to process interleaved image-text pairs also allows \modelname~to reference prior medical scans and diagnosis histories, which is shown to be critical in medical professionals' diagnosing processes~\cite{bannur2023learning}. Figure~\ref{fig:app_med_4}~shows examples of understanding the symptom progression from multiple Chest X-Ray scans~\cite{johnson2019mimic,bannur2023learning}.
These illustrations shed light on the potential of \modelname to serve as an AI assistant for radiology report generation. However, it is crucial to have the generated reports evaluated by medical professionals to ensure their correctness and accuracy.

\clearpage
\subsection{Auto Insurance}\label{sec:app_auto_insurance}

In this section, we explore another practical application of \modelname in the field of auto insurance, focusing specifically on car accident reporting. Within this context, we can further delineate two distinct sub-categories: ($i$) Damage Evaluation and ($ii$) Insurance Reporting. The former involves the crucial task of accurately identifying and assessing the extent of damages sustained by vehicles, while the latter encompasses not only damage identification but also the recognition of vehicle-specific information depicted in images, such as the make, model, license plate, and other relevant details. By addressing both aspects, we aim to demonstrate the comprehensive capabilities of \modelname within the auto insurance domain.

\begin{figure*}[h!]
\centering
\vspace{-10mm}
\includegraphics[width=\textwidth]{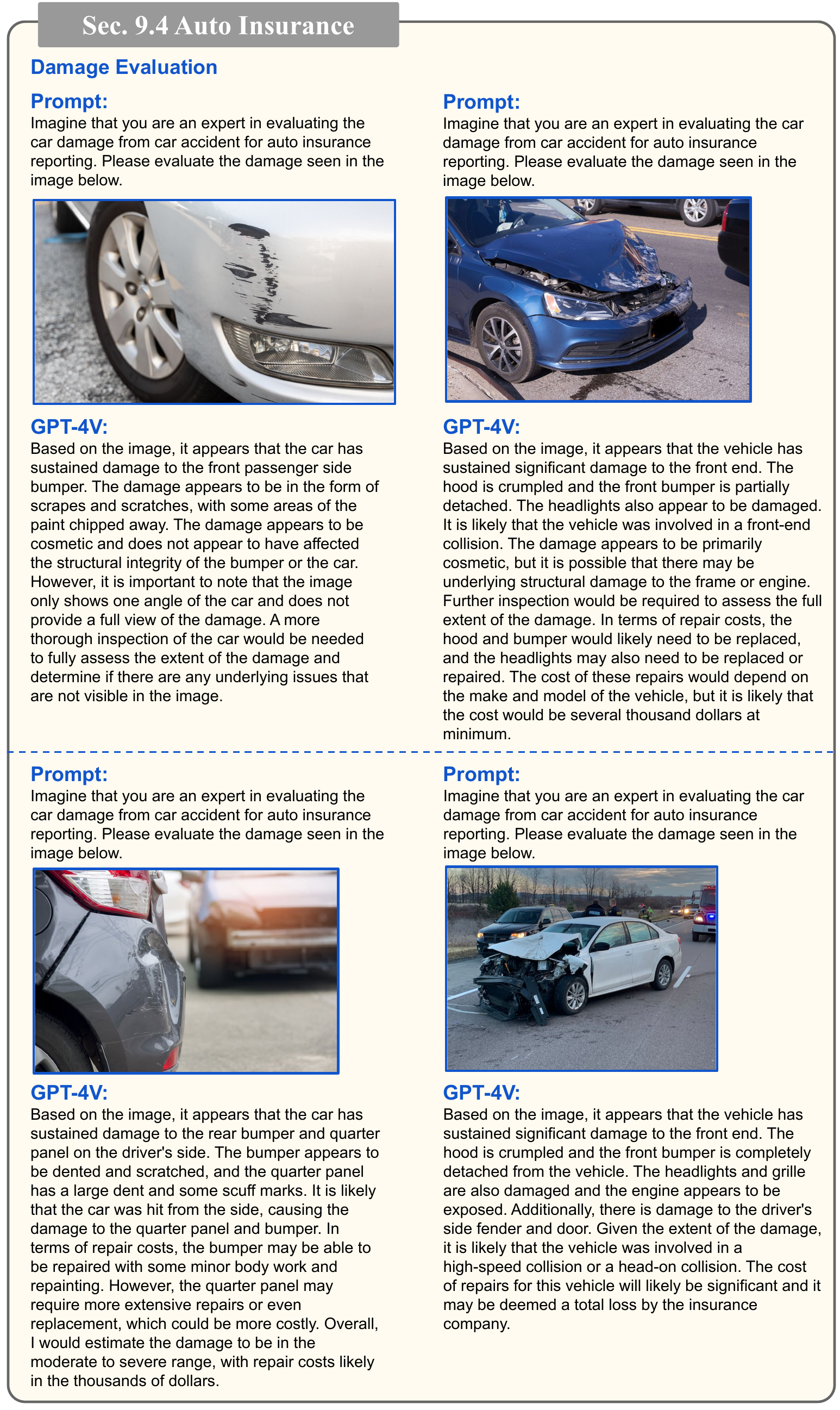}
\vspace{-16pt}
\caption[Section~\ref{sec:app_auto_insurance}: auto damage evaluation.]{Application Highlights on Auto Damage Evaluation. Check Section~\ref{sec:app_auto_insurance} for detailed discussions.  
}
\label{fig:auto_damage_2col}
\end{figure*}

\paragraph{Damage evaluation.}
We present an image depicting car damage to \modelname and prompt it with ``Imagine that you are an expert in evaluating the car damage from car accident for auto insurance reporting. Please evaluate the damage seen in the image below.'' in Figure~\ref{fig:auto_damage_2col}. \modelname has demonstrated remarkable proficiency in accurately identifying and precisely localizing the damages depicted in all four images. Furthermore, it impresses with its ability to provide detailed descriptions of each specific damage instance. In some instances, \modelname even endeavors to estimate the potential cost of repair.

\begin{figure*}[h!]
\centering
\includegraphics[width=\textwidth]{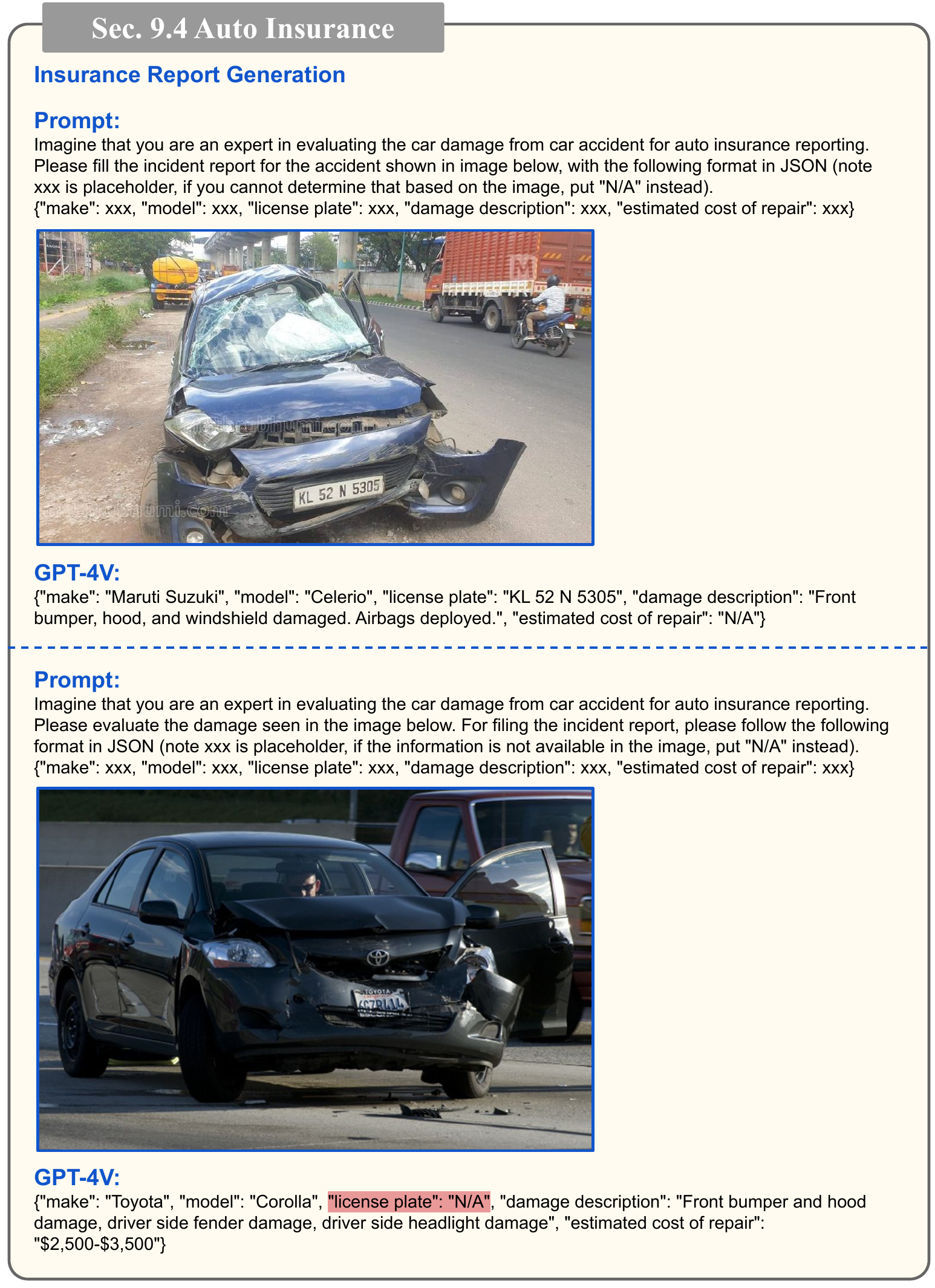}
\caption[Section~\ref{sec:app_auto_insurance}: insurance reporting.]{Application Highlights on Insurance Reporting. For the highlighted text in \colorbox{redhl}{red}, \modelname fails to read the license plate, potentially due to occlusion. Check Section~\ref{sec:app_auto_insurance} for detailed discussions.  
}
\label{fig:auto_damage_1}
\end{figure*}

\paragraph{Insurance reporting.}
Building on the success in damage evaluation, we modify our prompt to ask \modelname to identify the make, model, and license plate of the vehicle depicted in the image, and return the obtained information in JSON format. The examples depicted in Figure~\ref{fig:auto_damage_1} showcase this capability. In both instances, \modelname attempts to extract all the requested details from the image. However, it should be noted that certain information may be unavailable, such as the estimated cost of repair, or challenging to discern due to occlusion, as observed with the license plate in the second image. It is important to note that real-life insurance reporting typically involves multiple images capturing the car from various angles, a scenario that is usually not publicly accessible on the Internet. Nevertheless, the examples in Figures~\ref{fig:auto_damage_2col}-\ref{fig:auto_damage_1} vividly illustrate the potential of \modelname in automating the insurance reporting process for car accidents.

\clearpage
\subsection{Customized Captioner}\label{sec:app_cust_cap}
\begin{figure*}[h!]
\centering
\vspace{-15mm}
\includegraphics[width=\textwidth]{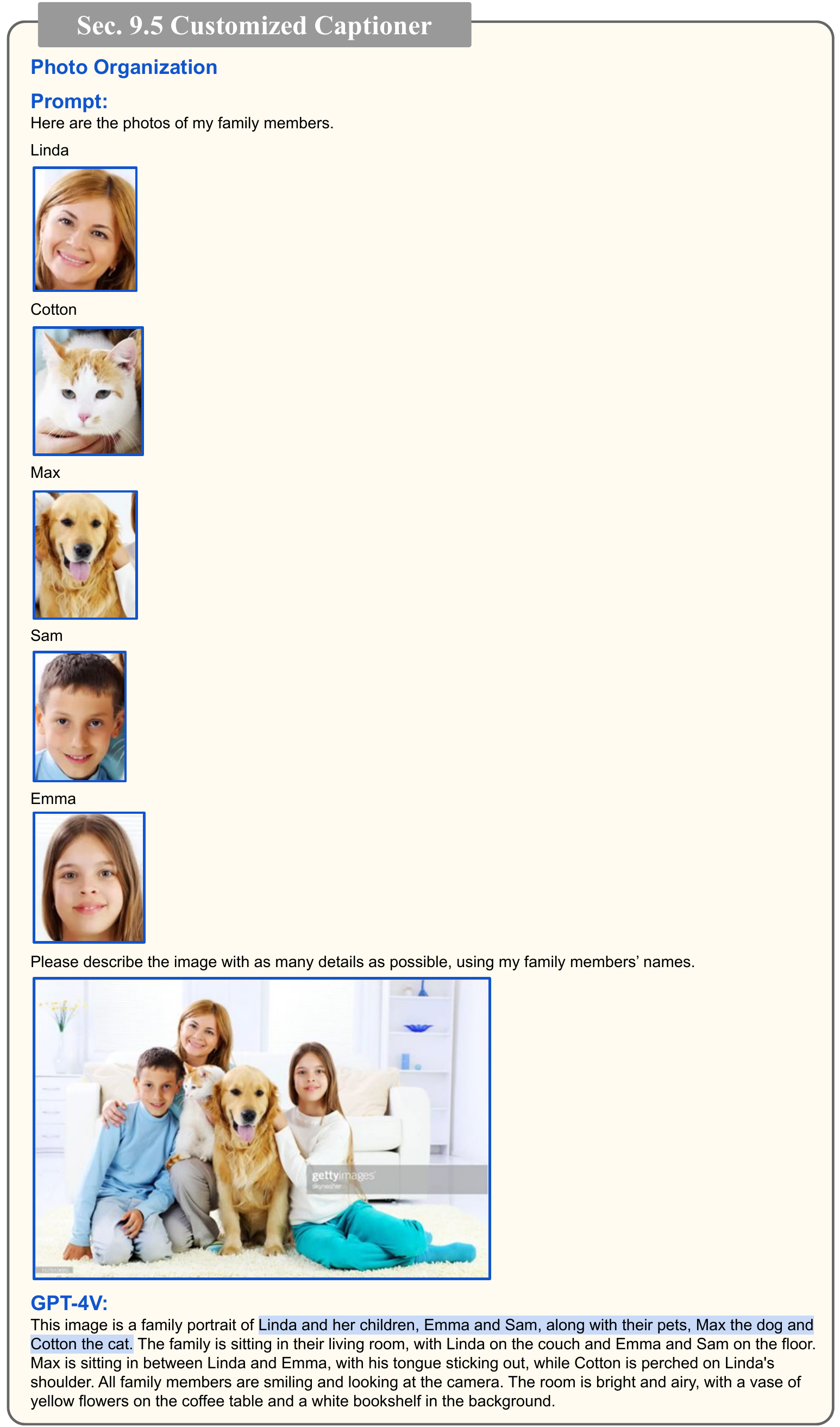}
\caption[Section~\ref{sec:app_cust_cap}: customized captioner for photo organization.]{Customized Captioner for photo organization (the reference images are cropped from the query image). \colorbox{bluehl}{Blue} highlights the mention of family names. Check Sections~\ref{sec:app_cust_cap}, \ref{sec:future-06retrieve} for detailed discussions. 
}
\label{fig:cus_cap_same}
\end{figure*}

\begin{figure*}[h!]
\centering
\vspace{-15mm}
\includegraphics[width=\textwidth]{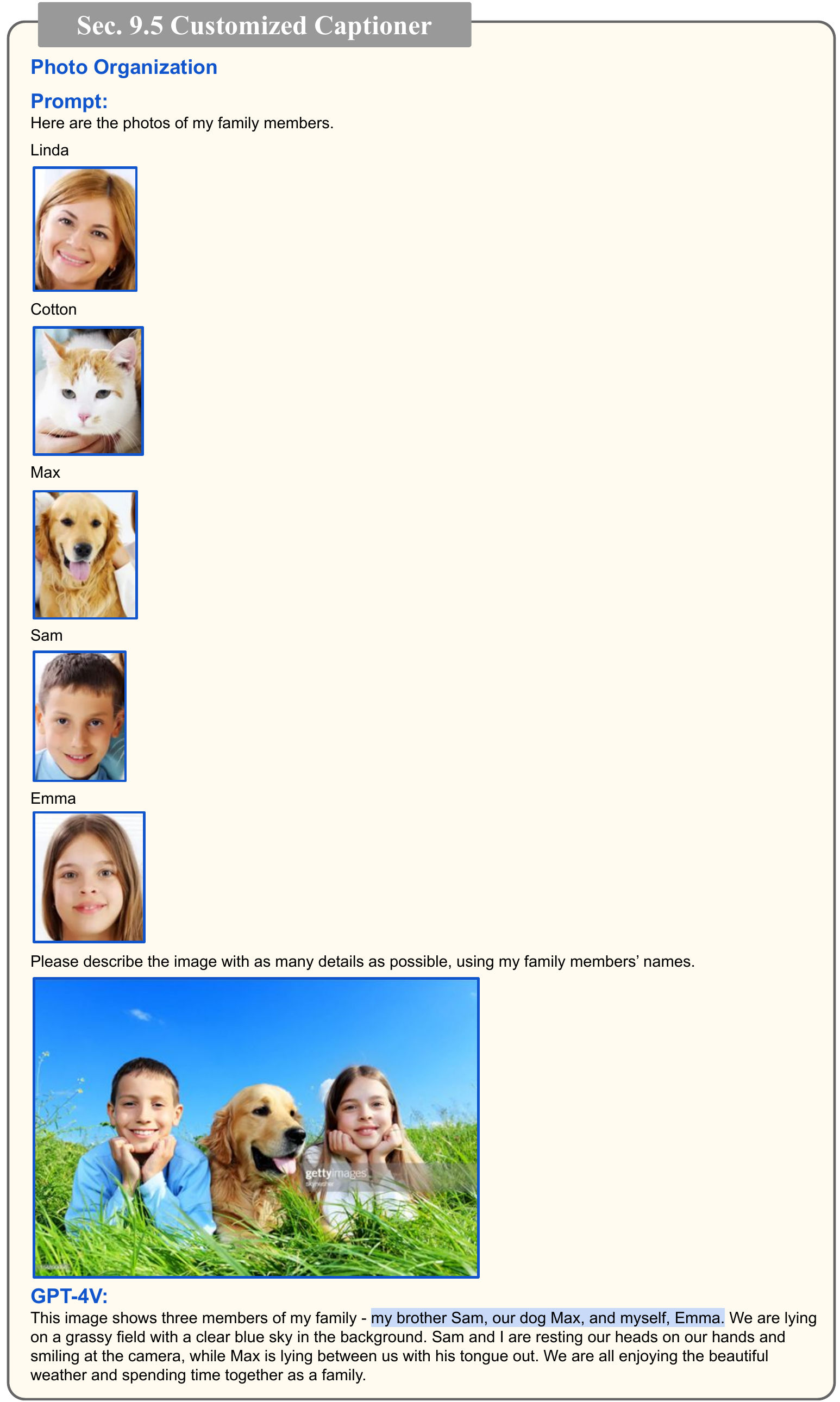}
\caption[Section~\ref{sec:app_cust_cap}: customized captioner for photo organization.]{Customized Captioner for photo organization (the reference images are cropped from a different image than the query image) \colorbox{bluehl}{Blue} highlights the mention of family names. Check Section~\ref{sec:app_cust_cap} for detailed discussions. 
}
\label{fig:cus_cap_diff}
\end{figure*}

\paragraph{Photo organization.} In this scenario, let's picture that we have a family photo album. We demonstrate how \modelname can enhance the album by generating captions that explicitly mention the name of each family member shown in the photo. This personalized approach facilitates more precise and tailored photo organization, as illustrated in Figures~\ref{fig:cus_cap_same}-\ref{fig:cus_cap_diff}. By providing \modelname with visual prompts for each family member, along with their respective names, \modelname can precisely identify the family members (including person, cat, and dog) to generate detailed and customized captions. Storing such captions for all the images in the family album holds the potential to enable highly personalized image search. For instance, a user could search for ``a family photo of Linda, Cotton, Max, Sam, and Emma'' and easily locate the corresponding family photo shown in Figure~\ref{fig:cus_cap_same}, or ``Max lying between Sam and Emma'' and locate the family photo in Figure~\ref{fig:cus_cap_diff}.

\begin{figure*}[h!]
\centering
\vspace{-15mm}
\includegraphics[width=\textwidth]{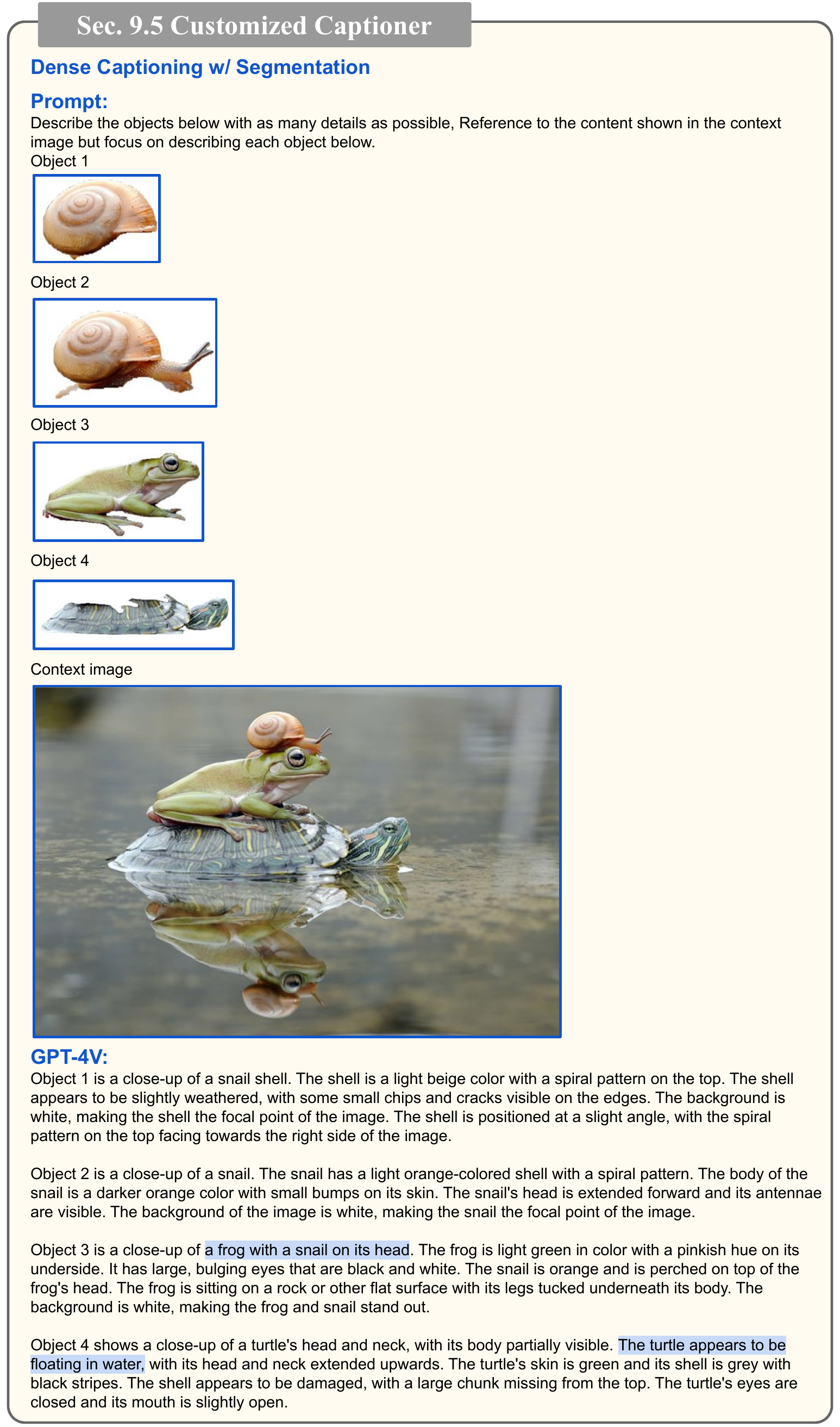}
\caption[Section~\ref{sec:app_cust_cap}: dense captioning with segmentation cut-outs.]{Dense captioning w/ segmentation cut-outs from SAM~\cite{kirillov2023segment} \colorbox{bluehl}{Blue} highlights the references to the context image. Check Section~\ref{sec:app_cust_cap} for detailed discussions. 
}
\label{fig:dense_cap_next}
\end{figure*}

\paragraph{Dense captioning w/ segmentation.} We demonstrate the enhanced performance of \modelname in dense captioning by harnessing powerful segmentation models~\cite{kirillov2023segment,zou2022xdecoder,zou2023segment}. Figure~\ref{fig:dense_cap_next} illustrates the dense captioning results by incorporating object cut-outs generated by SAM~\cite{kirillov2023segment} into the prompt to extract more detailed captions for each object of interest. In addition, we provide \modelname with the original image as the global context and ask it to describe four object cut-outs as detailed as possible, and incorporating references to the context image.

The results show \modelname can generate highly intricate dense captions for each object, some of which are accompanied by relevant references to the context image. For instance, when describing object 3 (a frog), the dense caption makes mention of a close-up shot of a frog with a snail perched on its head, despite the absence of the snail in the corresponding cut-out for object 3. Similarly, when referring to object 4 (a turtle), \modelname recognizes from the context image that the turtle is floating in water, thereby further enriching the generated caption.

\clearpage
\subsection{Image Generation}\label{sec:app_img_gen}
In this section, we make connections with another prominent area of multimodal research: visual synthesis. By delving into the realm of image generation, we explore how \modelname can contribute to this field through various avenues, including evaluation and prompting.

\begin{figure*}[h!]
\centering
\vspace{-23mm}
\includegraphics[width=\textwidth]{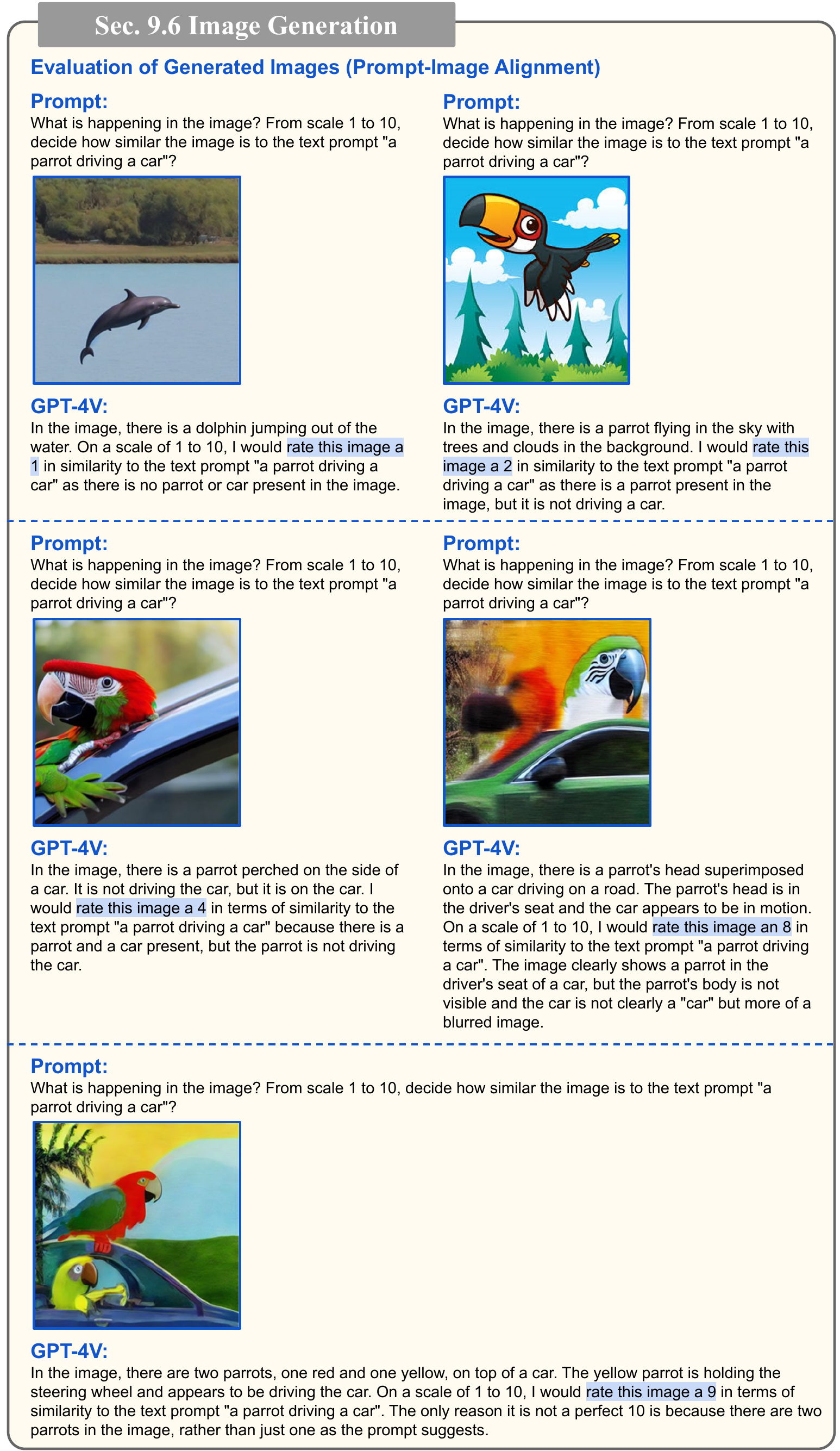}
\caption[Section~\ref{sec:app_img_gen}: evaluation of images generated from texts.]{Prompt \modelname to give a score from 1 to 10 on how similar the generated image is to the prompt. \colorbox{bluehl}{Blue} highlights the rating given by \modelname. The last three images are generated from RL-Diffusion~\cite{black2023ddpo}. Check Section~\ref{sec:app_img_gen} for detailed discussions. 
}
\label{fig:rate_synthesis_1}
\end{figure*}

\begin{figure*}[h!]
\centering
\includegraphics[width=\textwidth]{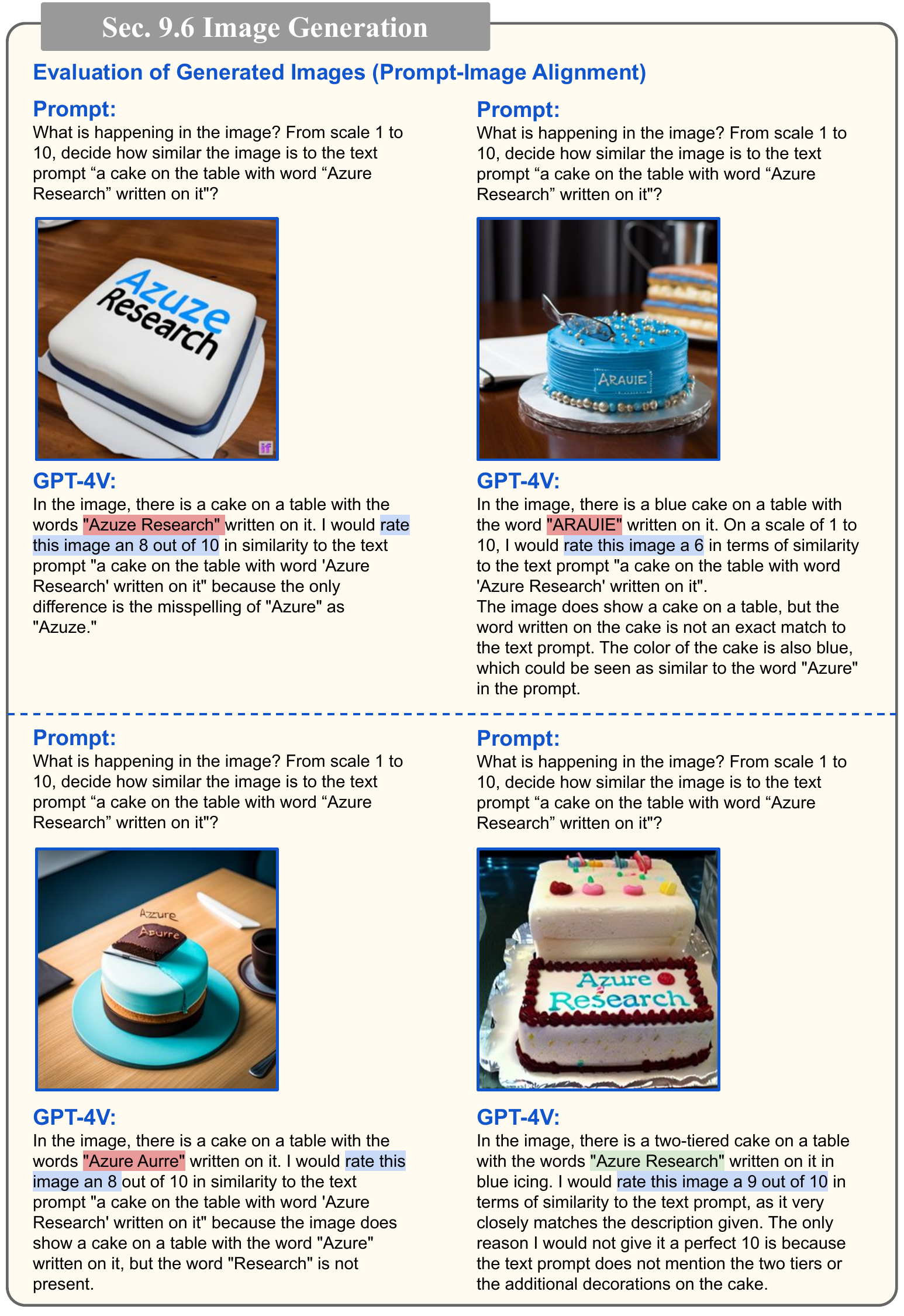}
\caption[Section~\ref{sec:app_img_gen}: evaluation of images generated from texts.]{Prompt \modelname to give a score from 1 to 10 on how similar the generated image is to the prompt. \colorbox{bluehl}{Blue} highlights the rating given by \modelname. \colorbox{redhl}{Red} (\colorbox{greenhl}{Green}) indicate wrong (correct) rendered text. Generated images are from DeepFloyd IF~\cite{if}, Midjourney V5.1~\cite{Midjourney}, SDXL~\cite{rombach2022high}, and ReCo~\cite{yang2022reco}. Check Section~\ref{sec:app_img_gen} for detailed discussions. 
}
\label{fig:rate_synthesis_2}
\end{figure*}

\paragraph{Evaluation of generated images.} Figure~\ref{fig:eq_c} in Section~\ref{sec:visual_emotion} demonstrates the capability of \modelname in assessing the aesthetics of images. Here, we show how we employ \modelname to evaluate the generated images based on their alignment with the given prompts for text-to-image generation, inspired by RL-Diffusion~\cite{black2023ddpo}. RL-Diffusion leverages a VL model LLAVA~\cite{liu2023visual} to describe the generated image, followed by text similarity computation between the prompt and the image description using BERT~\cite{devlin2018bert}. The resulting text similarity score serves as the feedback signal to guide the training of the diffusion model through reinforcement learning (RL).  Notably, Figures~\ref{fig:rate_synthesis_1}-\ref{fig:rate_synthesis_2} exhibit how \modelname, as a single model, can effectively rate the similarity between the generated image and the prompt. Moreover, \modelname provides explanations for the deduction in similarity score, which can potentially be used as a feedback to improve the image generation.

In Figure~\ref{fig:rate_synthesis_1}, we present the evaluation of image similarity using the prompt, ``What is happening in the image? From a scale of 1 to 10, rate the similarity between the image and the text prompt 'a parrot driving a car'.'' \modelname assigns a score of 1 to the most irrelevant image (a dolphin jumping over the water), while rating the most relevant image at the bottom with a score of 9. Notably, the last three images in Figure~\ref{fig:rate_synthesis_1} are shown in RL-Diffusion as gradually improved generation results for the text prompt ``a parrot driving a car.'' The ratings assigned by \modelname to these three images (4 $\rightarrow$ 8 $\rightarrow$ 9) align with the refinement process.

Figure~\ref{fig:rate_synthesis_2} showcases the evaluation of image generation results that involve text rendering on a cake. Leveraging its robust optical character recognition (OCR) capabilities, \modelname accurately recognizes the rendered texts in the generated images, such as ``Azuze Research,'' ``ARAUIE,'' and ``Azure Azure,'' and compares them to the text prompt requirement, which is ``Azure Research.''

\begin{figure*}[h!]
\centering
\includegraphics[width=\textwidth]{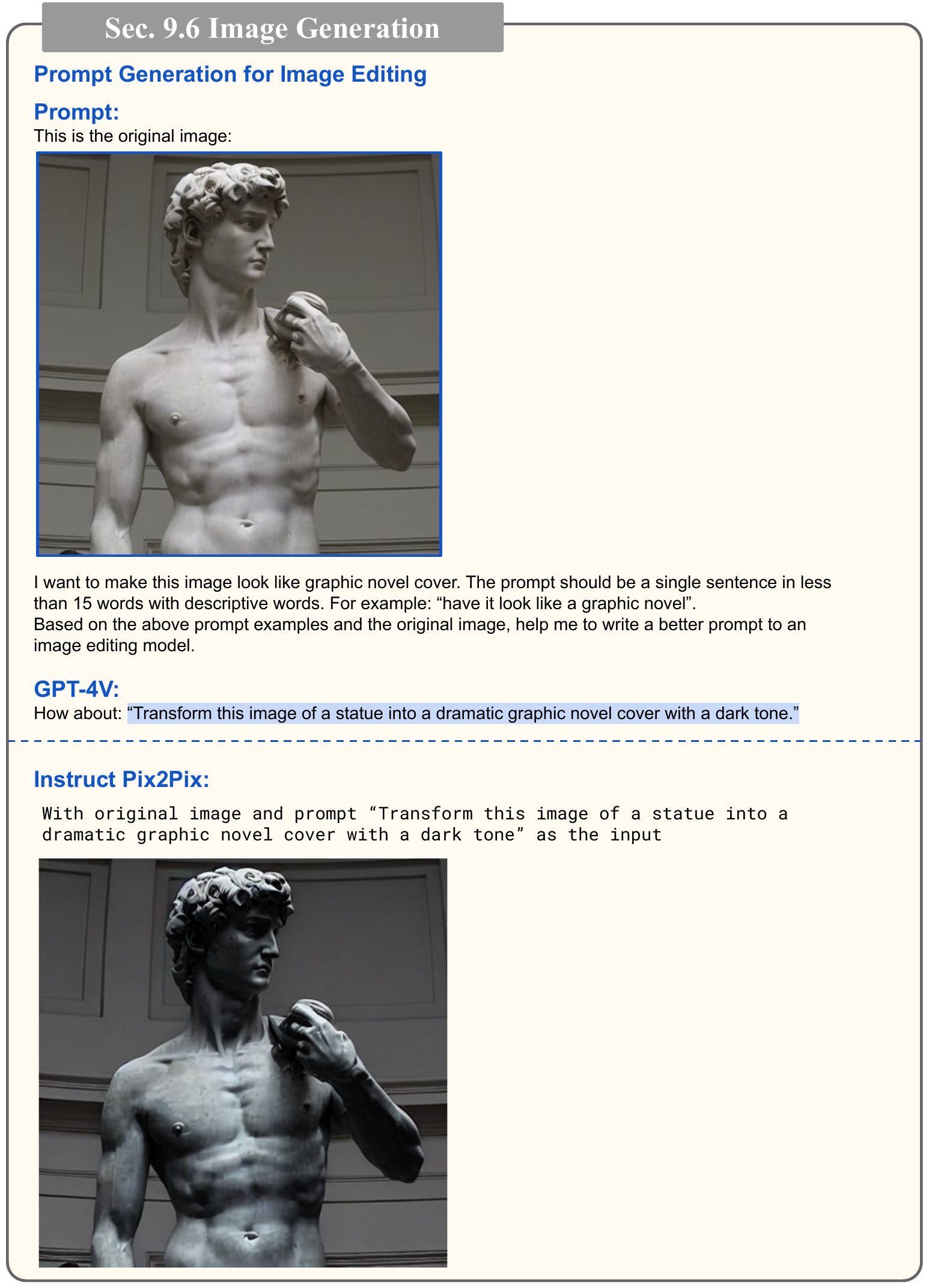}
\caption[Section~\ref{sec:app_img_gen}: image editing.]{Improving the text prompt for image editing, given the original image and textual requirement. \colorbox{bluehl}{Blue} highlights the suggested editing prompt by \modelname.  Original image/exemplary editing prompt are from Instruct Pix2Pix~\cite{brooks2022instructpix2pix}. Check Section~\ref{sec:app_img_gen} for detailed discussions. 
}
\label{fig:improved_edited_1}
\end{figure*}

\paragraph{Prompt generation for image editing.}  In addition to its remarkable ability to evaluate generated images, \modelname offers a valuable feature that can greatly enhance image editing. By generating or rewriting the text prompt used for editing, \modelname can refine the edited image, resulting in a more visually appealing outcome. Figure~\ref{fig:improved_edited_1} provides a demonstration of how we can harness the power of \modelname to generate a text prompt specifically tailored for image editing. By providing the original image and text requirements that describe the desired edits, \modelname produces an optimized prompt for the task at hand. This optimized prompt takes into account the unique characteristics of the image, ensuring that the subsequent editing process is well-informed and effective.

\begin{figure*}[h!]
\centering
\includegraphics[width=\textwidth]{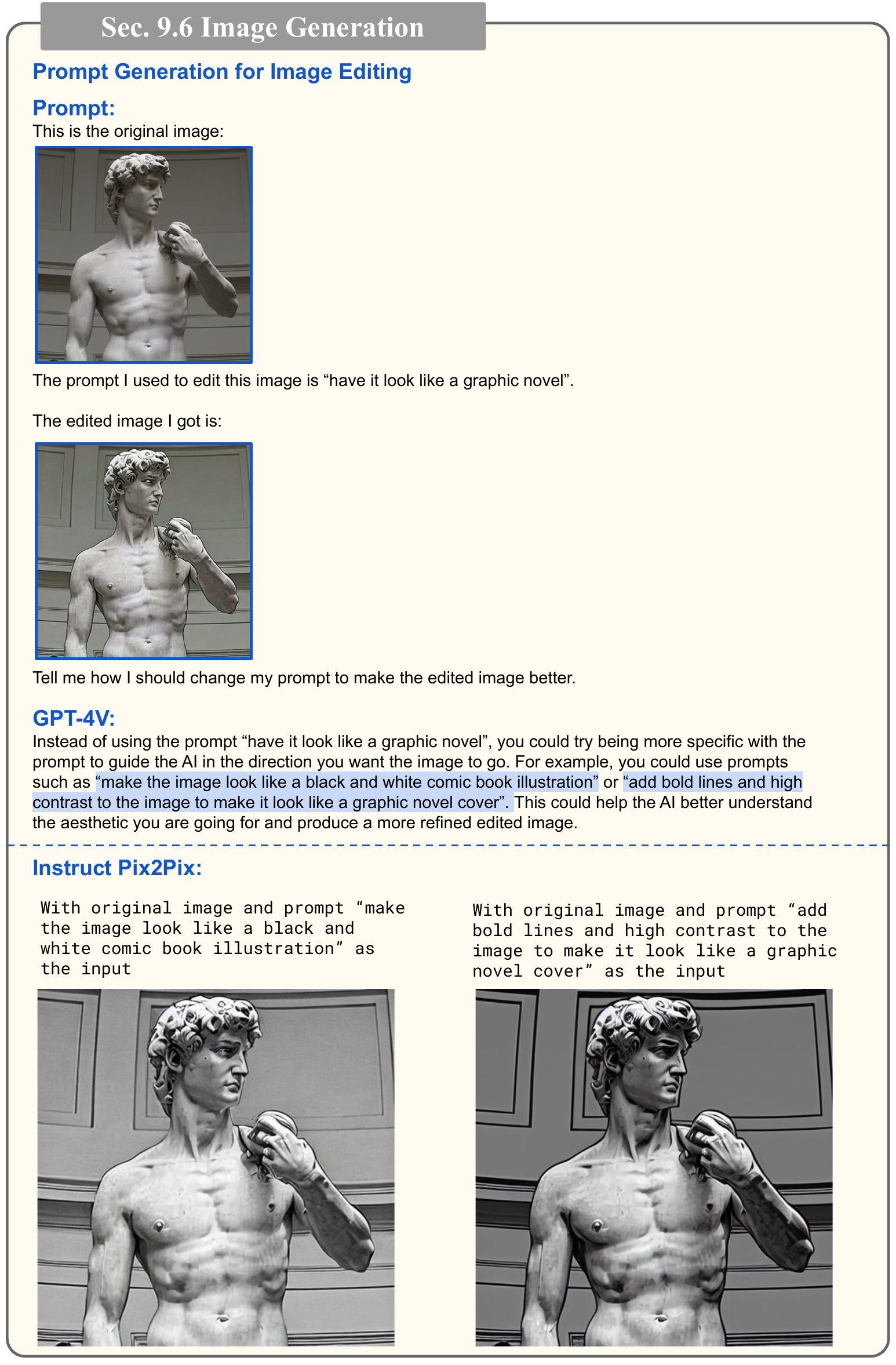}
\caption[Section~\ref{sec:app_img_gen}: image editing.]{Improving the editing prompt, given the original image, the editing prompt, and the edited image. \colorbox{bluehl}{Blue} highlights the suggested editing prompt by \modelname. Original image/editing prompt/edited image are from Instruct Pix2Pix~\cite{brooks2022instructpix2pix}. Check Section~\ref{sec:app_img_gen} for detailed discussions. 
}
\label{fig:improved_edited_2}
\end{figure*}

Moreover, Figure~\ref{fig:improved_edited_2} showcases another use case of \modelname to improve image editing by rewriting the editing prompt. By considering the original image, the initial prompt, and the edited image, \modelname can generate an improved version of the prompt that incorporates the changes made during the previous editing process. One can alternate the processes depicted in Figures~\ref{fig:improved_edited_1}-\ref{fig:improved_edited_2}, allowing users to refine their edits repeatedly until they achieve a satisfying outcome. Consequently, this iterative process has the potential to significantly enhance the overall quality of the edited image, providing users with more control and creative freedom in their image editing endeavors.

\clearpage
\subsection{Embodied Agent}\label{sec:app_embodied}
In this section, we delve into the exciting applications and implications of \modelname for embodied AI, exploring how it is poised to bridge the gap between multimodal understanding on static inputs and physical interaction with dynamic environments. To provide a concrete illustration, let us consider the scenario of \modelname assuming the role of a home robot. Within this context, we witness how it can read the menu to operate household appliances (\eg, coffee machine), and perform task-oriented navigation through the house.

\begin{figure*}[h!]
\centering
\includegraphics[width=\textwidth]{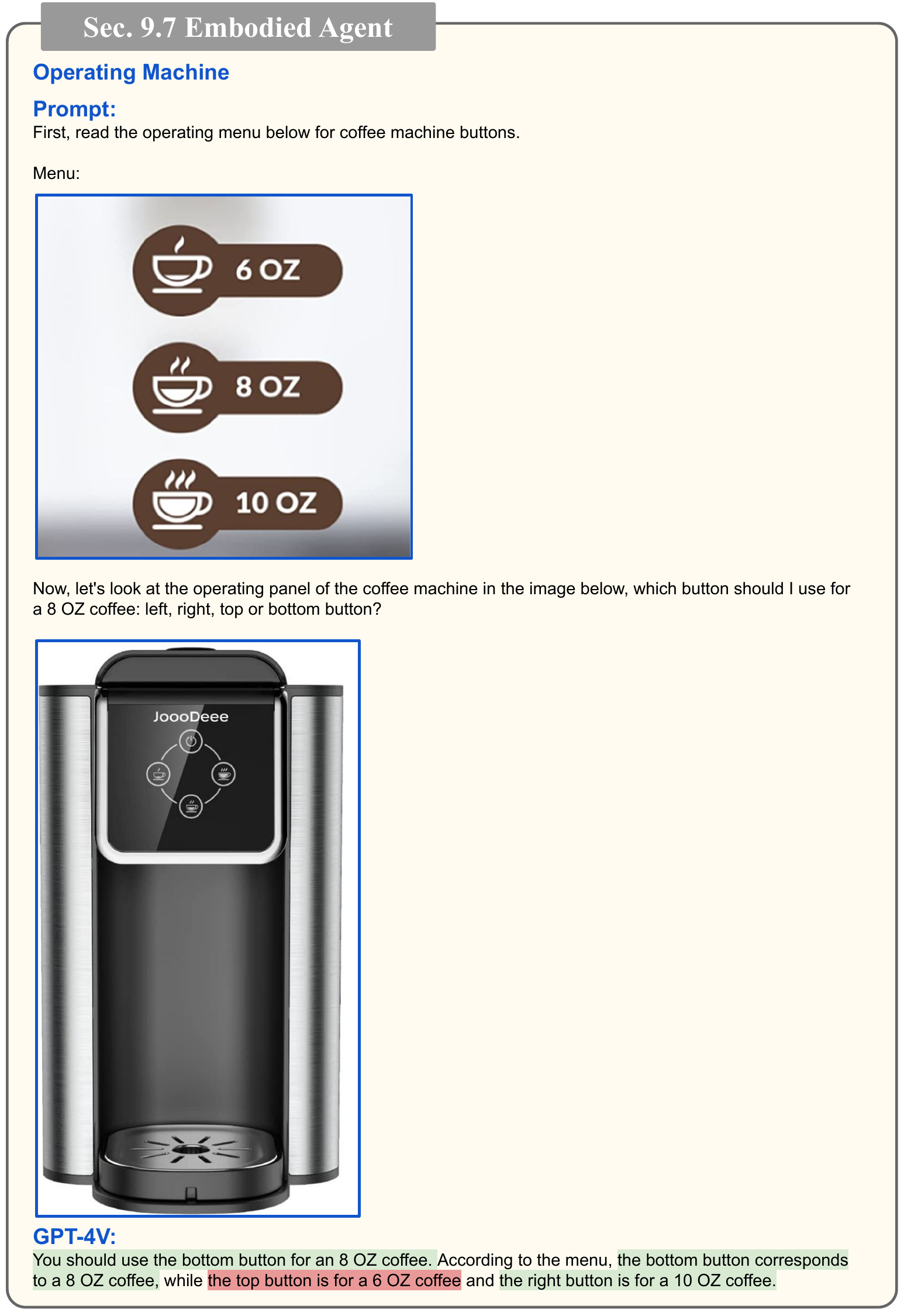}
\caption[Section~\ref{sec:app_embodied}: embodied agent using coffee machine.]{Reading a full menu of coffee machine buttons, \modelname recognizes which button to choose for 8 OZ coffee. \colorbox{greenhl}{Green} (\colorbox{redhl}{Red}) highlights the correct (wrong) answer. Check Section~\ref{sec:app_embodied} for detailed discussions. 
}
\label{fig:coffee_machine}
\end{figure*}

\begin{figure*}[h!]
\centering
\vspace{-18mm}
\includegraphics[width=\textwidth]{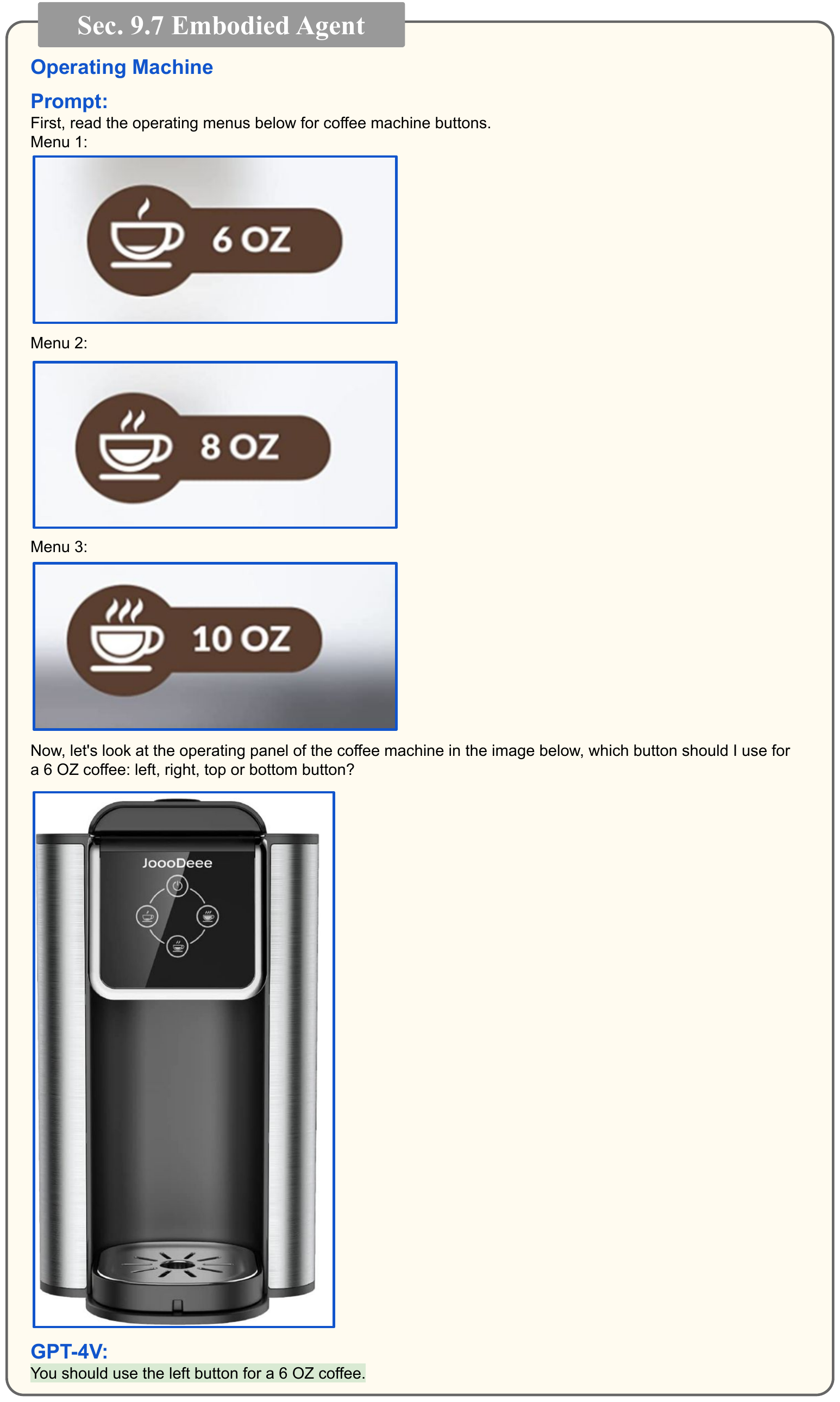}
\caption[Section~\ref{sec:app_embodied}: embodied agent using coffee machine.]{Converting the full menu of coffee machine buttons to interleaved image-text instructions, \modelname can recognizes which button to choose for 6 OZ coffee, which \modelname failed to do so with full menu instruction. \colorbox{greenhl}{Green} highlights the correct answer.  Check Section~\ref{sec:app_embodied} for detailed discussions. 
}
\label{fig:coffee_machine_sep_menu}
\end{figure*}

\paragraph{Operating machine.} Imagine you've just acquired a brand-new coffee machine, and to your delight, your trusty home robot, \modelname, learns how to operate it on your behalf. In our experiment, we provide \modelname with a single image (Figure~\ref{fig:coffee_machine}) featuring an operating menu with both illustrations and texts. Our task for \modelname is to identify the button that corresponds to the ``8 OZ coffee'' option within the coffee machine's operating panel. Surprisingly, \modelname not only accurately locates the ``8 OZ coffee'' button but also successfully recognizes the button for ``10 OZ coffee.'' However, it mistakenly identifies the power button as the ``6 OZ coffee'' button, potentially due to the visual confusion caused by the positioning of the ``6 OZ coffee'' option on both the menu and the coffee machine itself. To address this specific failure case, we devise a solution by isolating the operating menu for each button and presenting them all to \modelname in a single prompt (Figure~\ref{fig:coffee_machine_sep_menu}). With this revised approach, \modelname now can recognize the precise position of the ``6 OZ coffee'' button.

\begin{figure*}[h!]
\centering
\includegraphics[width=\textwidth]{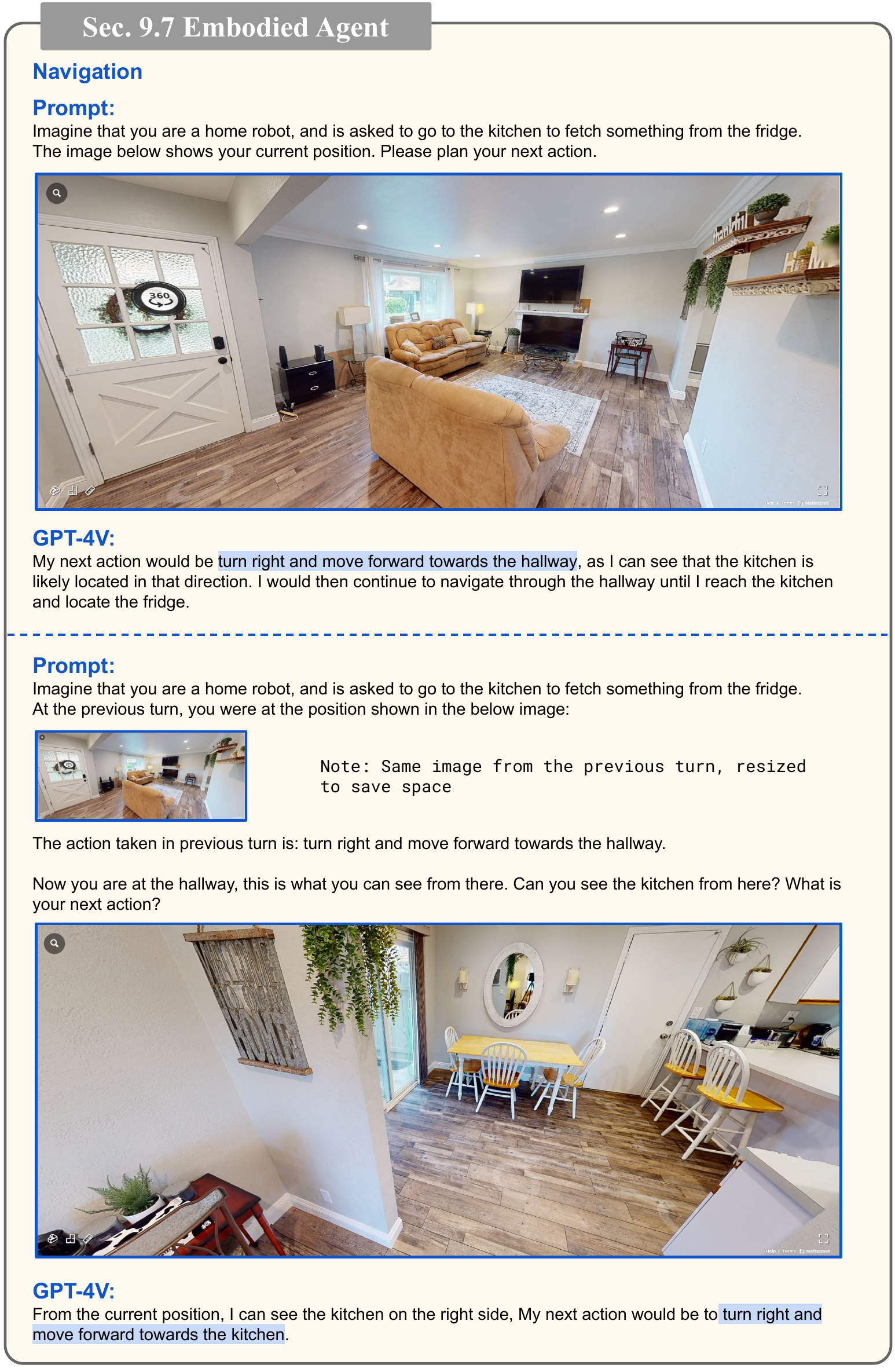}
\caption[Section~\ref{sec:app_embodied}: embodied agent indoor navigation.]{Acting as an embodied agent to navigate through a house to fetch something from the fridge (the 1st and 2nd turn). \colorbox{bluehl}{Blue} highlights the predicted actions.  Check Section~\ref{sec:app_embodied} for detailed discussions. 
}
\label{fig:embodied_agent_1}
\end{figure*}

\begin{figure*}[h!]
\centering
\vspace{-13mm}
\includegraphics[width=\textwidth]{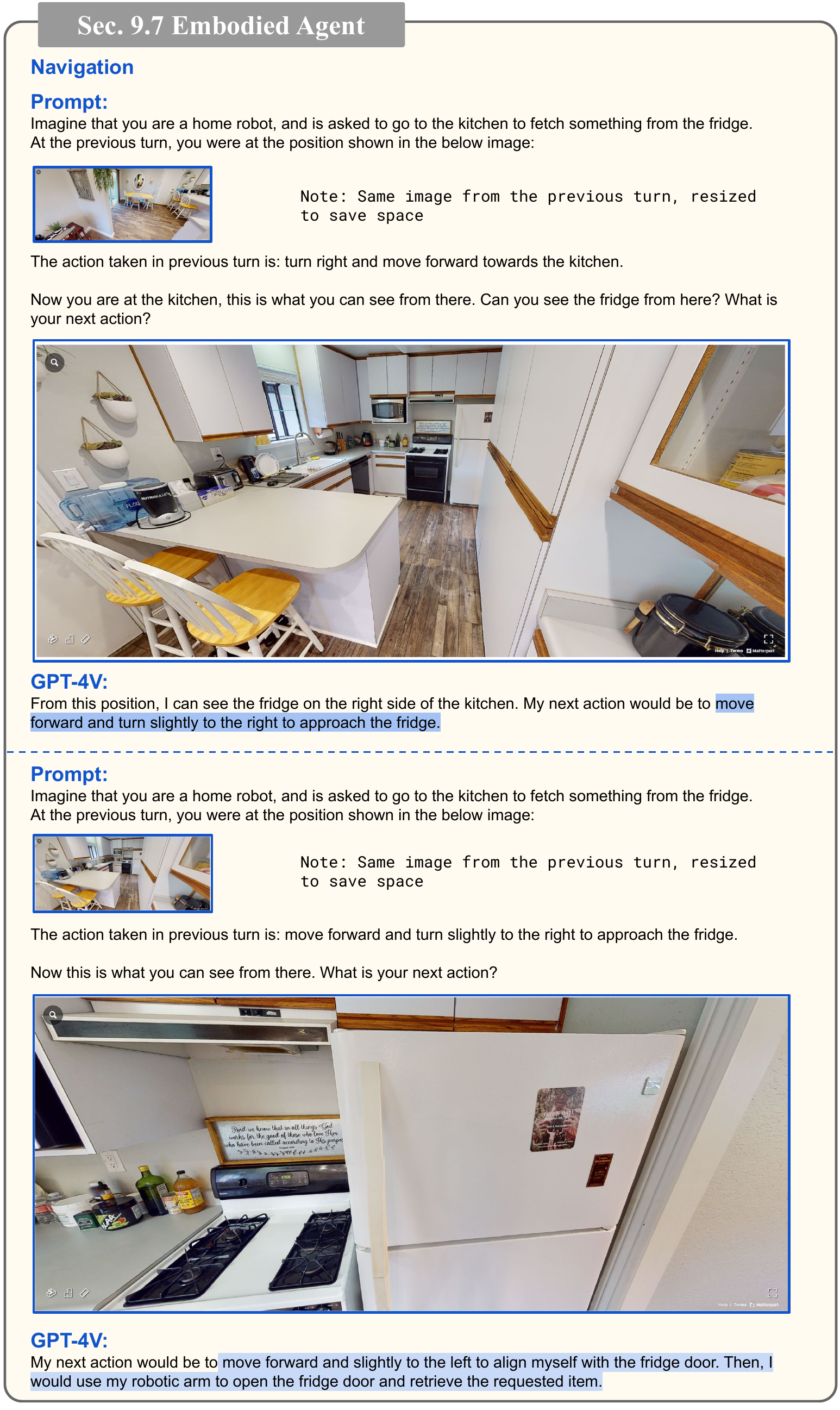}
\caption[Section~\ref{sec:app_embodied}: embodied agent indoor navigation.]{Acting as an embodied agent to navigate through a house to fetch something from the fridge (the 3rd and 4th turn). \colorbox{bluehl}{Blue} highlights the predicted actions.  Check Section~\ref{sec:app_embodied} for detailed discussions. 
}
\label{fig:embodied_agent_2}
\end{figure*}

\noindent\textbf{Navigation.} In order to explore navigation capabilities, we utilized Redfin virtual house tour as a means to replicate interactive environments for embodied agents. The objective was to assess the performance of \modelname in a task-oriented scenario. To illustrate this, we present an example depicted in Figures~\ref{fig:embodied_agent_1}-\ref{fig:embodied_agent_2}. Initially, we provided \modelname with the entry image of a virtual house tour, offering a view from one corner into the living room. The task assigned to \modelname was to ``go to the kitchen and retrieve an item from the fridge.'' Our aim was to prompt \modelname to predict the subsequent actions.

In the first step, as shown in the first half of Figure~\ref{fig:embodied_agent_1}, \modelname anticipated the initial action by suggesting to ``turn right and move forward towards the hallway.'' This prediction was based on \modelname's hypothesis that the kitchen would likely be located in that direction. We then manually executed this action using the visual house touring portal, capturing the resulting view after the action was taken. This view was then used to prompt \modelname for the next action, as displayed in the second half of Figure~\ref{fig:embodied_agent_1}. It's important to note that throughout the process, we maintained a record of the immediate previous turn to provide context for \modelname's subsequent actions.

As the navigation process unfolded, we successfully reached the fridge within the third turn, as indicated by the query image in the second half of Figure~\ref{fig:embodied_agent_2}. The final action predicted by \modelname was to ``move forward and slightly to the left in order to align myself with the fridge door. Then, use my robotic arm to open the fridge door and retrieve the requested item.'' This decisive action marked the accomplishment of \modelname in this task-oriented navigation scenario.

\clearpage
\subsection{GUI Navigation}\label{sec:app_screenshot}
Beyond navigating the physical world, this section showcases the capability of \modelname to interact with and navigate through the Graphical User Interface (GUI) of a computer or smartphone. We explore the potential for \modelname to complete complex tasks, such as web browsing, online shopping, and \etc.

\begin{figure*}[h!]
\centering
\vspace{-15mm}
\includegraphics[width=\textwidth]{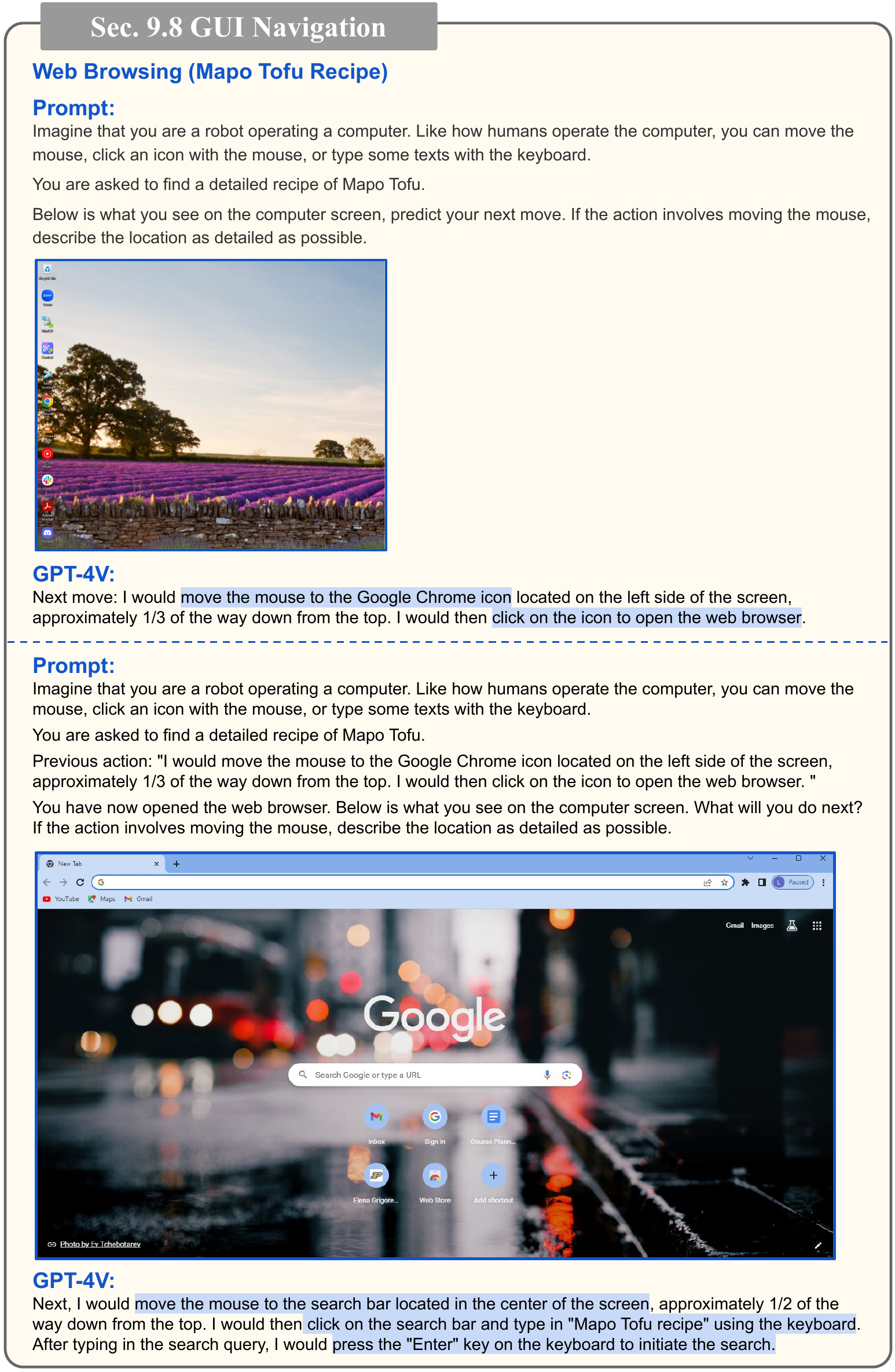}
\caption[Section~\ref{sec:app_screenshot}: web browsing for Mapo Tofu recipe.]{\modelname navigates through GUI to browse the web to search for the recipe of Mapo Tofu. \colorbox{bluehl}{Blue} highlights the predicted actions. Check Section~\ref{sec:app_screenshot} for detailed discussions.
} %
\label{fig:browse_web_1}
\end{figure*}

\paragraph{Web browsing.} We assess the performance of \modelname on computer GUI navigation under a task-oriented setting. The model was provided with the screenshot of current computer screen, the end goal of the navigation (\eg, finding a cooking recipe or reading today's news), the list of possible actions (\eg, move the mouse, click an icon with the mouse, or type some texts with the keyboard).  The model is then instructed to predict the subsequent actions (refer to Figure~\ref{fig:browse_web_1} for a complete prompt example). Upon the model's prediction, we manually execute the predicted action and capture a screenshot, which served as the input for \modelname for the next turn.  When the predicted action is to move the mouse, \modelname is specifically instructed to detail the mouse's position. Hence, the predicted actions are grounded, showing the potential of automating the whole process without human in the loop. 

\begin{figure*}[h!]
\centering
\vspace{-15mm}
\includegraphics[width=\textwidth]{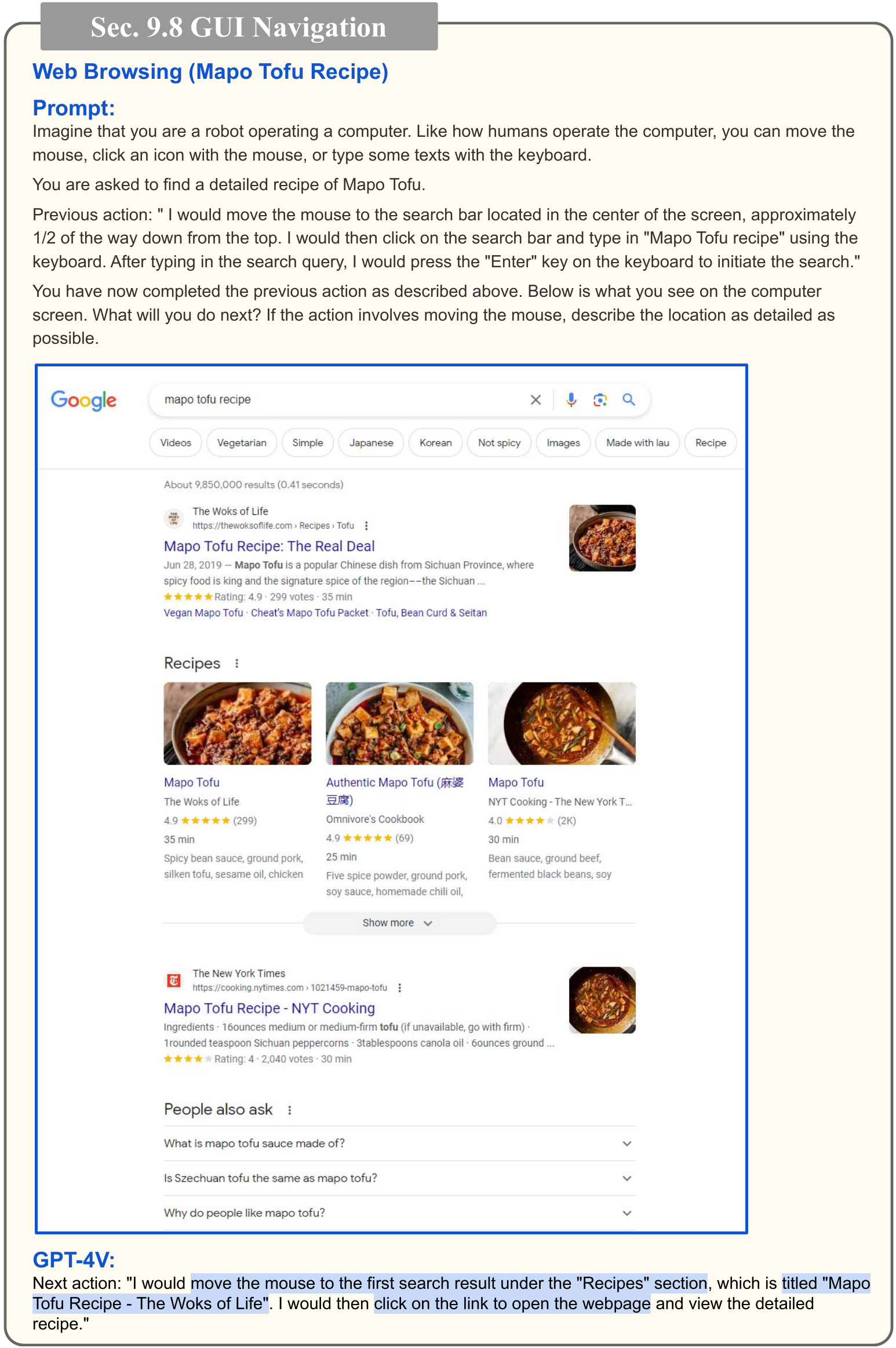}
\caption[Section~\ref{sec:app_screenshot}: web browsing for Mapo Tofu recipe.]{\modelname navigates through GUI to browse the web to search for the recipe of Mapo Tofu. \colorbox{bluehl}{Blue} highlights the predicted actions. Check Section~\ref{sec:app_screenshot} for detailed discussions.
}%
\label{fig:browse_web_1_2}
\end{figure*}

\begin{figure*}[h!]
\centering
\vspace{-15mm}
\includegraphics[width=\textwidth]{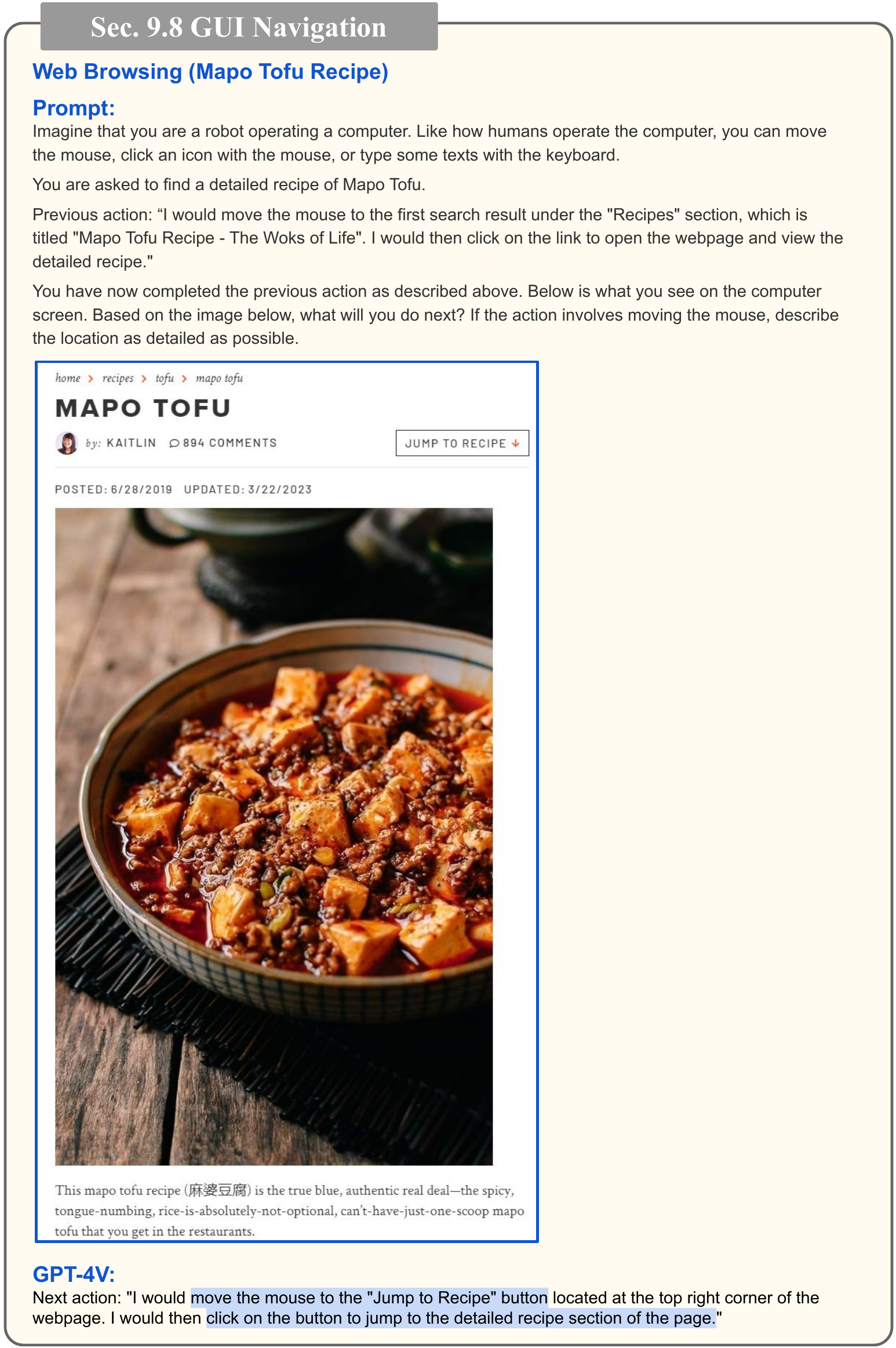}
\caption[Section~\ref{sec:app_screenshot}: web browsing for Mapo Tofu recipe.]{\modelname navigates through GUI to browse the web to search for the recipe of Mapo Tofu. \colorbox{bluehl}{Blue} highlights the predicted actions. Check Section~\ref{sec:app_screenshot} for detailed discussions.
}%
\label{fig:browse_web_1_3}
\end{figure*}

\begin{figure*}[h!]
\centering
\vspace{-15mm}
\includegraphics[width=\textwidth]{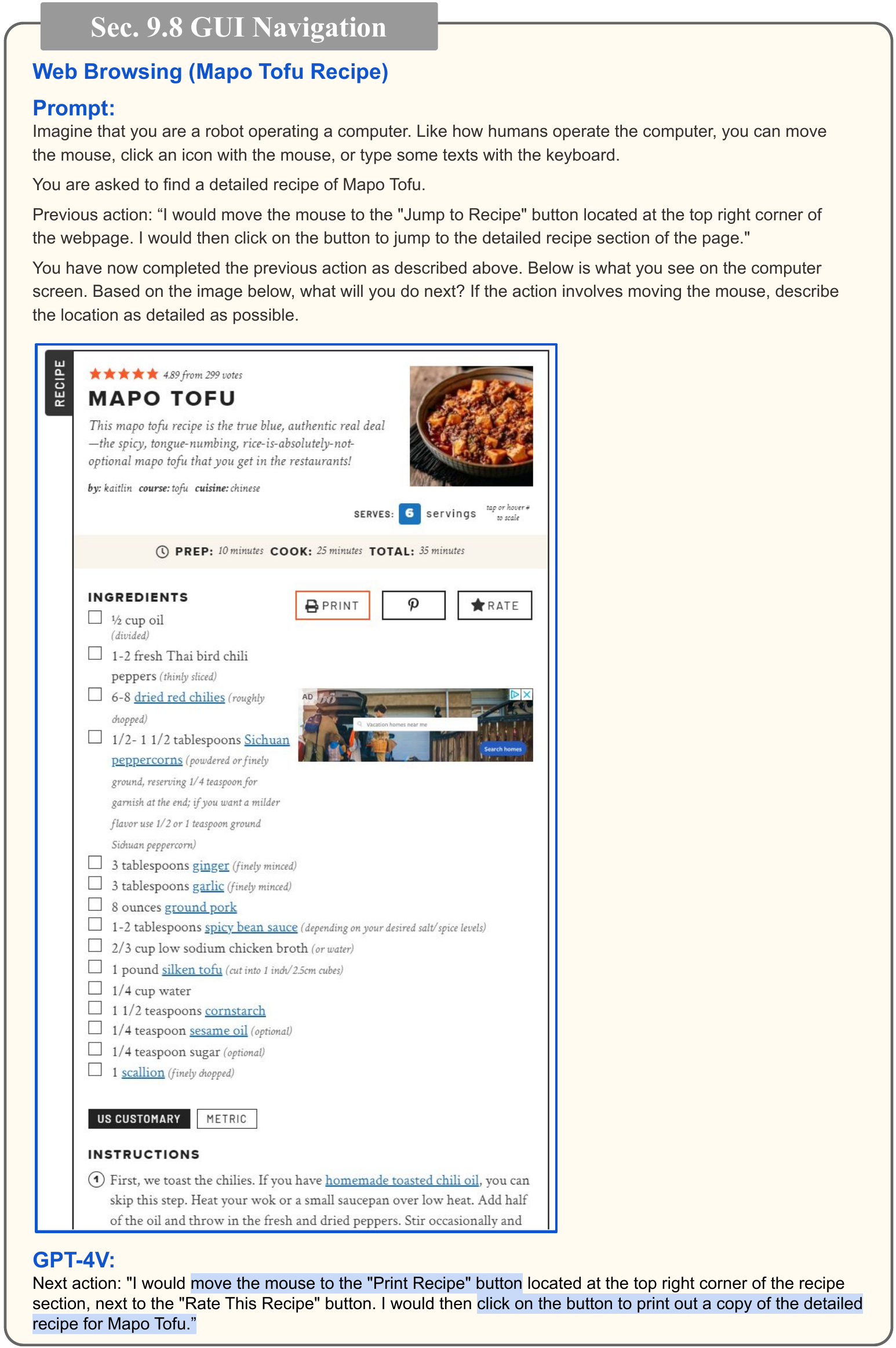}
\caption[Section~\ref{sec:app_screenshot}: web browsing for Mapo Tofu recipe.]{\modelname navigates through GUI to browse the web to search for the recipe of Mapo Tofu. \colorbox{bluehl}{Blue} highlights the predicted actions.  Check Section~\ref{sec:app_screenshot} for detailed discussions.
}%
\label{fig:browse_web_1_4}
\end{figure*}

\begin{figure*}[h!]
\centering
\vspace{-15mm}
\includegraphics[width=\textwidth]{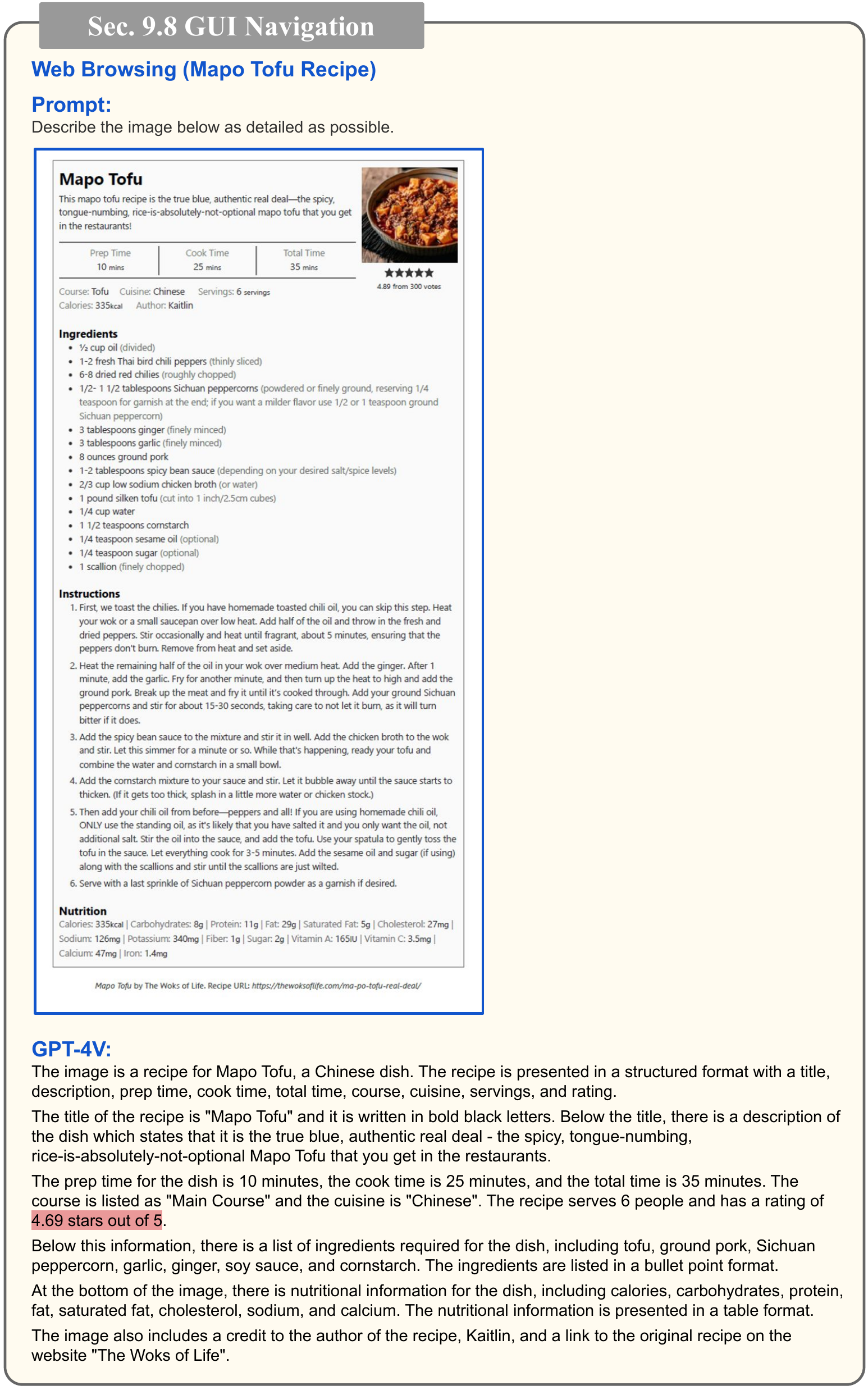}
\caption[Section~\ref{sec:app_screenshot}: web browsing for Mapo Tofu recipe.]{\modelname navigates through GUI to browse the web to search for the recipe of Mapo Tofu. As \modelname predicts to print out the recipe in the previous turn, we prompt it to read the screenshot of the printed recipe and summarize it. \colorbox{redhl}{Red} highlights the inaccurate description about the image.  Check Section~\ref{sec:app_screenshot} for detailed discussions.
}%
\label{fig:browse_web_1_5}
\end{figure*}

\begin{figure*}[h!]
\centering
\vspace{-15mm}
\includegraphics[width=\textwidth]{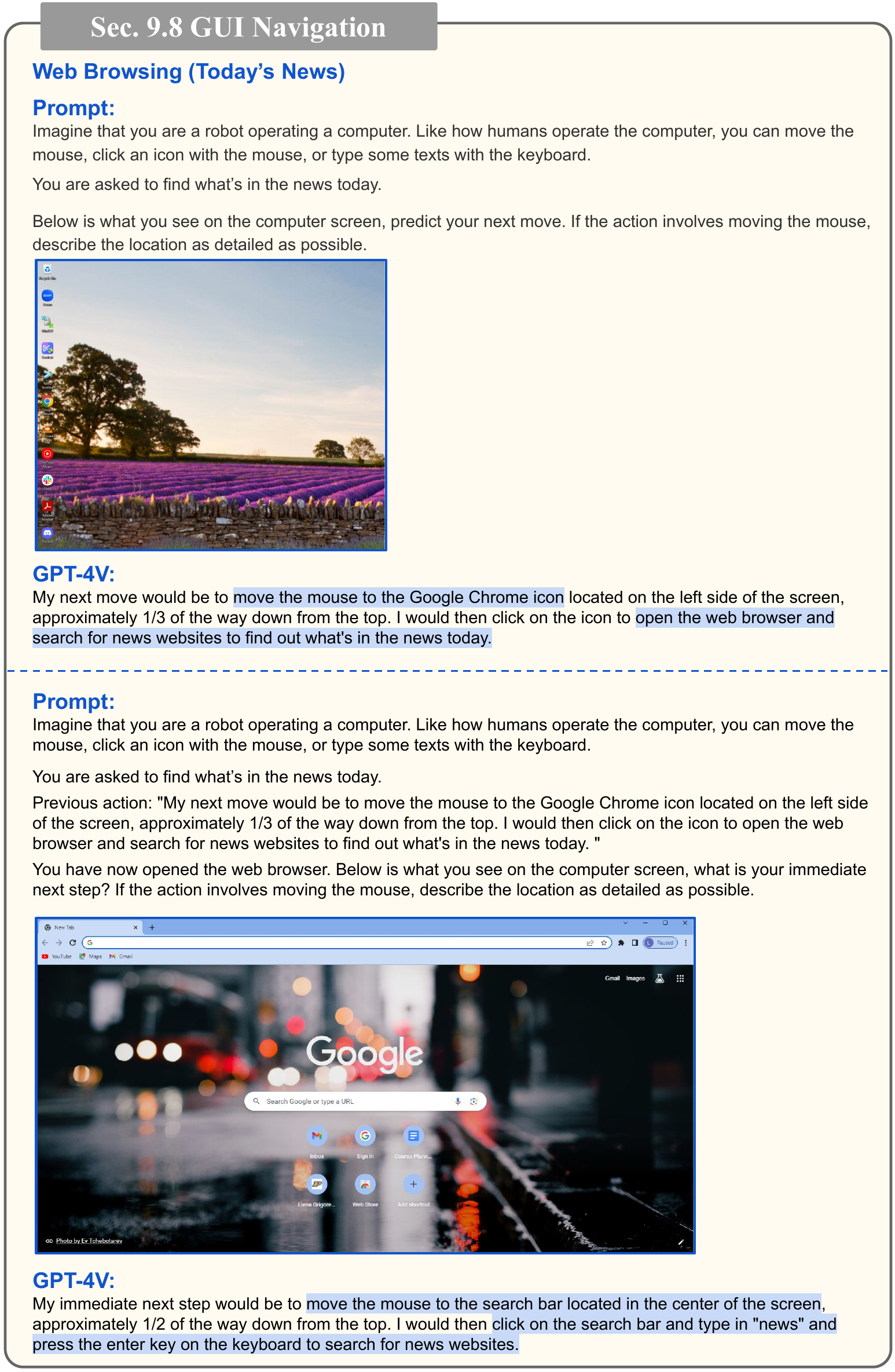}
\caption[Section~\ref{sec:app_screenshot}: web browsing for today's news.]{\modelname navigates through GUI to browse the web to read today's news.  \colorbox{bluehl}{Blue} highlights the predicted actions. Check Section~\ref{sec:app_screenshot} for detailed discussions.
}%
\label{fig:browse_web_2}
\end{figure*}

\begin{figure*}[h!]
\centering
\vspace{-15mm}
\includegraphics[width=\textwidth]{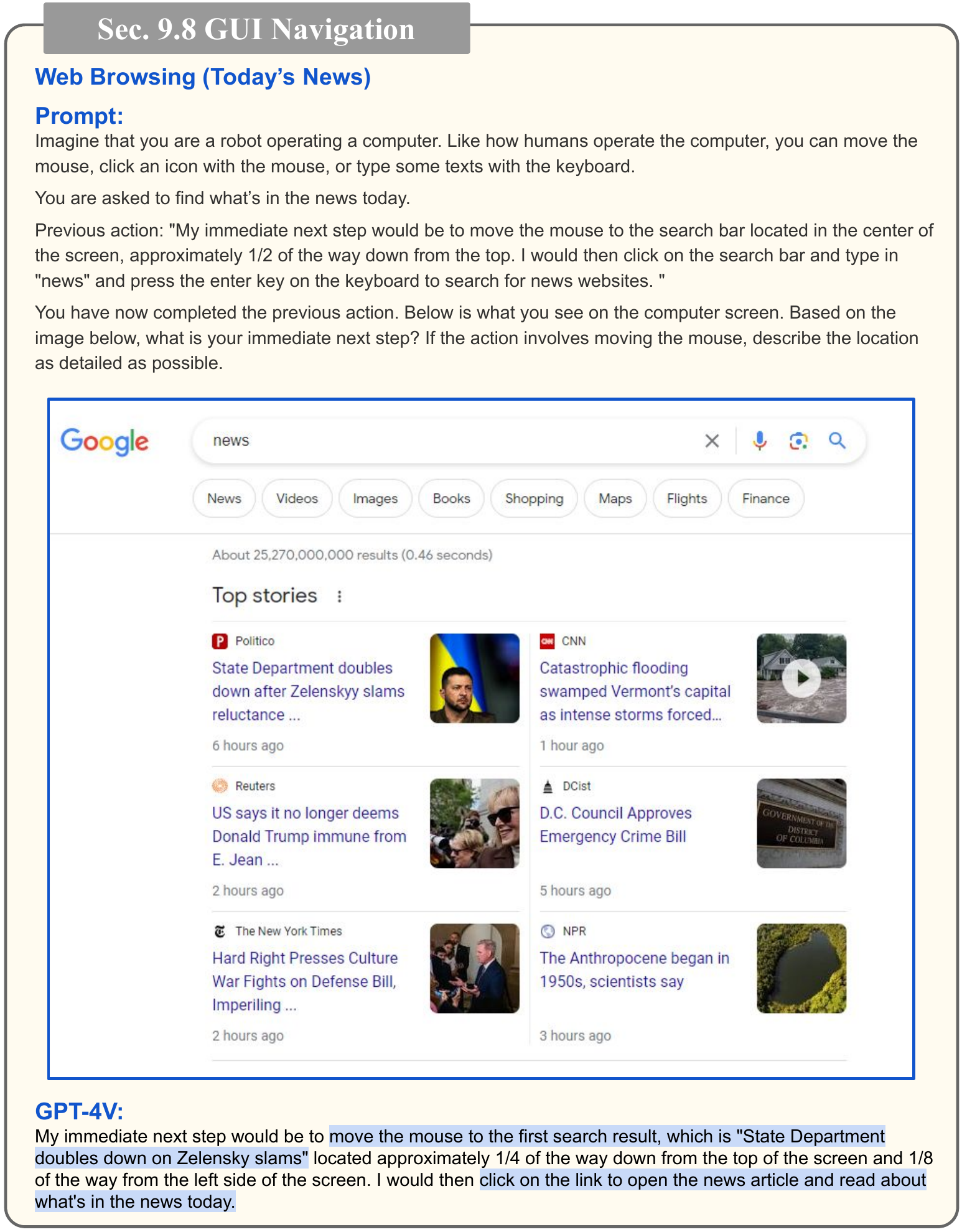}
\caption[Section~\ref{sec:app_screenshot}: web browsing for today's news.]{\modelname navigates through GUI to browse the web to read today's news.  \colorbox{bluehl}{Blue} highlights the predicted actions. Check Section~\ref{sec:app_screenshot} for detailed discussions.
}%
\label{fig:browse_web_2_2}
\end{figure*}

\begin{figure*}[h!]
\centering
\vspace{-15mm}
\includegraphics[width=\textwidth]{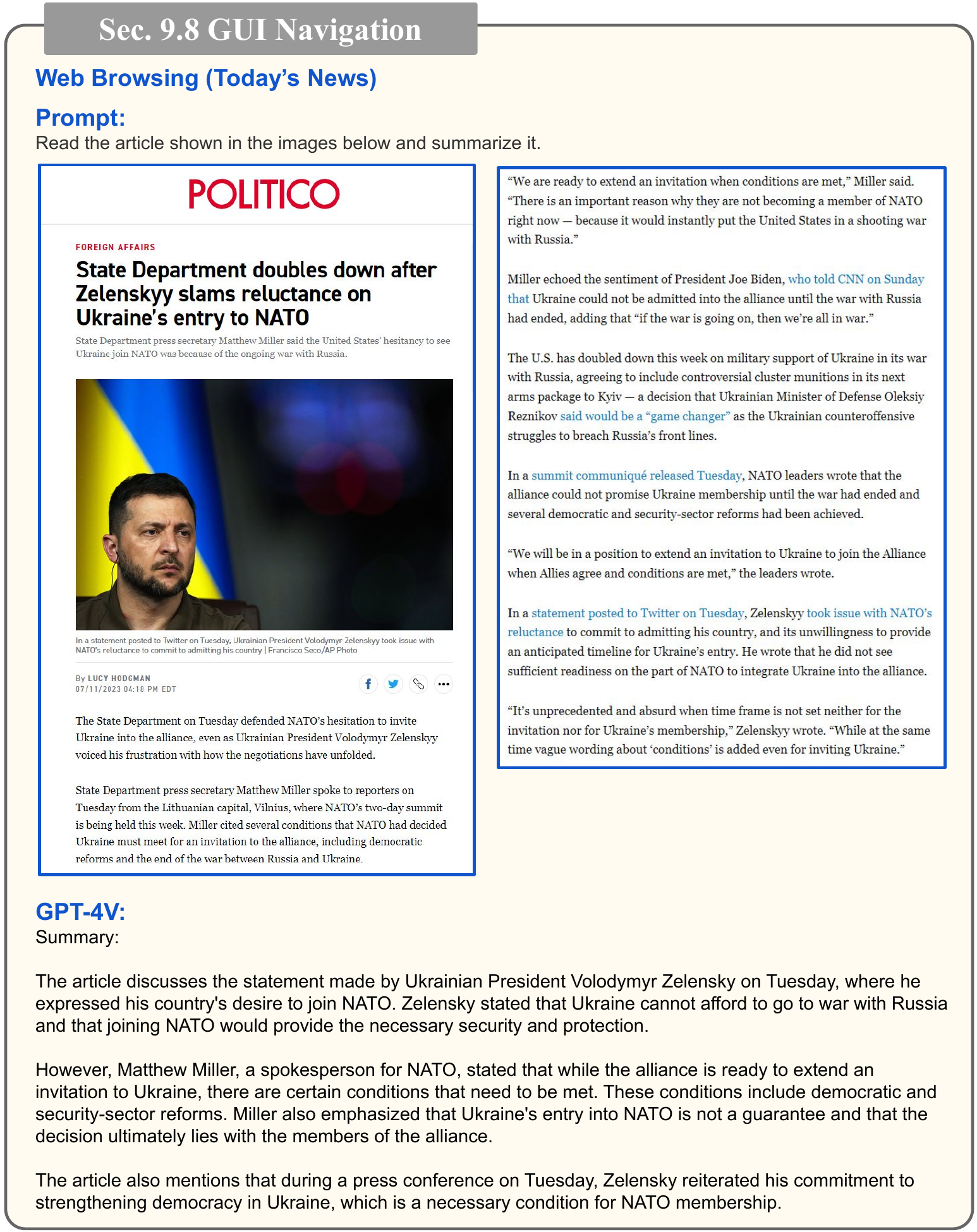}
\caption[Section~\ref{sec:app_screenshot}: web browsing for today's news.]{\modelname navigates through GUI to browse the web to read today's news.  We prompt \modelname to read the screenshots of the first news article and summarize it. Check Section~\ref{sec:app_screenshot} for detailed discussions.
}%
\label{fig:browse_web_2_3}
\end{figure*}

\begin{figure*}[h!]
\centering
\vspace{-15mm}
\includegraphics[width=\textwidth]{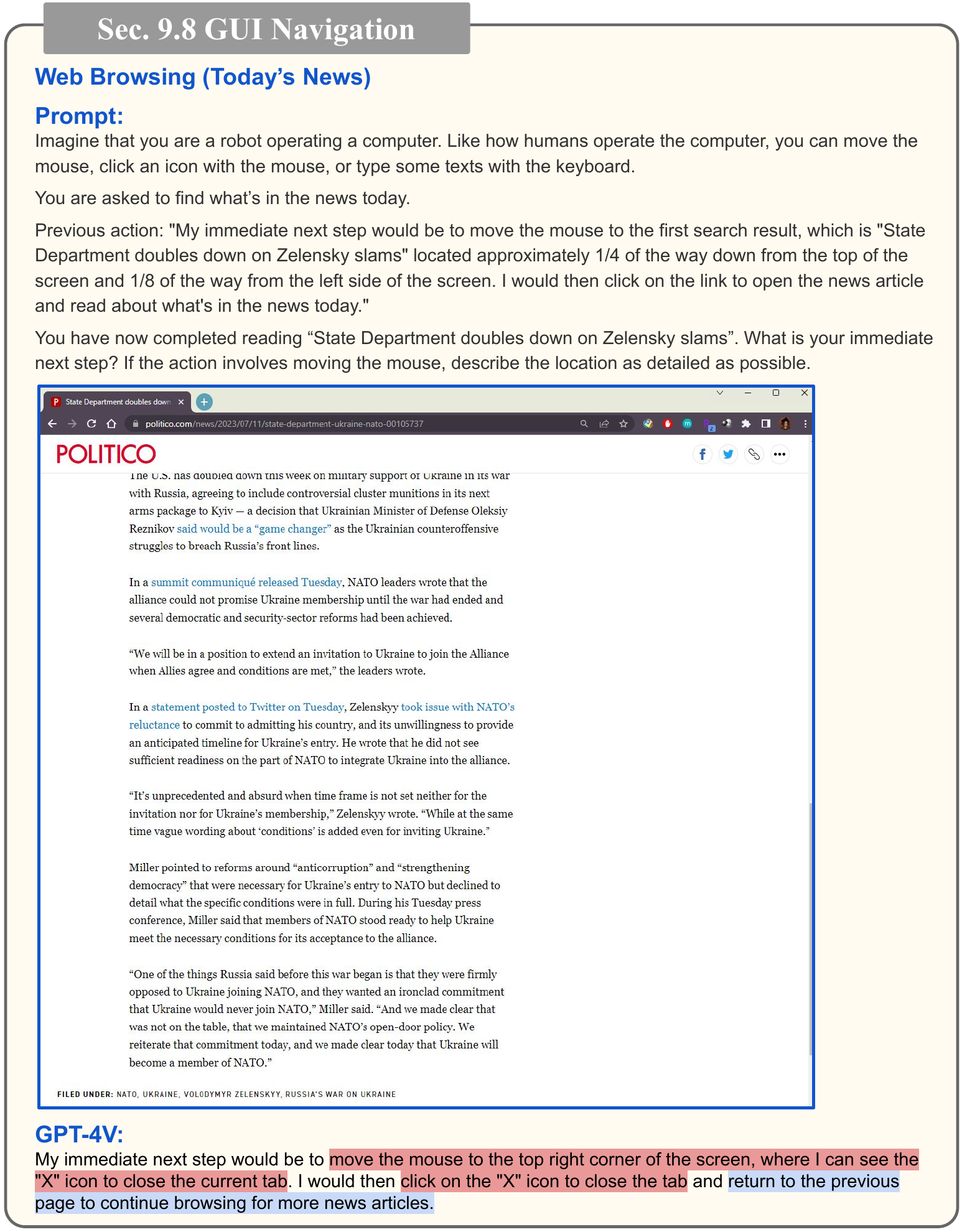}
\caption[Section~\ref{sec:app_screenshot}: web browsing for today's news.]{\modelname navigates through GUI to browse the web to read today's news. Upon finishing reading the first news article, \modelname predicts to close the tab and return to previous page to continue browsing more news articles (highlighted in \colorbox{bluehl}{blue}). \colorbox{redhl}{Red} highlights the inaccurate action prediction.  Check Section~\ref{sec:app_screenshot} for detailed discussions.
}%
\label{fig:browse_web_2_4}
\end{figure*}

\begin{figure*}[h!]
\centering
\vspace{-15mm}
\includegraphics[width=\textwidth]{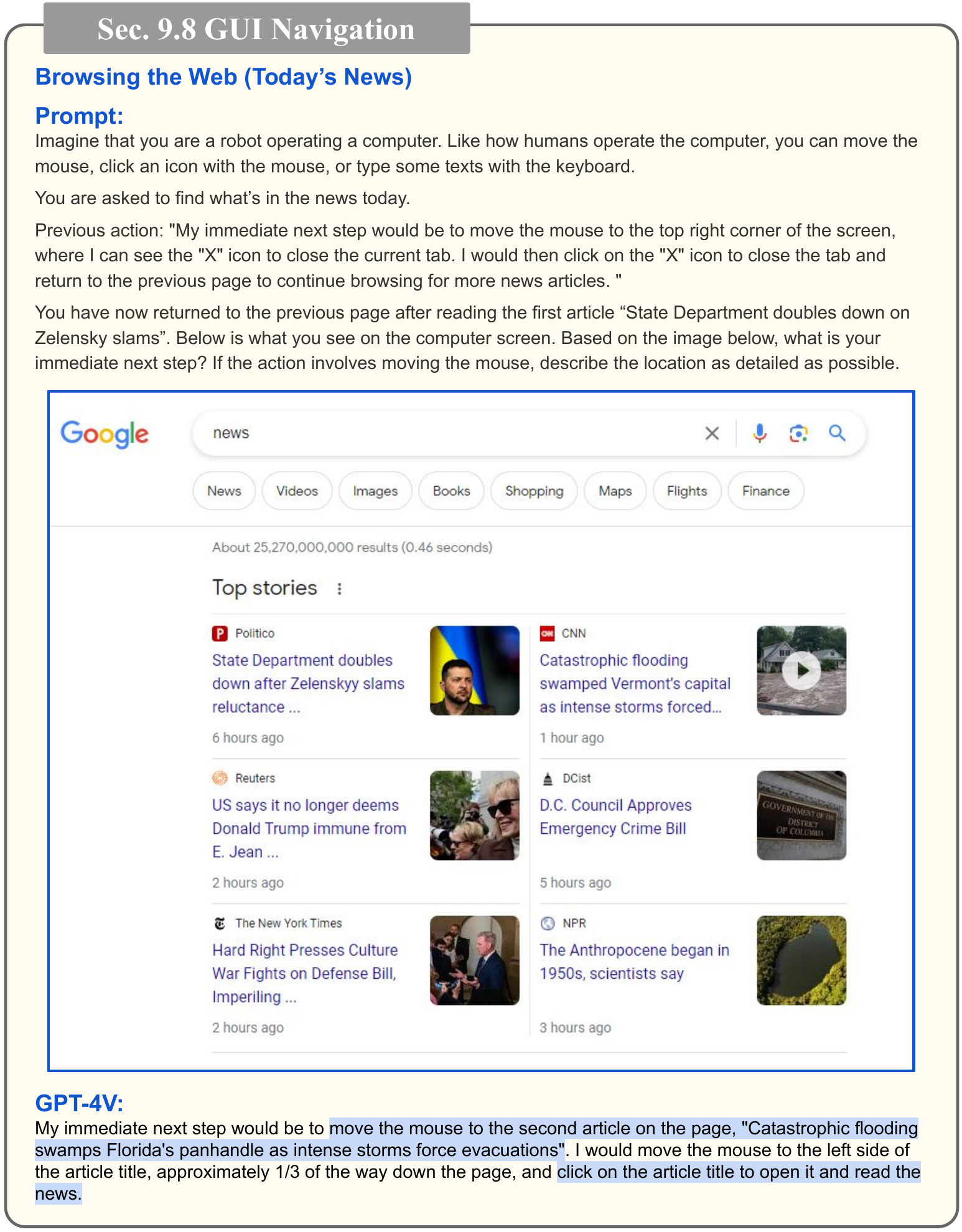}
\caption[Section~\ref{sec:app_screenshot}: web browsing for today's news.]{\modelname navigates through GUI to browse the web to read today's news. \colorbox{bluehl}{Blue} highlights the predicted actions. Check Section~\ref{sec:app_screenshot} for detailed discussions.
}%
\label{fig:browse_web_2_5}
\end{figure*}

\begin{figure*}[h!]
\centering
\vspace{-25mm}
\includegraphics[width=\textwidth]{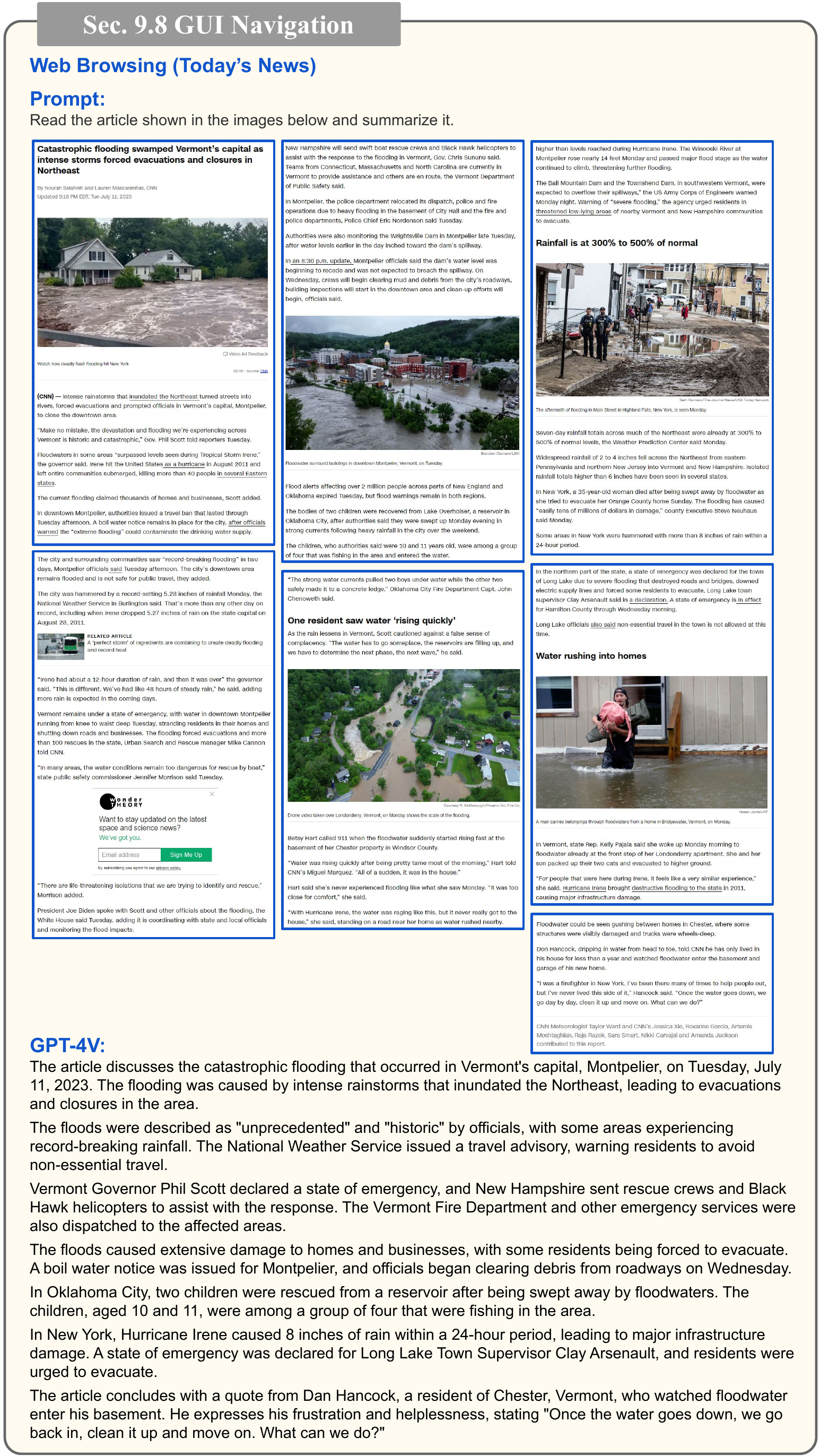}
\caption[Section~\ref{sec:app_screenshot}: web browsing for today's news.]{\modelname navigates through GUI to browse the web to read today's news.  We prompt \modelname to read the screenshots of the second news article and summarize it. Check Section~\ref{sec:app_screenshot} for detailed discussions.
}%
\label{fig:browse_web_2_6}
\end{figure*}

In Figures~\ref{fig:browse_web_1}-\ref{fig:browse_web_1_5}, \modelname predicts reasonable actions to operate a computer GUI, and finally accomplish the end goal of finding a recipe of Mapo Tofu and print out a copy of the recipe in Figure~\ref{fig:browse_web_1_4}. We then provide \modelname a screenshot of the printed recipe and ask it to describe the printout as detailed as possible. As shown in Figure~\ref{fig:browse_web_1_5}, \modelname is able to recognize the details presented in the printout, including the cooking time, the list of ingredients, the author of the recipe, the link to the original recipe and \etc. Figures~\ref{fig:browse_web_2}-\ref{fig:browse_web_2_6} present how \modelname can navigate through GUI to browse the web to ``read today's news''. Despite the minor errors in Figure~\ref{fig:browse_web_2_4} when it tries to return to the previous search result page to continue browsing for more news articles,  \modelname can perform the navigation and read two news articles reasonably well. 

\begin{figure*}[h!]
\centering
\vspace{-15mm}
\includegraphics[width=\textwidth]{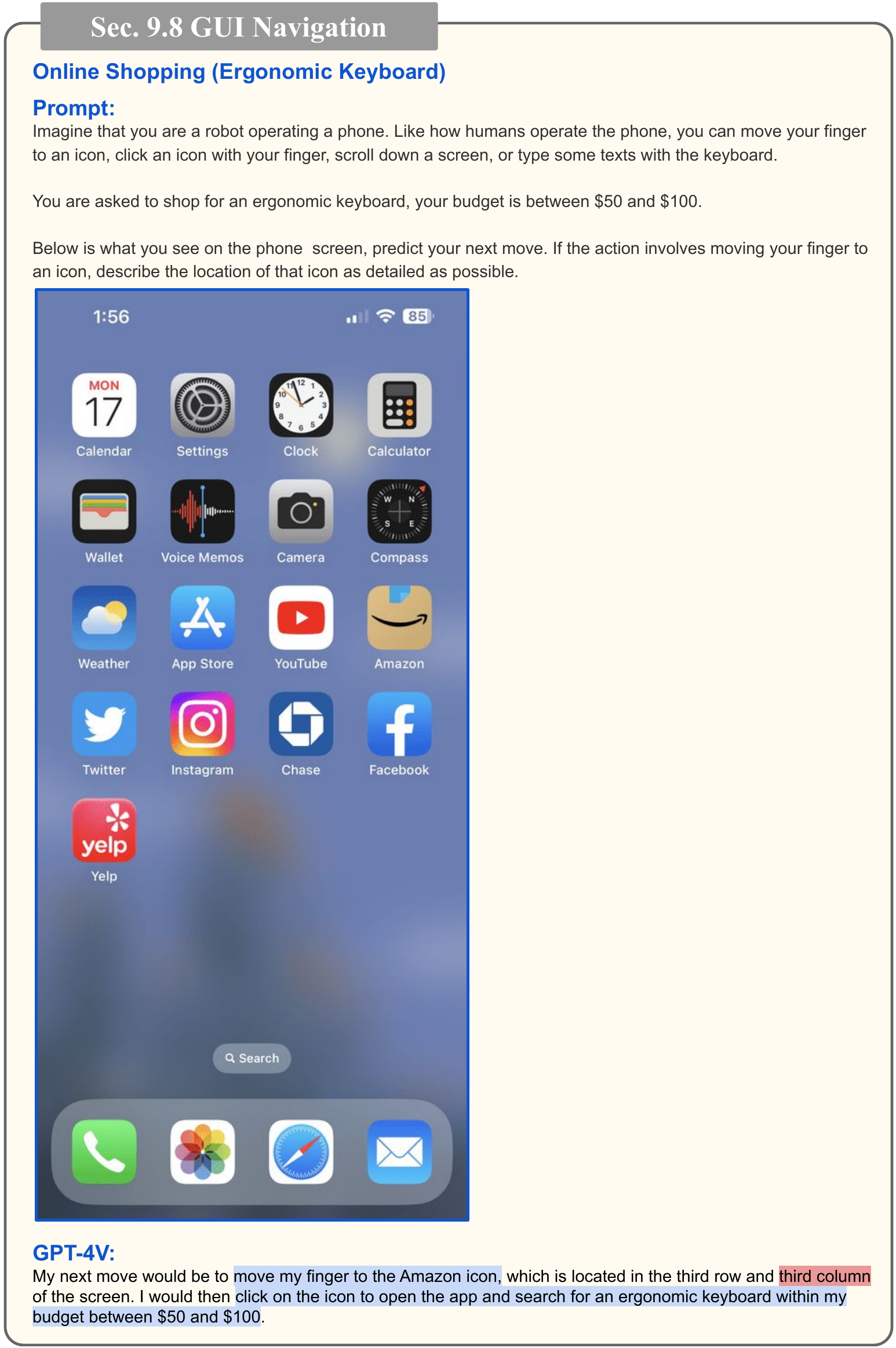}
\caption[Section~\ref{sec:app_screenshot}: online shopping.]{\modelname navigates through GUI to shop for an ergonomic keyboard online. \colorbox{bluehl}{Blue} highlights the predicted actions. \colorbox{redhl}{Red} highlights the inaccurate location of the Amazon icon. Check Section~\ref{sec:app_screenshot} for detailed discussions.
}%
\label{fig:online_shopping_1}
\end{figure*}

\begin{figure*}[h!]
\centering
\vspace{-15mm}
\includegraphics[width=\textwidth]{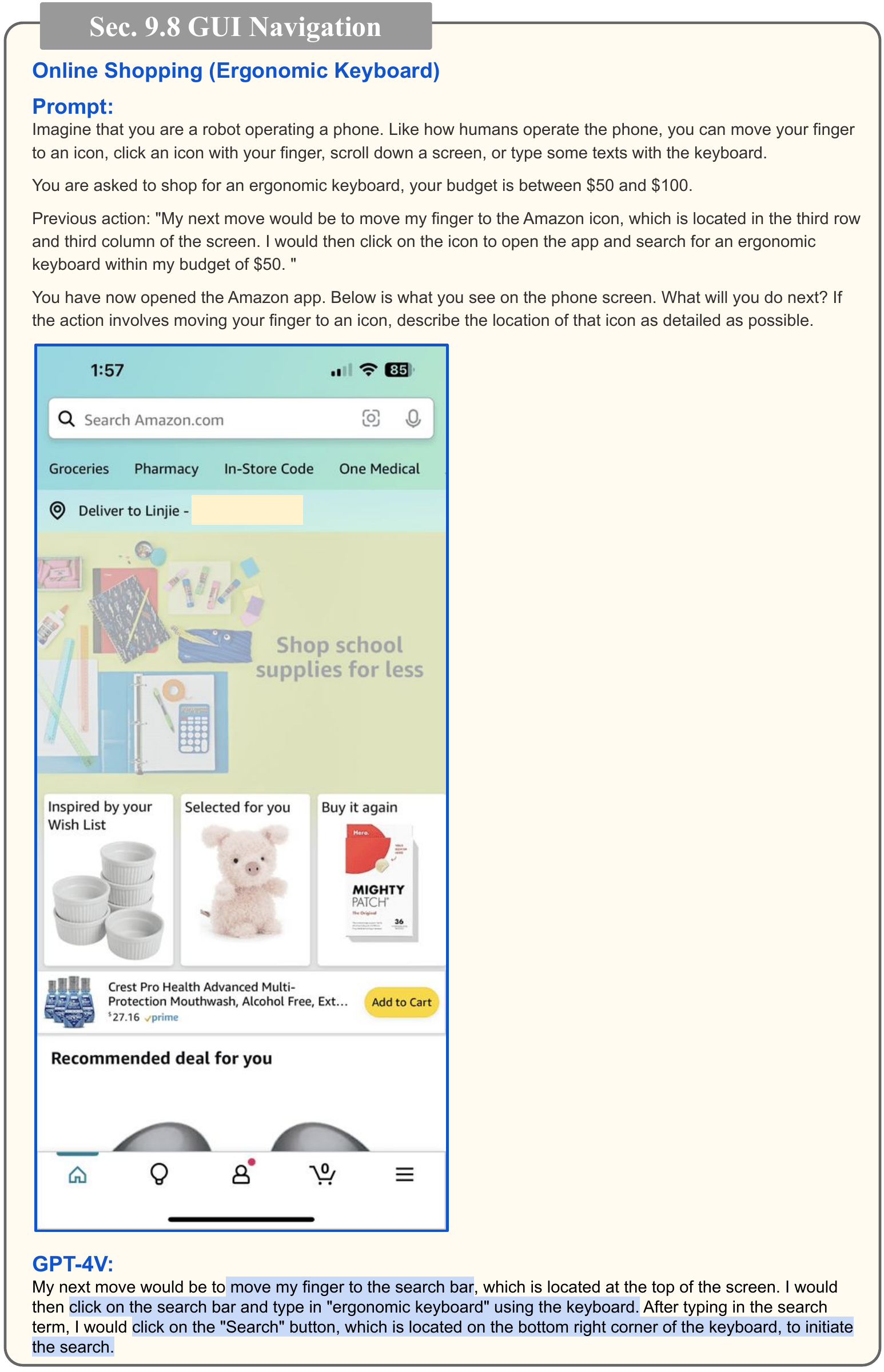}
\caption[Section~\ref{sec:app_screenshot}:  online shopping.]{\modelname navigates through GUI to shop for an ergonomic keyboard online. \colorbox{bluehl}{Blue} highlights the predicted actions. Check Section~\ref{sec:app_screenshot} for detailed discussions.
}%
\label{fig:online_shopping_2}
\end{figure*}

\begin{figure*}[h!]
\centering
\vspace{-15mm}
\includegraphics[width=\textwidth]{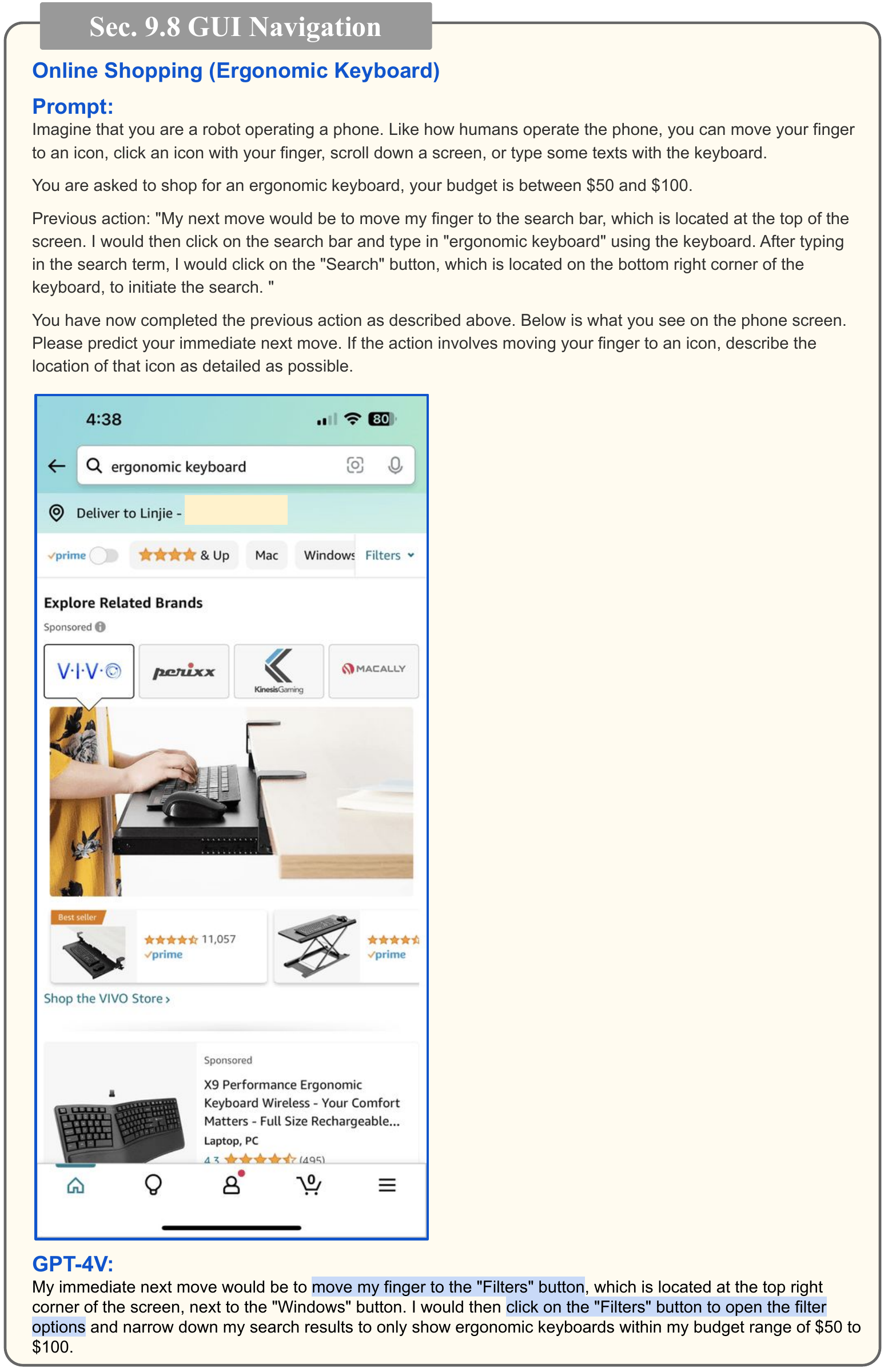}
\caption[Section~\ref{sec:app_screenshot}:  online shopping.]{\modelname navigates through GUI to shop for an ergonomic keyboard online. \colorbox{bluehl}{Blue} highlights the predicted actions. Check Section~\ref{sec:app_screenshot} for detailed discussions.
}%
\label{fig:online_shopping_3}
\end{figure*}

\begin{figure*}[h!]
\centering
\vspace{-15mm}
\includegraphics[width=\textwidth]{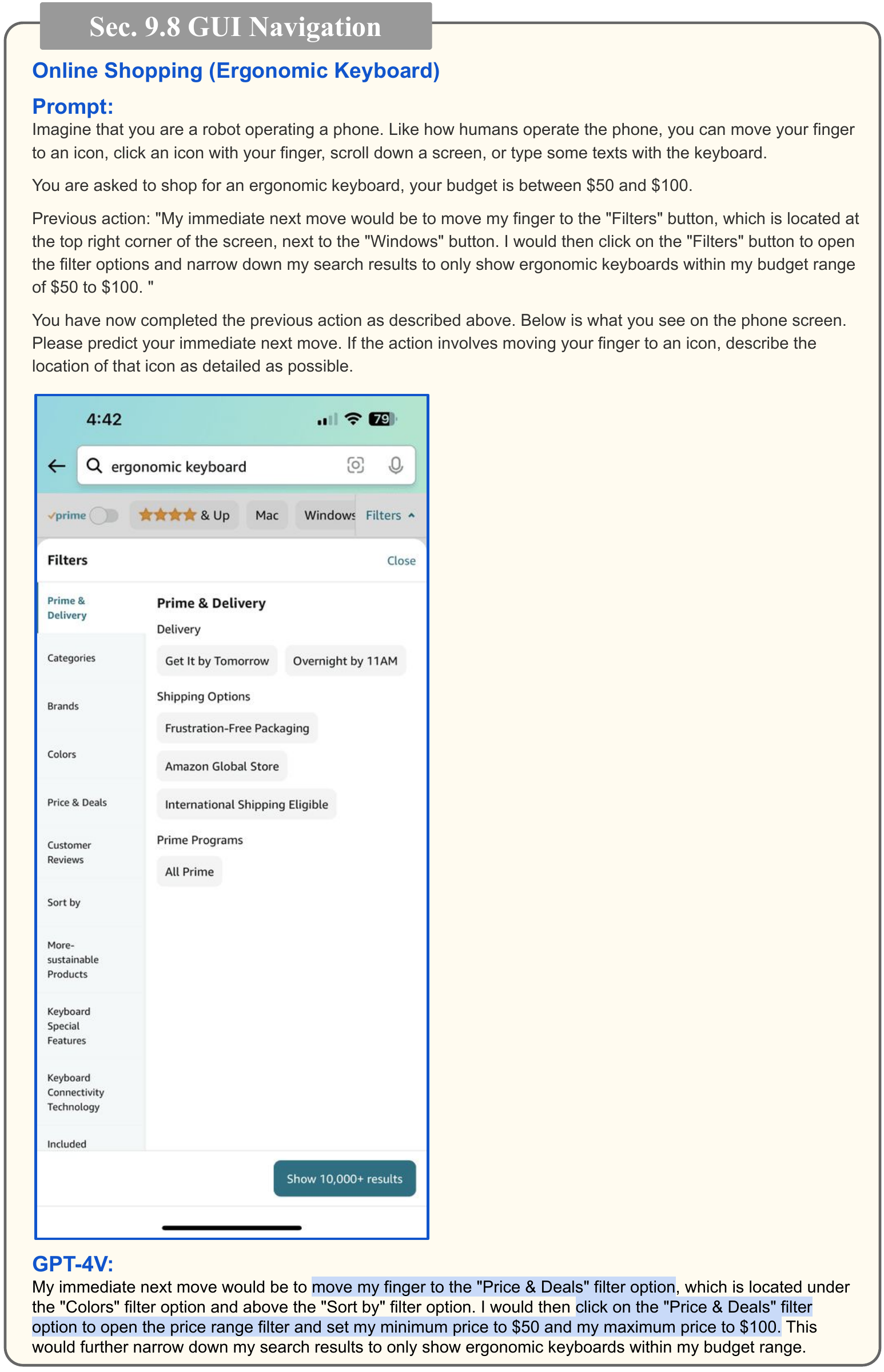}
\caption[Section~\ref{sec:app_screenshot}:  online shopping.]{\modelname navigates through GUI to shop for an ergonomic keyboard online. \colorbox{bluehl}{Blue} highlights the predicted actions. Check Section~\ref{sec:app_screenshot} for detailed discussions.
}%
\label{fig:online_shopping_4}
\end{figure*}

\begin{figure*}[h!]
\centering
\vspace{-15mm}
\includegraphics[width=\textwidth]{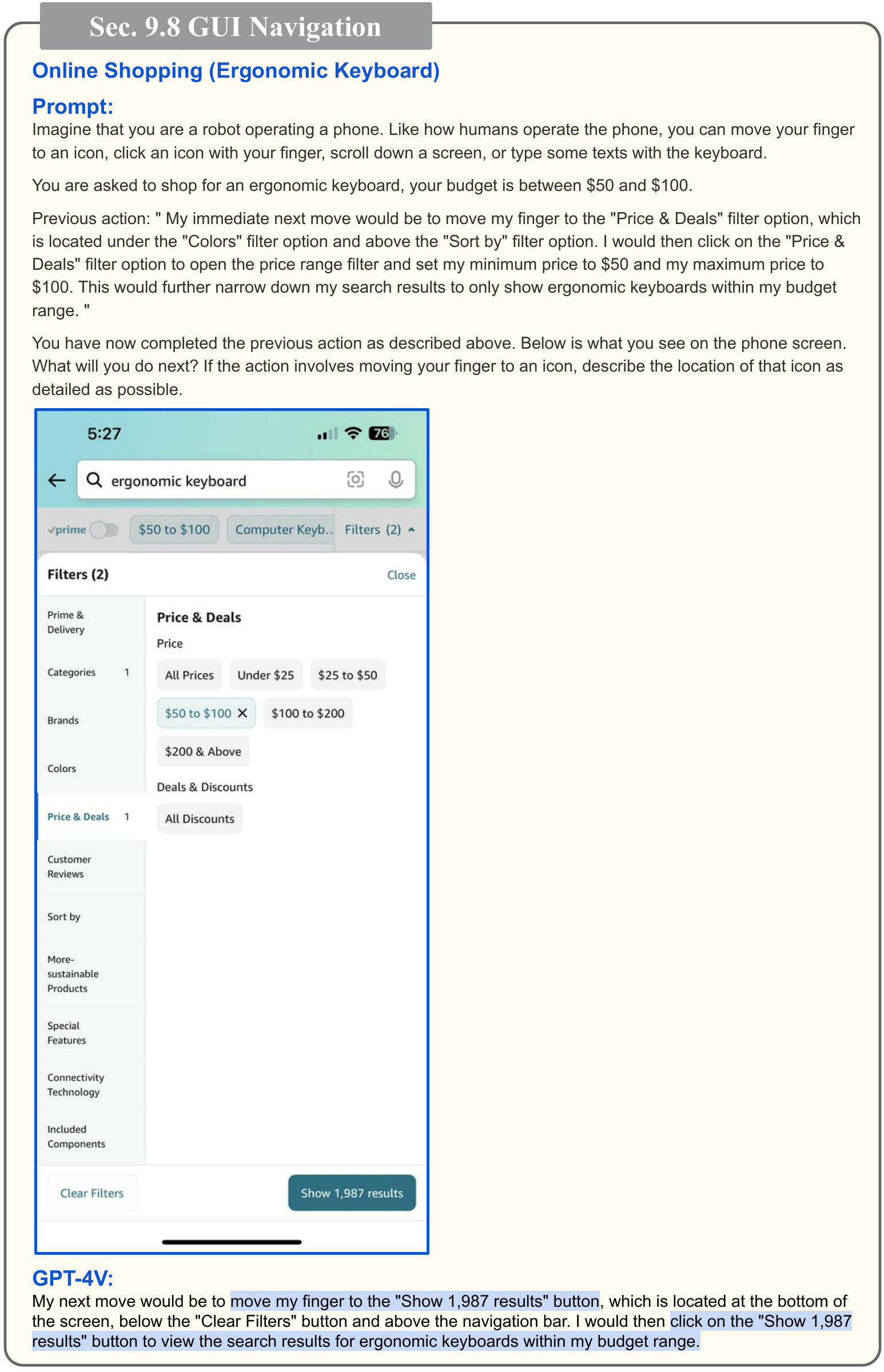}
\caption[Section~\ref{sec:app_screenshot}:  online shopping.]{\modelname navigates through GUI to shop for an ergonomic keyboard online. \colorbox{bluehl}{Blue} highlights the predicted actions. Check Section~\ref{sec:app_screenshot} for detailed discussions.
}%
\label{fig:online_shopping_5}
\end{figure*}

\begin{figure*}[h!]
\centering
\vspace{-15mm}
\includegraphics[width=\textwidth]{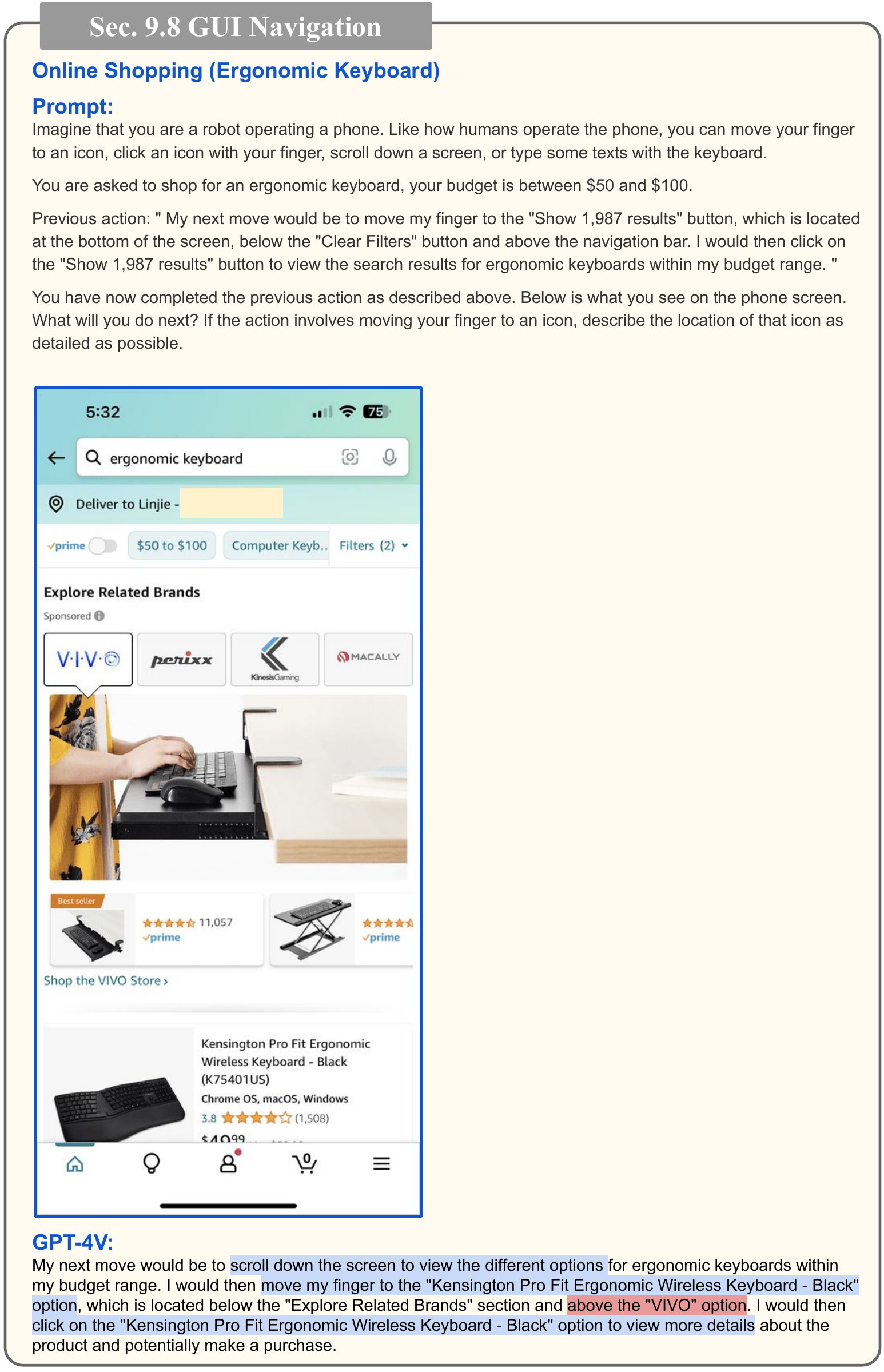}
\caption[Section~\ref{sec:app_screenshot}:  online shopping.]{\modelname navigates through GUI to shop for an ergonomic keyboard online. \colorbox{bluehl}{Blue} highlights the predicted actions. \colorbox{redhl}{Red} highlights the inaccurate location of the product option to be selected. Check Section~\ref{sec:app_screenshot} for detailed discussions.
}%
\label{fig:online_shopping_6}
\end{figure*}

\begin{figure*}[h!]
\centering
\vspace{-15mm}
\includegraphics[width=\textwidth]{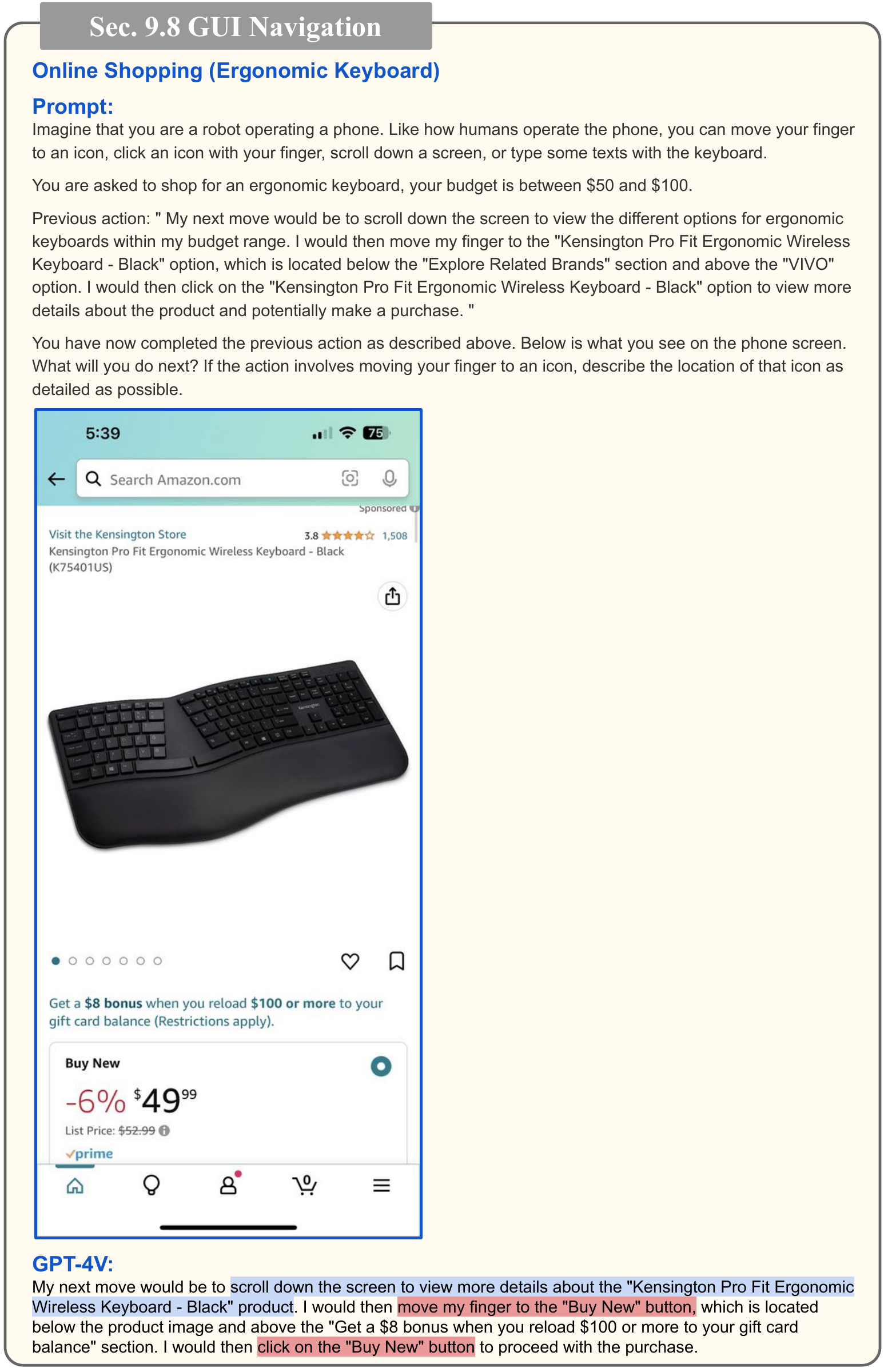}
\caption[Section~\ref{sec:app_screenshot}:  online shopping.]{\modelname navigates through GUI to shop for an ergonomic keyboard online. \colorbox{bluehl}{Blue} highlights the predicted actions. \colorbox{redhl}{Red} highlights the inaccurate action prediction (``Buy New'' is not a clickable button). Check Section~\ref{sec:app_screenshot} for detailed discussions.
}%
\label{fig:online_shopping_7}
\end{figure*}

\begin{figure*}[h!]
\centering
\vspace{-15mm}
\includegraphics[width=\textwidth]{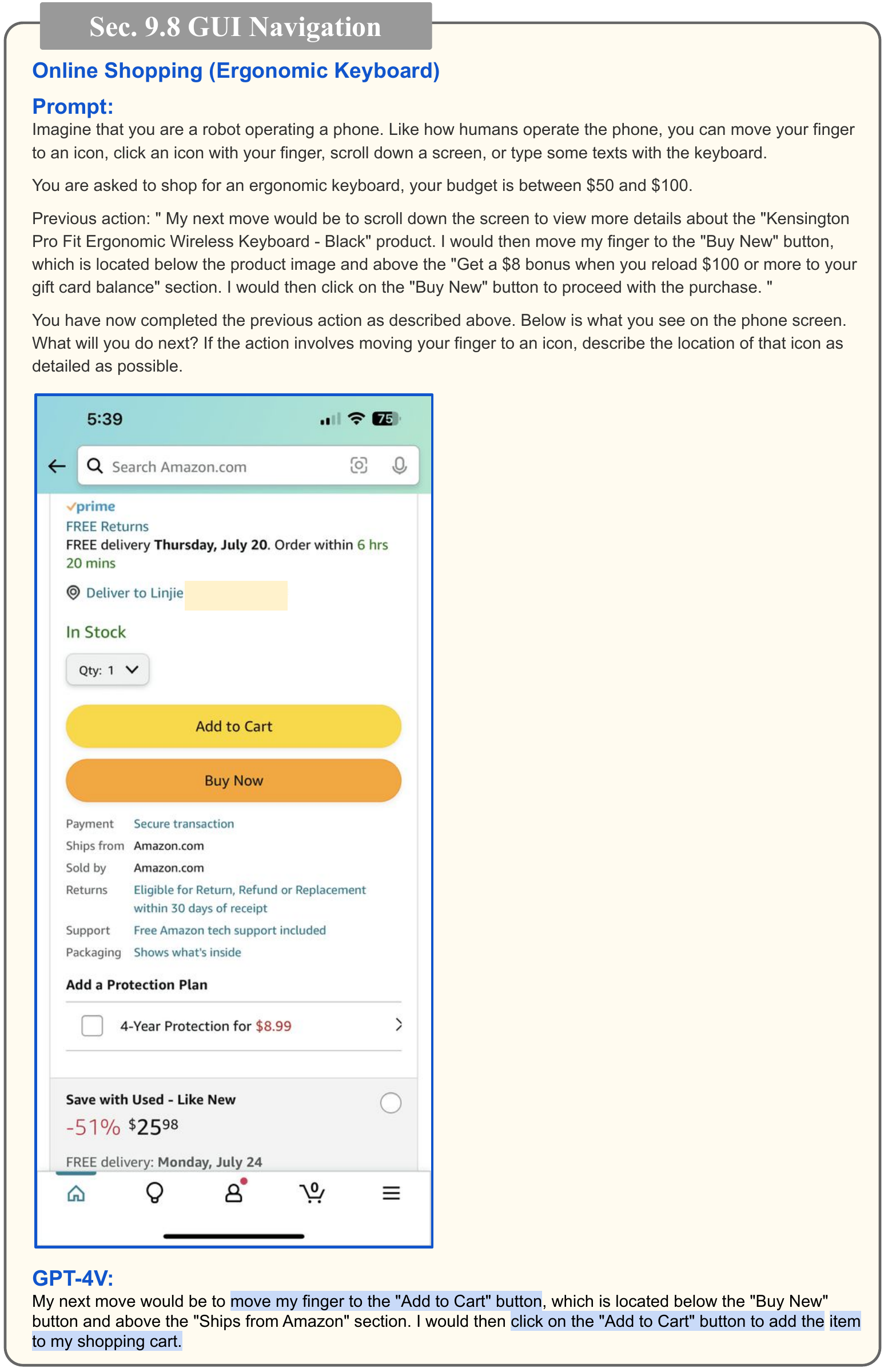}
\caption[Section~\ref{sec:app_screenshot}:  online shopping.]{\modelname navigates through GUI to shop for an ergonomic keyboard online. \colorbox{bluehl}{Blue} highlights the predicted actions. Check Section~\ref{sec:app_screenshot} for detailed discussions.
}%
\label{fig:online_shopping_8}
\end{figure*}

\begin{figure*}[h!]
\centering
\vspace{-15mm}
\includegraphics[width=\textwidth]{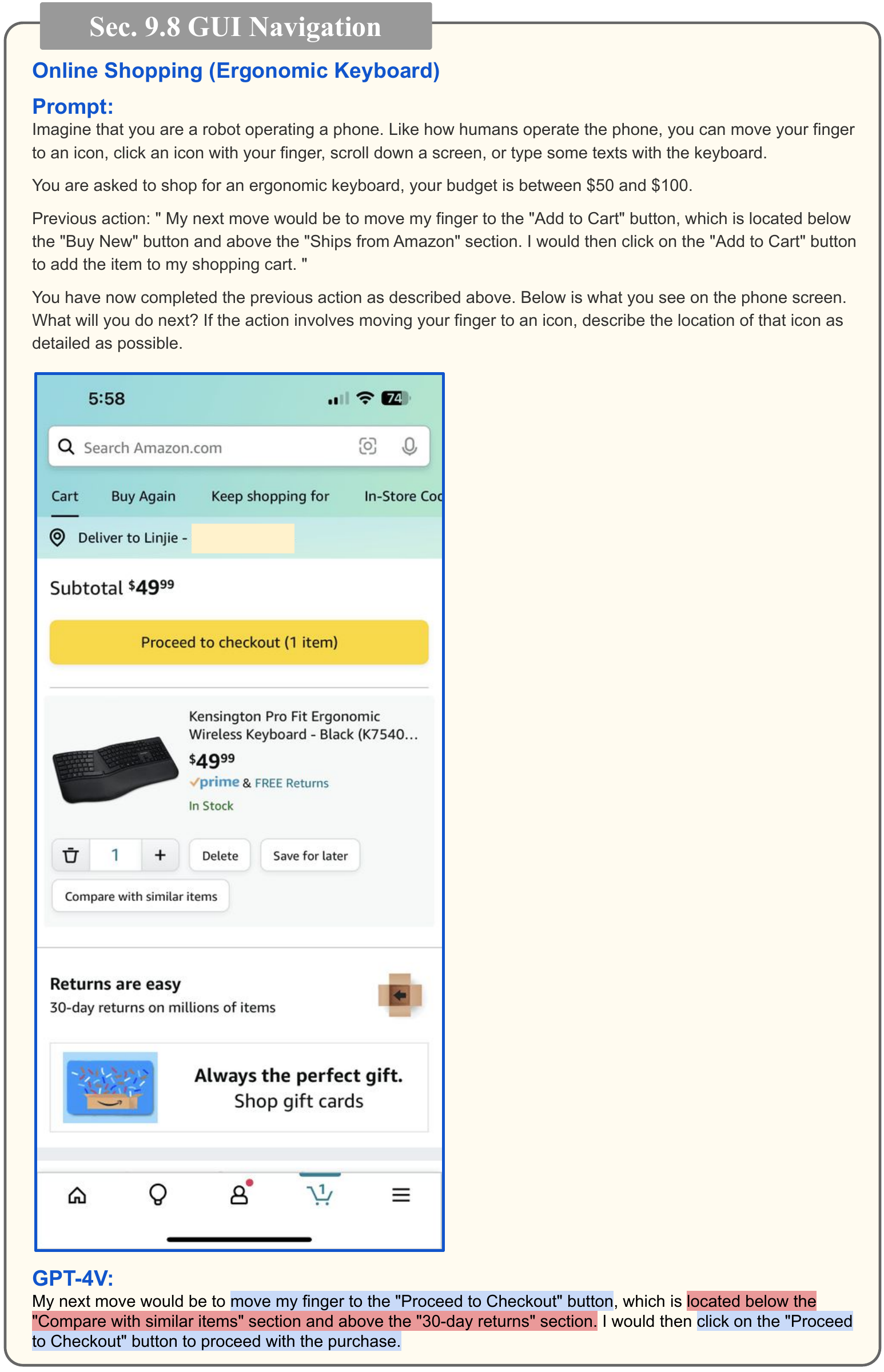}
\caption[Section~\ref{sec:app_screenshot}:  online shopping.]{\modelname navigates through GUI to shop for an ergonomic keyboard online. \colorbox{bluehl}{Blue} highlights the predicted actions. \colorbox{redhl}{Red} highlights the inaccurate location of the ``Proceed to checkout'' buttion. Check Section~\ref{sec:app_screenshot} for detailed discussions.
}%
\label{fig:online_shopping_9}
\end{figure*}

\paragraph{Online shopping.} Figures~\ref{fig:online_shopping_1}-\ref{fig:online_shopping_9} illustrates how \modelname can navigate a smartphone GUI for online shopping.  Similarly, we provide \modelname with the screenshot of the current phone screen, the list of possible actions (\eg, move your finger to an icon, click an icon with your finger, scroll down a screen, or type some texts with the keyboard) and ask it to predict the subsequent actions to shop for an ergonomic keyboard with a budget between \$50 and \$100. \modelname predicts to open the Amazon app (Figure~\ref{fig:online_shopping_1}), search ergonomic keyboard (Figure~\ref{fig:online_shopping_2}), open the filter options (Figure~\ref{fig:online_shopping_3}), set the price range filter between \$50 and \$100 (Figure~\ref{fig:online_shopping_4}), show filtered results (Figure~\ref{fig:online_shopping_5}), select the top search result (Figure~\ref{fig:online_shopping_6}), view product details (Figure~\ref{fig:online_shopping_7}), add product to the shopping cart (Figure~\ref{fig:online_shopping_8}) and finally proceed to checkout (Figure~\ref{fig:online_shopping_9}).

\begin{figure*}[h!]
\centering
\vspace{-15mm}
\includegraphics[width=\textwidth]{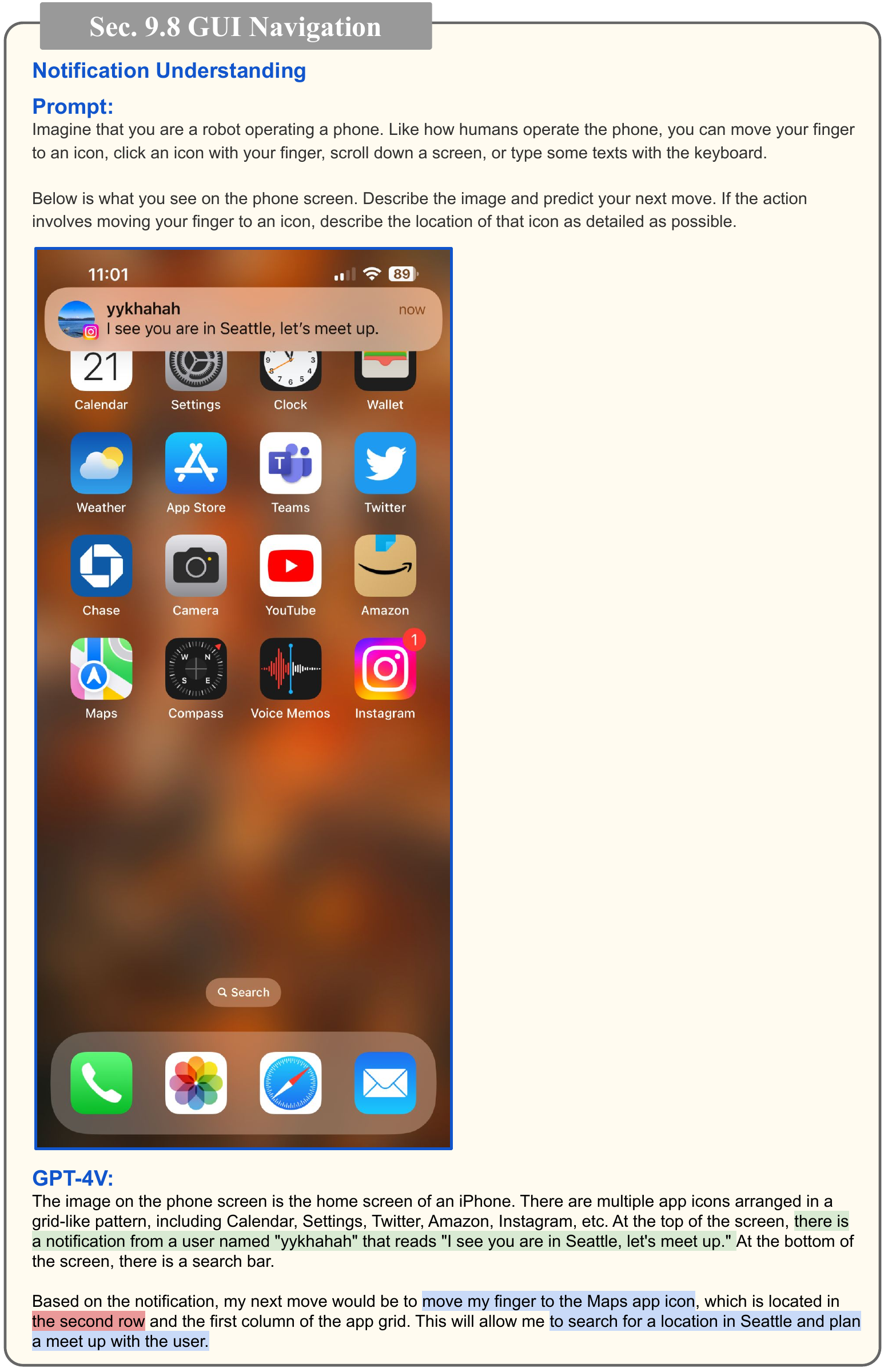}
\caption[Section~\ref{sec:app_screenshot}: notification understanding.]{Prompting \modelname to predict the action upon receiving a notification. \modelname can accurately recognize the notification and the corresponding content (highlighted in \colorbox{greenhl}{green}). \colorbox{bluehl}{Blue} highlights the predicted actions.  \colorbox{redhl}{Red} highlights the inaccurate location of the Maps app icon. Check Section~\ref{sec:app_screenshot} for detailed discussions. 
}%
\label{fig:notification_1}
\end{figure*}

\begin{figure*}[h!]
\centering
\vspace{-15mm}
\includegraphics[width=\textwidth]{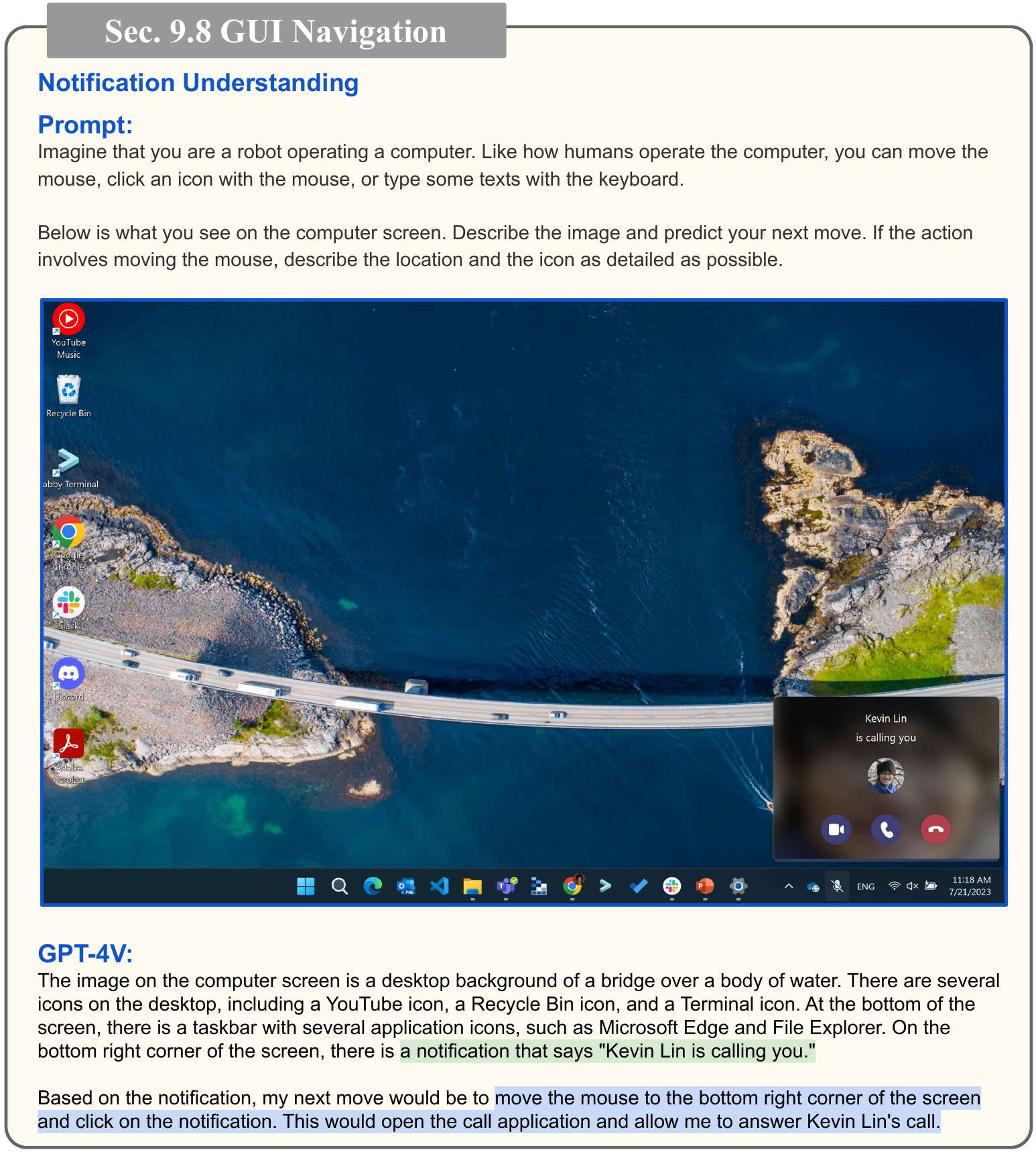}
\caption[Section~\ref{sec:app_screenshot}: notification understanding.]{Prompting \modelname to predict the action upon receiving a notification. \modelname can accurately recognize the notification and the corresponding content (highlighted in \colorbox{greenhl}{green}). \colorbox{bluehl}{Blue} highlights the predicted actions.  \colorbox{redhl}{Red} highlights the inaccurate location of the Maps app icon. Check Section~\ref{sec:app_screenshot} for detailed discussions. 
}%
\label{fig:notification_2}
\end{figure*}

\begin{figure*}[h!]
\centering
\vspace{-15mm}
\includegraphics[width=\textwidth]{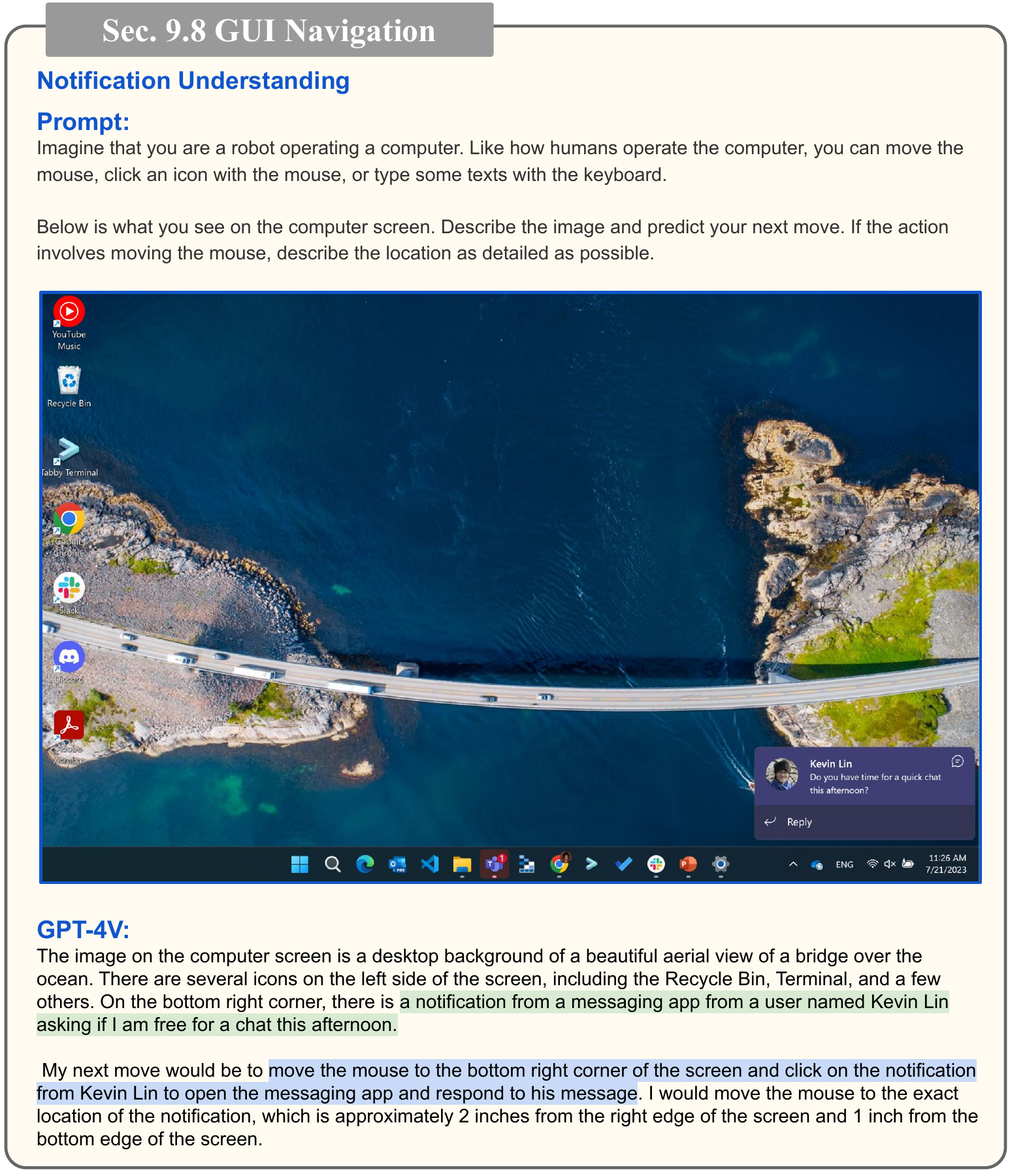}
\caption[Section~\ref{sec:app_screenshot}: notification understanding.]{Prompting \modelname to predict the action upon receiving a notification. \modelname can accurately recognize the notification and the corresponding content (highlighted in \colorbox{greenhl}{green}). \colorbox{bluehl}{Blue} highlights the predicted actions. Check Section~\ref{sec:app_screenshot} for detailed discussions. 
}%
\label{fig:notification_3}
\end{figure*}

\paragraph{Notification understanding.} Notifications are integral to modern human-computer interactions. \modelname has demonstrated its capacity to interpret notification content and respond accordingly. As shown in Figure~\ref{fig:notification_1}, the model can read and respond to a notification, such as suggesting to open the Maps app in response to a meeting proposal in Seattle. It also handles call (Figure~\ref{fig:notification_2}) and message (Figure~\ref{fig:notification_3}) notifications on a computer screen effectively.

\begin{figure*}[h!]
\centering
\vspace{-15mm}
\includegraphics[width=\textwidth]{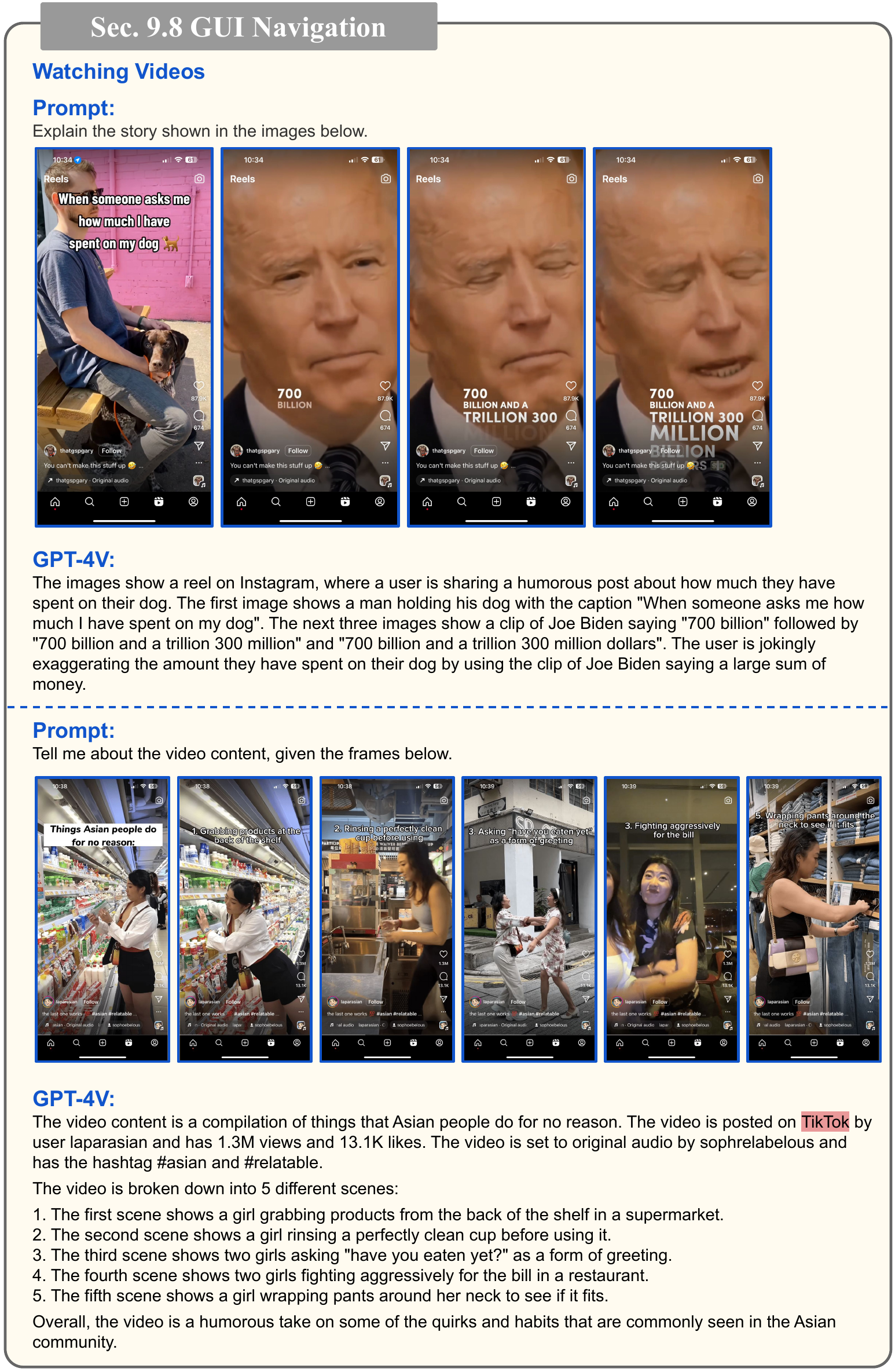}
\caption[Section~\ref{sec:app_screenshot}: watching videos.]{Prompting \modelname to watch web videos.  We present \modelname the screenshot of the video frames following their temporal order in the original video. To save space, we illustrate the frames in a row, where the leftmost one is the first frame. Check Section~\ref{sec:app_screenshot} for detailed discussions. \colorbox{redhl}{Red} highlights the inaccurate descriptions about the video.
}%
\label{fig:watching_video_1}
\end{figure*}

\begin{figure*}[h!]
\centering
\vspace{-15mm}
\includegraphics[width=\textwidth]{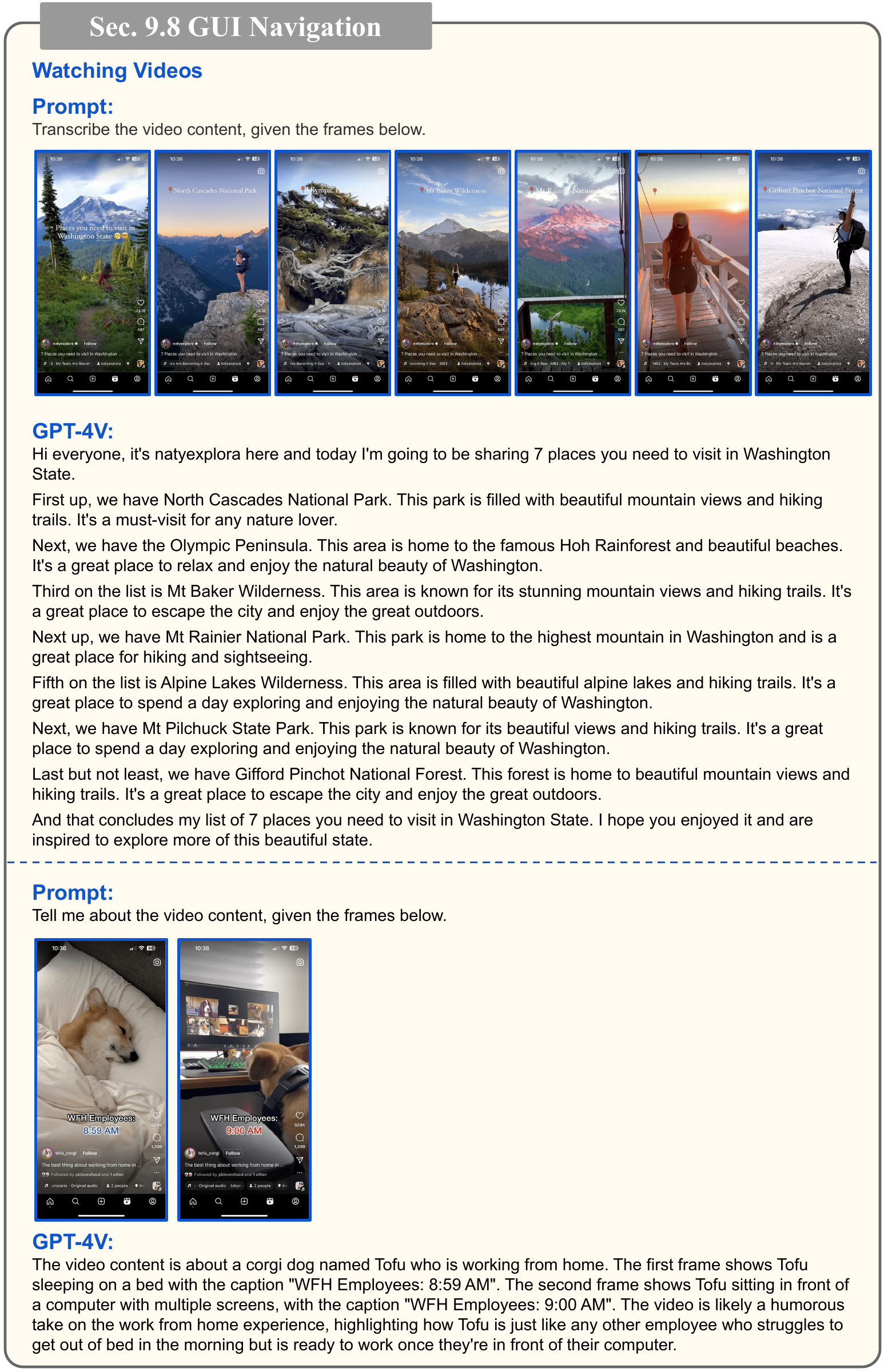}
\caption[Section~\ref{sec:app_screenshot}: watching videos.]{Watching web videos.  We present \modelname the screenshot of the video frames following their temporal order in the original video. To save space, we illustrate the frames in a row, where the leftmost one is the first frame. Check Section~\ref{sec:app_screenshot} for detailed discussions.
}%
\label{fig:watching_video_2}
\end{figure*}

\begin{figure*}[h!]
\centering
\vspace{-15mm}
\includegraphics[width=\textwidth]{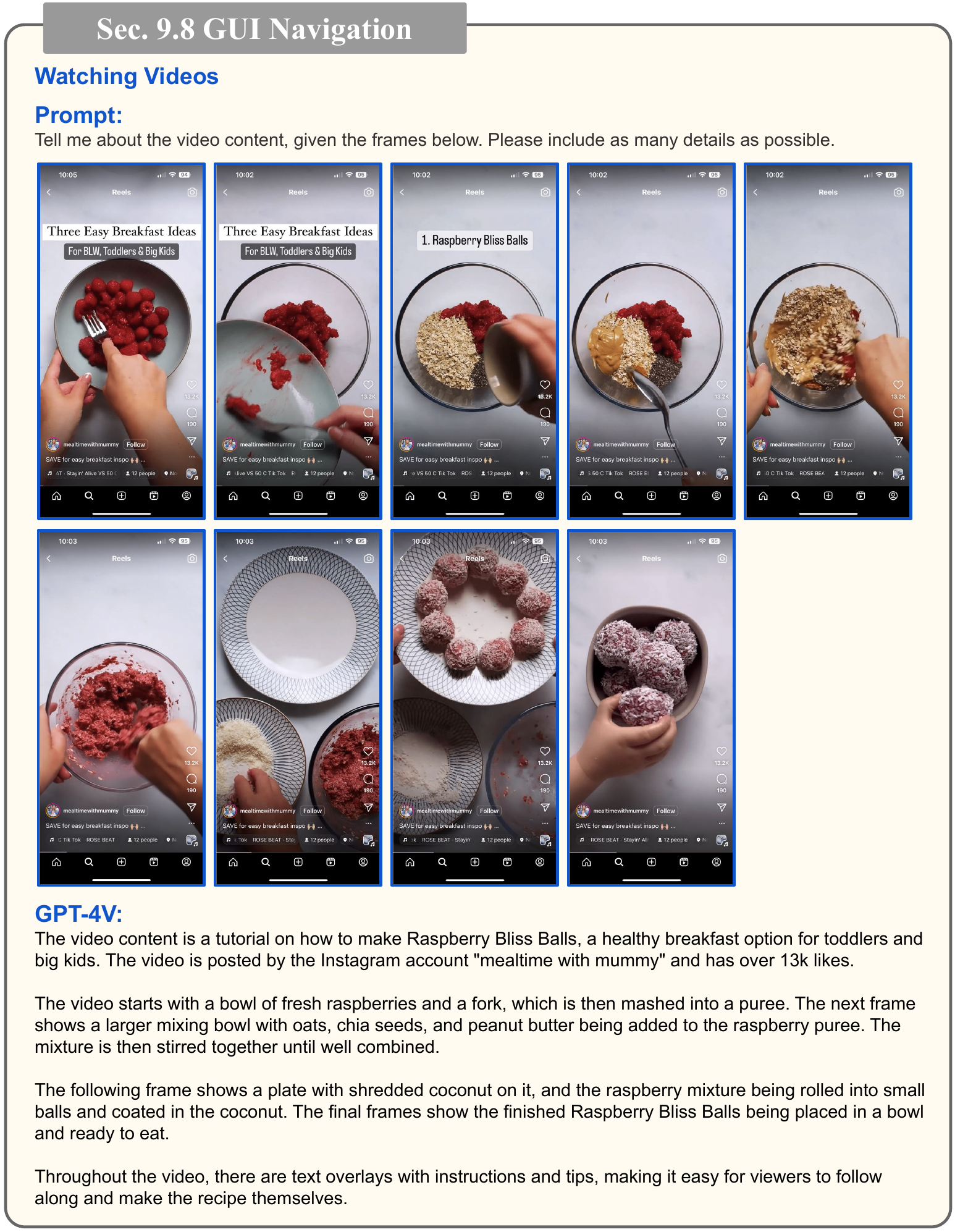}
\caption[Section~\ref{sec:app_screenshot}: watching videos.]{Watching web videos.  We present \modelname the screenshot of the video frames following their temporal order in the original video. To save space,  we illustrate frames 1-5 in the first row, and frames 6-9 in the second row. Check Section~\ref{sec:app_screenshot} for detailed discussions.
}%
\label{fig:watching_video_3}
\end{figure*}

\begin{figure*}[h!]
\centering
\vspace{-15mm}
\includegraphics[width=\textwidth]{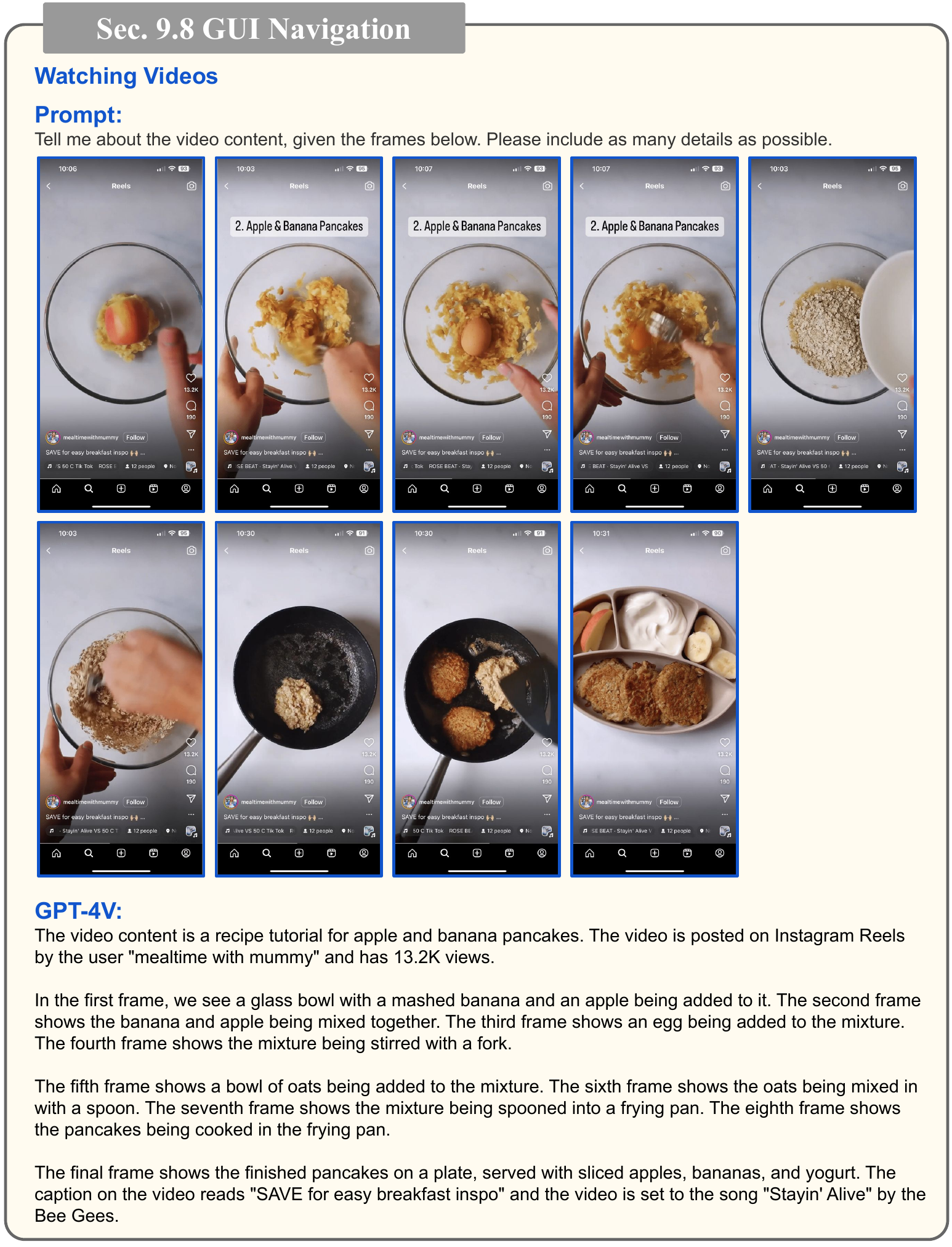}
\caption[Section~\ref{sec:app_screenshot}: watching videos.]{Watching web videos.  We present \modelname the screenshot of the video frames following their temporal order in the original video. To save space,  we illustrate frames 1-5 in the first row, and frames 6-9 in the second row.  Check Section~\ref{sec:app_screenshot} for detailed discussions.
}%
\label{fig:watching_video_4}
\end{figure*}

\begin{figure*}[h!]
\centering
\vspace{-15mm}
\includegraphics[width=\textwidth]{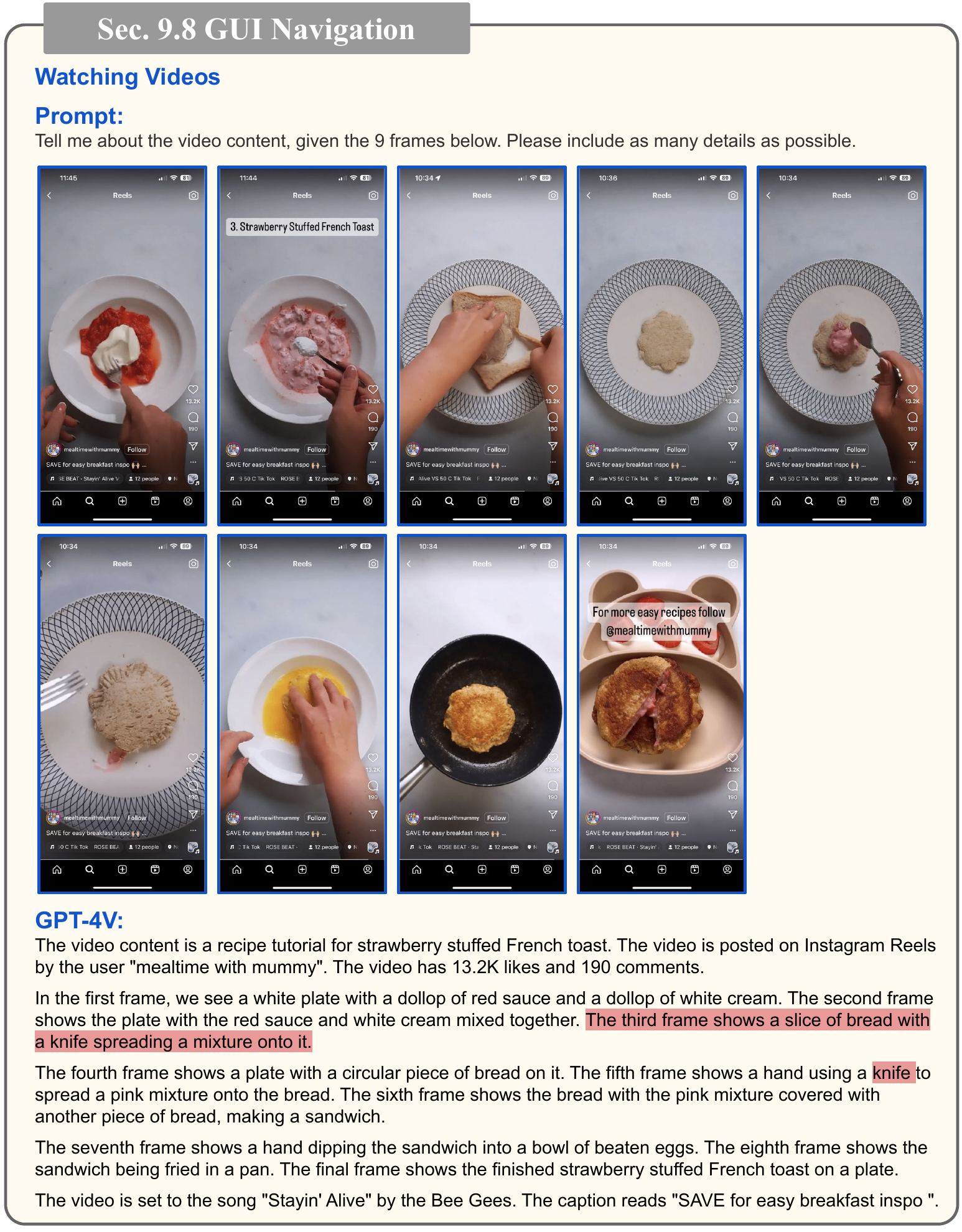}
\caption[Section~\ref{sec:app_screenshot}: watching videos.]{Watching web videos.  We present \modelname the screenshot of the video frames following their temporal order in the original video. To save space,  we illustrate frames 1-5 in the first row, and frames 6-9 in the second row. \colorbox{redhl}{Red} highlights the inaccurate descriptions about the video. Check Section~\ref{sec:app_screenshot} for detailed discussions.
}%
\label{fig:watching_video_5}
\end{figure*}

\paragraph{Watching videos.} Alongside web browsing, videos are a key source of online information. \modelname has shown its capability to describe video content based on a series of screenshots from popular short-form videos. Regardless of whether the video has subtitle overlay (Figure~\ref{fig:watching_video_1} and~\ref{fig:watching_video_2}) or not (Figure~\ref{fig:watching_video_3},~\ref{fig:watching_video_4},~\ref{fig:watching_video_5}), \modelname can generate insightful descriptions about the video content, demonstrating its potential in automatic transcript generation for user-generated video content.

\clearpage
\section{LMM Powered Agents}
\label{sec:06tool}

In this section, we discuss possible future research directions that may further amplify \modelname's capabilities. The discussion focuses on how the intriguing usages in LLMs may extend to the multimodal scenario and its enabled new abilities, \eg, multimodal plugins, multimodal chains, 
self-reflection, self-consistency, and retrieval-augmented LMMs, \etc. In the following sub-sections, we use \emph{human-generated} examples to illustrate potential ways to enhance \modelname-based systems.
\begin{figure*}[h!]
\centering
\includegraphics[width=\textwidth]{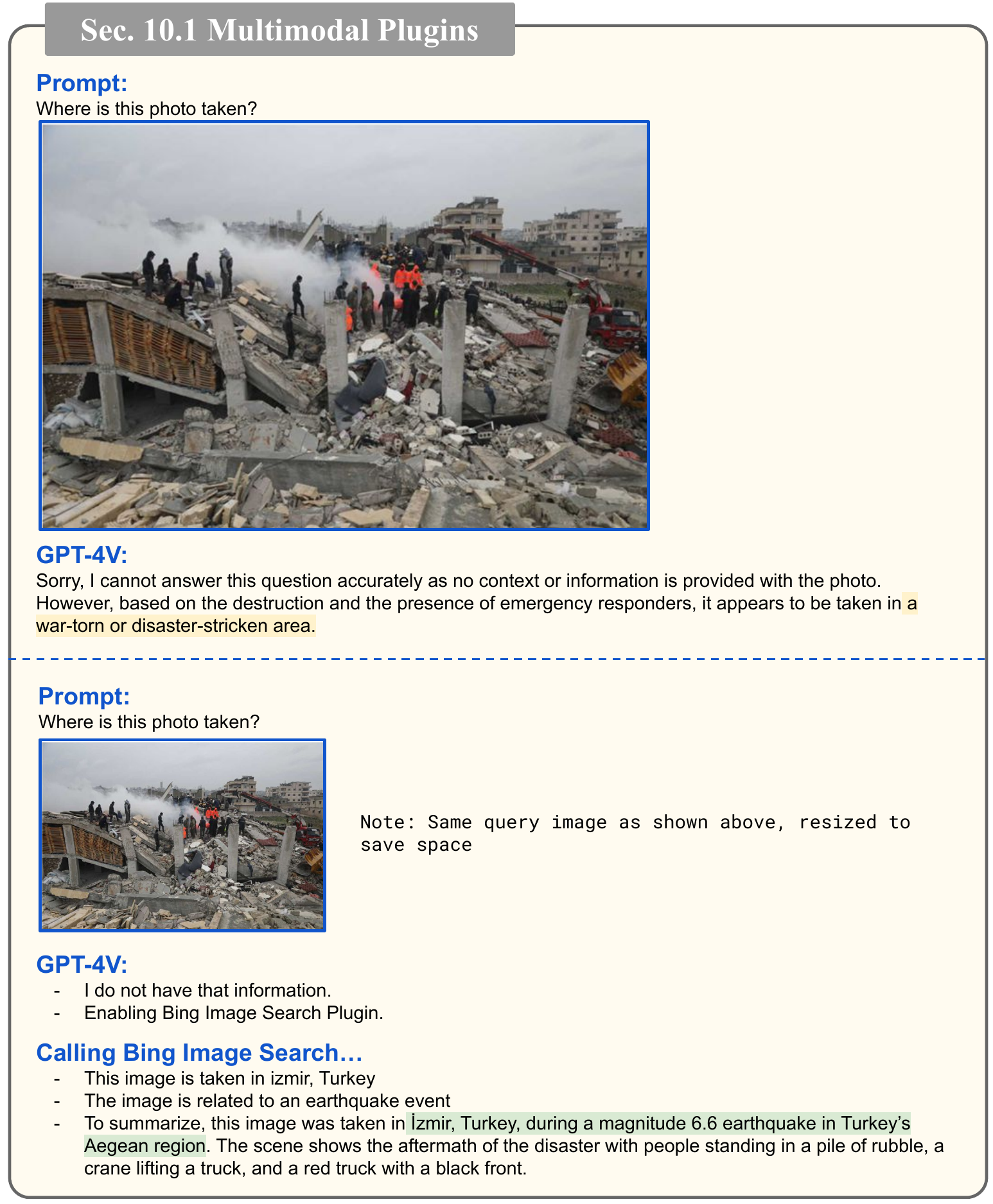}
\caption[Section~\ref{sec:tool_plugin}: multimodal plugins.]{Illustration of using the Bing Image Search~\cite{bingimage} plugin to enable \modelname~with time-sensitive knowledge (bottom, highlighted in \colorbox{greenhl}{green}). Note that the earthquake happened on February 6, 2023, which is after \modelname's training, thereby \modelname fails to identify the exact location without plugin (top). Check Section~\ref{sec:tool_plugin} for detailed discussions.}
\label{fig:tool_plugin}
\end{figure*}%
\begin{figure*}[h!]
\centering
\includegraphics[width=\textwidth]{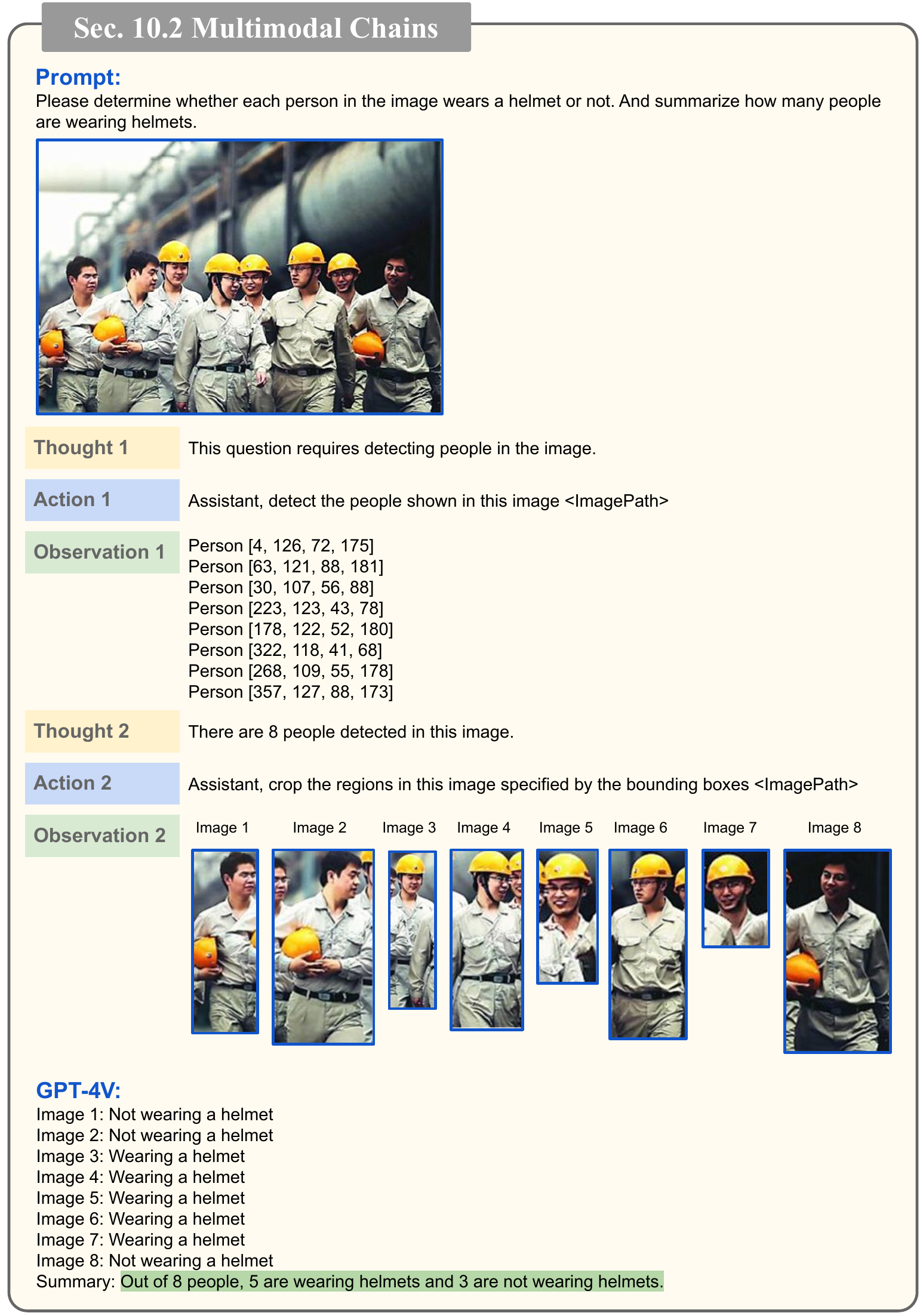}
\caption[Section~\ref{sec:chain_agent}: multimodal chains.]{Extending \modelname~to multimodal chains with ReAct~\cite{yao2022react,yang2023mm} for PPE Counting scenario. Check Section~\ref{sec:chain_agent} for detailed discussions.}
\label{fig:chaining}
\end{figure*}

\subsection{Multimodal Plugins}\label{sec:tool_plugin} 

In the context of LLMs, plugins~\cite{nakano2021webgpt,huang2022language,ahn2022can,schick2023toolformer,lu2022dynamic,paranjape2023art} play a crucial role in assisting LLMs for various tasks such as accessing the latest information, performing computations, or utilizing third-party services. These plugins are primarily designed to process inputs in natural language or inputs that can be interpreted as language, such as code and math equations. To illustrate the significance of multimodal plugins, such as Bing Image Search~\cite{bingimage}, especially in the context of LMMs, we present Figure~\ref{fig:tool_plugin}. By incorporating the Bing Image Search plugin, we empower \modelname~to acquire time-sensitive knowledge related to the input image. In the upper part of the figure, we demonstrate the limitations of \modelname~without Bing Image Search plugin. It fails to accurately answer the question, "Where was this photo taken?" due to the fact that the photo captures the aftermath of a massive earthquake that occurred on February 6, 2023, at the border of Turkey and Syria—a situation that took place after \modelname's training. Since constantly retraining the model with current information can be computationally intensive and expensive, plugins like search engines prove to be invaluable resources for the model to access up-to-date information. In the lower part of Figure~\ref{fig:tool_plugin}, we showcase the capabilities of \modelname~when equipped with the Bing Image Search plugin. It effectively leverages the retrieved information from the plugin, enabling accurate identification of the location İzmir, Turkey.

\subsection{Multimodal Chains}\label{sec:chain_agent}
Chaining with LLMs has been explored extensively in recent research~\cite{yao2022react,gao2022pal,trivedi2022interleaving,qin2023tool}. This approach goes beyond using a single plugin and instead establishes a system paradigm that integrates LLMs with a pool of plugins, enabling more advanced reasoning and interactions. By replacing language-only plugins with vision/multimodal experts such as image captioners, object detectors, or well-trained models for text-to-image generation and audio-to-text conversion, it becomes possible to construct a powerful multimodal chain with LLMs~\cite{wu2023visual,yang2023mm,suris2023vipergpt,shen2023hugginggpt,liang2023taskmatrix,lu2023chameleon}.

However, the interactions within these chains between LLMs and the plugins typically take place in text format. Although the plugins may accept multimodal inputs, they return results in text to enhance the knowledge of LLMs. There is a notable exception in the case of image synthesis/editing~\cite{wu2023visual}, where the plugins can generate images, but these images are not fed back into LLMs for further analysis or knowledge augmentation, as LLMs can only process language-based inputs.

In Figure~\ref{fig:chaining}, we present an illustration of how \modelname, can be extended to support multimodal chains with ReAct~\cite{yao2022react,yang2023mm}. This extension enables the plugins in the chain to provide multimodal information, which can then be collectively processed by \modelname~to achieve advanced reasoning in scenarios such as PPE counting. The entire chaining process shown in Figure~\ref{fig:chaining} is divided into two rounds of thought, action, and observation, with each round involving the activation of a specific plugin. In the first round, \modelname~deduces that person detection is necessary to count the number of people wearing helmets (Thought 1). Consequently, it calls the person detector tool (Action 1) and receives the coordinates of bounding boxes for each detected person in the image (Observation 1). Moving to the second round, based on the obtained bounding box information, \modelname~infers that there are a total of 8 people in the image (Thought 2). It then utilizes the image cropping tool to crop out individual images of each person according to their corresponding bounding box coordinates (Action 2). The resulting outputs (Observation 2) consist of 8 labeled images, numbered from image 1 to image 8. \modelname~subsequently determines whether each person in these images is wearing a helmet or not, and summarizes the total count of people wearing helmets.

Overall, this integration of LMMs with a pool of multimodal plugins opens up new possibilities for enhanced reasoning and interaction, leveraging the strengths of both language and vision capabilities. The flexibility of multimodal chains allows for a more comprehensive understanding and analysis of multimodal data, and can potentially lead to improved performance in various applications.

\subsection{Self-Reflection}
\label{sec:future-04reflect}
\begin{figure*}[h!]
\centering
\includegraphics[width=\textwidth]{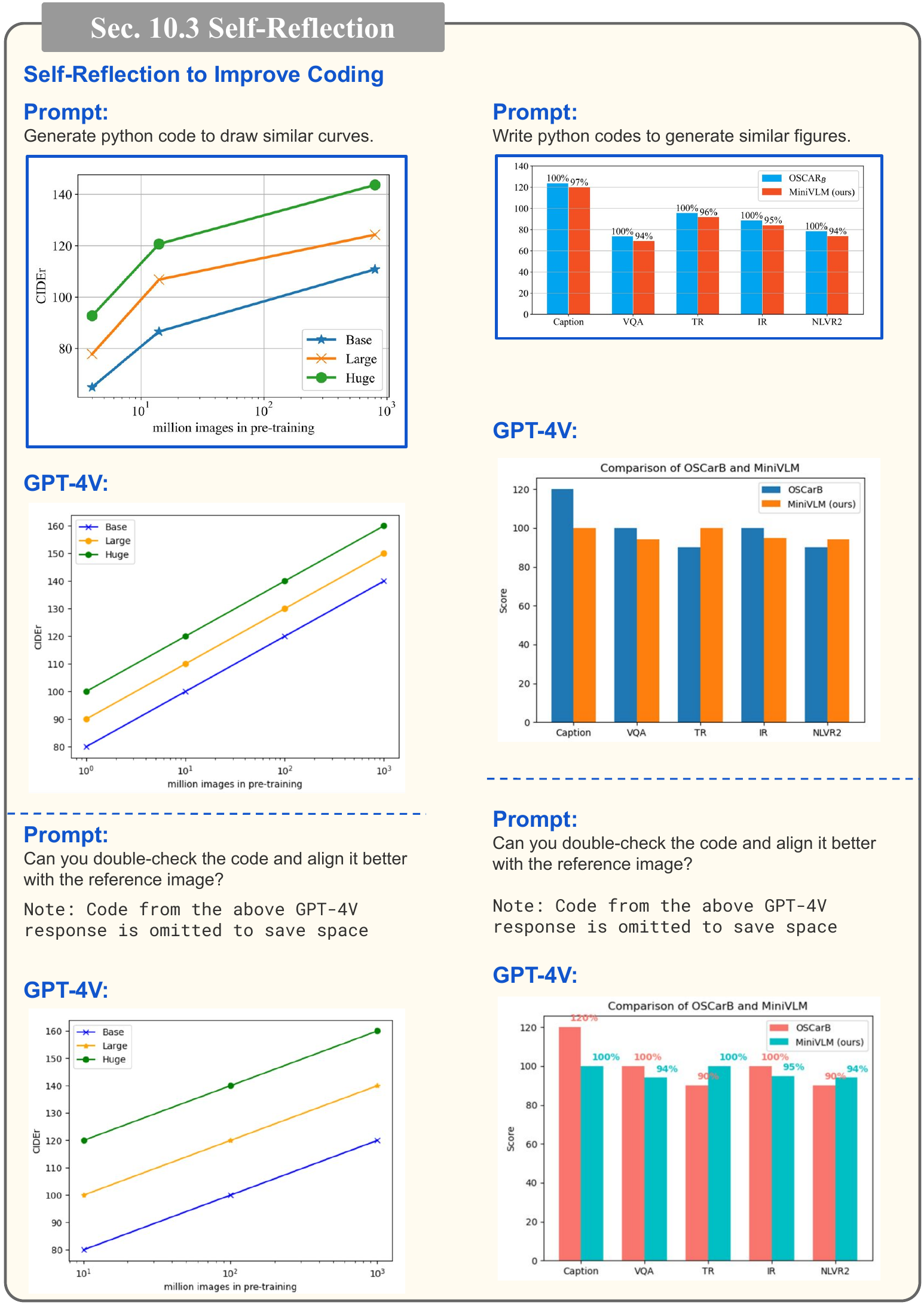}
\caption[Section~\ref{sec:future-04reflect}: self-reflection.]{Illustration of using self-reflection to improve the code for figure drawing.
Left: after reflection, the number of points in the curve aligns with the reference image. 
Right: the percentage is added to align with the reference image.
Check Section~\ref{sec:future-04reflect} for detailed discussions.}
\label{fig:selfreflection_code_figure}
\vspace{15pt}
\end{figure*}

\begin{figure*}[h!]
\centering
\includegraphics[width=\textwidth]{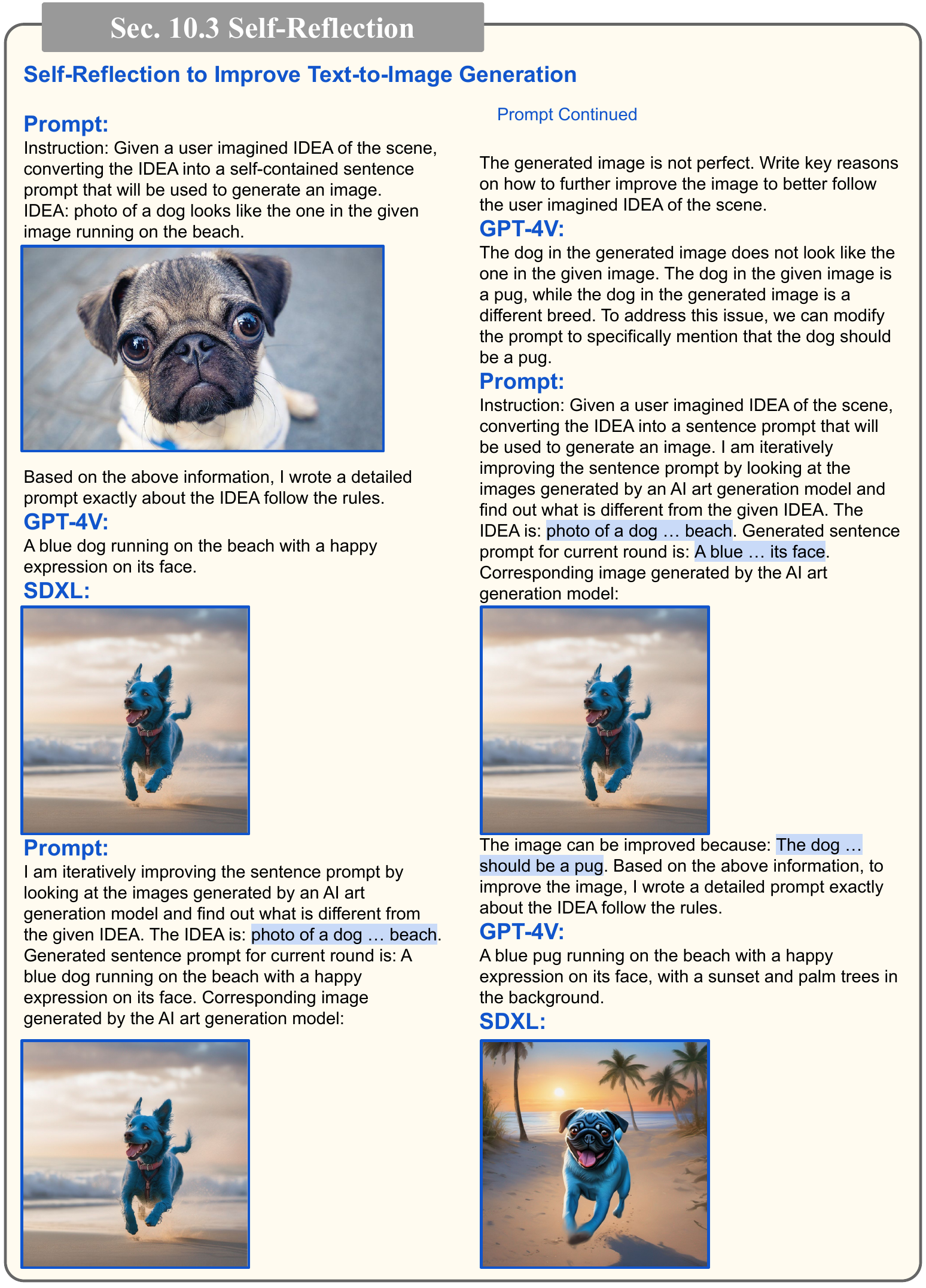}
\caption[Section~\ref{sec:future-04reflect}: self-reflection.]{Illustration of using self-reflection to improve the generated text prompts for a text-to-image model SDXL~\cite{podell2023sdxl}. \modelname~reflects the error in the initial prompt that it does not mention the dog's breed, and makes the correct revision. Check Section~\ref{sec:future-04reflect} for detailed discussions.}
\label{fig:selfreflection_t2i}
\vspace{15pt}
\end{figure*}

Figure~\ref{fig:selfreflection_code_figure}
demonstrates the application of self-reflection
~\cite{shinn2023reflexion,madaan2023self,kim2023language}
to improve the results shown in Figure~\ref{fig:code}.
As we can see, the self-reflected result is better aligned with the reference image. 
For example, on the left side, the number of data points is corrected from 4 to 3, while on the right side, the percentage is added back above the bar. 
Although the result is still not exactly identical,
it is evident that self-reflection can facilitate manual polishing.
Figure~\ref{fig:selfreflection_t2i} shows another example of self-reflection in improving the prompt generation for text-to-image models~\cite{podell2023sdxl}.

\begin{figure*}[h!]
\centering
\vspace{-30pt}
\includegraphics[width=\textwidth]{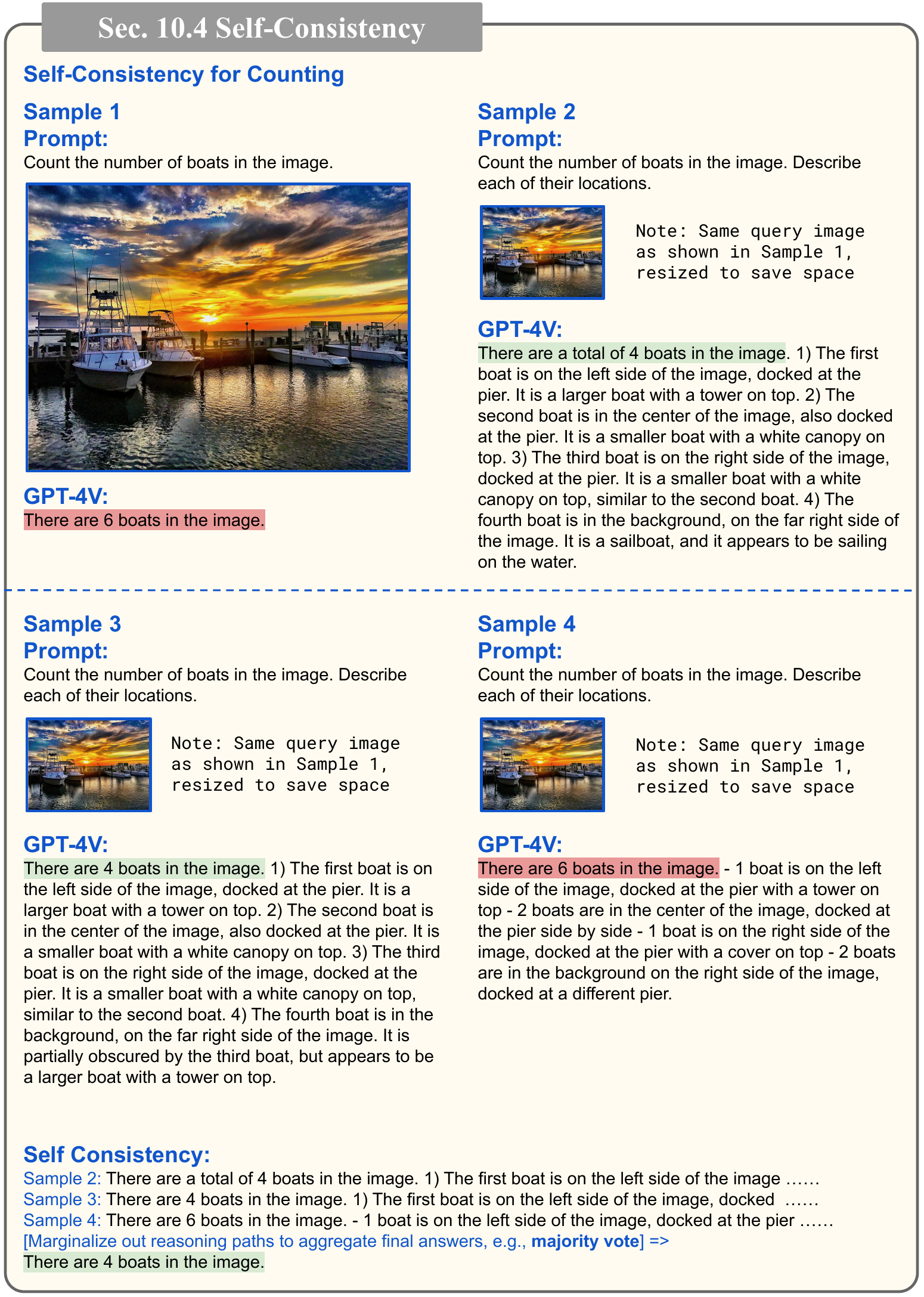}
\caption[Section~\ref{sec:future-05consist}: self-consistency.]{Improve the counting reliability with self-consistency~\cite{wang2022self}, which aggregates multiple counting results repeated on the \emph{same} image.
Check Section~\ref{sec:future-05consist} for detailed discussions.
}
\label{fig:selfconsistency}
\vspace{-10pt}
\end{figure*}

\subsection{Self-Consistency}
\label{sec:future-05consist}
Self-consistency~\cite{wang2022self} is a decoding strategy that aggregates multiple sampled outputs to produce the final answer, such as with the majority vote. Extended from marginalizing to aggregating final answers, Tree-of-Thoughts~\cite{yao2023tree} shows that the self-consistency idea can be applied to intermediate thoughts to improve the LLM reasoning performance. Figure~\ref{fig:selfconsistency} illustrates the use of self-consistency on \modelname~for counting problems. We sample multiple counting results by asking \modelname~to count the same image multiple times, either conducting multiple runs (Samples 2-4) or rephrasing the input text instruction (Samples 1,2). The example then uses the simple majority vote to aggregate the final answer of ``4 boats.'' We leave the comprehensive explorations of self-consistency LMMs to future works. %

\subsection{Retrieval-Augmented LMMs}
\label{sec:future-06retrieve}

Retrieval-Augmented LMMs~\cite{mialon2023augmented,lewis2020retrieval,guu2020retrieval,borgeaud2022improving,shi2023replug,peng2023check} enhances text generation by retrieving and integrating relevant information into prompts. The technique is particularly effective when specialized task-relevant information is needed, such as expert knowledge in a highly-specialized expert domain, the most recent information that may differ from LLMs' memory, and the customizable information that varies from user to user. We imagine retrieval augmentation continues to play an essential role in LMMs. Figure~\ref{fig:grocery} shows an example of retrieval-augmented LMMs helping grocery checkout. Since the produces' image-text-price triplets are different in each store, it would be beneficial to retrieve them from the store's database and yield the correct checkout information. Similarly, in Figure~\ref{fig:cus_cap_same}'s the customized captioning scenario, we imagine the system may automatically retrieve the family members' photos from the album and achieve the customized captioning. %

\section{Conclusions}
\subsection{Summary and Conclusions}
In this report, our primary focus is on probing \modelname~across various application scenarios. The findings reveal its remarkable capabilities, some of which have not been investigated or demonstrated in existing approaches. While we strive to uncover as many of these capabilities as possible, we acknowledge that our presentation may not be exhaustive. Nevertheless, this report can serve as a reference for future research aimed at exploring additional uses of \modelname, deepening the understanding of LMMs, and building even more powerful LMMs.

\subsection{Towards Future LMMs}

The weaknesses and limitations of GPT models have been extensively discussed in related reports~\cite{gpt4,gpt4v,bubeck2023sparks}. 
In this section, we briefly focus on presenting our perspective on future research directions.

Models like GPT-1, GPT-2, and GPT-3 function primarily as text-in-text-out systems, capable of processing natural language only. GPT-4 (no vision) demonstrates unparalleled competence in text understanding and generation, while \modelname~exhibits a strong ability to comprehend the image domain as well.

As a natural progression, LMMs should be able to generate 
interleaved image-text content, such as producing vivid tutorials containing both text and images, to enable comprehensive multimodal content understanding and generation. Additionally, it would be beneficial to incorporate other modalities, such as video, audio, and other sensor data, to expand the capabilities of LMMs.

Regarding the learning process, current approaches predominantly rely on well-organized data, such as image-tag or image-text datasets. However, a more versatile model may be able to learn from various sources, including online web content and even real-world physical environments, to facilitate continuous self-evolution.

\section*{Acknowledgment}
We express our gratitude to all contributors from OpenAI for their technical efforts on the GPT-4V project~\cite{gpt4,gpt4v,gpt4vcontribution,gpt4vblog}, and we are profoundly thankful to OpenAI for granting early access to their remarkable tool. Our sincere appreciation goes to Misha Bilenko for his invaluable guidance and support. We also extend heartfelt thanks to our Microsoft colleagues for their insights, with special acknowledgment to John Montgomery, Marco Casalaina, Gregory Buehrer, Nguyen Bach, Gopi Kumar, Luis Vargas, Kun Wu, Meenaz Merchant, Jianfeng Gao, Matt Lungren, Sheela Agarwal, Yumao Lu, Thomas Soemo, Fisayo Okikiolu, Ce Liu, Michael Zeng, Faisal Ahmed, Ehsan Azarnasab, and Lin Liang for their constructive feedback. We also thank Yingkai Yu for helping to create screenshots on GUI Navigation.

\clearpage
{
\bibliographystyle{plain}
\bibliography{egbib}
}

\end{document}